\documentclass[12pt, english, chapterprefix, numbers=noenddot]{book}

\usepackage{listings}
\usepackage{amsmath}
\usepackage{amsthm}
\usepackage[a4paper]{geometry}
\usepackage{color}

\usepackage{float}
\usepackage{graphicx}
\usepackage{nomencl}
\providecommand{\printnomenclature}{\printglossary}
\providecommand{\makenomenclature}{\makeglossary}
\makenomenclature

\makeatletter

\providecommand\textquotedblplain{%
  \bgroup\addfontfeatures{Mapping=}\char34\egroup}
\providecommand{\tabularnewline}{\\}
\floatstyle{ruled}
\newfloat{algorithm}{tbp}{loa}[chapter]
\providecommand{\algorithmname}{Algorithm}
\floatname{algorithm}{\protect\algorithmname}

\usepackage{url}

\usepackage{fancyhdr}
\makeatletter
\renewcommand*{\cleardoublepage}{\clearpage\if@twoside \ifodd\c@page\else
\hbox{}%
\thispagestyle{empty}%
\newpage%
\if@twocolumn\hbox{}\newpage\fi\fi\fi}
\makeatother

\usepackage{booktabs}

\usepackage{chngcntr}

\counterwithin{table}{chapter}
\counterwithin{figure}{chapter}
\counterwithin{algorithm}{chapter}
\numberwithin{equation}{chapter}

\AtBeginDocument{

}

\makeatother

\usepackage{array}
\usepackage[table]{xcolor}
\usepackage{colortbl}
\usepackage{csquotes}

\usepackage[english]{babel}

\usepackage[style=numeric,citestyle=numeric,sorting=nyt,sortcites=true,autopunct=true,autolang=hyphen,hyperref=true,abbreviate=false,backref=true,backend=biber,giveninits=true]{biblatex}
\usepackage{imakeidx}
\makeindex[columns=2, title=Index, options= -s indexStyle.ist]
\newcommand\gobbleone[1]{}

\newcommand{\seeonlyindex}[2]{\index{#1@#1\protect\gobbleone|seeonly{#2}}}

\usepackage{hyperref}
\hypersetup{colorlinks=true, allcolors=blue}

\frontmatter

\begin{document}

\begin{titlepage}

\begin{center}
\includegraphics[width=8cm]{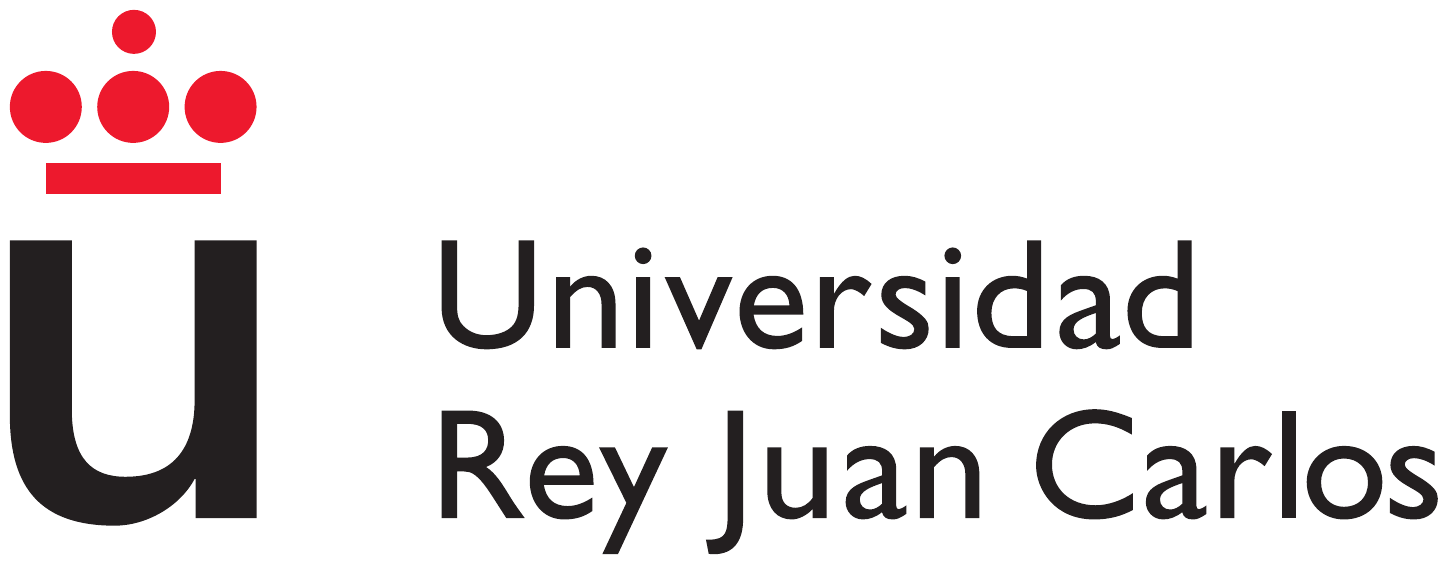}
\par\end{center}

\begin{center}
{\large{}\textbf{THESIS}}{\large\par}
\par\end{center}

\vspace{2cm}

\begin{center}
{\LARGE{}\textbf{Continuous Offline Handwriting Recognition using Deep Learning Models}}{\LARGE\par}
\par\end{center}

\vspace{2cm}

\begin{center}
{\large{}\textbf{Author}}{\large\par}
\par\end{center}

\begin{center}
{\large{}Jorge Juan Sueiras Revuelta}{\large\par}
\par\end{center}

\vspace{1cm}


\vspace{1cm}

\begin{center}
{\large{}Information technologies and Communications program}{\large\par}
\par\end{center}

\begin{center}
{\large{}Spain}{\large\par}
\par\end{center}

\vspace{1cm}

\begin{center}
{\large{}2021}{\large\par}
\par\end{center}

\end{titlepage}

\chapter*{Acknowledgements}

Una tesis es un largo viaje, quiero dar las gracias aquí a las personas que me han acompañado y apoyado a lo largo del mismo.

En primer lugar, quiero dar las gracias a mis directores de tesis José Vélez y Ángel Sánchez, sin los cuales este trabajo no hubiera sido posible. El buen talante siempre presente y la libertad que me disteis para perseguir mi curiosidad y mis intereses, han sido fundamentales para que este viaje haya sido una gran experiencia. José, gracias por compartir tu sabiduria sobre tantas áreas, pero especialmente en visión por computador, por tus consejos y por estar siempre ahí. Ángel, gracias por el apoyo y la crítica constructiva que tanto han ayudado en la publicación de los resultados de esta tesis.

Gracias a mi mujer Lua, que ha sido mi consejera y confidente y me ha apoyado incondicionalmente desde el principio hasta el final del viaje. Gracias también a mis hijos Mario, Diego y Tello que han aportado el contrapunto necesario al trabajo.

Gracias también a Victoria, con la que compartí una parte de este camino al inicio. Juntos aprendimos a resolver los problemas que van apareciendo durante el desarrollo de una tesis.

Gracias a Daniel Vélez, por presentarme a su hermano José. Ese fue realmente el punto de inicio de este viaje.

Gracias a Rocío por aportar su experiencia investigadora que me ha ayudado a enfocar mejor mi trabajo.

Gracias también a todas aquellas personas que he conocido durante esta andadura y que me han hecho el trabajo mas fácil y ameno, en particular a Juanjo.

\chapter*{Abstract}

Handwritten text recognition is an open problem of great interest in the area of automatic document image analysis. The transcription of handwritten content present in digitized documents is significant in analyzing historical archives or digitizing information from handwritten documents, forms, and communications. The problem has been of great interest since almost the beginning of the development of machine learning algorithms. In the last ten years, great advances have been made in this area due to applying deep learning techniques to its resolution.

This Thesis addresses the offline continuous handwritten text recognition (HTR) problem, consisting of developing algorithms and models capable of transcribing the text present in an image without the need for the text to be segmented into characters. For this purpose, we have proposed a new recognition model based on integrating two types of deep learning architectures: convolutional neural networks (CNN) and sequence-to-sequence (seq2seq) models, respectively. The convolutional component of the model is oriented to identify relevant features present in characters, and the seq2seq component builds the transcription of the text by modeling the sequential nature of the text. 

For the design of this new model, an extensive analysis of the capabilities of different convolutional architectures in the simplified problem of isolated character recognition has been carried out in order to identify the most suitable ones to be integrated into the continuous model. Additionally, extensive experimentation of the proposed model for the continuous problem has been carried out to determine its robustness to changes in parameterization. The generalization capacity of the model has also been validated by evaluating it on three handwritten text databases using different languages: IAM in English, RIMES in French, and Osborne in Spanish, respectively.

The new proposed model provides competitive results with those obtained with other well-established methodologies and opens the door to new lines of research focused on applying seq2seq models to the continuous handwritten text recognition (HTR) problem.

\chapter*{Resumen}

El reconocimiento de texto manuscrito es un problema abierto y de gran interés en área del análisis automático de documentos. La transcripción del contenido manuscrito que aparece escrito en documentos digitalizados es especialmente importante en el análisis de archivos históricos o en la digitalización de la información de documentos, formularios y comunicaciones manuscritas. El problema ha tenido un gran interés desde prácticamente los inicios del desarrollo de algoritmos de aprendizaje automático. En estos últimos 10 años se han producido grandes avances en el mismo a raíz de la aplicación de técnicas de aprendizaje profundo a su resolución.

En esta Tesis se aborda el problema del reconocimiento de texto manuscrito continuo, consistente en desarrollar algoritmos y modelos capaces de transcribir el texto presente en una imagen sin necesidad de que dicho texto esté segmentado en caracteres. Para ello, se ha propuesto un nuevo modelo de reconocimiento basado en la integración de dos tipos de arquitecturas de aprendizaje profundo, las redes convolucionales y los modelos sequence-to-sequence (seq2seq), respectivamente. El componente convolucional del modelo se orienta a identificar las caracteristicas relevantes de los caracteres presentes en el texto y el componente seq2seq construye la transcripción del mismo modelando la naturaleza secuencial del texto. 

Para el diseño de este nuevo modelo se ha realizado un extensivo análisis de las capacidades de distintas arquitecturas convolucionales en el problema simplificado de reconocimiento de caracteres aislados con el objetivo de identificar las más adecuadas para integrarlas en el modelo continuo. Adicionalmente, se ha realizado una extensiva experimentación del modelo propuesto para el problema de reconocimiento de texto continuo a nivel de palabras, lo que ha permitido determinar su robustez frente a cambios en la parametrización.  También se ha validado la capacidad de generalización del modelo mediante la evaluación del mismo usando tres bases de datos de texto manuscrito en diferentes idiomas: IAM en inglés, RIMES en francés y Osborne en español, respectivamente.

El nuevo modelo proporciona resultados competitivos con los obtenidos aplicando otras metodologías bien establecidas y abre nuevas puertas a líneas de investigación centradas en la aplicación de modelos seq2seq al problema del reconocimiento de texto manuscrito continuo.


\tableofcontents{}

\listoftables

\listoffigures

\listof{algorithm}{Algorithms}

\settowidth{\nomlabelwidth}{notepad}
\printnomenclature{}

\mainmatter


\chapter{Introduction}
\label{chapter:introduction}

This chapter describes the main motivations for the research carried out in this Thesis. It also details the objectives for this research and outlines the Thesis chapters.

\section{Motivation}

\begin{figure}[!ht]
\centering 
\includegraphics[width=14cm]{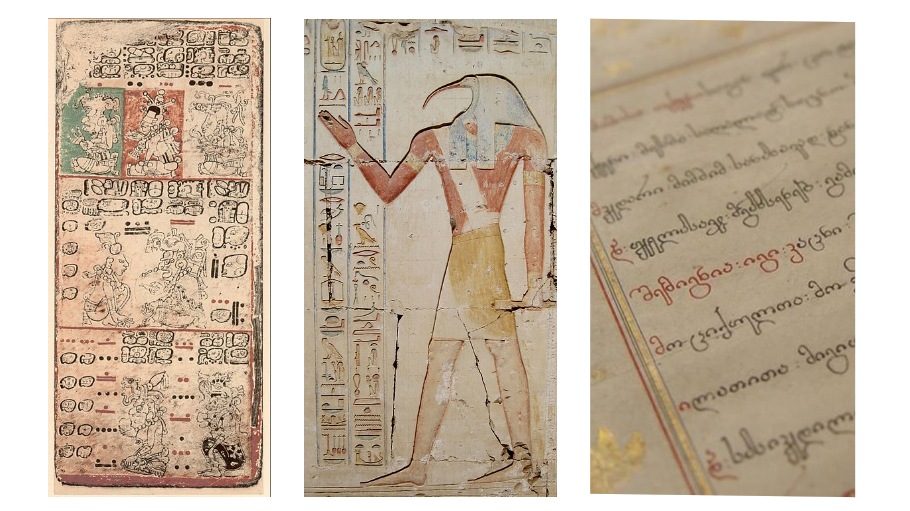} 
\caption{Historical writing examples. Maya  (12th century AC), Egyptian hieroglyphs (30th century BC) and Georgian manuscript (6th century AC). (Source: Wikipedia)}
\label{fig:hitoricalWritting}
\end{figure}

Literacy is the ability to read and write in, at least, one method of writing, and the first and principal writing method that is learned 'through' education is handwriting. The ability to write is one of the essential knowledge of human beings, recognized as a fundamental right in 1997 by the Hamburg Declaration\footnote{\textit{Literacy, broadly conceived as the basic knowledge and skills needed by all in a rapidly changing world, is a fundamental human right}} \cite{HamburgDeclaration}. Until the invention of the printing press in the 15th century, handwriting text was the only method of transmitting and perpetuating knowledge, stories and beliefs over time. Writing has been present in all cultures and has taken many forms throughout the history, see Fig. \ref{fig:hitoricalWritting}. In fact, the appearance of writing is precisely one of the essential factors in determining the beginning of human history.

Handwriting is present in our daily lives, usually in notes, lists or other short texts in everyday life. But it is also used more systematically in other areas, such as taking notes in academic classes or in business meetings. Moreover, despite the advent of new technologies such as computers, tablets or smartphones, handwriting is still the preferred method for many people to capture their ideas or thoughts, at least initially. Handwriting can also be done virtually anytime and anywhere with a minimum of technology: with a pencil and a notebook, chalk on a wall or the hand in the sand.

In fact, the main disadvantage of the handwriting text is that it must be digitized to facilitate its preservation, arrangement, and dissemination. Now, we are in the digitization era. All the information is stored and indexed in digital formats, and all the business processes must be digital. With all data and knowledge stored in digital databases, multiple advantages are achieved in terms of accessibility and analysis of the information. In this context, the capacity to recognize and digitize the content of handwriting text is necessary to extract shareable knowledge from it. 

\begin{figure}[!ht]
\centering 
\includegraphics[width=10cm]{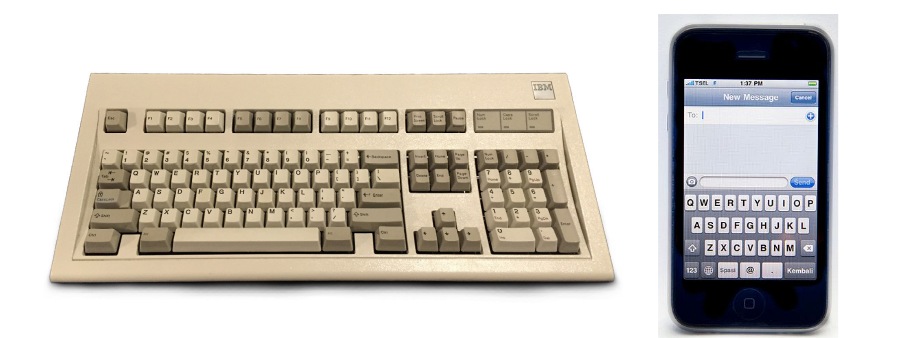} 
\caption{Computer and mobile keyboards. (Source: Wikipedia)}
\label{fig:IBM_Model_M}
\end{figure}

It is very likely that if the handwritten text could be converted into digital text in a simple way, its use would increase significantly. However, at present, we are far from having a simple and accurate way to convert handwritten text into a digital format. The current common way to generate digital text is through a keyboard like the ones in Fig. \ref{fig:IBM_Model_M}. The keyboard is not natural and it is effective only for experienced people. An alternative for generating digital text is the use of voice. Speech recognition has advanced significantly in accuracy in recent years and it works well with a continuous and linear stream of content \cite{chorowski2015attention}. But translating the thoughts into text is rarely a continuous and linear process. This unnatural operation is a drawback in popularizing speech as a digital text generation tool. So is the fact that it cannot be used in all contexts, such as for taking notes in an academic class or in work meetings.

Automatic handwritten Text Recognition (hereafter abbreviated as HTR)\nomenclature{HTR}{Handwritten Text Recognition} is a problem that has been active for a long time \cite{plamondon2000online}. It is a relatively simple task for people but very difficult to model. It is currently considered in its general form as an unsolved problem yet and under active research. Several conferences and periodic publications are specialized in handwriting, as the International Conference on Frontiers in Handwriting Recognition (ICFHR) or DAS (Document Analysis Systems), and others dedicate specific tiers to the handwriting recognition problem such as the International Conference on Document Analysis and Recognition (ICDAR). There are also specialized journals in this field like The International Journal of Document Analysis and Recognition (IJDAR).
  \nomenclature{DAS}{Document Analysis Systems}
  \index{Conferences!DAS}
  \nomenclature{ICFHR}{International Conference on Frontiers in Handwriting Recognition}
  \index{Conferences!ICFHR}
  \nomenclature{ICDAR}{International Conference on Document Analysis and Recognition}
  \index{Conferences!ICDAR}
  \nomenclature{IJDAR}{The International Journal of Document Analysis and Recognition}
  \index{Journals!IJDAR}

\begin{figure}[!ht]
\centering 
\includegraphics[width=14cm]{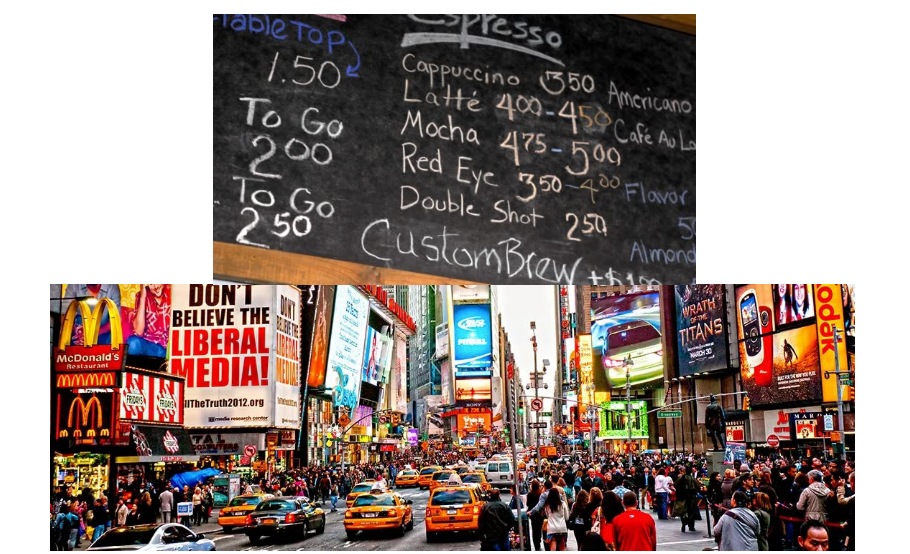} 
\caption{Natural scenes with handwritten and typographic text. (Source: OpenImages)}
\label{fig:naturalSceneText}
\end{figure}

The need to recognize text automatically appears in several contexts. The main problem is recognizing the text in digital images obtained by scanning paper pages from documents, books, letters, or forms. However, the problem also appears with images of real-world scenes that can include posters, labels, or nameplates among others. This need may also appear in retrieving text from videos. For example, in order to provide the autonomous driving systems the capacity to understand the names that appear in informative signs of the road. In the cases of managing real-world scene images, the problem is named Scene Text Recognition. and is an active research area \cite{wang2011end} \cite{jaderberg2015spatial} \cite{shi2016end}. Fig. \ref{fig:naturalSceneText} shows some examples of natural scenes that contains text.
  \index{Scene text recognition}
  
The offline continuous HTR problem \cite{graves2009offline} consists of the automatic transcription of images that contain handwritten text. The problem has been extensively studied in the literature \cite{sayre1973machine} \cite{plamondon2000online} \cite{graves2009offline} \cite{yuan2012offline}. This specific problem presents relevant applications such as bank cheque processing \cite{impedovo1997automatic}, automatic address reading \cite{manisha2016role} or content extraction from digitized databases of historical documents \cite{sanchez2014handwritten}. Despite recent advances, significant differences in the writing of individuals and in the imprecise nature of handwritten characters make this recognition task hard, and it remains as an open research problem \cite{kang2021candidate}.

\begin{figure}[!ht]
\centering 
\includegraphics[width=14cm]{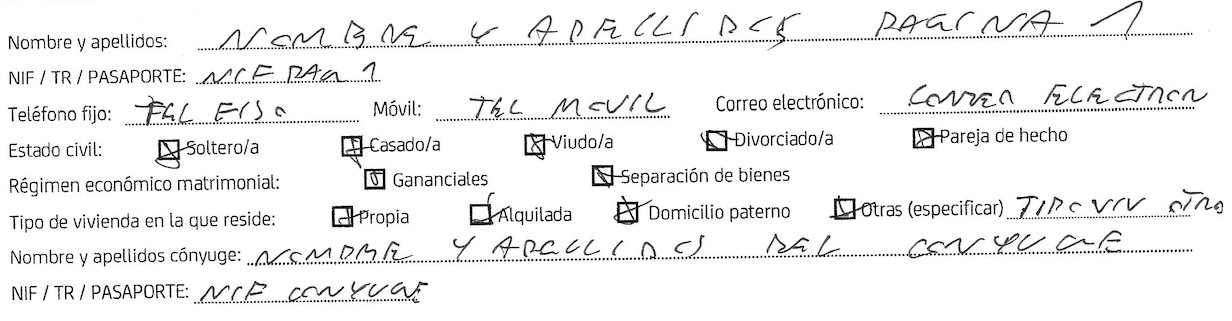} 
\caption{Example of personal data collection form containing both handwritten and printed text.}
\label{fig:formSample}
\end{figure}

Another relevant application of HTR is the automatic processing of forms such as the one in Fig. \ref{fig:formSample} . In countless administrative tasks, forms are usually completed with handwritten text and often have to be digitized afterwards. This digitization is done manually. It is a repetitive and tedious task that consumes large amounts of work time. The automation of the task requires high precision HTR systems, because personal data or contact info like phone number or email usually requires precise transcription. 

In the industry, there have been, for many years, solutions to solve the handwriting problem in some specific areas. For example, the French company A2iA\footnote{https://www.a2ia.com/en} offers a solution to read bank check amounts. In these cases, where the recognition of handwritten text is performed on a limited domain such as numerical amounts, the accuracy of these industrial systems is sufficiently high to surpass manual management. 
  \index{Company!A2IA}
  
\begin{figure}[!ht]
\centering 
\includegraphics[width=14cm]{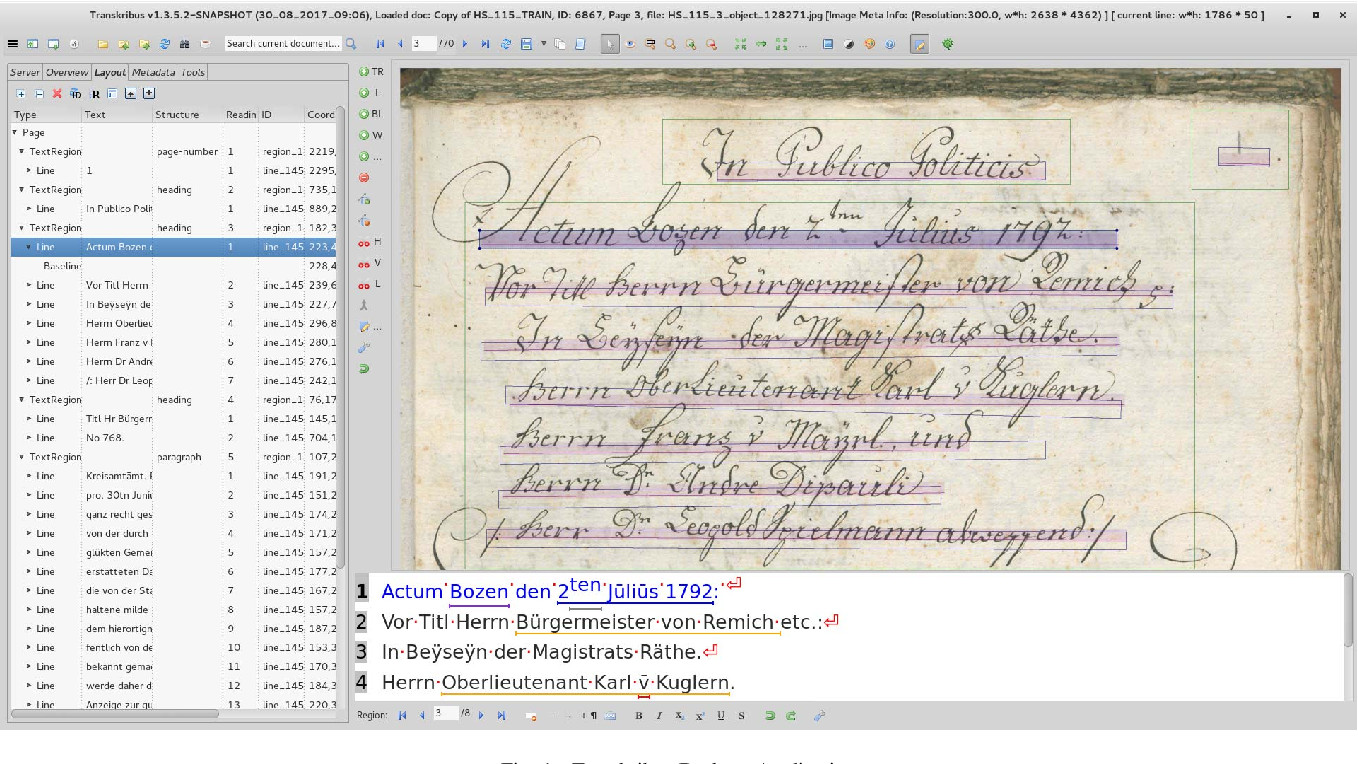} 
\caption{Transkribus interface example image.}
\label{fig:transcribusExample}
\end{figure}

In recent years industrial products and services have appeared to ally in the general handwriting problem for localizing and read handwriting text in any type of documents. The most relevant are Azure Cognitive Services from Microsoft \footnote{https://westus.dev.cognitive.microsoft.com/docs/services} and the Detect Handwriting in Images service of the Google Cloud Vision API \footnote{https://cloud.google.com/vision/docs/handwriting}. Another initiative coming from the academic world is Transkribus, \cite{kahle2017transkribus} to recognize and index historical handwritten documents. This last one has led to the development of a popular software\footnote{https://readcoop.eu/transkribus} for automatic recognition of text in historical documents. Fig. \ref{fig:transcribusExample} shows an example of the transcription interface in Transkribus desktop software.
  \index{Company!Google}
  \index{Company!Microsoft}
  \index{Transkribus}

But these general-purpose systems are limited by the current state-of-the-art of the general HTR problem in terms of error. That is why they can only be applied when a certain significant margin of error can be considered acceptable. For example, in the digitization of documents for the application of statistical text analysis models. However, their use is very complicated or not feasible in problems that require high accuracy, for example, the previous case of transcription of form fields.

\begin{figure}[!ht]
\centering 
\includegraphics[width=10cm]{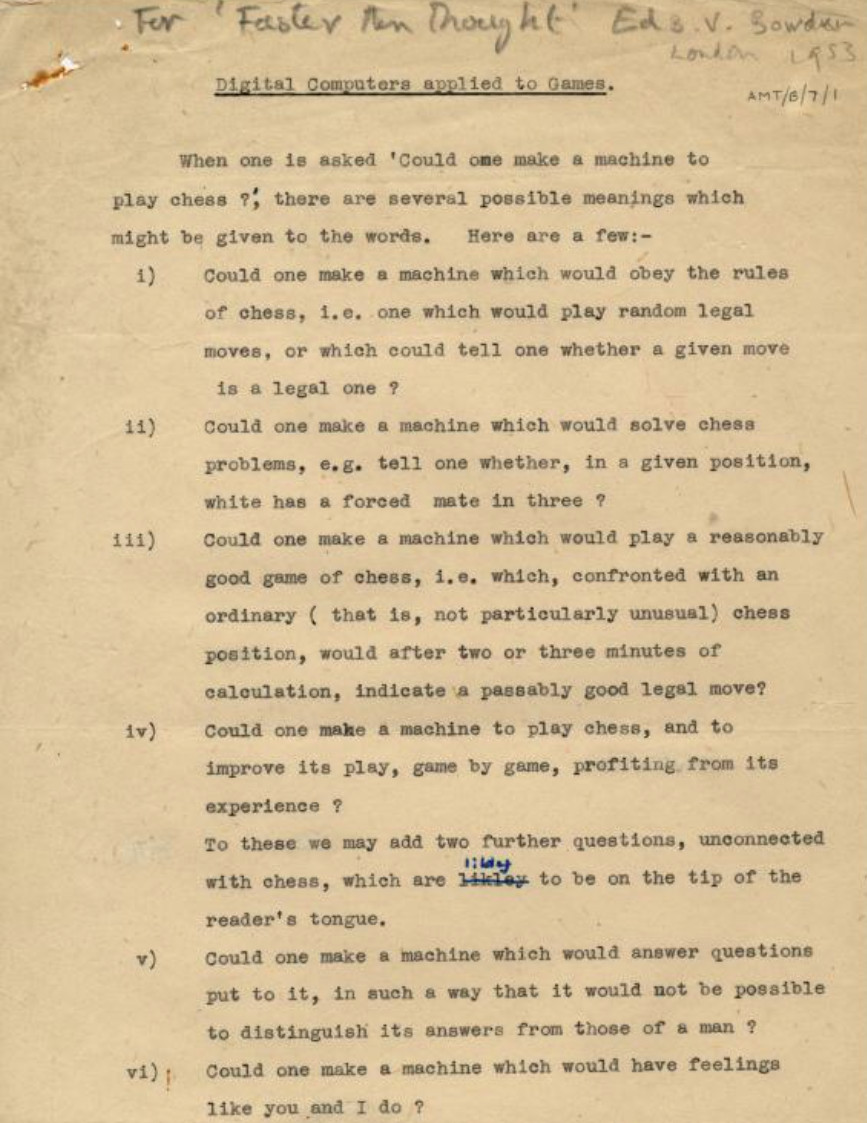} 
\caption{Image of an original document of the paper 'Digital computers applied to games' from The Alan Turing Digital Archive. The image is also an example of document that combines handwritten and typographic text.}
\label{fig:TuringText}
\end{figure}

Arthur Samuel \cite{5389202} defines \textit{Machine Learning} as the field of study that gives computers the ability to learn without being explicitly programmed. More specifically machine learning consists of the automatic development of computer algorithms capable of reproducing a complex task from previous experiences encoded in data \cite{bishop2006pattern}. For example, to teach a machine to play chess, make sure that it follows the game's rules. This can be easily encoded through a standard algorithm. But it is also necessary that the machine can play games reasonably well, and it is not obvious that standard algorithms can achieve this. Machine learning algorithms built by analyzing previous games encoded and stored in digital data are used in cases like this. Alan Turing well defined the different levels at which a machine can approach a complex problem like chess in \cite{turing1953digital}. An excerpt from the first page of the original typed document of this paper with handwritten notes by the author is included in Fig. \ref{fig:TuringText}.
  \index{Machine learning}
  \nomenclature{ML}{Machine Learning}
  
For years Machine Learning has been successfully applied to structured data in the form of tables. Each table record corresponded to a learning example and each column was a numerical attribute with which to build the algorithm in the form of a mathematical formula. It has been applied for a long time to business problems such as credit scoring \cite{lessmann2015benchmarking} or fraud detection \cite{Wang2010Survey}. However, machine learning algorithms failed to learn from unstructured data, such as images or text. This has changed drastically in recent years with the emergence of \textit{Deep Learning}. Deep Learning are neural models in which the learning function is defined as the composition of a large number of simpler functions called layers \cite{Goodfellow-et-al-2016}. Understanding function composition in the mathematical sense of $(f\circ g)(x) = f(g(x))$. These Deep Learning models are being successfully used in especially complex machine learning problems such as image classification \cite{krizhevsky2012imagenet}, speech recognition \cite{chorowski2015attention} or machine translation \cite{bahdanau2014neural}.
  \index{Deep learning}
  \index{Automatic speech recognition}
  
The continuous HTR problem is a decoding problem \cite{graves2009novel}. An image containing a sequence of strokes is decoded into the sequence of characters that form those strokes. In this sense, it is similar to speech recognition problem, where an audio sequence is decoded into the sequence of the words of this audio. The \textit{machine translation} problem, where a sequence of words in one language is decoded into its translation in another language, present also similarities with the HTR problem. Recent advances in the above problems through Deep Learning models specially oriented to sequence decoding, named as sequence-to-sequence (seq2seq) models \cite{sutskever2014sequence}, can be very relevant to solve the HTR problem.
\index{Sequence-to-sequence}

\section{Research objectives}

This Thesis aims to propose a solution to general handwritten text recognition at the word level. This problem is defined as the development of algorithms capable of transcribing into text any image in a handwritten word, regardless of its content, author, or nature. Additionally, to achieve the above objective, it will be analyzed first the problem of isolated character recognition. To propose this approach, we have taken inspiration from how people learn to read, who first learn to identify each letter separately and then learn the words.

This general objective can be detailed into the following specific sub-objectives: 

\begin{itemize}

  \item Conduct a detailed historical study of the different approaches to the offline continuous HTR problem over the years. This study deepen into the relevant contributions of each approach and identifies the main limitations and criticisms. Having a deep understanding of how this problem has been proposed to be solved allows making the best decisions to address it with a new approach.

  \item Explore new normalization algorithms of handwritten text images to reduce the variability of the text. In this way, it is intended to facilitate the construction of word recognition models, with particular emphasis on the optimization of algorithms that correct the italic feature of the text.  

  \item Contribute to research in character recognition through the construction of a new image database of handwritten characters. This database will be oriented to evaluate recognition models on the broadest possible set of characters present in handwritten text. It will thus complement the existing character databases.

  \item Study the application to the problem of character image classification of deep learning models that have worked well in other image classification problems. In particular, we will analyze convolutional network models, which are the ones that have obtained the best results in this type of problems in the last years \cite{krizhevsky2012imagenet}.

  \item Study the word recognition problem under the perspective that it is a problem of transforming an input data sequence of word image column pixels into an output sequence of the ordered characters that form the word. Conduct experiments to determine whether this problem could be solved with algorithms used in other similar problems which have more active research, like language translation or speech recognition. These problems have in common with the HTR problem that, in all cases, it is a matter of decode an arbitrary length sequence of input data into another arbitrary length sequence of output data.

  \item Build sequence-to-sequence (seq2seq) network models applied to the HTR problem. This type of model has achieved essential breakthroughs in the problems mentioned above of language translation \cite{sutskever2014sequence} or speech recognition \cite{chorowski2015attention}. Experiment with different deep learning architectures of such models to determine the optimal configurations when applied to the general HTR problem.
  \index{Sequence-to-sequence}
    
\end{itemize}

\section{Thesis outline}

This chapter has introduced the motivation of our work. We also propose the main objective and detail the specific objectives of this Thesis.

In Chapter \ref{chapter_analysis} we provide an overview of the offline HTR problem putting it in the context of text recognition analysis in general. Next, we include the details about the experiment design and evaluation metrics. 

In Chapter \ref{chapter_Handwriting_Databases} we review the primary databases at character and word level used to evaluate the results of the experiments of the recognition models.

In Chapter \ref{chapter_background} we include a detailed review of the works in offline HTR general, focused on the more active research lines of the last years.

Chapter \ref{c_character} includes our proposed solutions to the offline handwritten recognition of isolated characters. 

Chapter \ref{chapter_word_models} details the solutions and seq2seq model architectures that we propose to solve the handwriting recognition problem at the word level. We also analyze the difficulties existing in the literature to identify comparable results between authors, proposing a framework of standardization of results to facilitate such comparisons.

Finally, in Chapter \ref{chapter_conclusions} we include our conclusions from this work and the future work is proposed.


\chapter{Analysis of the offline handwritten text recognition problem}
\label{chapter_analysis}

This chapter analyzes the general offline HTR problem. This is defined and contextualized in the more general framework of text analysis and recognition. Section \ref{Sect:Overview_of_the_problem} introduces the problem, contextualizes and narrows it to define the framework of this Thesis. Next, all the elements necessary for the experiments are defined. In Section \ref{c2-section:Image_processing} image processing techniques applied as image preprocessing are introduced. Section \ref{SubSect:Data augmentation strategies} details the data augmentation strategies used. The different alternatives to decoding predictions are reviewed in Section \ref{SubSect:Decoding_predictions}. Finally, the evaluation metrics are detailed in Section \ref{SubSect:Evaluation_metrics}.

\section{Overview of the problem}
\label{Sect:Overview_of_the_problem}

The handwriting recognition problem (HTR) consists of developing models and algorithms able of transcribing the handwritten text into a digital format. The main difficulty of the problem lies in the significant variability of the handwritten text. Each person's handwriting is different from others (interpersonal variability), and even the same person writes the same word in many different ways (intrapersonal variability). Factors such as writing speed, typography size, type of paper and pen used, and even emotional state further increase this variability.

\begin{figure}[!ht]
\centering 
\includegraphics[width=7cm]{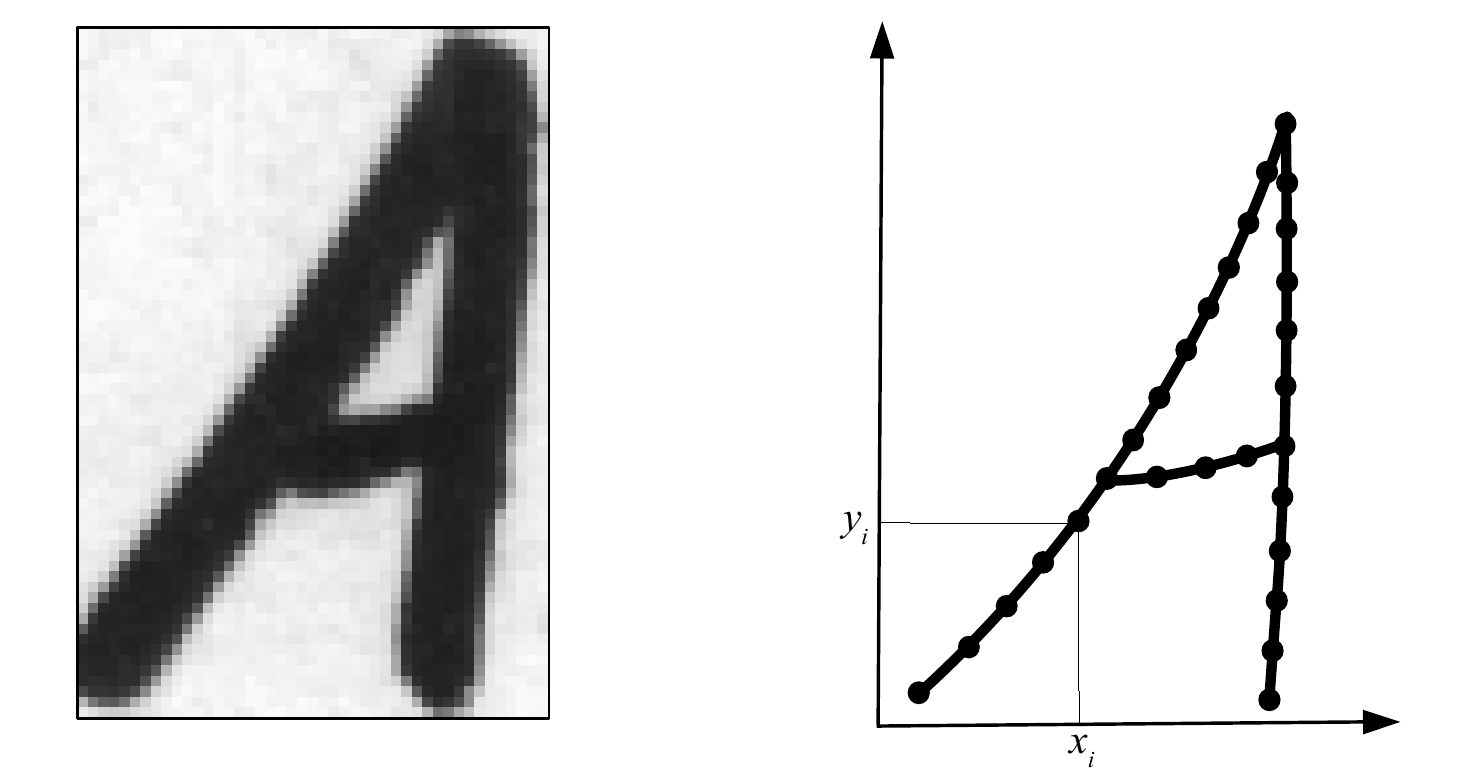} 
\caption{Example of offline (left) and online(right) handwriting data.}
\label{fig:offline_vs_online}
\end{figure}

There are two main different lines of research in the HTR problem, depending on the nature of the source data. On the one hand, the problem of recognizing handwritten text on a paper page scanned into a digital image, called \textit{offline handwriting recognition}\index{Offline handwriting recognition}. On the other hand, the so-called \textit{online handwriting recognition}\index{Online handwriting recognition}, which consists of recognizing the handwritten text from the continuous data of the $(x,y)$ positions of the pen obtained while it is written \cite{graves2008unconstrained}. These data are usually obtained by typing directly on a touchscreen or similar device. This Thesis focuses exclusively on the offline handwriting recognition problem. In Fig. \ref{fig:offline_vs_online} we include an example of offline versus online handwriting types of data.

The offline handwriting recognition problem keeps open several research lines that are far from being entirely solved. To recognize the handwritten text in a document is necessary that the algorithms identify which parts of the documents contain handwriting text, optionally segment lines, and words and recognize the words converting the images into sequences of characters. The image of the text can appear into tables or forms or overlap with other elements of the page such as: pictures, schemes, or graphs. The text can be written using different languages with different character sets (i.e., scripts), such as Chinese, Arabic, or Japanese.

The first step to recognize the text consist in identifying which parts of the image contains a text. The problem is trivial in some cases; for example, in a well-centered scanned image of a textbook page. However, in other situations is complex, for example, in a commercial street photo with many posters or in the scanned image of a scientific paper that can include text, images, tables, graphs, or equations. 

\begin{figure}[!ht]
\centering 
\includegraphics[width=16cm]{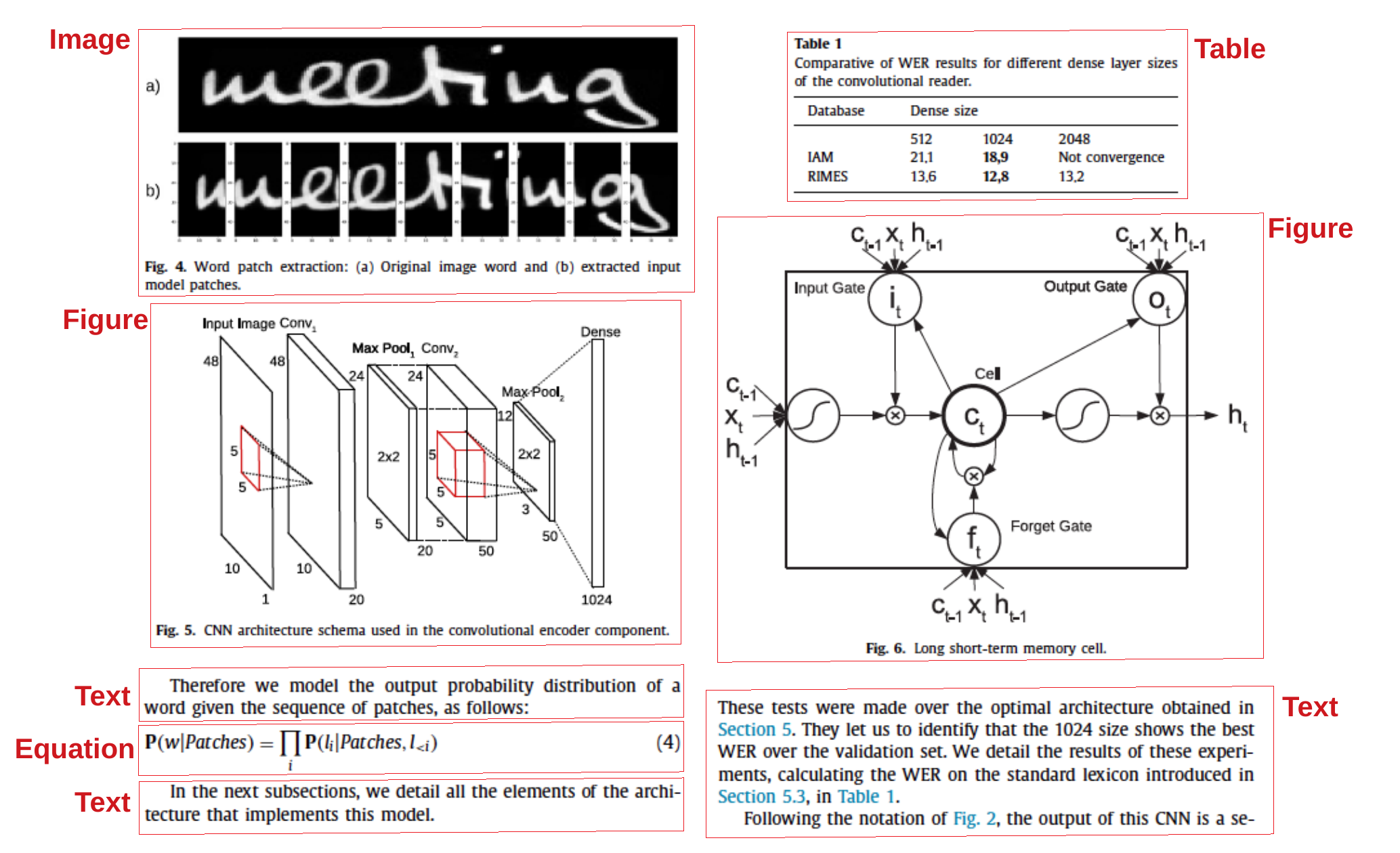} 
\caption{Example of layout analysis over a scientific paper.}
\label{fig:layoutAnalysis}
\end{figure}

The identification of the different types of elements (e.g., handwritten text, printed text, graphics, etc.) in scanned images of documents is named \index{Layout analysis}\textit{layout analysis} \cite{binmakhashen2019document} (see example in Fig. \ref{fig:layoutAnalysis}). This type of analysis is usual in the automation of document management processes, such as forms processing or automated invoice management. It is also necessary to extract certain elements from the image that needs to be processed separately, such as handwritten signatures.

\begin{figure}[!ht]
\centering 
\includegraphics[width=12cm]{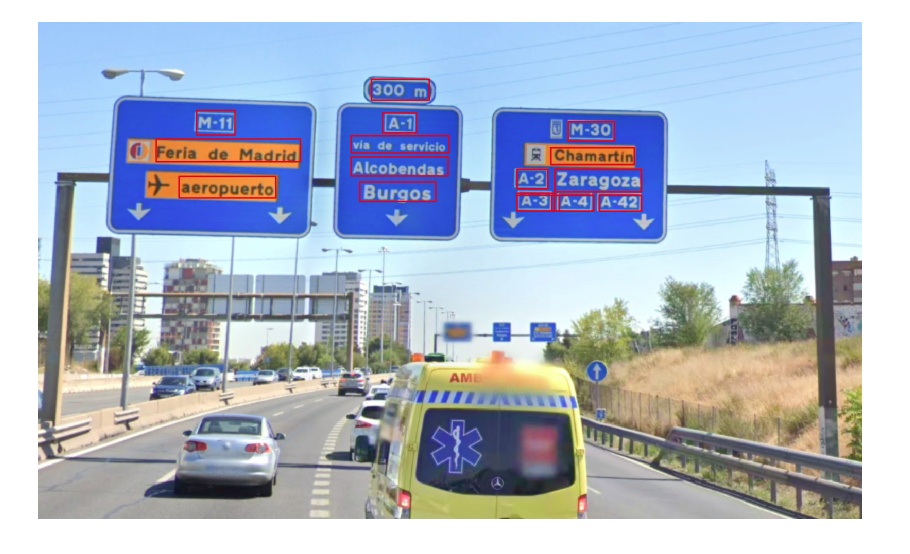} 
\caption{Example of scene text detection. Source: maps.google.com.}
\label{fig:sceneTextDetection}
\end{figure}

The problem of detecting and cropping the text parts in natural scene images is usually analyzed under the general object detection\index{Object detection} \cite{liu2020deep} or image segmentation\index{Image segmentation} \cite{ZAITOUN2015797} frameworks. An example of scene text detection is provided in Fig. \ref{fig:sceneTextDetection} This problem is particularly complex because the text in natural images is subject to extensive variability. For example, the use of a wide variety of typographies, the relative orientation and position of the text to the camera, the presence of multi-textured backgrounds or the frequent use of WordArt in posters and signs \cite{wang2011end}.

\begin{figure}[!ht]
\centering 
\includegraphics[width=14cm]{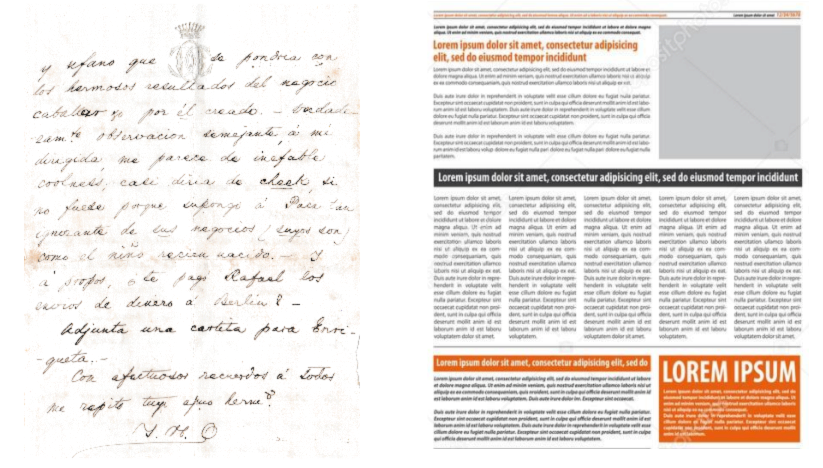} 
\caption{Examples of handwriting (left) and typographic (right) text.}
\label{fig:handwritting_vs_typogrphic}
\end{figure}

In all of the previous cases, the text detected could be of two types: typographic or handwritten. See differences in Fig. \ref{fig:handwritting_vs_typogrphic}. As it can be seen, in many real problems, the extraction of the image of a handwritten text on which to apply transcription algorithms is not trivial. Depending on the source, the image of the handwritten text may: have more or less noise, be distorted, include artifacts coming from the elements surrounding the text, or present a partially cropped text. All this adds difficulties to the subsequent recognition steps and opens multiple lines of research regarding this problem \cite{ye2014text}.

\begin{figure}[!ht]
\centering 
\includegraphics[width=14cm]{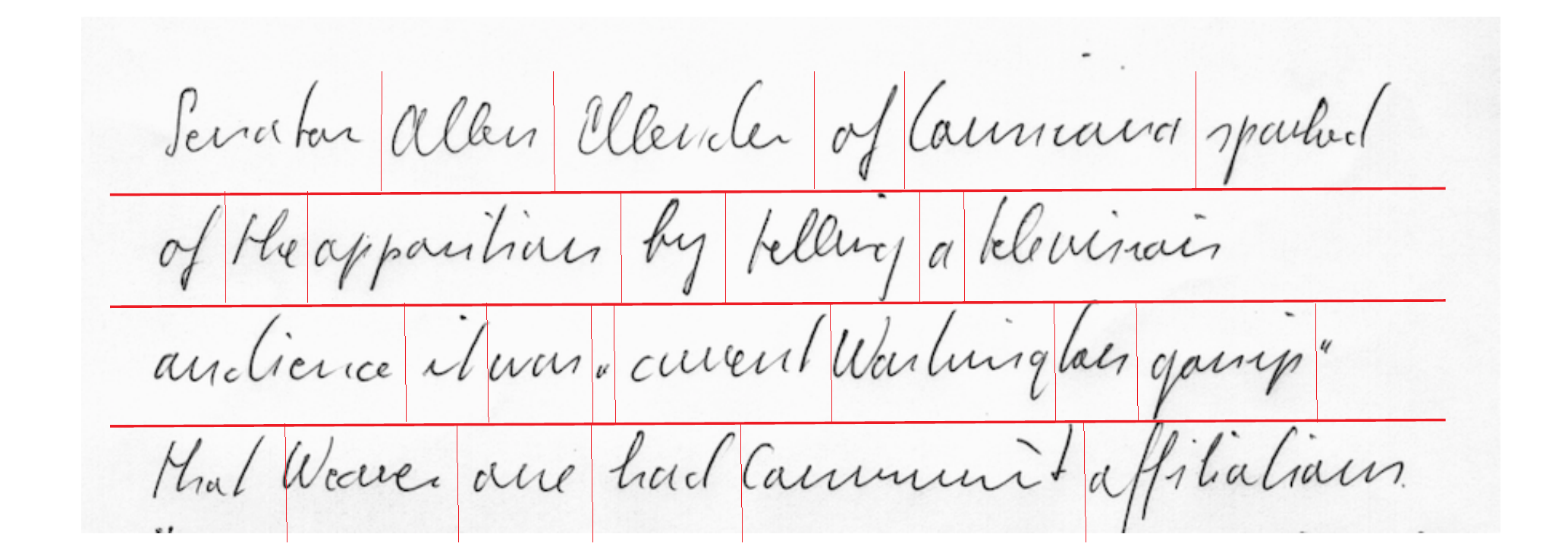} 
\caption{Example of line and word segmentation performed on an image of the IAM database.}
\label{fig:lineSegmentationExample}
\end{figure}

Once the text has been extracted and trimmed, its recognition usually includes several steps \cite{plamondon2000online}. In the most general case, the text image may contain a paragraph composed of multiple lines. In this case, it is most common to segment the text into lines \cite{likforman2007text}. However, the line segmentation is not trivial due to the slope of the lines, to the slant of characters, and to the fact that some characters in consecutive lines may overlap with others. Text lines can be applied directly to recognition algorithms, or they can be segmented into words, see Fig. \ref{fig:lineSegmentationExample}, and recognition can be performed at the word level. Most of the current HTR algorithms can be applied either on lines or on words, which are still a particular case of short lines. In this Thesis, all the experiments performed have been done at the word level.

The training of handwritten text identification models using machine learning techniques requires handwritten text image databases properly annotated with a transcription of the text present in each image. Several such public databases exist; some of them have images of isolated characters, and others contain transcriptions of words, lines, and paragraphs. In any case, the volume of annotated data available to train the models is limited and makes it necessary to consider using data augmentation strategies\index{Data augmentation} for model training.

In the development of HTR models, the text to be recognized is not a random sequence of characters, but belongs to a given language and is therefore subject to the syntactic and grammatical rules of that language. The step that converts the direct output of an HTR visual model into the sequence of characters that make up the message in a given language is called \textit{decoding}. It can be performed at different levels. A basic level can be used to make sure that the words in the text belong to a given dictionary or lexicon \index{Lexicon}; this is common in the word recognition problem. Advanced decoding models can also be built to learn the syntactic and grammatical rules of a language and are more common in line or paragraph level recognition models.

Finally, another element that must be defined in detail is the metrics chosen to determine the quality and accuracy of the models obtained, both at character and word recognition levels. The metrics must be selected to understand the model behavior and its recognition capacity, but the metrics used by other authors must also be considered to establish comparisons with the state-of-the-art techniques.

In the following sections, we analyze the considered elements for performing the different character and word recognition modeling experiments in detail. The Subsection \ref{c2-section:Image_processing} describes some image processing techniques applied to reduce the inherent variability existent in handwriting images. Data augmentation techniques are summarized in Subsection \ref{SubSect:Data augmentation strategies}. In the Subsection \ref{SubSect:Decoding_predictions}, we analyze the problem of decoding the predictions provided for the visual model to convert it into words. Finally, Subsection \ref{SubSect:Evaluation_metrics} details the common metrics used for evaluating the experimental results.

\section{Handwritten text image preprocessing}
\label{c2-section:Image_processing}
\index{Image preprocessing}

As explained in the previous section, handwritten text exhibits a very high variability, which not only depends on the writer. Additionally, when the text is digitized, the quality and age of the paper and the digitization process itself can add significant noise to the image. That is why it is usual to pre-process the images of handwritten text \cite{graves2008unconstrained} \cite{puigcerver2017multidimensional} before using them in models and recognition algorithms, which allows to eliminate noise and normalize the images to be analyzed as much as possible.

Some ablation \footnote{The ablation study aims to measure the performance of an AI model by removing components from it, in order to understand their contribution to the overall performance of the model. The concept was introduced by Newell et al. in \cite{newell1975tutorial} p. 43} studies of handwritten text recognition models, which include the analysis of the impact of image preprocessing on their results, support the importance of this step for the construction of a competitive recognition system (see for example the work of Dutta et al. in \cite{dutta2018improving}).
  \index{Calligraphy zones}

\begin{figure}[!ht]
\centering 
\includegraphics[width=14cm]{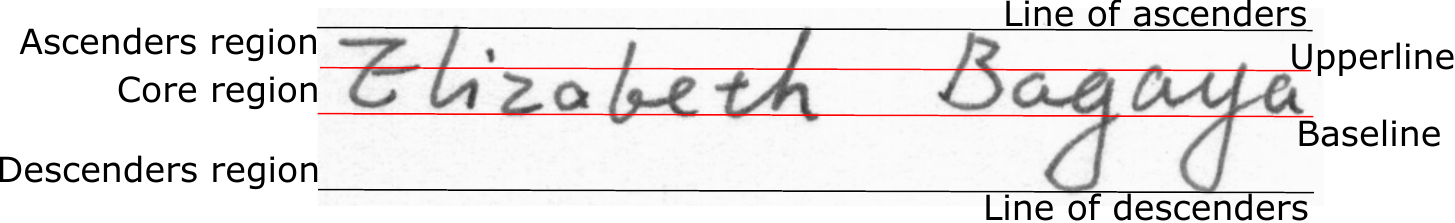} 
\caption{Calligraphy zones definition.}
\label{fig:calligraphy_zones}
\end{figure}

To describe in detail the different preprocessing techniques applied to handwritten text, it is important to define the elements that characterize a calligraphy. As shown in Fig. \ref{fig:calligraphy_zones} a handwritten line or word has three defined zones \cite{pesch2012analysis}, separated by two horizontal lines called baseline and upperline or header. The central area, called the core-region, is where the characters that do not have ascending or descending strokes are located, such as vowels or some consonants like: 'c', 'm', 'n', 'r', 's', 'v', and 'x'. The height of this zone is called $x$-height. The upper area, called the ascenders area, includes the upper strokes of characters such as: 'b', 'd', 'f', 'h', and 'l'. The lower area, called the descenders area, contains the lower strokes of characters such as: 'f', 'g', 'j', 'p', and 'q'.

\begin{figure}[!ht]
\centering 
\includegraphics[width=6cm]{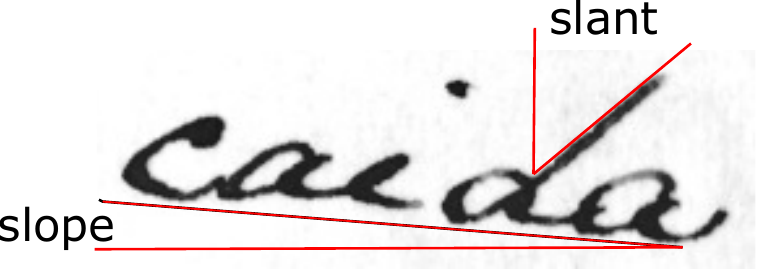} 
\caption{Slant and slope in a handwritten word image.}
\label{fig:slant_slope_definition}
\end{figure}

Although different authors \cite{graves2008unconstrained} \cite{bluche2013tandem} \cite{kozielski2013improvements} perform this task with different processes, there are, in most cases, a series of common steps. These are:
\begin{itemize}
  \item Remove the noise of the image, usually by contrast enhancement techniques.
  \item Correct of the slope and slant of the text. Fig. \ref{fig:slant_slope_definition} illustrates with an example the slant and slope definitions.
  \item Normalize the height of the ascenders and descenders areas.
  \item Resize and frame the text. For example, to a fixed height or by fitting the text to the edges of the image or by adding fixed size empty borders.
\end{itemize}

Next we describe each of the above four processings in detail in the context of the HTR problem.

\subsection{Remove the noise}
\index{Image denoising}

\begin{figure}[!ht]
\centering 
\includegraphics[width=14cm]{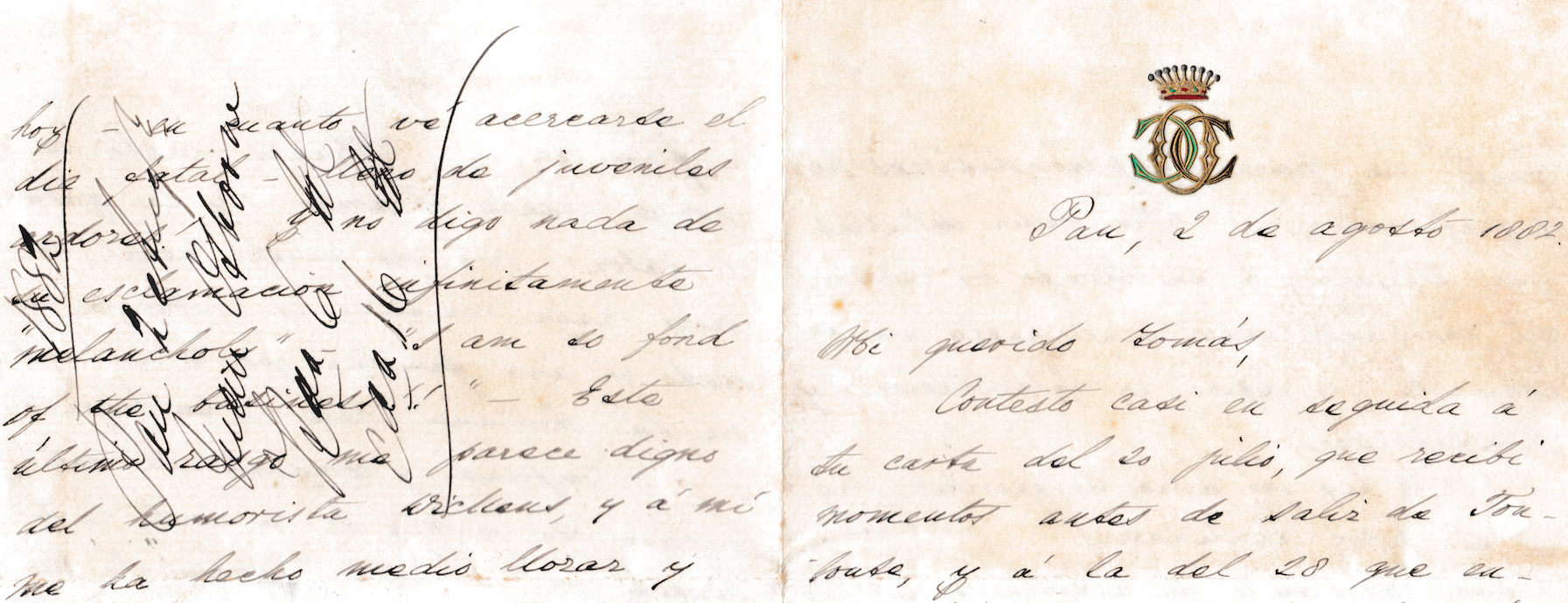} 
\caption{Example of image with noise take from Osborne database.}
\label{fig:noiseImage}
\end{figure}

The digitized image of a handwritten text is subject to multiple sources of noise that make it difficult to recognize, even for a human. The paper may contain marks, may not be completely white, or are degraded. If it is thin, it is also possible to see what is written on its back (i.e., bleed-through effect). The digitization process can introduce artifacts in the image caused, for example, by dirt on the scanner. The final result can be noised images like the examples shown in Fig. \ref{fig:noiseImage}.

Image denoising is the first step in image processing and standardization. The objective is to have a grayscale image in which the text has sharp, well-defined black strokes and the background is white with no elements that could affect the model.

\subsection{Slope correction}
\index{Slope correction}

\begin{figure}[!ht]
\centering 
\includegraphics[width=15cm]{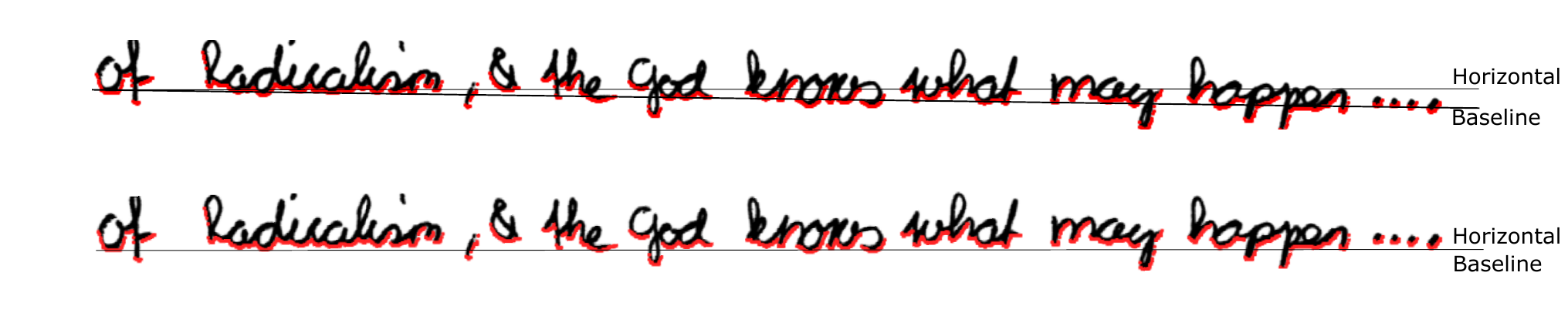} 
\caption{Example of slope angle estimation used in this Thesis.}
\label{fig:slope_example2}
\end{figure}

The slope or also called skew is the inclination of the lines of text with respect to a completely horizontal baseline. It usually appears when the text is written on a blank page without a pre-printed pattern of lines or boxes. The deviation of the line can have a positive angle if the writer tends to raise the line when he/she writes it, or it can also have a negative angle with respect to the horizontal if the line tends to descend. Fig. \ref{fig:slope_example2} shows an example line slope with a positive angle corresponding to the RIMES database \cite{augustin2006rimes}. Slope correction can be performed at the page level, as well as at the line level or even at the word level.

Likewise, the slope correction is sometimes carried out simultaneously with the slant correction since they are related, because a slope angle correction modifies the slant angle of the words by the same amount (see Fig. \ref{fig:slant_slope_definition}).

\subsection{Slant correction}
\label{c2_section:Slant correction}
\index{Slant correction}

The slant (see Fig. \ref{fig:slant_slope_definition}) is present in multiple handwritten texts, especially in the ascending and descending strokes. Usually, the text presents an inclination to the right with the strokes forming an angle of less than 90 degrees with the horizontal, but left-handed writers tend to write with a slant to the left, that is, with an angle greater than 90 degrees with respect to the horizontal.

The slant is a common feature of cursive-type calligraphy used in the learning process of writing. That is why it is very widespread and is one of the main sources of variability in the handwritten text that makes it difficult to recognize text using automatic systems.

The slant identification and correction is a critical aspect in HTR since many recognition algorithms that use handwritten text images as inputs usually have an approach that manages the image column by column. For example, those that use dense neural network-type models such as multilayer perceptron or recurrent neural networks that handle the image as a sequence of columns. Models using convolutional networks do not have this problem.
  \index{Multilayer perceptron}
  \index{Recurrent neural network}

In any case, the slant correction provides a variability reduction of the handwritten text that facilitates its recognition regardless of the type of model used. And it is, along with denoising, the most common preprocessing technique when tackling the problem of HTR.

Slant analysis is especially relevant when analyzing databases of historical handwritten documents since many calligraphies from past centuries have a very pronounced slant.

The slant correction is usually done in two steps \cite{puigcerver2017multidimensional}: firstly, identifying the inclination angle of the lines and then applying a transformation to correct that angle.

\subsection{Normalize ascenders and descenders regions}
\label{section:Normalize ascenders and descenders regions}
\index{Ascenders and descenders regions}

Fig. \ref{fig:calligraphy_zones} shows how the writing area is divided into three horizontal regions: the core region that contains most of the strokes, the ascending region that includes the strokes of the upper characters and the descending region that includes the lower strokes of certain characters. A very important source of variability is precisely the relative size of the ascending and descending regions, depending on the stroke width of each calligraphy.

To standardize the ascending and descending regions sizes it is necessary to identify and delimit the ascenders and descenders zones and scale them in some way.

\subsection{Crop and resize input images}

Handwritten text recognition models based on direct image modeling (i.e., segmentation free), which are practically all used in recent years, usually need that input images to be fixed size, specifically fixed height. For this reason, a series of transformations are applied to a previously standardized line or word images to fit them into this size. These transformations usually include: centering the text in the image, adding borders around the text and resizing the image, keeping the aspect ratio (for which the image is usually completed with columns of white pixels on the right).

The centering of the text in the image is done by removing the empty borders that do not contain any image strokes. Sometimes, a frame of blank rows and columns of fixed size is added around the text image to be centered. The image is then rescaled to a fixed height.

Usually, the selection of the rescaling height, as in the previous reference, reduces the size of the original image by lowering its resolution. This is done because sometimes the scanning resolution of the text image is high, and it is possible to reduce it without affecting the human ability to read the text. It is also done because, in the case of using deep models, the size of the input image to the model has a significant impact on the memory requirements and training times of the model, especially if it is a deep model with multiple layers.

If the model used supports input images of variable width, no additional steps are necessary. If, on the other hand, the model only supports inputs of a fixed size, the last step is to set the width of the image to a fixed value without changing the aspect ratio of it. To perform this stage, maximum image width is selected and all images are adjusted to that width by completing each one of them with empty columns on the right.

\section{Data augmentation strategies}
\label{SubSect:Data augmentation strategies}
\index{Data augmentation}

Data augmentation is a technique in which a training set is expanded with class-preserving transformations.\cite{dao2019kernel}. In the case of handwritten text recognition, it consists of slightly modifying the image of the text to be recognized so that the text remains legible and has the same transcription as the original. For this purpose, some image transformations such as translations, slight rotations and, in general, other affine transformations that meet the above requirement are used. These transformations are usually applied in combination (e.g., translate 5 pixels with a positive rotation of 3º of the image text). 
\index{Affine transformation}

There are two data augmentation strategies: data augmentation during training and data augmentation in the inference or evaluation process called test-time data augmentation. Test-time data augmentation is a practice that allows the improvement of the accuracy metrics of models by evaluating multiple versions of test data. In general, it can be applied in a standard way on any predictive model.

\section{Decoding predictions}
\label{SubSect:Decoding_predictions}

This Thesis aims to develop models for the recognition of general-purpose handwritten text. This implies that such models must be able to recognize any word that appears written in an image. For this reason, it is not possible to employ strategies in which the system's output is a closed set of words from a vocabulary. Therefore, the systems proposed here are capable of generating any sequence of characters as output.

However, not all character sequences are valid from a grammatical point of view and not all word sequences are valid from a semantic point of view. Therefore, the character sequence resulting from applying a visual recognition model to the text image is processed to be considered a valid word. This is a common step in text recognition systems as described in Subsection \ref{Sect:Overview_of_the_problem} 

This decoding can be performed at several levels. The simplest one is at the vocabulary level, for which a lexicon is used as a closed set of words that define the vocabulary where it must be the resulting character sequence of the visual model. If the visual model result is in the lexicon, it is determined that the word is identified in the image. If, on the contrary, it is not, the nearest lexicon word is assigned using a word proximity metric. The most common metric in this case is the Levenshtein distance \cite{levenshtein1966binary} which is described in detail in the Subsection \ref{SubSect:Evaluation_metrics}. 
  \index{Lexicon}
  \index{Levenshtein distance}

There are some contexts with a limited vocabulary where this lexicon-based approach is the most appropriate for validating the visual model output. For example, in the case of recognizing form fields with a closed list of valid options such as addresses and postal codes \cite{srihari1993recognition}, in recognition of telephone numbers \cite{kermorvant2009isolated} or in the processing of bank cheques \cite{dimauro1997automatic}. In some more general application areas, the main drawback of this method is that it does not consider the word context to refine its recognition. On the other hand, this is an easy-to-reproduce and straightforward method.

When applying this decoding method, the primary consideration is to define which words will be part of the lexicon. In the case of closed output recognition systems, words that are not included in the lexicon, called Out-of-Vocabulary words (OOV), can never be correctly identified. On the other hand, if an extensive lexicon is used when decoding the output character sequence of the visual model, the error rates increases.
  \index{Out of vocabulary}

The direct search for the nearest word in the vocabulary has a complexity that grows linearly with the vocabulary size. More efficient strategies have been proposed because many words share characters in the same positions. A review of different techniques of this type can be consulted in \cite{koerich2003large}. Currently, there are very efficient search implementations using Levenshtein distance; for example, the one provided by the open-source software Apache Lucene \cite{bialecki2012apache}. 
  \index{Software!Apache Lucene}

In addition to the fact that the word to be decoded belongs to a vocabulary, the context of that word (i.e., the words that are before and after it in the sentence) imposes grammatical and semantic restrictions on the word to be identified, and it can be used to make a more accurate identification of it. This fact introduces a second level at which decoding can be carried out. The most common way of introducing this context into the decoding process, in the area of handwriting recognition, is the statistical \textit{n-gram language models} \cite{bertolami2008hidden}, \cite{kozielski2013improvements}. 
  \index{n-gram language models}

An $n$-gram language model is a probabilistic model that predicts the next word in a sequence as a function of the previous $n$-1 words: \(P(x_{i}\mid x_{{i-(n-1)}},\dots ,x_{{i-1}})\), regardless of the position of the word in the sequence. The calculation of the maximum plausible estimator is made in frequency terms as indicated in the equation \ref{eq:n_gram}, by taking into account that the probability of a word conditioned to the previous word sequence follows a categorical distribution also known as \textit{multinomial distribution}.
  \index{Multinomial distribution}

\begin{equation}
p(x_i|x_{i-1}, . . . ,x_{i-n+1}) = \frac{Count(x_i,x_{i-1}, . . . ,x_{x-i+1})}{\sum_x Count(x,x_{i-1}, . . . ,x_{i-n+1})}
\label{eq:n_gram}
\end{equation}

These models are generally constructed from large corpora that allow sufficient repetition of the most common word sequences in a language. However, as $n$ increases, the probability of an $n$-gram not appearing in the corpus also grows. It results in many sequences having an estimated probability of zero. To solve this problem, the most common way is to add a small smoothing value $\delta$ to all frequency counts as indicated in the equation \ref{eq:n_gram_smoothing}.

\begin{equation}
p(x_i|x_{i-1}, . . . ,x_{i-n+1}) = \frac{Count(x_i,x_{i-1}, . . . ,x_{x-i+1})+\delta}{\sum_x (Count(x,x_{i-1}, . . . ,x_{i-n+1})+\delta)}
\label{eq:n_gram_smoothing}
\end{equation}

A comparative review of different ways of applying smoothing to $n$-gram based language models can be found in \cite{chen1999empirical}.

\section{Evaluation metrics}
\label{SubSect:Evaluation_metrics}

In order to evaluate the quality of the models built, metrics must be defined to assess the usefulness and applicability of the models and their limitations. Likewise, the usual metrics in the literature should be provided so the results obtained can be compared precisely with those from other authors.

In the case of character recognition models, the most common metric used is \textit{accuracy} (i.e., the percentage of cases correctly identified). However, for word or line recognition, this metric cannot distinguish between an error produced by a single character from another error in which all the characters of the word are incorrect. To solve this inconvenience, the \textit{Levenshtein edit distance} metric \cite{levenshtein1966binary} is used. This metric counts the minimum number of operations required to transform a character sequence into another one by considering substitutions, deletions, and insertions of characters.
  \index{Levenshtein distance}
  \index{Metrics!Accuracy}

\subsection{Metrics to evaluate character recognition models}
\label{section:Metrics to evaluate character recognition models}

Although the metric almost exclusively used in the evaluation of this type of multi-category classifier is accuracy or its complement, the \textit{error rate} (percentage of cases incorrectly identified), these metrics have their limitations and drawbacks. Being $N$ the total number of cases and $c_{ij}$ the number of cases of true category $i$ assigned to the predicted category $j$, the accuracy is defined according to the equation \ref{eq:accuracy}.
  \index{Metrics!Error rate}

\begin{equation}
accuracy = \frac{1}{N}\sum_{i=j}^{} c_{ij}
\label{eq:accuracy}
\end{equation}

\textit{Accuracy} is a metric that should always be provided since it is the one that allows comparing the results with other authors. However, accuracy alone does not provide details about the limitations and types of errors that each model has, which provide clues for its optimization and improvement. Additionally, accuracy is influenced by the number of different target values and the distribution of these values' frequencies. For example, the accuracy of a random model based on the target distribution for a binary target with an unbalanced distribution of 9 to 1 for the most frequent category would be 90\%, and for a target of 10 balanced categories, it would be 10\%. 

In unbalanced samples, the less frequent categories may obtain a low allocation rate by the model even if high overall accuracy is obtained. To monitor this, \textit{recall} metric is used to measure how many of the actual cases in each category are correctly predicted by the model. This metric is calculated individually for each of the target categories. To follow the usual notation used in the case of binary targets, given a reference category $i$, we will adopt the following notation:
\begin{itemize}
  \item True Positive (TP) represent the cases of the category $i$ correctly classified
  \item False Positive (FP) represent cases from other categories assigned to category $i$
  \item False Negative (FN) represent cases in category $i$ assigned to other categories
  \item True Negative (TN) represent cases from other categories assigned to other than the $i$
\end{itemize}

Following this, the \textit{recall} metric is defined according to the equation \ref{eq:recall}
  \index{Metrics!Recall}

\begin{equation}
recall = \frac{TP}{TP + FN} 
\label{eq:recall}
\end{equation}

In the case of unbalanced samples also the most frequent category or categories are usually assigned cases of other minority categories that could be similar. To analyze this effect, the category \textit{precision} metric is calculated. It estimates the percentage of cases correctly assigned by the model to that category (see the equation \ref{eq:precision});

\begin{equation}
precision = \frac{TP}{TP + FP} 
\label{eq:precision}
\end{equation}

Accuracy and recall metrics provide good information about the model behavior in each category. To obtain a single unified measure of the model goodness in each category, the metric called \textit{F1 score}\cite{powers2020evaluation} is usually used, which is mathematically defined as the harmonic mean of precision and recall:
  \index{Metrics!F1 score}

\begin{equation}
F1 score = 2 * \frac{precision * recall}{precision + recall}
\label{eq:F1_score}
\end{equation}

The metrics above quantify the volume of model error but do not provide information on what specific allocation errors the model produces (i.e., what categories the model is confusing). The \textit{confusion matrix} is used to evaluate the marginal distributions of the error cases in each category. This square matrix, of size equal to the number of categories in the target, is constructed by including in each cell $(i, j)$ the number of cases in category $i$ that the model has assigned to category $j$, $c_{ij}$. Therefore, the main diagonal of this matrix includes the number of cases correctly assigned to each target category, and the cells outside this diagonal include all the cases erroneously assigned to different categories.
\index{Metrics!Confusion matrix}

With the different metrics previously indicated, it is possible to analyze the results of a classification model in detail. However, it must be remembered that all of them are influenced by the target distribution and that the assignment is made to the category for which the model provides a higher probability conditioned on the input variables. If one wants to obtain a metric that measures the quality of the predictions in relative terms and not from their absolute values, independent of the target distribution and the classification thresholds of each category, one must consider the \textit{Area Under the ROC curve} (AUC).
\nomenclature{AUC}{Area under the ROC curve}
\index{Metrics!AUC}

The \textit{Receiver Operations Characteristics} (ROC) curve \cite{bradley1997use}, is a line graph showing the performance of a classification model for each of the possible classification thresholds of a given category. The graph is calculated for each category of the target and the curve represents the relationship between two parameters:
\begin{itemize}
  \item The so-called \textit{true positive rate} (TPR), calculated as the number of cases in the category correctly classified (TP) divided by the total number of cases in the category, which includes those correctly classified plus those incorrectly classified in other categories (TP + FN). This is the recall of the category.
  \item The so-called \textit{false positive rate} (FPR), calculated as the number of cases in the other categories misclassified in the category (FP) divided by the total number of cases in the other categories, which includes those misclassified in the category plus those classified in the other categories (FP + TN)
\end{itemize}
\nomenclature{ROC}{Receive Operations Characteristics}
\index{ROC curve}
\index{Metrics!True positive rate}
\index{Metrics!False positive rate}
\nomenclature{TPR}{True Positive Rate}
\nomenclature{FPR}{False Positive Rate}

\begin{equation}
TPR = \frac{TP}{TP + FN} 
\label{eq:TPR}
\end{equation}

\begin{equation}
FPR = \frac{FP}{FP + TN} 
\label{eq:FPR}
\end{equation}

\begin{figure}[!ht]
\centering 
\includegraphics[width=8cm]{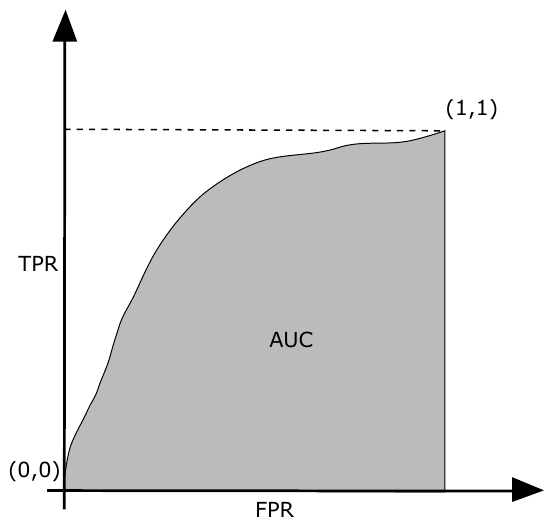} 
\caption{Example of ROC curve and AUC area.}
\label{fig:ROC_curve}
\end{figure}

The ROC curve represents TPR versus FPR for each of the different possible classification thresholds for the analyzed category. It is starting on the threshold zero, which does not assign any case to the category and therefore provides values of 0 to FPR and TPR, and ends in the threshold one that assigns all cases to the category and provides a value of 1 to both TPR and FPR. An example of such a curve is shown in Fig. \ref{fig:ROC_curve}. 

The ROC curve provides detailed information on the classification behavior in a category as a function of the classification thresholds. This information is usually summarized in the area metric under the ROC curve (AUC) \cite{bradley1997use} which measures all the two-dimensional area under the curve between the points (0.0) and (1.1) as shown in Fig. \ref{fig:ROC_curve} 

A perfect model, without errors, would provide a TPR of 1 and an FPR of 0 and therefore an AUC=1. A purely randomized model that assigns cases to the category at random following the proportion of that category in the target, assigns values of TPR = FPR for each possible classification threshold, producing a ROC curve equal to the bisector of the first quadrant and an AUC of 0.5 that is used as baseline reference of a randomized model.

\subsection{Metrics to evaluate word recognition models}
\label{section:Metrics to evaluate word recognition models}

The two main metrics commonly used to evaluate word and line-level handwriting recognition models are the \textit{Character Error Rate (CER)} and the \textit{Word Error Rate (WER)} \cite{frinken2014continuous}. The CER measures the Levenshtein distance \cite{levenshtein1966binary} between the predicted and the real character sequence of the word. The Levenshtein distance, also sometimes called \textit{edit distance}, is a metric to measure the difference between two sequences. Informally, the Levenshtein distance between two words (the model's prediction and the real one) is the minimum number of insertions, deletions, or substitutions needed to transform the prediction into the real word divided by the length of the real word, as indicated in the equation \ref{eq:edit_distance}. In this Thesis, the CER results provided have been calculated with the Python library {\fontfamily{cmtt}\selectfont editdistance} that implements the algorithm efficiently by means of dynamic programming as indicated in \cite{hyyro2001explaining}.
\nomenclature{CER}{Character Error Rate}
\nomenclature{WER}{Word Error Rate}
\index{Levenshtein distance}
\index{Metrics!Character error rate}
\index{Metrics!Word error rate}

\begin{equation}
CER(prediction, real) = \frac{substitutions + insertions + deletes}{len(real)}
\label{eq:edit_distance}
\end{equation}

The \textit{word error rate} is defined in a similar way to CER by calculating the minimum number of insertions, substitutions and deletions of words needed to go from a text string predicted by the model to the real text string. In the case that the recognition is done at the level of individual words, the WER represents the percentage of words correctly identified by the model and therefore matches the accuracy of the model.

The two previous metrics can be directly calculated from the estimation of the visual model or on the result of applying a subsequent decoding process by searching lexicons or language models. This Thesis provides evaluation results of the proposed models using these two previous metrics, both applied to the direct estimation of the visual model and then using a decoding process by searching in standard lexicons for reference datasets.

It is important to highlight that the above metrics are strongly influenced by the set of characters you decide to select to encode the texts of the reference corpora. That is, the different corpora used as benchmarks in handwriting provide an exact transcription of the text images included in them, differentiating capitalization, lower case and punctuation. However, some authors, when training and evaluating their models, eliminate punctuation marks or do not distinguish between upper and lower case letters. Without discussing which approach may be the most appropriate, it is clear that these are different problems not comparable to each other in terms of the above metrics. This is one of the first sources of problems when it comes to comparing results from different lines of research in the area of handwriting, since simplifying the coding masks all errors in recognition of punctuation marks or apostrophes and all the confusion that may exist between upper and lower case forms of the same letter.

In order to determine that there are statistically significant differences between different methods, the comparison must consider not only the estimated value of the error metric obtained from an experiment, but also a measure of the estimation error of that metric. Confidence intervals are usually used to determine whether differences between two values of a metric can be considered statistically significant or not. The most common way to construct these confidence intervals is to assume that the metric under consideration follows a known statistical distribution. For a rate-of-error type metric, the usual assumption is that it follows a binomial or normal distribution. An alternative, if this assumption about the distribution of error cannot be guaranteed, is to use a non-parametric estimation method such as non-parametric boostrapping \cite{diciccio1996bootstrap}. In the case of this Thesis, the confidence intervals shown have been calculated using this non-parametric technique.

In the case of handwritten text recognition, few authors provide confidence intervals for the error measures of the proposed models (for example, \cite{puigcerver2017multidimensional} or \cite{bluche2015deep}).

Finally, one must also consider that handwritten text recognition models are becoming more computationally complex since they are built with increasingly complex deep learning architectures. In addition to the cost of the visual recognition model, we must add the computational cost of the decoding process, which includes context information in the results. If this is done by searching in lexicons, the cost increases significantly with the size of the lexicon as indicated in the Subsection \ref{SubSect:Decoding_predictions}, and if a language model does it, these models, especially those based on neural networks, also have a very high computational cost. Reporting these computational time and cost metrics is also important to evaluate the feasibility of the practical application of the proposed algorithms when using deep learning architectures. It is also relevant to provide these efficiency metrics if CPU-based or GPU-based hardware is used because the computation time and cost of deep architectures in these two types of hardware platforms are very different. Some authors like \cite{liu2015study} provide these metrics, although this is not common in most publications.


\chapter{Handwriting Databases}
\label{chapter_Handwriting_Databases}

In this chapter, we describe the databases used for the experiments carried out in this Thesis. First, the handwriting character databases in Section \ref{section:Handwriting characters databases} and, then, the general databases at word and line level in the Section \ref{section:General handwriting databases}. Finally, Section \ref{section:Databases summary} present a summary of the analyzed databases.

\section{Handwriting characters databases}
\label{section:Handwriting characters databases}

In this section we provide a brief description of the research handwriting isolated character databases used in this Thesis. For a complete list of databases available and a more detailed description, Hussain et al. \cite{hussain2015comprehensive} provides a comprehensive survey about handwritten databases.

The following four databases have been selected for the isolated character recognition experiments carried out in this Thesis:
\begin{itemize}
    \item The NIST database is the largest database of isolated characters but the images are binary.
    \item The MNIST digits database is the most widely used database but only contains digits. 
    \item The TICH database provides grayscale images of letters and digits.
    \item The new COUT database includes special characters and punctuation marks. 
\end{itemize}

In the following sections, we describe these databases, as well as example images and the distributions of the available data by partition and available characters.

\subsection{NIST database}
\index{Databases!NIST}

We provide some details about the second edition of the NIST Special Database 19, Hand printed Forms and Characters \cite{grother2016nist}. This database provides isolated characters segmented from 3,669 handwriting sample forms written each one by a different person.

\begin{figure}[!ht]
\centering 
\includegraphics[width=8cm]{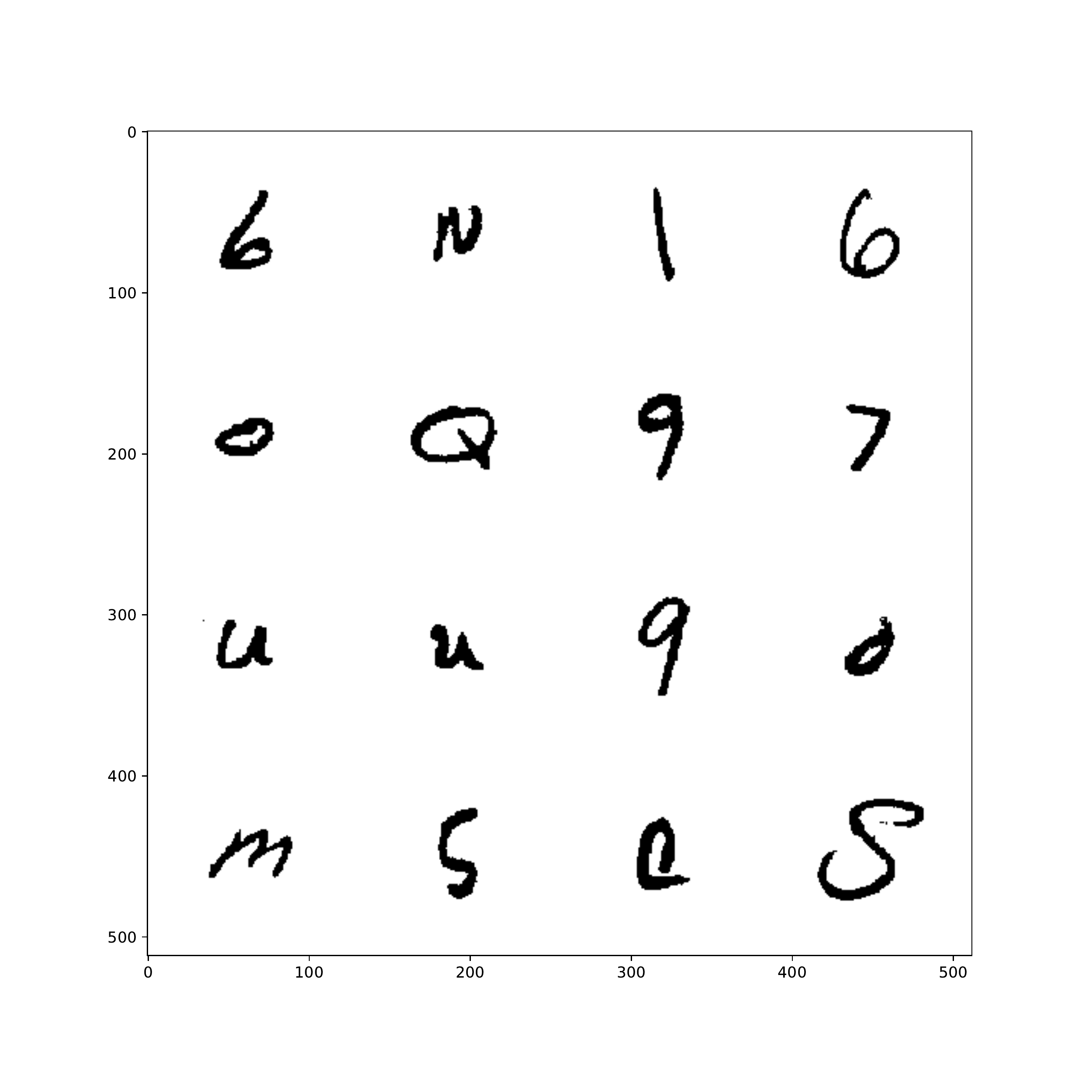} 
\caption{Sample images of the NIST database.}
\label{fig:databasesNISTSample}
\end{figure}

The isolated character database includes a total of 814,225 images of letters and digits characters corresponding to '0'- '9', 'A'- 'Z' and 'a'- 'z'. Al the images are saved in PNG binary format with a resolution of 128$\times$128 pixels per image. The characters are well centered on the image with large margins. We show some examples of the character images in Fig. \ref{fig:databasesNISTSample}.
  \index{PNG}

The database has eight partitions: the partitions {0, 1, 2, 3, 6, 7, and 8} were obtained from Census Bureau employees in Maryland and partition 4, proposed as test partition, was completed by the Bethesda high school students.

The distribution of the available characters in the train partition is biased with high digit frequency and low word frequency. This is because the template form used to get the data includes 28 boxes for numbers but only 2 for alphabet characters and 1 for a 53 words paragraph. For digits, the distribution is similar in the train and test partitions. In Table \ref{table:NISTFreqTableDigits} we provide the frequencies of each digit in the partitions.

\begin{table}[!ht]\footnotesize
\centering
\begin{tabular}{lrrrr}
 \toprule
 Char &  \#Train &  \%Train &  \#Test &  \%Test \\
 \midrule
 0 & 34,803 & 10.11 & 5,560 &  9.48 \\
 1 & 38,049 & 11.05 & 6,655 & 11.35 \\
 2 & 34,184 &  9.93 & 5,888 & 10.04 \\
 3 & 35,293 & 10.25 & 5,819 &  9.92 \\
 4 & 33,432 &  9.71 & 5,722 &  9.76 \\
 5 & 31,067 &  9.02 & 5,539 &  9.44 \\
 6 & 34,079 &  9.90 & 5,858 &  9.99 \\
 7 & 35,796 & 10.40 & 6,097 & 10.40 \\
 8 & 33,884 &  9.84 & 5,695 &  9.71 \\
 9 & 33,720 &  9.79 & 5,813 &  9.91 \\ 
 \bottomrule
\end{tabular}
\caption{Digits frequency in the NIST train and test partitions.}
\label{table:NISTFreqTableDigits}
\end{table}

For letters, the distribution is clearly different in the test partition, where the available examples are evenly distributed. We provide the frequencies of each letter in the train and test partitions of the database in Table \ref{table:NISTFreqTable}.

\begin{table}[!ht]\footnotesize
\centering
\begin{tabular}{lrrrr|lrrrr}
\toprule
Char &  \#Train &  \%Train &  \#Test &  \%Test & Char & \#Train &  \%Train &  \#Test &  \%Test \\
\midrule
 A & 7,010  & 1.81 & 459 & 1.92 &  a & 11,196 & 2.89 & 481 & 2.01 \\
 B & 4,091  & 1.06 & 435 & 1.82 &  b & 5,551  & 1.43 & 461 & 1.93 \\
 C & 11,315 & 2.92 & 518 & 2.16 &  c & 2,792  & 0.72 & 494 & 2.06 \\
 D & 4,945  & 1.28 & 396 & 1.65 &  d & 11,421 & 2.95 & 439 & 1.83 \\
 E & 5,420  & 1.40 & 365 & 1.52 &  e & 28,299 & 7.31 & 424 & 1.77 \\
 F & 10,203 & 2.63 & 419 & 1.75 &  f & 2,493  & 0.64 & 468 & 1.95 \\
 G & 2,575  & 0.66 & 389 & 1.62 &  g & 3,839  & 0.99 & 437 & 1.83 \\
 H & 3,271  & 0.84 & 402 & 1.68 &  h & 9,713  & 2.51 & 504 & 2.11 \\
 I & 13,179 & 3.40 & 815 & 3.40 &  i & 2,788  & 0.72 & 364 & 1.52 \\
 J & 3,962  & 1.02 & 426 & 1.78 &  j & 1,920  & 0.50 & 293 & 1.22 \\
 K & 2,473  & 0.64 & 377 & 1.57 &  k & 2,562  & 0.66 & 395 & 1.65 \\
 L & 5,390  & 1.39 & 496 & 2.07 &  l & 16,937 & 4.37 & 916 & 3.83 \\
 M & 10,027 & 2.59 & 460 & 1.92 &  m & 2,634  & 0.68 & 475 & 1.98 \\
 N & 9,149  & 2.36 & 439 & 1.83 &  n & 12,856 & 3.32 & 460 & 1.92 \\
 O & 28,680 & 7.40 & 459 & 1.92 &  o & 2,761  & 0.71 & 454 & 1.90 \\
 P & 9,277  & 2.39 & 467 & 1.95 &  p & 2,401  & 0.62 & 415 & 1.73 \\
 Q & 2,566  & 0.66 & 452 & 1.89 &  q & 3,115  & 0.80 & 384 & 1.60 \\
 R & 5,436  & 1.40 & 446 & 1.86 &  r & 15,934 & 4.11 & 491 & 2.05 \\
 S & 23,827 & 6.15 & 445 & 1.86 &  s & 2,698  & 0.70 & 438 & 1.83 \\
 T & 10,927 & 2.82 & 469 & 1.96 &  t & 20,793 & 5.37 & 434 & 1.81 \\
 U & 14,146 & 3.65 & 458 & 1.91 &  u & 2,837  & 0.73 & 475 & 1.98 \\
 V & 4,951  & 1.28 & 482 & 2.01 &  v & 2,854  & 0.74 & 524 & 2.19 \\
 W & 5,026  & 1.30 & 475 & 1.98 &  w & 2,699  & 0.70 & 465 & 1.94 \\
 X & 2,731  & 0.71 & 472 & 1.97 &  x & 2,820  & 0.73 & 472 & 1.97 \\
 Y & 5,088  & 1.31 & 453 & 1.89 &  y & 2,359  & 0.61 & 387 & 1.62 \\
 Z & 2,698  & 0.70 & 467 & 1.95 &  z & 2,726  & 0.70 & 450 & 1.88 \\
 \bottomrule
\end{tabular}
\caption{Letters frequency in the NIST train and test partitions.}
\label{table:NISTFreqTable}
\end{table}

\subsection{MNIST database}
\label{section:MNIST database}
\index{Databases!MNIST}

The MNIST\footnote{\url{http://yann.lecun.com/exdb/mnist/}} database \cite{lecun1998gradient} is a subset of the previously discussed NIST database. It includes only images of digits characters, where the digits have been size-normalized and centered in a fixed-size image. This dataset is extensively used in the literature to compare deep learning models.

The MNIST character database contains a balanced sample of 70,000 digits characters from '0' to '9' selected from the previous NIST database. The database is available in several ways. It is included in different Python packages as {\fontfamily{cmtt}\selectfont scikit-learn}, {\fontfamily{cmtt}\selectfont tensorflow} or {\fontfamily{cmtt}\selectfont pytorch}. Each digit image has a resolution of 28$\times$28 pixels in grayscale format of 128 bits.
\index{Software!Tensorflow}
\index{Software!Scikit-learn}

To build it, the original binary images from the NIST database were cropped and resized to 20$\times$20 pixels while preserving the aspect ratio. The grey levels appear as a result of the anti-aliasing technique used. Finally, the images were centered over a 28$\times$28 image, including margins of 4 pixels.

The database is distributed with a standard train test partition of 60,000 images to train and 10,000 to test. In each partition, the distribution of characters is balanced, including 6,000 samples of each digit in the training partition and 1,000 samples of each digit in the test partition. This partition is different from the NIST original one, but it is guaranteed that the partitions contain different writers who produced the digits.

\begin{figure}[!ht]
\centering 
\includegraphics[width=10cm]{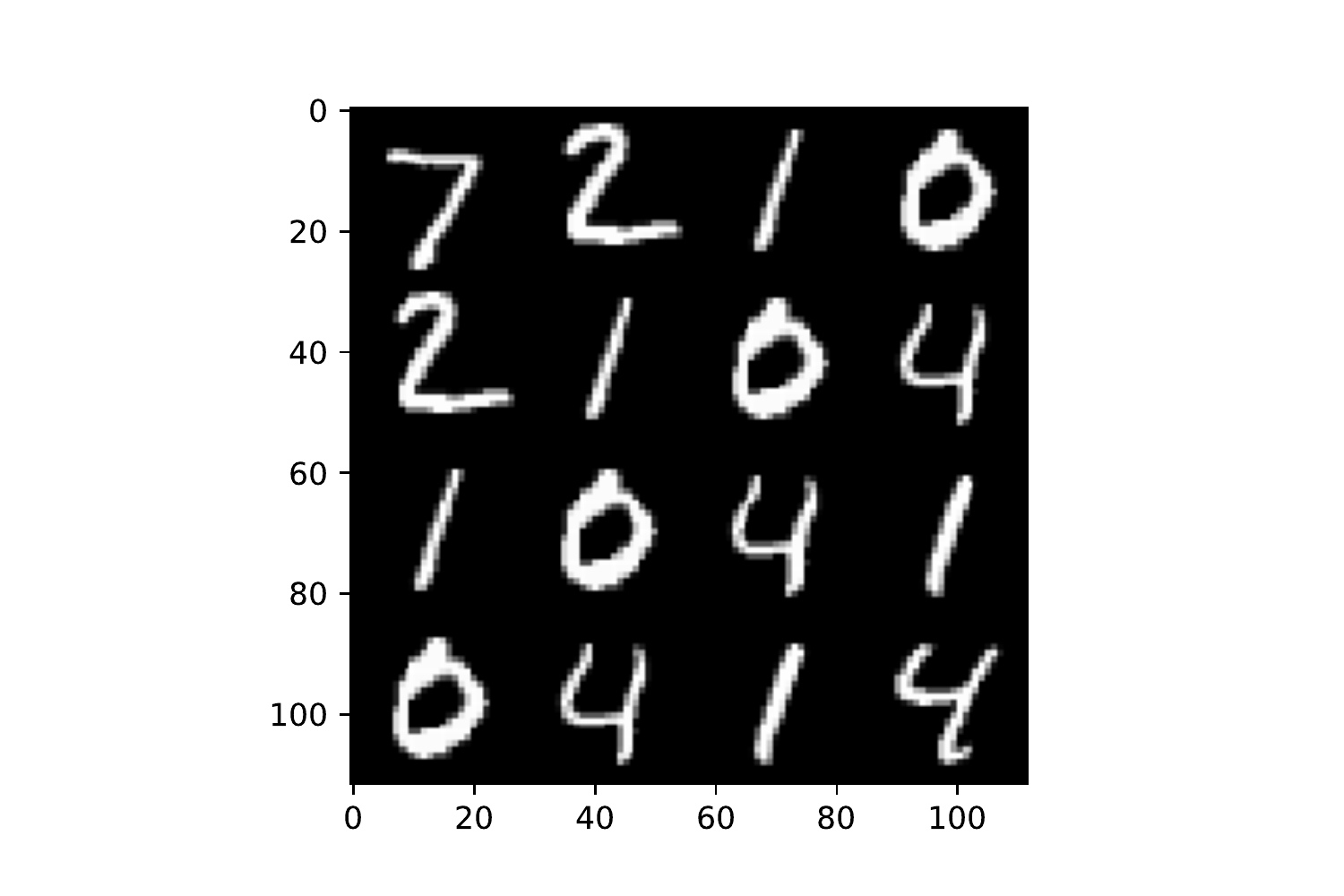} 
\caption{Sample images of the MNIST database.}
\label{fig:databasesMNISTSample}
\end{figure}

\subsection{TICH database}
\label{section:TICH database}
\index{Databases!TICH}

The Tilburg Characters data set (TICH) contains labeled characters, cut out from the Firemaker collection \cite{schomaker2000forensic} using a semi-automatic approach and curated them by a human labeler. The result includes a total of 40,141 images of isolated characters. The database includes only uppercase characters and digits and is described extensively by van der Maaten in \cite{van2009new}.

The database is available to download in \url{https://lvdmaaten.github.io/publications/misc/characters.zip}\footnote{last access May 2021} in Matlab format.

\begin{figure}[!ht]
\centering 
\includegraphics[width=10cm]{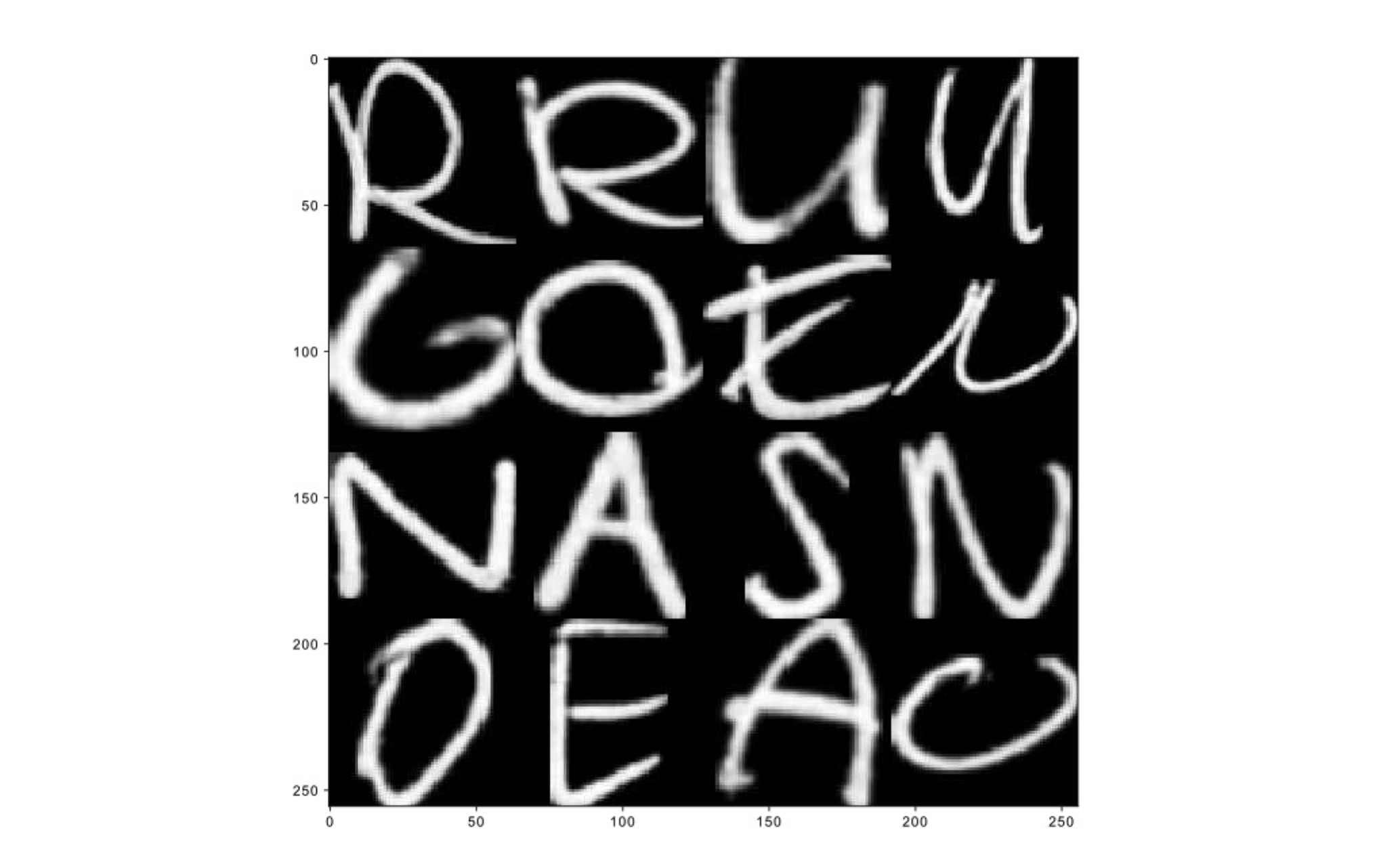} 
\caption{Sample images of the TICH database.}
\label{fig:databasesTICHSample}
\end{figure}

The distributed database includes the set of original images cropped from the Firemaker collection scanned in grayscale at 300 dpi with 8 bits per pixel. In addition, it provides two versions of normalized images. In the first one, the original images are normalized by resizing them to fit into a box of 50$\times$50 pixels keeping the aspect ratio and then 3 pixels width are added to pad the borders resulting in a 56$\times$56 pixels images. The second normalization centered and normalized the characters in a box of 90$\times$90 pixels without borders. In this Thesis all the experiments and results were developed with this second normalization. Some samples of images of the TICH database are shown in Fig. \ref{fig:databasesTICHSample}

This database also includes a list of character labels, a list of writer labels, a default partition in train and test subsets, and a baseline benchmark for the isolated character recognition problem.

The distribution of available characters is biased with the high frequency of English most common characters and they are detailed in Table \ref{table:TICHcharacters} for the train partition.

\begin{table}[ht!]\footnotesize
\centering
\begin{tabular}{lrr|lrr|lrr} 
 \toprule
 Char & N & Freq. & Char & N & Freq. & Char & N & Freq. \\ [0.5ex] 
 \midrule
   0 &    375 &   1.21\% &   C &     432 &   1.39\% &   O &   2,177 &   7.02\% \\
   1 &    171 &   0.55\% &   D &   1,704 &   5.49\% &   P &     317 &   1.02\% \\
   2 &    177 &   0.57\% &   E &   2,428 &   7.83\% &   Q &     171 &   0.55\% \\
   3 &    185 &   0.60\% &   F &     304 &   0.98\% &   R &   2,279 &   7.35\% \\
   4 &    168 &   0.54\% &   G &     454 &   1.46\% &   S &   1,286 &   4.15\% \\
   5 &    124 &   0.40\% &   H &     591 &   1.91\% &   T &   1,267 &   4.08\% \\
   6 &    156 &   0.50\% &   I &     668 &   2.15\% &   U &   2,049 &   6.61\% \\
   7 &    190 &   0.61\% &   J &     104 &   0.34\% &   V &   1,525 &   4.92\% \\
   8 &    135 &   0.44\% &   K &     696 &   2.24\% &   W &     785 &   2.53\% \\
   9 &    152 &   0.49\% &   L &     903 &   2.91\% &   X &       0 &   0.00\% \\
   A &  2,707 &   8.73\% &   M &   1,163 &   3.75\% &   Y &     364 &   1.17\% \\ 
   B &    610 &   1.97\% &   N &   3,469 &  11.18\% &   Z &     733 &   2.36\% \\
  \bottomrule
\end{tabular}
\caption{Character frequency in the train partition of TICH database.}
\label{table:TICHcharacters}
\end{table}

\subsection{New COUT database}
\label{section:New COUT database}
\index{Databases!COUT}
\index{Databases!UNIPEN}

The new COUT (Characters Offline from Unipen Trajectories) database contains images of isolated handwriting characters and is built using the UNIPEN online handwriting database \cite{guyon1994unipen}.

\begin{figure}[!ht]
\centering 
\includegraphics[width=8cm]{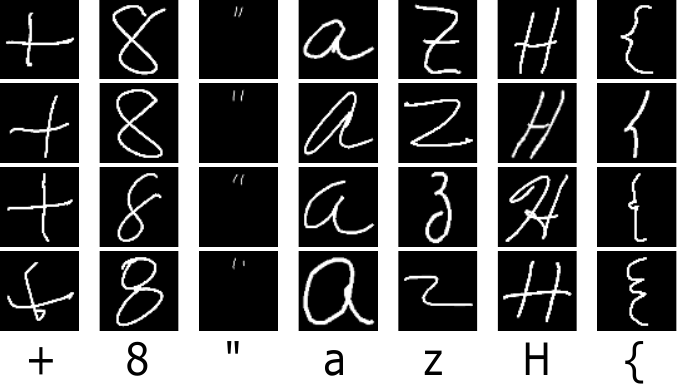} 
\caption{Sample images of the COUT database.}
\label{fig:SynteticUnipenSampleImages}
\end{figure}

To generate the COUT database, the point trajectories obtained from the original UNIPEN online handwriting database were used. Variable-thickness strokes were used depending on the original resolution of characters to ensure that all final characters have a similar thickness. The images from different available categories (i.e., different types of alphabet characters) were generated: uppercase, lowercase, digits and punctuation marks. These generated images are resized to 64$\times$64 pixels without changing their aspect ratio. Finally, the generated images were curated manually, one by one, to ensure that they are assigned to their correct category and that they are human-legible. Some samples of the database images can be viewed in Fig. \ref{fig:SynteticUnipenSampleImages}.

This database contains 93 categories and a total of 62,382 image characters in grayscale organized in 93 folders, one for each different character. The folder name is assigned using the ASCII numeric code of the character. The included characters are all between the ASCII code 33 (character: !) and the ASCII code 126 (character: \textasciitilde), both included. As for as we know, it is the handwriting character database that includes the largest number of different characters. The frequency of each character is detailed in Table \ref{table:SynteticUnipenFreqTable}.

\begin{table}[!ht]\footnotesize
\centering
\begin{tabular}{lrr|lrr|lrr|lrr} 
\toprule
Char & N & Freq. & Char & N & Freq. & Char & N & Freq. & Char & N & Freq. \\ [0.5ex] 
\midrule
A     & 923 & 1.49    & a     & 2,219 & 3.58   & 0 & 1,103 & 1.78    & ;     & 223 & 0.36 \\
B     & 505 & 0.81    & b     & 624 & 1.01    & 1 & 1,036 & 1.67    & \textless & 168 & 0.27 \\
C     & 644 & 1.04    & c     & 880 & 1.42    & 2 & 1,056 & 1.70     & =     & 254 & 0.41 \\
D     & 553 & 0.89    & d     & 1,074 & 1.73   & 3 & 1,047 & 1.69    & \textgreater & 162 & 0.26 \\
E     & 1180 & 1.90    & e     & 2,962 & 4.78   & 4 & 1,028 & 1.66    & ?     & 194 & 0.31 \\
F     & 486 & 0.78    & f     & 608 & 0.98    & 5 & 957 & 1.54     & @     & 83 & 0.13 \\
G     & 453 & 0.73    & g     & 760 & 1.23    & 6 & 943 & 1.52     & [     & 111 & 0.18 \\
H     & 557 & 0.90     & h     & 990 & 1.60     & 7 & 1,031 & 1.66    & ]     & 104 & 0.17 \\
I     & 898 & 1.45    & i     & 2,035 & 3.28   & 8 & 1,055 & 1.70     & \textasciicircum & 88 & 0.14 \\
J     & 476 & 0.77    & j     & 427 & 0.69    & 9 & 1,049 & 1.69    & \_    & 80 & 0.13 \\
K     & 508 & 0.82    & k     & 557 & 0.90     & ! & 207 & 0.33     & `     & 42 & 0.07 \\
L     & 703 & 1.13    & l     & 1,415 & 2.28   & " & 267 & 0.43     & \{    & 73 & 0.12 \\
M     & 496 & 0.80     & m     & 879 & 1.42    & \#   & 152 & 0.25   & \textbar & 91 & 0.15 \\
N     & 689 & 1.11    & n     & 1,906 & 3.08   & \$    & 192 & 0.31    & \}    & 77 & 0.12 \\
O     & 1,126 & 1.82   & o     & 2,048 & 3.30    & \%    & 190 & 0.31    & \textasciitilde & 59 & 0.10 \\
P     & 614 & 0.99    & p     & 786 & 1.27    & \&    & 104 & 0.17 &&\\
Q     & 413 & 0.67    & q     & 427 & 0.69    & '     & 276 & 0.45 &&\\
R     & 816 & 1.32    & r     & 1,708 & 2.76   & (     & 346 & 0.56 &&\\
S     & 799 & 1.29    & s     & 1,557 & 2.51   & )     & 359 & 0.58 &&\\
T     & 738 & 1.19    & t     & 1,781 & 2.87   & *     & 128 & 0.21 &&\\
U     & 461 & 0.74    & u     & 1,319 & 2.13   & +     & 146 & 0.24 &&\\
V     & 407 & 0.66    & v     & 555 & 0.90     & ,     & 320 & 0.52 &&\\
W     & 528 & 0.85    & w     & 680 & 1.10     & -     & 447 & 0.72 &&\\
X     & 368 & 0.59    & x     & 463 & 0.75    & .     & 486 & 0.78 &&\\
Y     & 457 & 0.74    & y     & 680 & 1.10     & /     & 259 & 0.42 &&\\
Z     & 459 & 0.74    & z     & 505 & 0.81    & :     & 287 & 0.46 &&\\\bottomrule
\end{tabular}
\caption{Character frequency in train partition of COUT database.}
\label{table:SynteticUnipenFreqTable}
\end{table}

The database is free available to download for research purposes from: \url{https://github.com/sueiras/handwritting_characters_database}

\section{General handwritten text databases}
\label{section:General handwriting databases}

This subsection describes in detail the databases used at word level. These datasets do not include any information that would allow these words to be segmented into their component characters. The databases considered provide segmented images of words as well as their transcriptions. In some cases, these databases have images and transcriptions at line or page level. In general, the datasets considered include text written by multiple authors with a high variability regarding writing styles and they cover three of the main languages with Latin script characters. In particular:

\begin{itemize}
    \item IAM database corresponds to English text,
    \item RIMES database corresponds to French text, and
    \item The Osborne database is written in Spanish
\end{itemize}

\subsection{IAM Database}
\label{chapter:2 - subsect:IAM}
\index{Databases!IAM}
\index{PNG}

\begin{figure}[!ht]
\centering 
\includegraphics[width=12cm]{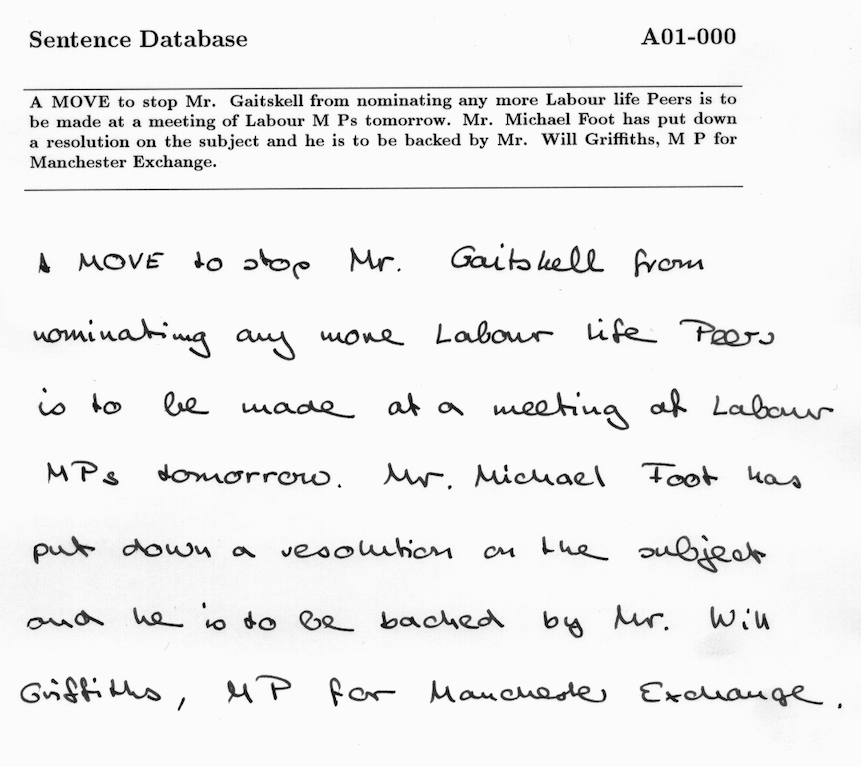} 
\caption{Sample image of a page of the IAM database.}
\label{fig:databasesIAMSamplePage}
\end{figure}

IAM database was introduced by Marti et al. \cite{marti2002iam} and it consists of handwritten English sentences extracted from the Lancaster-Oslo/Bergen (LOB) corpus \cite{johansson1986tagged}. This database is widely used in the scientific literature; see for example Doetsch et al. \cite{doetsch2014fast}, Bluche et al. \cite{bluche2016scan}, or Graves et al. \cite{graves2009offline}. This is due to its large diversity of styles contained and its size (as it includes more than one hundred thousand annotated words). IAM database also does not impose any restrictions on the annotated character set, including not only capital and lowercase but also the most common punctuation marks. This database provided images of forms, lines and words, with a resolution of 300 DPI that were saved as PNG with 256 gray levels. Fig. \ref{fig:databasesIAMSamplePage} show an example of one IAM form. 
  \index{Corpus!LOB}

\begin{figure}[!ht]
\centering 
\includegraphics[width=10cm]{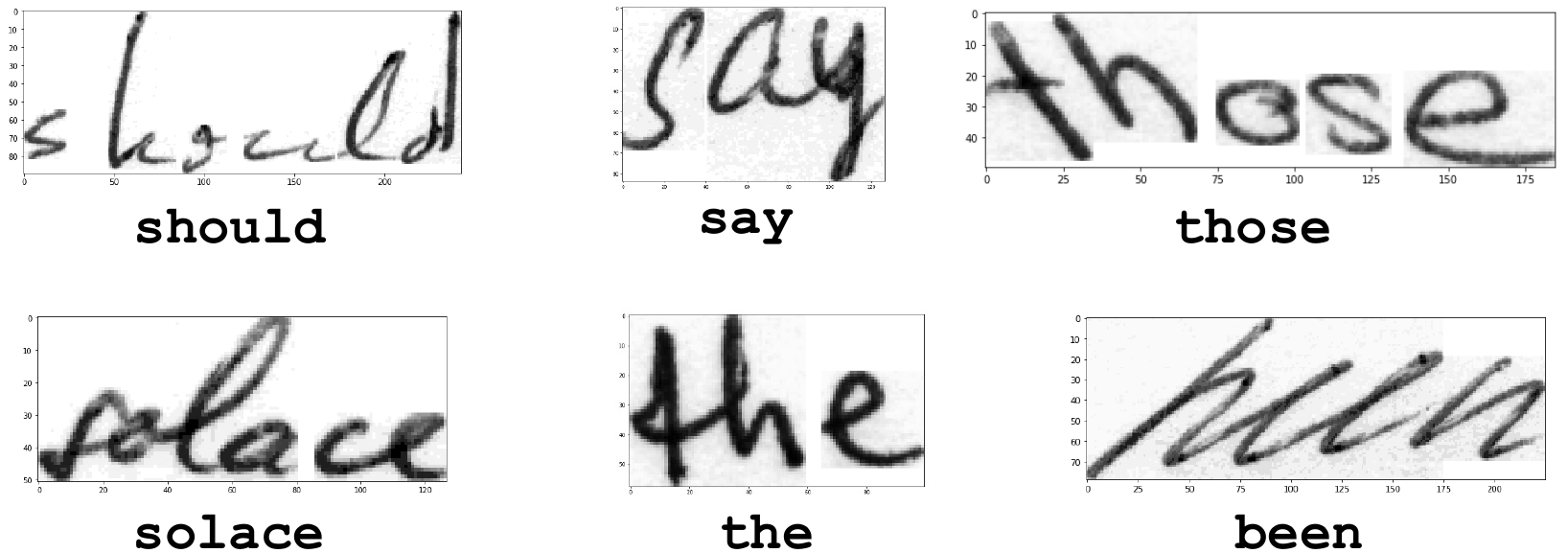} 
\caption{Sample images of the IAM database.}
\label{fig:databasesIAMSample}
\end{figure}

The complete database can be downloaded from \url{https://fki.tic.heia-fr.ch/databases/iam-handwriting-database} \footnote{Last access May 2021} previous registration. A total of 657 writers contributed with samples of their handwriting and the database includes a total of 1,539 text pages, 13,353 isolated and labeled text lines and 115,320 isolated and labeled words. In this Thesis we use the isolated words images and their corresponding annotations. Fig. \ref{fig:databasesIAMSample} includes some examples of several isolated words.

The authors also provided a standard partition per writer in a training set, two validation sets (val1 and val2) and one test set. In the problem of handwritten text recognition, both at word and line level, the train, val2 and test partitions are used, and the val1 partition is ignored. This selection of partitions contains a total of 8,962 lines from 454 different authors distributed as shown in Table \ref{table:IAM_official_partition_stats}

\begin{table}[ht!]
\centering
\begin{tabular}{lrrrr} 
 \toprule
 Partition & \#Authors & \#Pages & \#Lines & \#Words \\
 \midrule
 train & 283 & 747 & 6,161 & 53,841 \\
 val1  &  43 & 115 &   940 &  8,566 \\
 test  & 128 & 232 & 1,861 & 17,616 \\
 \bottomrule
\end{tabular}
\caption{Distributions of authors, pages, lines and words in the IAM official partition.}
\label{table:IAM_official_partition_stats}
\end{table}

This official partition was used by many different authors (see references \cite{bluche2013tandem}, \cite{poznanski2016CVPR}, \cite{dutta2018improving}, \cite{Chowdhury2018AnEE}, or \cite{huang2020end} for the development of models both at line and at word levels).

Additionally, there exist a second unofficial partition, usually named Aachen or RWTH split, widely used by most authors (see references \cite{kozielski2013improvements}, \cite{bluche2014comparison}, \cite{doetsch2014fast}, \cite{voigtlaender2016handwriting} or, \cite{sueiras2018offline}) who used the database for handwritten text recognition. To our best knowledge, this partition was first used in \cite{kozielski2013improvements}. This partition included a total of 747 pages to train, 116 for validation and 336 for test. The lines and words are distributed as shown in Table \ref{table:IAM_aachen_partition_stats}, The partition is available in \url{http://www.openslr.org/56/}\footnote{Last access May 2021}.

\begin{table}[ht!]
\centering
\begin{tabular}{lrrr} 
 \toprule
 Partition & \#Pages & \#Lines & \#Words \\
 \midrule
 train      & 747 & 6,482& 55,081 \\
 validation & 116 &   976&  8.895 \\
 test       & 336 & 2,915& 25,920 \\
 \bottomrule
\end{tabular}
\caption{Distributions of pages, lines and words in the IAM Aachen partition.}
\label{table:IAM_aachen_partition_stats}
\end{table}

\begin{figure}[!ht]
\centering 
\includegraphics[width=10cm]{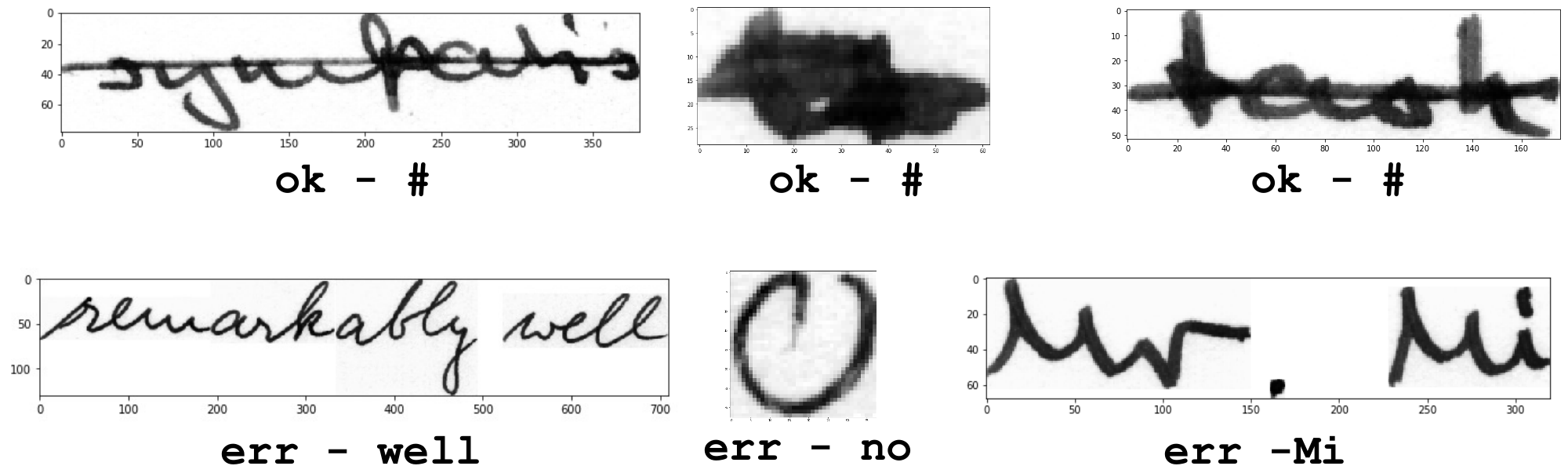} 
\caption{Sample error images of the IAM database.}
\label{fig:databasesIAMSampleErrors}
\end{figure}

The database includes a metadata file with the details of each isolated word segmented, including the necessary label transcription of each word. The file also includes a column that indicates if the word was correctly segmented or not. These segmentation errors are marked at the line level, even though only one word in that line may have segmentation errors. Since it is not possible to identify in the data which words of each line are incorrect, a valid option is to remove the wrong marked lines from the different partitions. This is because the IAM database was originally annotated only at line level. In a later work of the same authors \cite{zimmermann2002automatic}, a bounding box and a word level transcript were provided, based on a segmentation algorithm of the identification of related components, when it was later validated manually at line level to identify erroneous cases. Altogether, there was 2,009 error marked lines out of a total of 13,353 available lines. In addition, it has been identified that the words annotated with the character '\#', although they are identified as correct segmentation, correspond to crossed-out words that must also be discarded (there are a total of 33 words in this situation). Some examples of images of these words are included in Fig. \ref{fig:databasesIAMSampleErrors}.
In summary, after filtering the words that are marked as correctly segmented and with different annotations for the character '\#', were obtained a final 47,952, 7,558 and 20,306 words for the training, validation and test sets, respectively, over the Aachen partition. These are the final datasets that we have originally used in the publication \cite{sueiras2018offline} which is discussed in detail in Chapter \ref{chapter_word_models}.

The set of characters that appears in IAM database texts are composed of a total of 77 different symbols, including all the English lowercases and capital characters, and also the punctuation marks included in Table \ref{table:IAMcharacters}. In addition, it also includes the symbol '\#' to represent garbage (basically crossed-out words) and also the 33 occurrences of the blank character (the ASCII code 32), mainly used to separate capital letters from acronyms and abbreviations.

\begin{table}[!ht]
\centering 
\begin{tabular}{lrr|lrr|lrr} 
 \toprule
 Char & N & Freq. & Char & N & Freq. & Char & N & Freq.f \\ [0.5ex] 
 \midrule
 . & 5,005 & 1.244\% & , & 4,418 & 1.098\% & '  & 1,379 & 0.343\% \\ 
 - & 1,184 & 0.294\% & " & 1,080 & 0.268\% & !  &   185 & 0.046\% \\
 ? &   162 & 0.040\% & ; &   159 & 0.040\% & :  &   137 & 0.034\% \\
 ( &   103 & 0.025\% & ) &    96 & 0.024\% & \& &    47 & 0.012\% \\
 / &    10 & 0.002\% & * &     5 & 0.001\% & +  &     5 & 0.001\% \\
 \bottomrule
\end{tabular}
\caption{Frequency of punctuation symbols that appears as a segmented words in the IAM database.}
\label{table:IAMcharacters}
\end{table}

\subsection{RIMES database}
\label{chapter2:The_RIMES_database}
\index{Databases!RIMES}

RIMES (Reconnaissance et Indexation de données Manuscrites et de fac similÉS / Recognition and Indexing of handwritten documents and faxes) database \cite{augustin2006rimes} consists of images of handwritten letters in exchange of gift vouchers. The database is maintained by the company A2IA and the complete version contains 12,723 pages corresponding to three types: handwritten letters, forms and fax covers. This database was designed to cover several tasks such as layout analysis, handwriting recognition, writer identification, logo identification and information extraction. The database is available at \url{http://www.a2ialab.com/doku.php?id=rimes_database:start}\footnote{Last access May 2021}.

\begin{figure}[!ht]
\centering 
\includegraphics[width=8cm]{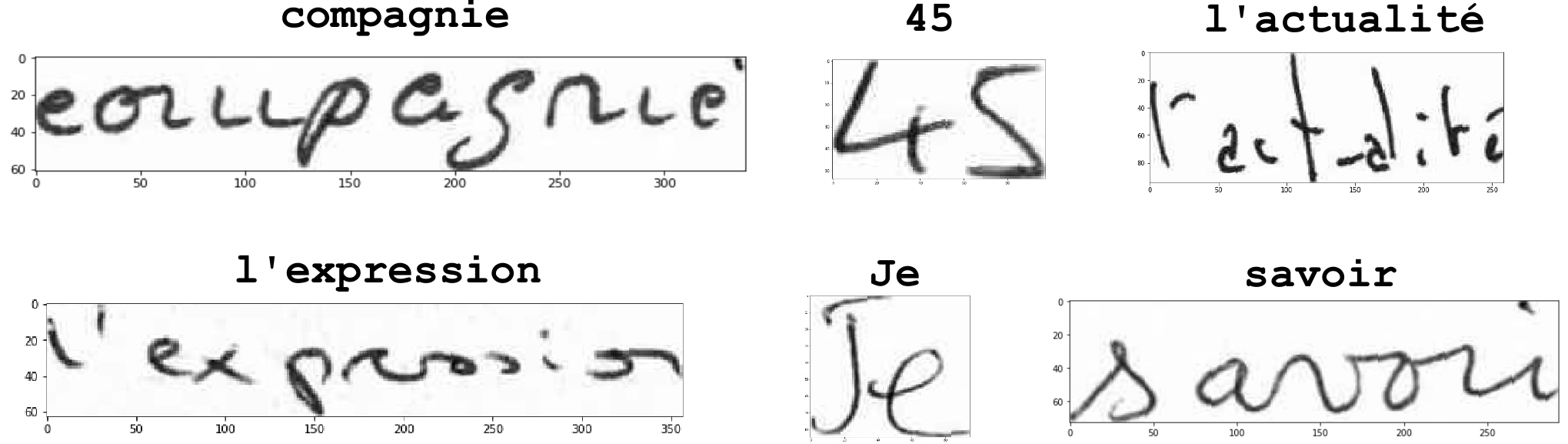} 
\caption{Sample images of the RIMES database.}
\label{fig:databasesRIMESSample}
\end{figure}

RIMES database has been used for several handwriting competitions (such as ICFHR 2008, ICDAR 2009, ICDAR 2011) with different tasks and datasets. In this Thesis we have used the ICDAR 2011 competition on word recognition. This version of the database includes 51,739 words images to train, 7,464 for validation and 7,776 to test, respectively. Fig. \ref{fig:databasesRIMESSample} includes some examples of several isolated words at RIMES.
\index{Conferences!ICFHR}
\index{Conferences!ICDAR}

The set of characters in words of the RIMES database include all the lowercase and capital characters contained in standard ASCII, the ten digits and some punctuation marks and specific French characters detailed in Table \ref{table:RIMEScharacters}

\begin{table}[!ht]
\centering 
\begin{tabular}{lr|lr|lr|lr} 
 \toprule
 Char & N & Char & N & Char & N & Char & N \\ [0.5ex] 
 \midrule
 é & 5,155 & ' & 2,646 & è & 762  & - & 360 \\
 à  & 178  & ê &   158 & ç & 94   & ô & 69  \\
 î & 67    & / &   58  & û & 52   & â & 23  \\
 ù & 15    & ° &   13  & ë & 3    & ï & 2   \\
 \% & 1    & ² &   1   & É & 1    &   &     \\
 \bottomrule
\end{tabular}
\caption{Frequency of accented vowels and punctuation symbols in the train partition of the RIMES database.}
\label{table:RIMEScharacters}
\end{table}

\subsection{Osborne database}
\index{Databases!Osborne}

\begin{figure}[!ht]
\centering 
\includegraphics[width=14cm]{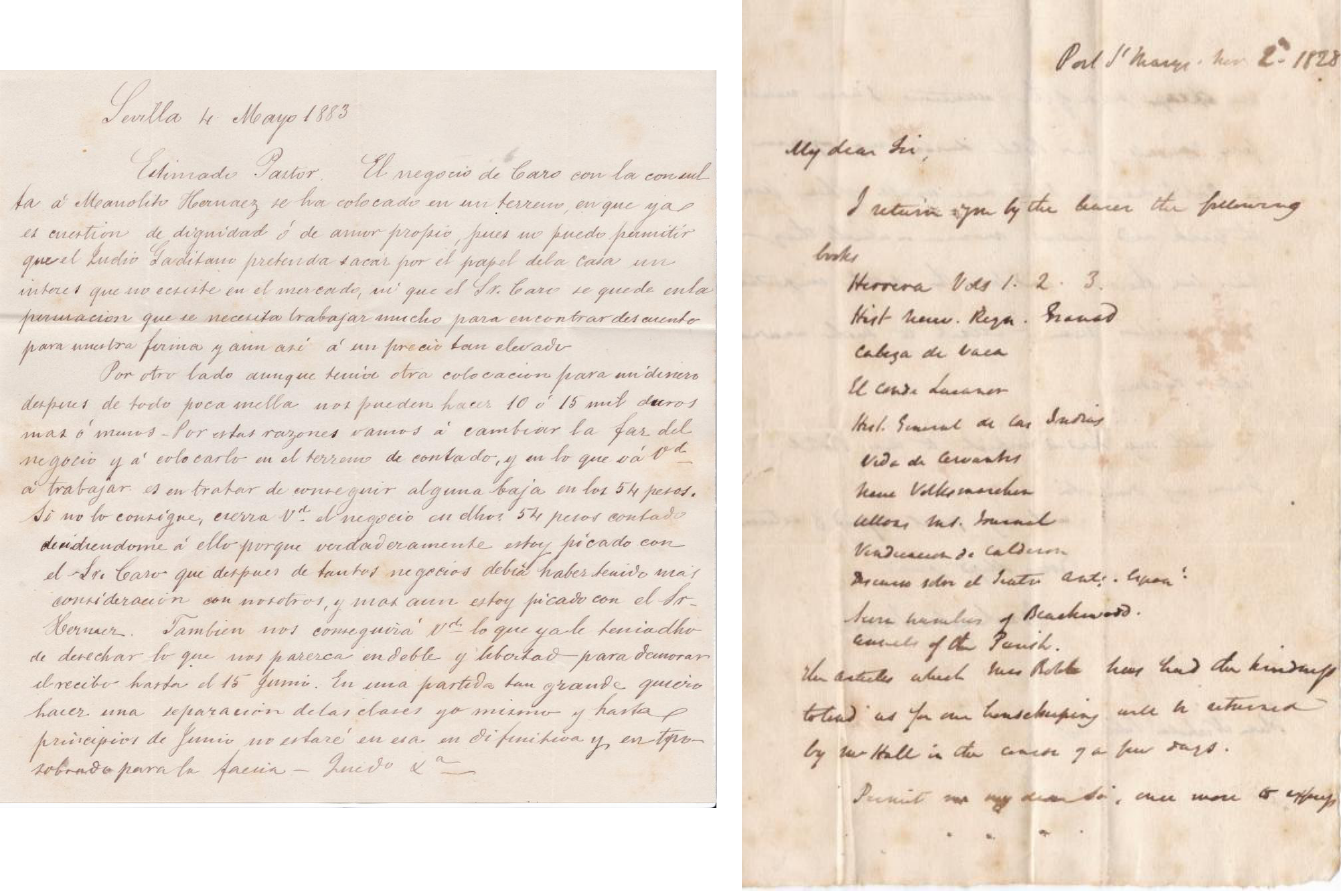} 
\caption{Sample images of pages of the Osborne database.}
\label{fig:databasesOsborneSamplePages}
\end{figure}

Founded in 1772, Bodegas Osborne is one of the world’s oldest companies in active. Along their history they provide wines to several illustrious clients in Europe like the British Royal Family or the Vatican. This has caused that the company had have to exchange mail with relevant historical figures from the 18th century onwards. The Osborne historical archive counts with several hundreds of documents including personal and commercial letters or counting sheets. Fig. \ref{fig:databasesOsborneSamplePages} contains two documents examples of this database. All the available document images and transcriptions can be accessed at \url{https://www.fundacionosborne.org/es/simple-search}\footnote{Last access May 2021}.

\begin{figure}[!ht]
\centering 
\includegraphics[width=10cm]{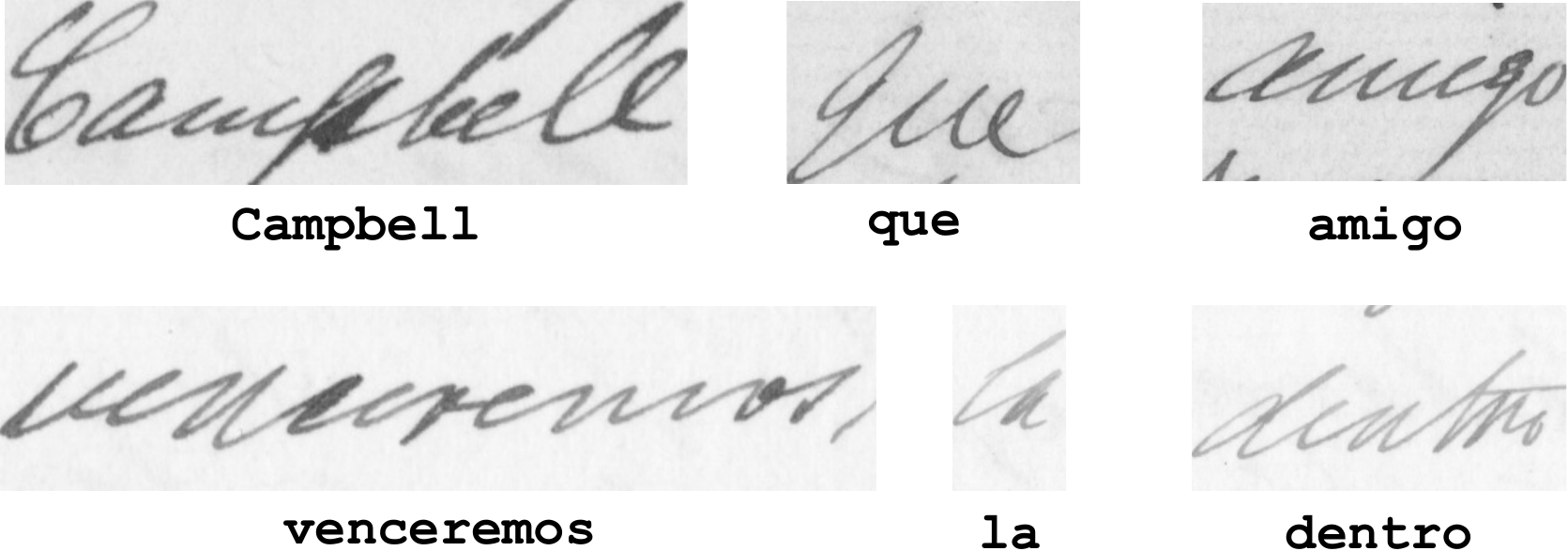} 
\caption{Sample images of the Osborne database.}
\label{fig:databasesOsborneSampleWords}
\end{figure}

The dataset used in this Thesis includes a subset of images corresponding to isolated words extracted from a total of 138 pages from 58 documents digitized from this archive. A total of 6 documents are from the year 1877, other 7 ones are from 1882 and 45 were dated in the year 1883. These documents were scanned using RGB 400 DPI resolution. The annotation process was developed by a specialized documentation team and the isolated word images are obtained by applying an automatic process that includes line segmentation and word segmentation. The result is a human curated dataset to ensure that the cropping and translation are correct. The final images were converted to grayscale with 8-bits pixels depth and saved in PNG format. Fig. \ref{fig:databasesOsborneSampleWords} includes some image examples from Osborne database.

The dataset includes a default partition in train, validation and test which was used in this Thesis. The partition has been built randomly by selecting all pages of complete scanned documents and all the words of each page in one partition. The train partition has a total of 7,149 word images extracted from 129 pages of 54 different documents. The validation partition has 288 words images extracted from 5 pages of 2 documents, and the test partition covers 194 word images extracted from 4 pages of 2 different documents. 

The target annotations are included in the name of each image file. In this name it is registered: the year of the document, the document identifier, the page of the document, the word position in the page and the transcription of the word encoded in hexadecimal format. The number of different words in the database is 2,893 and the most frequent word is 'de' with a total of 358 occurrences. 

Because the segmentation into words has been done by an autonomous process, in some cases, the segmented image corresponds to a part of a word or to two or more words whose strokes are close and have not been identified as different words. In these cases, the blank character ' ' may be part of the transcription of the image text. It has been manually verified that all the transcriptions are correct.

Table \ref{table:OsborneCharacters} details the frequency of characters that appears in the database.

\begin{table}[!ht]\footnotesize
\centering
\begin{tabular}{lrr|lrr|lrr}
\toprule
char &     N &  Freq. & char &     N &  Freq.& char &      N &  Freq.\\
    \midrule
     &   468 &  1.42 &    F &     9 &   0.03 &    j &    112 &  0.34 \\
   ! &     3 &  0.01 &    G &     6 &   0.02 &    k &     14 &  0.04 \\
   \&&     6 &  0.02 &    H &    26 &   0.08 &    l &  1,549 &  4.69 \\
   ' &    52 &  0.16 &    I &    15 &   0.05 &    m &    898 &  2.72 \\
   ( &    16 &  0.05 &    J &    24 &   0.07 &    n &  2,002 &  6.06 \\
   ) &    13 &  0.04 &    L &    52 &   0.16 &    o &  2,919 &  8.83 \\
   , &   264 &  0.80 &    M &    78 &   0.24 &    p &    752 &  2.28 \\
   - &   284 &  0.86 &    N &    25 &   0.08 &    q &    445 &  1.35 \\
   . &   111 &  0.34 &    O &    20 &   0.06 &    r &  2,022 &  6.12 \\
   0 &    19 &  0.06 &    P &    44 &   0.13 &    s &  2,216 &  6.71 \\
   1 &    58 &  0.18 &    Q &    10 &   0.03 &    t &  1,318 &  3.99 \\
   2 &    49 &  0.15 &    R &    18 &   0.05 &    u &  1,240 &  3.75 \\
   3 &    36 &  0.11 &    S &    98 &   0.30 &    v &    364 &  1.10 \\
   4 &    18 &  0.05 &    T &    43 &   0.13 &    w &     22 &  0.07 \\
   5 &     7 &  0.02 &    U &    77 &   0.23 &    x &     20 &  0.06 \\
   6 &     7 &  0.02 &    V &    23 &   0.07 &    y &    373 &  1.13 \\
   7 &    14 &  0.04 &    W &     1 &   0.00 &    z &     80 &  0.24 \\
   8 &    67 &  0.20 &    Y &    17 &   0.05 &    ¡ &      7 &  0.02 \\
   9 &     7 &  0.02 &    a &  3843 &  11.63 &    ¿ &      2 &  0.01 \\
   : &     5 &  0.02 &    b &   392 &   1.19 &    á &     55 &  0.17 \\
   ; &    11 &  0.03 &    c &  1163 &   3.52 &    é &     50 &  0.15 \\
   ? &     7 &  0.02 &    d &  1588 &   4.81 &    í &     13 &  0.04 \\
   A &    83 &  0.25 &    e &  4319 &  13.07 &    ñ &     54 &  0.16 \\
   B &    32 &  0.10 &    f &   200 &   0.61 &    ó &     17 &  0.05 \\
   C &    76 &  0.23 &    g &   387 &   1.17 &    ú &      7 &  0.02 \\
   D &    89 &  0.27 &    h &   340 &   1.03 &      &        &       \\
   E &    87 &  0.26 &    i &  1785 &   5.40 &      &        &       \\
\bottomrule
\end{tabular}
\caption{Character frequency in the Osborne database.}
\label{table:OsborneCharacters}
\end{table}

\section{Databases summary}
\label{section:Databases summary}

This subsection includes two summary tables of the databases studied. For the character databases we include the character set and the cardinality of the target, The number of images, the number of writers (if it is knows) and the resolution of the images in the Table \ref{table:characterDatabases}.

\begin{table}[!ht]
\centering 
\begin{tabular}{llllr}
 \toprule 
 Database & Char set (N) & Writers & Resolution & Size \\
 \midrule 
 MNIST& Digits(10)                  &          & 28$\times$28&    70,000   \\
 TICH & Digits-upper(36)            &       250& 90$\times$90&    40,141   \\
 NIST & Digits-upp-low(52)          &     3,669& 128$\times$128& 814,225 \\
 COUT & Digits-upp-low-symbols(92)  &          & 64$\times$64&    62,382   \\
 \bottomrule 
\end{tabular}
\caption{Summary for the isolated character databases.}
\label{table:characterDatabases}
\end{table}

For the continuous text databases, data on the number of different writers, number of pages, number of lines, and number of words are included for each partition in the Table \ref{table:wordDatabases}. The RIMES data are calculated from the RIMES words database of the ICDAR 2011 competition \footnote{\url{http://www.a2ialab.com/doku.php?id=rimes_database:data:icdar2011:word:icdar2011competitionword}}

\begin{table}[!ht]
\centering 
\begin{tabular}{lllrrrr}
 \toprule 
 Database & & Partition & \#writers & \#pages & \#lines & \#words  \\
 \midrule
                && Train     &   &  747 & 6,482 & 55,081 \\
 IAM (Aachen)   && Validation&   &  116 &   976 &  8.895 \\
                && Test      &   &  336 & 2,915 & 25,920 \\
 \midrule
                && Train     &   283&   747 & 6,161& 53,841 \\
 IAM (Official) && Validation&    43&   115 &   940&  8.566 \\
                && Test      &   128&   232 & 1,861& 17,616 \\
 \midrule
                && Train     &   &  1,344&  8,807&  51,739 \\
 RIMES (2011)   && Validation&   &    196&  1,268&   7,464 \\
                && Test      &   &    179&  1.250&   7,776 \\
 \midrule
                && Train     &    &  112&   & 6.340 \\
 Osborne        && Validation&    &   13&   &   646 \\
                && Test      &    &   13&   &   645 \\
 \bottomrule 
\end{tabular}
\caption{Summary for the continuous handwriting text databases.}
\label{table:wordDatabases}
\end{table}


\chapter{Review of Handwriting recognition approaches}
\label{chapter_background}

This chapter reviews the different approaches to solving the offline handwritten text recognition problem over time. In the last twenty years under review, the problem has been actively investigated, and several approaches to the problem have emerged, which have prevailed at different stages and have generally replaced the approaches of the previous stages. For a review of previous years, see Plamondon and Srihari's survey paper \cite{plamondon2000online}.

In the first Section \ref{c3:section:Introduction}, the different stages that have been identified over time are summarized. The next Section \ref{section: Data_normalization_background} reviews the various input image preprocessing and normalization algorithms used. Section \ref{section: Data_augmentation_background} describes data augmentation techniques considered. This is followed by details of the HMM-based algorithms that have been predominant up to 2013 (Section \ref{section:Classical approaches based on HMM}). Subsequently, the solutions using neural networks, predominant from 2013 onwards, are analyzed in Section \ref{section:Neural networks and deep learning architectures} and the different types of networks used. The next Section \ref{section:Connectionist Temporal Classification - CTC} reviews the use of Connectionist Temporal Classification (CTC) in the offline handwriting recognition problem, usually employed with previous architectures based on neural networks. This is followed by a review of k-NN-based solutions in Section \ref{section:Solutions based on Nearest neighbors}. The next Section \ref{section:seq2seq architectures} details the approaches employing seq2seq architectures. Section \ref{section: synthetic data and transfer domain} reviews the works that have employed synthetic data and transfer domain strategies which are mostly concentrated in the last three years. Finally, in Section \ref{c4 - section - Language models} we explain the historical applications of language models in the HTR problem.

\section{Introduction}
\label{c3:section:Introduction}

The problem of automatic Handwritten Text Recognition (HTR) persists since document digitization started. Text recognition is a simple task for humans, but it has been proved to be complex for automatic systems. In fact, it is considered an unsolved problem and under active research \cite{kang2021candidate}, \cite{Poulos2021character}. The high variability between writers and the cursive nature of the handwriting text are the main difficulties presented by this problem. These difficulties have meant that historically, the practical applications of offline handwriting recognition technologies have been quite limited \cite{cheriet2009handwriting}.

The first Offline HTR models were aimed at recognizing isolated characters \cite{lecun1989backpropagation}. The MNIST database \cite{lecun1998gradient}, described in Subsection \ref{section:MNIST database}, consisted of fixed-size images of handwritten digits from 0 to 9, is well known. This database, created in 1998, is still widely used today as a problem benchmark for nonlinear models. 
  \index{Databases!MNIST}

Subsequently, the analysis was focused on the recognition of certain handwritten content restricted to some specific application domains. For example, recognize numerals \cite{suen1992computer}, characters \cite{impedovo1991optical}, bank checks \cite{impedovo1997automatic} or postal addresses \cite{brakensiek2004handwritten}, among others. This Thesis focuses on the general HTR problem, and the goal is to develop systems capable of recognizing any image in a handwritten text, regardless of its content, author, or nature. The publication of the IAM database in 2002 \cite{marti2002iam} settles a turning point to detect an increased number of publications focused on this specific problem.
  \index{Databases!IAM}

In this chapter, the main approaches to the general HTR problem are reviewed, including a description of their evolution along time as well as the main contribution of each one. In certain cases, the relationships of each model with the experiments described in this Thesis and their application are also indicated. The review of these methods and their evolution over time provide adequate context to understand the suggested models and the carried out experiments. 

The different general HTR models are typically composed of three steps and different approaches are considered for each one. First, the strategy to extract features from the handwriting image. Second, the model to convert these features into an output signal. Third, the way to decode the output signal to predict the text that appears in the input image. It is useful to provide a brief review of the evolution of the developed approaches in each step to understand well the evolution of the different approximations to the HTR problem.

Initially, the strategies to extract image features were rather complex and elaborated because the recognition models were not expressive enough to analyze the image pixel values directly. For example, Doetsch et al. \cite{doetsch2014fast} and Kozielski et al. \cite{kozielski2013improvements} employed \textit{Principal Component Analysis} (PCA) to extract components over fixed-size frames of pixels. Bideault et al. \cite{bideault2015spotting} also extracted features, in this case using \textit{Histograms of Oriented Gradients} (HOG). Graves et al. \cite{graves2009offline} considered features extracted of each column pixel, such as mean and other moments, center-of-gravity, transitions and other aggregations.
  \index{Principal component analysis}
  \nomenclature{HOG}{Histograms of Oriented Gradients}
  \index{Histograms of oriented gradients}

The change arose when more complex and expressive models, with deep \textit{Neural Networks} (NN) architectures, composed of \textit{Recurrent Neural Networks} (RNN) and \textit{Convolutional Neural Networks} (CNN), appeared. The use of RNNs, and specially CNNs, make the extraction of features unnecessary and pixel features are used directly as input of the handwriting model. The works by Graves et al. \cite{graves2009offline} and Bluche et al. \cite{bluche2016scan} are two of the first examples of this strategy. The direct use of the pixels from the image as an input to the model had become a standard over the past few years, and this is the strategy selected in this Thesis.
  \index{Recurrent neural network}
  \index{Convolutional neural network}

The second component, the recognition model, is the core of each solution to the HTR problem. These models are the most developed and researched ones, and the evolution (and the main contributions) of these models are detailed in the next sections. This is linked to the important development of neural models in recent years and the emergence of deep architectures. The progress in the architectures for the HTR problem has frequently relied on the advances in Computer Vision, Speech Recognition or Language Modeling \cite{graves2006connectionist} \cite{bahdanau2016end}. 

These models were initially based on \textit{Hidden Markov Models} (HMM), first considered separately and later combined with a neural network, which acted as a feature extractor for the HMM. These networks are usually \textit{Multilayer Perceptrons} (MLP) or \textit{Recurrent Neural Networks} (RNN). 
  \index{Hidden Markov model}
  \index{Multilayer perceptron}

As an alternative to HMM, the first models were based on the \textit{Multidimensional Recurrent Neural Networks} (MDRNN) . They were introduced in 2007 by Graves et al. \cite{graves2007multidimensionalrecurrent}, and its architecture exploits the bi-dimensional nature of the handwritten images. In all cases, the authors propose a deep architecture with several \textit{Multidimensional Long Short Term Memory} (MDLSTM) networks. Some examples are included in the works \cite{graves2009offline}, \cite{voigtlaender2016handwriting} and \cite{bluche2016scan}.
  \index{Multidimensional recurrent neural network}
  \nomenclature{MDLSTM}{Multidimensional Long Short Term Memory}

Later on, the MDLSTM models were progressively outdated due to their high computational cost, whereas more efficient architectures offered similar or even better results. At that time, an architecture based on one sequence of CNNs and RNNs became popular and it continues to be used nowadays.

Next, the models named as \textit{sequence-to-sequence} (seq2seq), described in Section \ref{section:seq2seq architectures}, were proposed. These are also the ones used in this Thesis. These models become popular after the papers of Sueiras et al. \cite{sueiras2018offline}, and Kang et al. \cite{kang2018convolve}, both published in 2018 and they persist as an active research line for the HTR problem.
  \index{Sequence-to-sequence}
  \nomenclature{seq2seq}{sequence-to-sequence}

In the third step of output decoding, the handwriting prediction output is transformed into the sequence of text characters that appear in the image. When the model was a HMM, this step is developed with an alignment algorithms like the \textit{Viterbi} one \cite{forney1973viterbi} (see for example \cite{bluche2013tandem} or \cite{doetsch2014fast}). This approach based on HMM with a subsequent alignment was complex because it required several steps as well as quite manual tunning. 
  \index{Viterbi algorithm}

A second option for decoding was provided by the \textit{Connectionist Temporal Classification} (CTC), introduced in 2006 by Graves et al. \cite{graves2006connectionist}. This layer, specialized in the alignment of sequences of different lengths, allowed the training end-to-end of an HTR model without the need for several steps (as it happened with HMM). The first application to the HTR problem was carried out by Graves et al. \cite{graves2009offline} in 2009. Since 2013 its use became popular (see for example references \cite{bluche2016scan} and \cite{voigtlaender2016handwriting}), and it continues to be so actually.
  \index{Connectionist temporal classification}

Finally, with the emergence of the seq2seq models, described in the Section \ref{section:seq2seq architectures} of this chapter, the third type of decoding algorithm arises. In these models, the decoding is performed by a decoder component of the model, usually composed of a Recurrent Neural Network. This RNN generates the sequential decoding, character by character, of the text contained in the input image.

\section{Data normalization background}
\label{section: Data_normalization_background}
   \index{Image preprocessing}

The main difficulty of the HTR problem lies in the large variability of handwritten text. Each individual has his/her own particularities when handwriting. Even the same person writes differently on different situations, for example writing faster or slower or using different sizes of letters. Additionally, different types of paper or pens incorporate additional variations. This is why a great deal of effort has long been devoted to the development of text normalization algorithms to reduce this variability. 

In this section we introduce the main concepts of handwritten text image normalization, such as the baseline and upperline detection, and the slant or slope corrections. Next, we review the main contributions of the literature in this area.

Data normalization has been of great importance in HMM-based models. This is due to the limited ability of such models to describe all the variability existing in unnormalized text. This was also the case with other NN-based approaches where the size of the model was limited by the existing computational capabilities of the time when they were proposed. It is at that time that the normalization techniques described below were most developed.

In recent years, the size of NN models that can be trained in reasonable times has increased significantly. Therefore, publications such as \cite{kang2020unsupervised} or \cite{huang2020end} in which no prior normalization of the input images is applied to the model, except for a rescaling and adjustment to a fixed size, have appeared. However, some authors have demonstrated their effectiveness \cite{dutta2018improving} also in this type of larger models. In this Thesis, data normalization techniques are used and their impact in terms of accuracy are analyzed in Chapter \ref{chapter_word_models}.

The main normalizations carried out on handwritten images are the following ones:
\begin{itemize}
  \item Removal of background noise to improve the image contrast.
  \item Text slope correction.
  \item Text slant or cursive correction.
  \item Normalization in size of the ascenders and descenders zones in characters.
  \item Image resizing keeping its aspect ratio.
\end{itemize}

The application of the above techniques in the offline HTR literature is quite variable, although three main groups can be distinguished:

\begin{itemize}
  \item There are authors who perform a 'complete' image normalization and use all of the above normalizations with minor variations, such as: \cite{bertolami2008hidden}, \cite{graves2009novel} o \cite{bluche2014comparison}.
  
  \item Other authors perform a basic normalization, focusing on the two aspects identified as most relevant, which are contrast normalization and slant correction. In this case they are: \cite{doetsch2014fast}, \cite{bluche2013tandem} or \cite{kozielski2013improvements}
  
  \item Some authors do not apply any standardization, except a resize to fit the data into the input dimension of the model, such as: \cite{dreuw2011hierarchical}, \cite{kang2020unsupervised} or \cite{huang2020end}.
\end{itemize}

The main contributions of the literature for each of the above normalizations are summarized below. It is important to note that the data normalization procedures described in this section are applied to databases of text written in Latin alphabets. The analysis of normalization techniques in non-Latin languages, such as Arabic or Chinese, is beyond the scope of this Thesis.

\subsection{Noise removal}
  \index{Image denoising}

The objective of the denoising process is to obtain an image with a white background and well defined characters. For this, the usual thing is to use contrast enhancement techniques.

The simplest way to increase the contrast is to binarize the image. That is, a grayscale image with pixel values in range $[0, 255]$ is converted into a binary image. For this, a threshold is selected so that if the pixel value is lower than the threshold, the pixel is encoded as black (pixel value 0) and if the value is higher or equal, it is encoded as white (pixel value 255). The most common way to select this threshold is to use the algorithm proposed by Otsu in 1979 \cite{otsu1979threshold}, which assumes the existence of a bimodal distribution in the pixel image histogram and selects a threshold so that the quotient of the variance between the two segments in distribution and the variance within each segment is maximized.

Some authors use two thresholds, for example, \cite{pesch2012analysis} and \cite{bluche2015deep} start from the hypothesis that text strokes pixel values are lower than the background pixels and establish a threshold that maps 5\% of the darkest pixels to black and another threshold to map 70\% of lighter pixels to white. Intermediate pixels are linearly mapped to the range of values $(0, 255)$ that represent the entire gray scale.

Selecting a single threshold for the entire image can be ineffective, especially if the noise level of the image is different in separate sections of the image. In these cases the use of adaptive thresholds methods such as \cite{sauvola2000adaptive}, which calculate neighborhood-based thresholds for each pixel, provide better results.

In any case, the conversion of the handwritten text image into a binarized image is not the best option for its subsequent recognition by an automatic method, because the binarization to two black and white values produces irregularities and artifacts on the edges of lines that distort them. 

Hybrid methods which combines the advantages of local binarization and preservation of the gray scale provided better results \cite{villegas2015modification}. The authors modify the Saviola's algorithm \cite{sauvola2000adaptive}, so that instead of binarizing each pixel based on its neighborhood, it converts the pixel into a gray value through a linear transformation.

\subsection{Slope correction}
  \index{Slope correction}

One of the most popular slope correction techniques is the \textit{Hough transform} \cite{duda1972use} due to its robustness and simplicity. However, this technique is computationally expensive. Therefore, several authors have proposed variants that reduce the size of the Hough space \cite{pal1996improved} \cite{boukharouba2017new}, although the computational cost remains high. Boudra et al. \cite{boudraa2017improved} proposed a more efficient method, called \textit{Progressive Probabilistic Hough Transform} (PPHT), which computes the global slope angle estimation of a document with a more adjusted computational cost.
  \nomenclature{PPHT}{Progressive Probabilistic Hough Transform}
  \index{Progressive probabilistic Hough transform}

Other widely used method, both to identify the slope and the slant, are based on projection profiles, such as \cite{kavallieratou2002skew} or \cite{pastor2004projection}. The horizontal projection of the text pixels is used to identify the slope, and the vertical projection is used to identify the slant angle. These methods are quite sensitive to image noise, and for applying them it is necessary first to use contrast enhancement techniques such as those described in previous section.

\begin{figure}[!ht]
\centering 
\includegraphics[width=15cm]{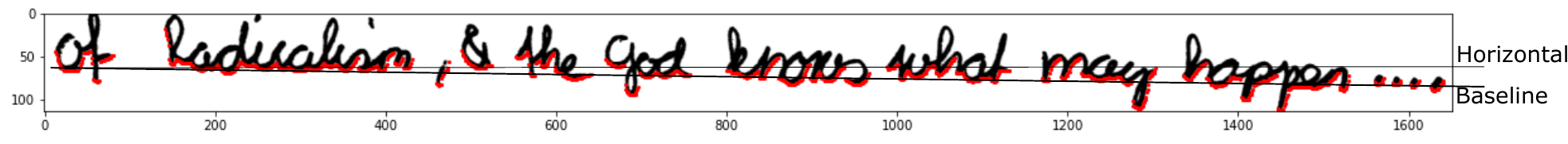} 
\caption{Example of slope correction.}
\label{fig:slope_example_c3}
\end{figure}

Alternatively, other authors identify the slope angle by estimating the line that best fits a selection of pixels in the image. For example, Papandreou and Gatos \cite{papandreou2014slant} use the pixels of the so-called core-region which is the area between the upper and lower baselines of a line or word image. Fig. \ref{fig:slope_example_c3} shows these lines that separate the central region of the text from its ascenders and descenders. The estimation can be done in multiple ways, for example, Kumar et al. \cite{bera2018one} propose the adjustment of an ellipse on the image to the word that allows the simultaneous estimation of the slope and the slant from the axes of the ellipse, Okun et al. \cite{okun1999document} estimate the best fit line using the maximum eigenvectors of the covariance matrix and Gupta and Chanda \cite{gupta2014efficient} use linear regression of the ${x, y}$ positions of pixels in the core region.

\subsection{Slant correction}
  \index{Slant correction}

As indicated in previous section, methods based on projection profiles are commonly used to identify the slant angle (see for example \cite{kavallieratou2002skew} and \cite{pastor2004projection}). In these cases, the algorithm performs an angle systematic search in a given range by applying the correction transformation for each possible value and obtaining a measure of verticality to be maximized. In the case of work by Kavallieratou et al. \cite{kavallieratou2002skew}, the maximum intensity of the \textit{Wigner-Ville Distribution} \cite{cohen1966generalized} of the document image horizontal histogram is used as criterion for its skew angle estimation. Pastor et al. \cite{pastor2004projection} propose a method based on the hypothesis that the distribution of the vertical projection profile histogram presents a maximum variation for the deslanted text. 

In the work by Gupta and Chanda \cite{gupta2012novel}, an algorithm for identifying the slant angle is proposed, which is based on the ability of the \textit{Gabor filter} to detect directional textures. To do this, it applies a Gabor filter of a size equal to the height of text core-region and rotated at a certain angle in a given search range. The angle for which the absolute sum of all pixel values is maximum, is the overall slant angle of the text image. 
  \index{Gabor filter}

Another example can be found in Vinciarelli and Luettin \cite{vinciarelli2001new} that propose a deslanting technique based on the hypothesis that the word image is deslanted when the number of columns that contains continuous strokes is maximum. In the case of this last algorithm, Pastor et al. \cite{pastor2006improving} have shown that it is very sensitive to the usual sources of noise such as salt and pepper, so these authors proposed a variant that softens the requirement of analyzing vertical strictly strokes, extending it to quasi-continuous strokes.

Alternatively, other authors have proposed a direct estimation of the slant angle without the need for a systematic search \cite{marti2001using} \cite{you2002slant}. These methods usually have a lower computational cost than the previous ones. Gupta and Chanda \cite{gupta2012novel} proposed a second slant angle estimation method in addition to the one described above, which uses the \textit{Fourier transformation}  to convert the image of the binary word into the Fourier spectrum. The repetition of points along a given direction activates the frequency space in its perpendicular direction, and this direction corresponds to the skew angle. To identify the most relevant line orientation in the frequency space, the Hough transformation was used. 
  \index{Fourier transformation}

Other methods of direct estimation of the slant angle are based on the analysis of 8-directional chain code of the thinned image, such as \cite{you2002slant} or \cite{kimura1993improvements}. In the case of work \cite{marti2001using}, the characters contours were used instead of the thinned image.

\subsection{Normalize ascenders and descenders}
  \index{Ascenders and descenders regions}
  \index{Calligraphy zones}

To standardize the ascending and descending regions sizes, it is necessary to identify the baseline and the upperline to delimit the ascenders and descenders zones and scale it in some way. This is actually done in the slope correction step described above. In particular, by means of projection profiles methods \cite{kavallieratou2002skew} \cite{pastor2004projection} or by estimating the baseline and upperline lines \cite{toselli2004integrated}. 

In most cases, the algorithms used for the identification of ascender and descender zones are more or less standard and delimit the zones by lines. However, some authors propose more advanced alternatives. For example, Pastor-Pellicer et al. \cite{pastor2015combined} train a CNN to classify the pixels of the scanned text line into the main body area and into the ascenders and descenders areas. 

Finally, it is worth mentioning the work by Wigington et al. \cite{wigington2017data} that introduces a normalization technique to compensate differences in word size between writers by adjusting the sizes according to the distances between the upper and lower baselines. The published ablation study indicates that this technique provided significant improvements in the model. Although the use of test-time data augmentation and the use of the test lexicon in the decoding (and several different character sets) make it difficult to interpret the results and compare them with other authors.

\section{Data augmentation}
\label{section: Data_augmentation_background}
  \index{Data augmentation}

An additional problem to the high variability of HTR is the fact that existing databases for model development are limited in size. Data augmentation is a technique that tries to mitigate this problem by generating additional transformed data from existing ones. In particular, transformations invariant to the text content present in the image are applied to the input image. This strategy is regularly used in other Computer Vision problems such as image classification or object detection.
  \index{Computer vision}

There are two data augmentation strategies: data augmentation during training and data augmentation in the inference (or evaluation) process, also called test-time data augmentation.

In the first case, for algorithms based on neural networks trained iteratively by means of backpropagation, it is a common practice to introduce the data augmentation process in each iteration. This integration makes possible that, in each epoch of the model training algorithm, a set of input data different from those of the previous epoch is obtained. This allows to perform the training with a theoretically infinite sample of data. This is the usual way of incorporating data augmentation into the HTR problem; for example, see references \cite{dutta2018improving} or \cite{michael2019evaluating}.

Train data augmentation contributes to improving the generalization capacity and accuracy of models. For example, Coquenet et al. \cite{coquenet2019have} report an improvement of between 0.94\% and 1.3\% in Character Error Rate (CER) using IAM and RIMES dataset. Word recognition using data augmentation consisting of: contrast modification, sign flipping, long and short scale modifications, and width and height dilations.

Although data augmentation is usually introduced in the training as a first transformation step prefixed at the beginning of each iteration, in some cases the data augmentation is configured as another layer of the model with trainable parameters. For example, Krishnan et al. \cite{krishnan2018word} introduced an initial \textit{Spatial Transformer Network} (STN) \cite{jaderberg2015spatial} layer configured to perform a data-driven augmentation. So that the parameters of this layer are trained together with the network parameters to adapt the data augmentation parameterization to the training data epoch by epoch, like the rest of the model components.
  \index{Spatial transformer network}

Another relevant contribution is presented by Wigington et al. in \cite{wigington2017data}. These authors introduce a new data augmentation technique by applying grid-based local distortions. This technique is based on one previously used by Simard et al. \cite{simard2003best} to construct random elastic distortion for single character images. It is performed by applying random distortion on a grid of control points defined for each image and aligned on the baseline of the text. In \cite{wigington2017data}, this technique is used both dynamically in the training process and in the evaluation process (test-time data augmentation). 

Finally, it is worth noting that another technique improving performance is \textit{sign flipping} \cite{cubuk2019autoaugment}, motivated by the fact that the text content is invariant with color or texture. Text and background color inversion is a simple transformation, which works \cite{yousef2020accurate} and \cite{cubuk2019autoaugment} have pointed out that improve generalization, even if all the training and test data have the same text and background colors.

In the case of test-time data augmentation, the process of generating new modified images occurs at inference time, so in order to evaluate a handwritten text image, multiple variations of the image are evaluated and the result is combined to obtain the final evaluation. For example, Poznanski et al. \cite{poznanski2016CVPR} propose rotations and shear transformations in predefined angles to be used in both train and test time. Wigington et al. \cite{wigington2017data} use a random distortion on a predefined grid that apply local distortions over the original image and it is applied both in train and test time. Finally, Dutta et al. \cite{dutta2018improving} applied affine transformations, elastic distorsions and multi-scale transformations.

Test-time data augmentation is not a common practice in HTR research and, as discussed in Section \ref{section:On the difficulties of comparing results}, it generates many difficulties when comparing results from different authors.

\section{Classical approaches based on HMM}
\label{section:Classical approaches based on HMM}
  \index{Hidden Markov model}
  \index{Automatic speech recognition}

Continuous HTR problem, where no information is available to segment the text into its constituent characters, has been studied for quite some time \cite{plamondon2000online}. Initially, the predominant approach to the problem was the use of \textit{Hidden Markov Models} (HMM) as a technique that allowed the decoding of variable-length transcribed text. This same technique had previously been used successfully in the speech recognition problem \cite{gales1998maximum}, which has significant similarities with the continuous HTR problem.

In recent years, the use of HMM for decoding has fallen into disuse in the face of other alternatives based on Neural Networks. However, understanding how it works and how it has been applied to the HTR problem, and especially how it has been combined with models based on Neural Networks, provides a necessary perspective to understand many relevant works on HTR. 

In the next subsection, the main concepts related to HMMs are introduced, and in the Subsection \ref{subsection: Hidden Markov Models in handwriting recognition} their application to continuous HTR systems is detailed.

\subsection{Overview of Hidden Markov Models (HMMs)}
\label{section:Hidden Markov Models intro}

HMMs are a technique that has been widely used over time, especially on the speech recognition problem \cite{trentin2001survey}. It was also used extensively on the continuous HTR problem until the years 2012-2014 \cite{bertolami2008hidden} \cite{dreuw2011hierarchical} \cite{doetsch2014fast}. The technique is well developed and explained in several authors, particularly in Theodore Bluche's Thesis \cite{bluche2015deep} is detailed, both in general and in its particularization for the HTR problem. Therefore, only a general summary of it is included here. 
  \index{Speech recognition}

An HMM represents a stochastic process that models the behavior of a sequence called \textit{observed values} and modeled by a random variable denoted by $X$. The $X$ values are produced by a sequence of \textit{states} modeled by a second random variable denoted by $S$. The sequence of states remains hidden and acts in the background on the $X$ sequence of output observed values \cite{rabiner1989tutorial}.

The sequence $S={s_1, ..., s_n}$ represents a set of states connected to each other by transitions subject to a probability distribution $P$ that models the transitions between states: $P(s_{t+1}|s_t)$. This function is called the \textit{transition function}.

The values of each element of the sequence observed values $X={x_1, ..., x_n}$ could be emitted with a given likelihood based on a probability mass or density function of the observed value based on the associate state $p(x_t|s_t)$. This function is called the \textit{emission function}.

Additionally, an HMM must fulfill two requirements:

\begin{itemize}
    \item The sequence of states $S$ must satisfy the condition that it is a first-order Markov process, in which the probability of a future state depends only on the current state.
        \begin{equation}
        P(s_1,...,s_n) = P(s_1) \prod_{t=1}^{n}P(s_{t+1}|s_t)
        \end{equation}
    \item The value of observed values sequence $X$ at time $t$ depends only on the value of the sequence of states $S$ at the same time $t$.
        \begin{equation}
        P({x_1,..., x_n}|{s_1, ...,s_n}) = \prod_{t=1}^{n}P(x_t|s_t)
        \end{equation}
    
\end{itemize}

To train an HMM we traditionally use the \textit{Maximum Likelihood Estimation} (MLE) \cite{Brox2014} criterion applied over the observed values of the train data. In other words, it is identified the set of parameters that maximizes the probability that the feature vectors generated by the training data produce the reference target sequences. The set of parameters to be optimized includes both the parameters of the transmission function and the emission function.
  \index{Maximum likelihood estimation}
  \nomenclature{MLE}{Maximum Likelihood Estimation}

Since the above optimization problem it is not possible to solve it exactly for complex probabilistic models with hidden variables, a numerical method of finding a local optimum is used. The method used is the \textit{Expectation-Maximization} algorithm \cite{dempster1977maximum}. More particularly, in the case of HMM, we use the variant called \textit{Baum-Welch algorithm} \cite{baum1967inequality} \cite{baum1970maximization}. The most popular HMM toolkits implement these algorithms or variations, such as \textit{Viterbi's algorithm} \cite{young2002htk} \cite{povey2011kaldi}. 
  \index{Expectation-maximization}
  \index{Baum-Welch algorithm}
  \index{Viterbi algorithm}

The following subsection describes the application of HMMs to the HTR problem. These models were very popular for solving the HTR problem because they provided many relevant advantages. In particular, they produced good results for low-volume data samples, and it was not necessary to segment words into characters to model word recognition \cite{doetsch2014fast}. It was also easy to integrate HMMs with language models for application during decoding as a prior probability.

\subsection{Hidden Markov Models in handwriting recognition}
\label{subsection: Hidden Markov Models in handwriting recognition}

Historically, HMMs have been used assiduously in the Automatic Speech Recognition (ASR) problem \cite{trentin2001survey} , which is a problem with many similarities with the continous HTR problem. In both cases it is a matter of obtaining the transcription of a coded text message, either as an audio or as an image. This, together with the fact that efficient training and decoding algorithms are available for their application, has made them very popular in handwriting as well \cite{bercu1993line} \cite{starner1994line} \cite{hu1996hmm} \cite{marti2001using} \cite{schenkel1995line} \cite{el1999hmm} \cite{bertolami2008hidden} \cite{doetsch2014fast}. 
  \index{Automatic speech recognition}
  \nomenclature{ASR}{Automatic Speech Recognition}

In HTR, an output sequence of characters or words with variable length which includes the text transcription in the input image must be obtained. It has to be done from a sequence of input signals extracted from that image that has a length usually different from the length of the output sequence, so it is necessary to perform a decoding step to align the two sequences and build the transcript. HMMs come to solve this alignment and decoding problem.

\begin{figure}[!ht]
\centering
  \includegraphics[width=0.75\textwidth]{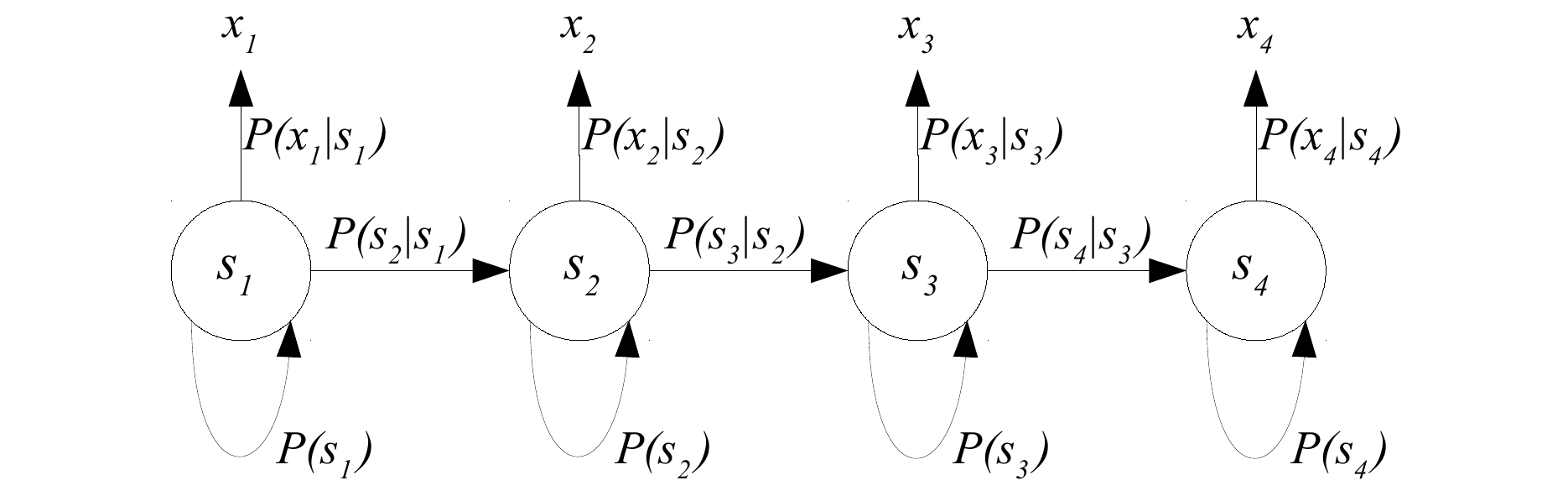}
\caption{Example of typical HMM topology for the HTR problem.}
\label{fig:HMM_schema}
\end{figure}

When an HMM is applied to the HTR problem, some structural decisions are made to reduce the number of model parameters. Usually, a sequential topology with unidirectional left-to-right connections is adopted. In this way, each emitting state has two transitions, the first is a transition to the next stage, and the second is a self-transition over the same state. In Fig. \ref{fig:HMM_schema} we include an example of this topology.

Most of the work on continuous HTR models each character employing an individual HMM. The number of states can be fixed for each character or variable since there are shorter or longer characters requiring more or less states to be modeled correctly \cite{bertolami2008hidden} \cite{zimmermann2006offline}. Regarding the emitting states, Gaussian Mixture Models (GMM) with diagonal covariance matrices as density function to model $P(x_t/s_t)$ are used. A complete analysis of different HMM topology optimization methods applied to the HTR problem is provided by Günter et al. in \cite{gunter2004hmm}.
  \nomenclature{GMM}{Gaussian Mixture Models}
  \index{Gaussian mixture models}

The text image has to be previously transformed into a sequence of feature vectors to apply the HMM model. The most common technique is to construct handcrafted geometrical features using a left-to-right sliding window over the handwriting image \cite{bertolami2008hidden}. Sometimes the Principal Component Analysis (PCA) \cite{dunteman1989principal} technique is applied \nomenclature{PCA}{Principal Component Analysis} to reduce the feature vector dimension, as in \cite{dreuw2011confidence}. 
  \index{Principal component analysis}

It is also possible to construct the sequence of feature vectors from a visual model applied directly to the input image. The most recent works on the application of HMM to the HTR problem employs visual neural network type models, especially RNNs \cite{bluche2014comparison} \cite{zamora2014neural}, but also CNNs \cite{bluche2013tandem}. There are two approaches to combine a discriminative visual model with a generative HMM model, the \textit{hybrid approach} and the \textit{tandem approach}.

The \textit{hybrid approach} was introduced by Bourlard and Morgan in \cite{bourlard2012connectionist} and it tries to model the dependence of a directly constructed HMM using the visual model output, usually a neural network. It is done in two steps: first, the original features provided by the visual model are used to obtain an initial alignment; second, the posterior estimates are used as scaled emission probabilities in the HMM, which is already trained in the previous step.

The \textit{tandem approach} was introduced by Hermansky et al. in \cite{hermansky2000tandem} with the idea of using a neural network visual model output in HMMs. In this case, the output of the visual model are used as a new vector of features, and a completely new HMM is trained using these features. Again, the process is performed in two steps. In the first step, the original features obtained directly from the image are used to build an initial alignment, allowing the training of the visual model. In a second step, the estimations of the visual model are used as a feature stream to train a new \textit{GMM-based} system used in the HMM model.


HMMs were for a long time the best alternative for modeling the continuous HTR problem because of their ability to segment and recognize continuous text simultaneously. However, they had several well-known drawbacks. The main one was that it assumed that the probability of each observation depended only on the current state, which made contextual factors challenging to model. Another problem was that HMMs are a generative model, whereas discriminative models generally perform better in supervised classification problems \cite{graves2009novel}.

In 2009, Graves et al. published the paper \cite{graves2009novel}, where a fully discriminative model was introduced to solve the continuous HTR problem using the CTC algorithm to perform the decoding. This opened the door to full-discriminative models based on neural networks to solve the HTR problem. These models, which are currently used, are described in the following sections.

\section{Neural networks and deep learning architectures}
\label{section:Neural networks and deep learning architectures}

Until the years 2012-2013, the approaches to the problem of recognizing handwritten words were based on building text features from the images that usually required domain expertise, and after that modeling them through Hidden Markov Models (HMM)\nomenclature{HMM}{Hidden Markov Models}, as it is described in the Section \ref{section:Classical approaches based on HMM}. One improvement over the previous approach was the introduction of a Neural Network (NN)\nomenclature{NN}{Neural Network} \cite{mcculloch1943logical} that improves the HMM capacity to decode the correct sequence of characters. Neural networks are used to extract relevant characteristics of the image or to transform them as a previous step to the use of HMM to align and decode the transcriptions. This usually improves the HMM capacity to decode the correct sequence of characters.

More recently, deep neural networks \cite{krizhevsky2012imagenet}, which learn a multilayer hierarchy of features, have been the type of model massively selected to analyze unstructured data such as images or texts. These models achieved significant improvements in several complex problems such as image recognition \cite{krizhevsky2012imagenet} or speech recognition \cite{chorowski2015attention}. The case of handwriting recognition is not an exception and several deep architectures has been proposed for the word and line recognition problems with significant improvements in the results, as shown in \cite{graves2009offline}, \cite{bluche2015deep} or \cite{puigcerver2017multidimensional}. 

The deep neural networks became more popular due to their success in the problems of Computer Vision \cite{krizhevsky2012imagenet} and fueled by the increased computing capacity of the computers. Their high popularity is also linked to the appearance of software frameworks which allow the efficient and simple implementation of the neural models with a high number of parameters and multiple layers.

Since then, neural networks have become the most commonly used methods to address the problem of offline handwriting recognition. Different authors are proposed multiple architectures and use different types of networks to handle this problem.

In the next sections, we describe in detail the use of this different neural networks architectures to solve the problem of handwriting recognition. But first we include two subsections to introduce the historical background and introduce the most relevant concepts of neural networks applied to the HTR problem.

\subsection{Historical background}

This subsection includes an introduction to neural networks, starting with a brief summary of its history. Further details about this history can be found in \cite{kurenkov2020briefhistory}

The first appearance of neural networks as a mathematical model, which mimics the functioning of neurons in the brain, was presented by McCulloch and Pitts \cite{mcculloch1943logical} in 1947. These authors introduced a model of simplified neurons as a circuit that can perform computational tasks. Neural networks begin to gain interest until 1969 when Minsky and Papert, in their book Perceptrons \cite{minsky1969perceptrons}, expose several problems and limitations of such models. 

It was not until the 1980s, with the application of the \textit{backpropagation} technique for their training, that neural networks again gained the attention of the scientific community. The backpropagation technique was proposed by several authors in the 1960s and in 1974. Paul Werbos proposed its use in neural networks in his Thesis \cite{werbos1974beyond}. But it did not become popular until the publication in 1986 by David Rumelhart, Geoffrey Hinton and Ronald Williams of two papers: "Learning representations by back-propagating errors" \cite{rumelhart1986learning} that clearly details its practical implementation as it is known today, and "Learning internal representations by error propagation" \cite{rumelhart1985learning} that addresses the problems evidenced by Minsky almost 20 years earlier and shows how neural networks can be successfully trained to solve complex problems. Only a few years later, and by applying the techniques proposed in these two papers, one of the first practical applications of neural networks appeared, precisely for the problem of reading handwritten text \cite{lecun1989backpropagation}. 

Neural networks were once again a topic of interest for the scientific community thanks to the application of backpropagation for their training and to the publication in 1989 of the paper "Multilayer feedforward networks are universal approximators" \cite{hornik1989multilayer}. In this work, it was mathematically demonstrated that a multilayer neural network can theoretically implement any function, which eliminated the main problem that Minsky had raised to them twenty years earlier.

But soon arises a problem, precisely related to backpropagation, that makes impossible to train multilayer neural networks with several layers. In short, the fact that when the backpropagation algorithm distributes the responsability for the error among the layers of the network and it assigns gradients to try to reduce the error in the next iteration. These gradients either decay to zero exponentially with the number of layers or they explode. This is known as the 'vanishing or exploding gradient problem' \cite{279181}. Additionally, in 1992, a new algorithm called \textit{Support Vector Machines} (SVM) \nomenclature{SVM}{Support Vector Machines} \cite{boser1992training} appeared. In a very basic way, it can be described as a mathematically optimal way of training an equivalent of a two-layer neural network. The SVM begins to show itself as superior to the difficulty of working with neural networks, as indicated by the first comparisons precisely in the handwritten text recognition problem \cite{lecun1995comparison}.  
  \index{Support vector machines}

The interest in neural networks declined again, until the publication in 2006 of the paper: 'A fast learning algorithm for deep belief nets' \cite{hinton2006fast}. It shows that it is possible to correctly train the multilayer neural networks provided when their weights are initialized intelligently rather than randomly and proposes an initialization procedure by means of unsupervised layer-by-layer learning. This paper marks a turning point in the interest in neural networks, which is now being rekindled. From this moment until today, there is an escalation in the application of neural networks to a multitude of problems, as well as in the number of layers and in their accuracy. All this fed by the increase in both the size of the datasets and the computational capacity of computers. 

Nowadays, neural networks completely dominate a multitude of research areas and in both academic and business environments it is one of the most important topics of intense research \cite{SCHMIDHUBER201585} \cite{Goodfellow-et-al-2016}.


\subsection{Multilayer perceptrons}
  \index{Multilayer perceptron}
  \nomenclature{MLP}{Multilayer Perceptron}

\begin{figure}[!ht]
\centering
  \includegraphics[width=0.75\textwidth]{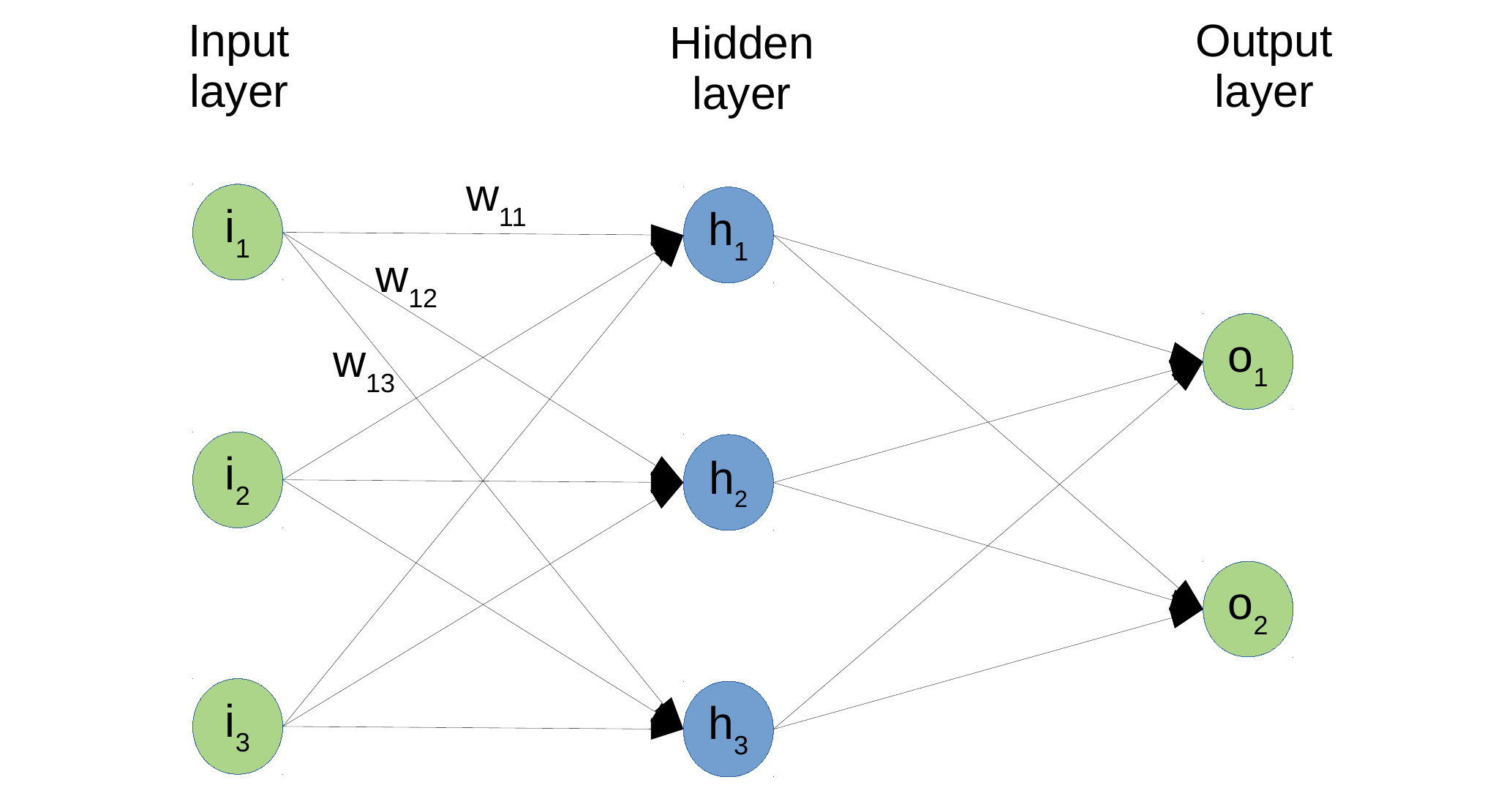}
\caption{Feedforward neural network basic architecture.}
\label{fig:neural_network_schema}
\end{figure}

An \textit{Artificial Neural Network} (ANN)\nomenclature{ANN}{Artificial Neural Network} is a computing system based on a collection of connected units or nodes organized in layers. The connections between nodes are called edges, and they are directed and have an associated parameter called weight, which modulates the signal as it passes through each edge. There are multiple network architectures, oriented both to the construction of supervised and unsupervised models. In this section we focus on the so-called feedforward neural networks or Multilayer Perceptrons (MLP). For these cases, the network receives an input signal $x$ through a set of initial nodes called input layer and it produces an output signal $y$ through other nodes called output layer. The intermediate layers, through which the signal flows and is transformed, are called hidden layers $h$. Fig. \ref{fig:neural_network_schema} shows a basic neural network scheme with the above components. These models are called feedforward because the input information $x$ follows a single forward path until it is transformed into the output information $y$. When feedback connections are added to such a network in which the output is fed back into itself then a particular type of network called \textit{Recurrent Neural Networks} (RNN)\nomenclature{RNN}{Recurrent Neural Networks} is obtained.

\begin{figure}[!ht]
\centering
  \includegraphics[width=0.6\textwidth]{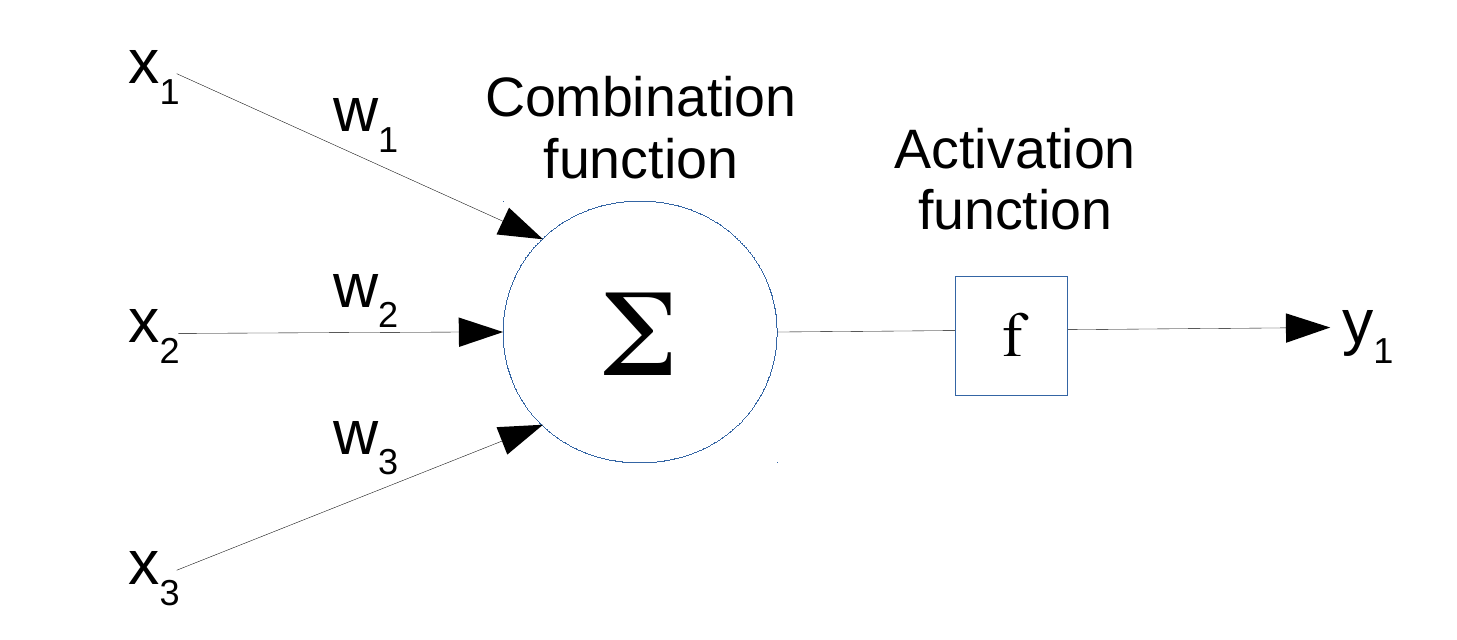}
\caption{Neural network node detail.}
\label{fig:neural_network_node}
\end{figure}

Fig.\ref{fig:neural_network_node} shows a detail of the operation of a particular node of the network. Each node receives multiple input signals $x_i$ through its incoming edges, modulated by their respective weigths $w_i$. It merges them by means of a combination linear function, which is usually the sum, and generates an output signal $y$ which is a real number. This output is modulated by an non-linear activation function before being sent to the next nodes in the network. This activation is a monotonic function that usually limits the range of variation of the output signal.

Formally, the input signals $x_1,...,x_n$ coming from the previous layer are multiplied by their respective weights $w_1,...,w_n$ and combined at the node, in this case with the sum function. To the previous result a bias $b$ is added and to this the activation function $f$ is applied as indicated in the equation \ref{eq:neura_network_node}.

\begin{equation}
y = f(b + \sum_{i=1}^{n}w_i x_i)
\label{eq:neura_network_node}
\end{equation}

The combination function is usually standard, calculated as given in the equation \ref{eq:neura_network_node} as a linear combination of the inputs $x$ multiplied by the weights $w$. The activation function must be monotonic. Traditionally the most common choices were the hyperbolic tangent function, which rescaled the output in the interval $(-1,1)$ or the logistic sigmoid which rescaled it in $(0,1)$. But with the advent of deep architectures, the preferred activation function is the Rectified Linear Unit (ReLU) \cite{jarrett2009best} calculated as: $f(x)=max{(0,x)}$.

The goal of a neural network is to model the relationship between the input data $x$ and the output data $y$ by means of some function $f$ that approximates it. This function depends on the structure of the network that has been defined and a set of parameters $theta$ that characterize it, which are basically the weights $w$ of the edges of the network. To make the adjustment, the neural network defines a mapping $y=f(x,\theta)$ and through a process called \textit{training}, it learns the values of the $\theta$ parameters that best approximate the function $f$ to the data. To do this, a learning algorithm is employed that uses the experience provided by the known data $(x,y)$ to iteratively search for the value of the $\theta$ parameters that optimize the fit of $f$ to these data.

The common learning algorithm used is \textit{backpropagation}. This algorithm, based on gradient descent, allows to optimize the parameters of a network through an iterative process that seeks to minimize the error of the network by representing a set of known data $(x,y)$, which are called \textit{training data}. More formally, the backpropagation algorithm is defined by three elements:

\begin{itemize}
 \item A neural network $\hat{y}=f(x,\theta)$, where $\theta$ represents the set of parameters of the neural network.
  \item A data set $X={(x_1, y_1), ..., (x_n,y_n)}$ where $x_i$ is the input and $y_i$ is the desired output of the network with $x_i$ input.
  \item An error function $E(y, \hat{y})$ that measures the error between the desired output $y$ and the calculated output $\hat{y}$ for a given value of the parameters $\theta$.
\end{itemize}

At each iteration $t$, the algorithm performs two steps:

\begin{itemize}
  \item A first step called \textit{forward} in which, given the value of the current parameters $\theta^t$, the estimated value of $\hat{y}$ is calculated as $\hat{y} = f(x,\theta^t)$. And then the error $E(y,\hat{y})$ is calculated as a function of the difference between the calculated values $\hat{y}$ and the expected values $y$.

  \item In a second step called \textit{backward}, the $\theta$ parameters are updated as a function of the gradient of the error function $E$ with respect to the $\theta$ parameters as shown in the next equation:
\end{itemize}

\begin{equation}
\theta^{t+1} = \theta^t - \lambda \frac{\partial E(X,\theta^t)}{\partial \theta}
\label{eq:neura_network_node2}
\end{equation}

where $lambda$ is a training parameter called \textit{learning rate} that determines the step size at each iteration of the algorithm and allows to control how quickly the model adapts to the training data.

In the HTR problem, the MLP networks have been mostly used as part of the solutions based on the Hidden Markov Models. In certain cases, they have been used to predict the HMM states within a hybrid approach like \cite{espana2010improving}. In other cases, they have been applied to extract features which can be used later in models based on GMM and HMM, following the tandem approach like the one included in \cite{dreuw2011hierarchical}.
  \index{Hidden Markov model}

In the hybrid approach \cite{bourlard2012connectionist}, a NN replace the traditional Gaussian Mixture Models (GMM) as emission models for the HMMs. This approach has been implemented with diverse neural networks. For example, \cite{dreuw2011hierarchical} and \cite{espana2010improving} use MLP, while \cite{bengio1994globally} and \cite{bluche2013tandem} use convolutional networks. In the tandem approach \cite{hermansky2000tandem} the NN and the HMM are combined using the NN as a features extractor for the GMM-HMM system. Similarly to the previous case, the network is also trained to predict the HMM states or characters, although the probabilities are not further directly used in the visual model, the network works as a feature extractor. For example, the work \cite{kozielski2013open} uses the activations of a \textit{Long Short Term Memory} (LSTM) network, described in the Subsection \ref{LSTM}, as the features to build a GMM-HMM model. Considering that the activations of a NN are highly correlated to themselves, a transformation is usually applied to eliminate such correlation, frequently Principal Component Analysis (PCA) \cite{dunteman1989principal} or Linear Discriminant Analysis (LDA) \cite{mclachlan2004discriminant}. 
  \index{Gaussian mixture models}
  \index{Principal component analysis}
  \index{Linear discriminant analysis}
  \nomenclature{LDA}{Linear Discriminat Analysis}

The wide variety of alternatives to use the outputs of the NNs is a great advantage of the tandem approach. However, the main inconvenient lays on the fact that it requires a complete training of the HMM after the outputs of the NN are extracted. Besides, the non-correlation and the reduction of the characteristics introduce additional parameters which have to be manually tuned. Nonetheless, the comparative studies indicate that this strategy is superior to the hybrid approach \cite{dreuw2011hierarchical}.

In the next sections, it will be shown that the construction of HTR systems exclusively based on the NN will not use MLP networks. They are based on the \textit{Recurrent Neural Networks} as well as on \textit{Convolutional Neural Networks} \nomenclature{CNN}{Convolutional Neural Network}.

\subsection{Recurrent Neural Networks}
\label{section:Recurrent Neural Networks}

\begin{figure}[!ht]
\centering
  \includegraphics[width=0.75\textwidth]{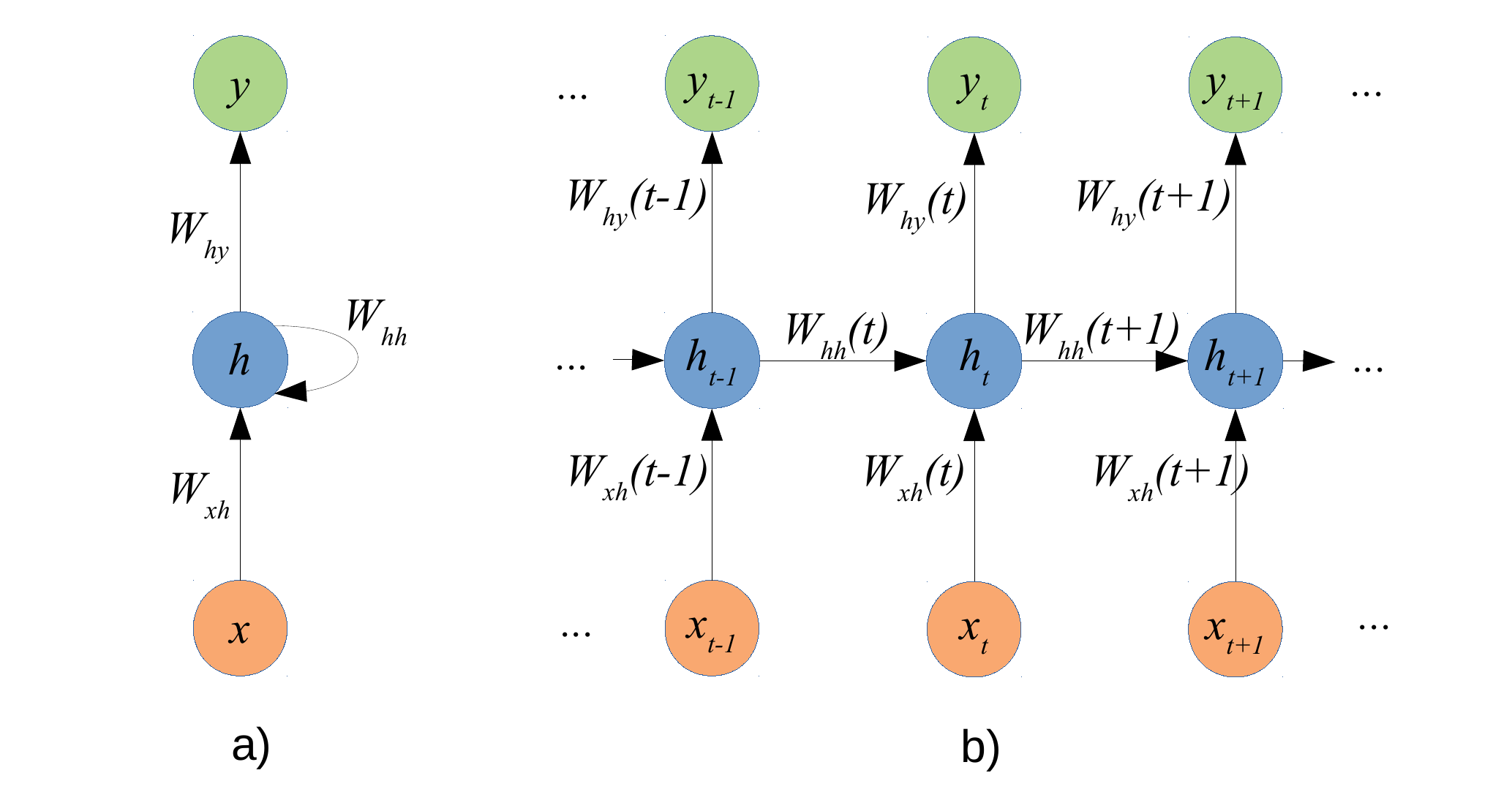}
\caption{RNN schema: a) compact, and b) unfold-over-time.}
\label{fig:RNN_basics}
\end{figure}

\textit{Recurrent Neural Networks} \cite{rumelhart1986learning} are one type of networks aimed at sequential data processing ${x_1, ...x_n}$. These networks can model the sequential relationships in data when they are relevant for the goal of the training. For this, they rely on the idea of sharing parameters between the different nodes of the recurrent layer while the input data sequence is processed. Thus, in a RNN each output element $h_t$ is function of the previous output element $h_{t-1}$ and, based on this recurrent definition, is a function of the input elements ${x_1,...,x_t}$ . Fig. \ref{fig:RNN_basics} shows the scheme of a basic RNN.

According to the notation of Fig. \ref{fig:RNN_basics} the equations of a basic RNN are defined as follows:

\begin{equation}
\begin{array}{lcl} 
h_{t} =\tanh(W_{hh}h_{t-1} + W_{xh}x_t) \\
y_{t} = W_{hy}h_t 
\end{array}
\label{eq:RNN}
\end{equation}

These equations describe a basic strategy in order to apply the recurrence and the activation \textit{tanh} in a recurrent connection. This model is the most common one, although the recurrent networks accept multiple variations, and some of them are introduced in the next sections.

The training of a RNN can be done with a modified version of the backpropagation algorithm, named \textit{Back-Propagation Through Time} (BPTT), which is an optimization of the backpropagation applied on a RNN unfolded. The BPTT includes a truncation strategy in the calculation of the gradients which ensures the numeric stability and, for efficient computation, intermediate values are cached. 
  \index{Back-propagation through time}
  \nomenclature{BPTT}{Back-Propagation Through Time}

The basic recurrent network described previously has a causal structure because in time $t$ only information from the past $x_1, ..., x_{t-1}$ and the current time $x_t$ are used. However, sometimes the prediction $y_t$ has to depend on the whole context, both previous and after the moment $t$. For example, in the HTR offline problem, the identification of a specific character in the center of the text can benefit from analyzing both the previous and subsequent characters.

\begin{figure}[!ht]
\centering
  \includegraphics[width=0.75\textwidth]{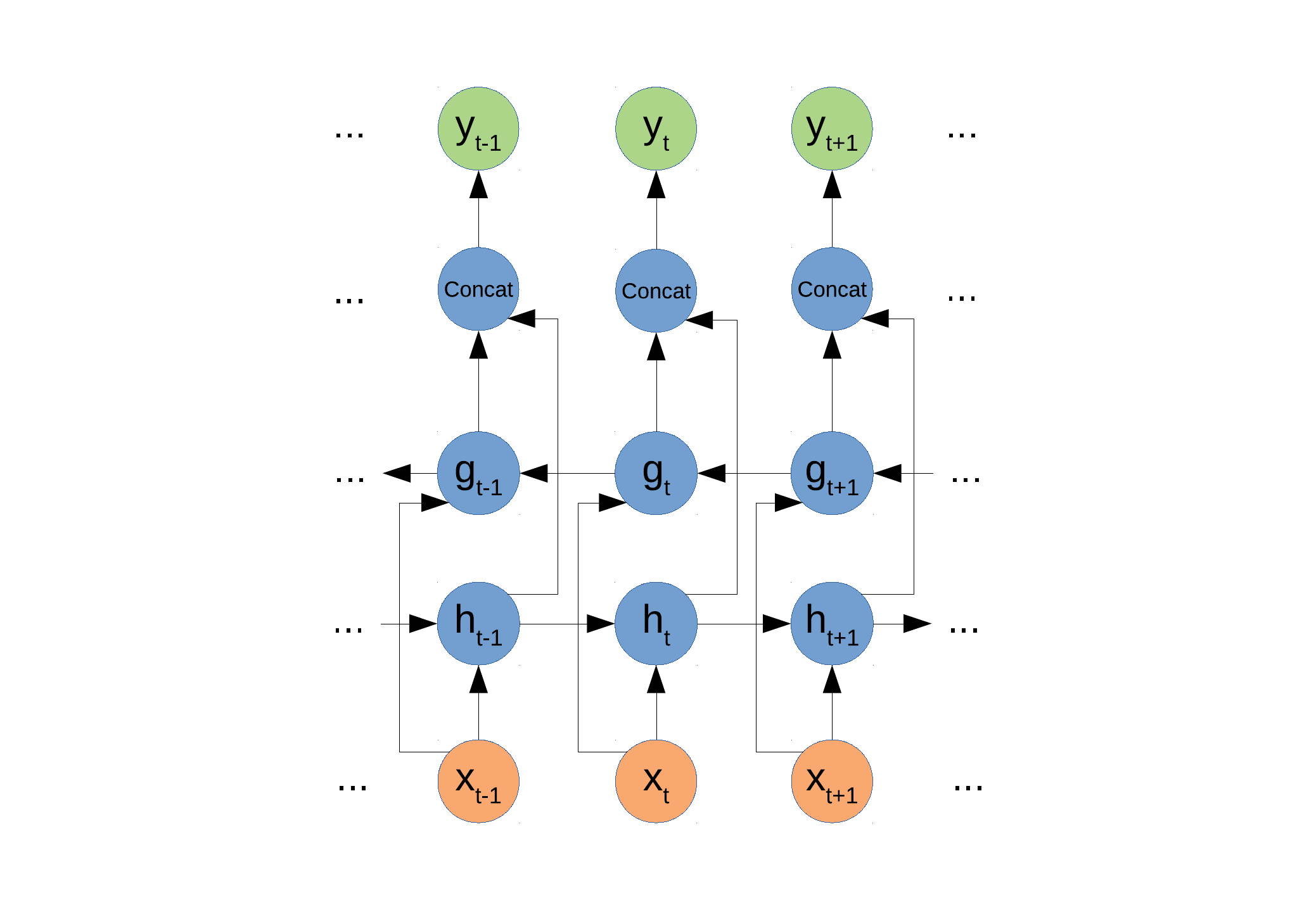}
\caption{Bidirectional RNN schema.}
\label{fig:Bidirectional_RNN_schema}
\end{figure}

In order to include the whole text in the model \textit{Bidirectional Recurrent Neural Networks} (BRNN) \cite{schuster1997bidirectional} are used. They have been successfully applied to the HTR problem during a long time \cite{graves2008unconstrained} \cite{graves2009offline}. These networks blend one RNN executed in an increased direction of $t$. From the beginning of the sequence, with other RNN executed in a decreasing direction of $t$, from the end of the sequence towards its beginning. Fig. \ref{fig:Bidirectional_RNN_schema} shows a typical scheme of a BRNN where $h_t$ is the RNN state that moves forwards through time and $g_t$ is the RNN state that moves backwards through time. The output sequences of both RNN are combined at each step using an aggregation function, being concatenation the most common one, although the sum or the mean applied elementwise are also occasionally used.
  \index{Bidirectional recurrent neural network}
  \nomenclature{BRNN}{Bidirectional Recurrent Neural Networks}

This idea can be naturally extended to two dimensions or more, as it happens for images, generating the \textit{Multidimensional Recurrent Neural Network} (MDRNN) architecture, which are discussed in Subsection \ref{section:Multidimensional Recurrent neural netwoks} and which have been frequently use for the HTR problem, as mentioned in \cite{graves2009offline} or \cite{pham2014dropout}.
  \index{Multidimensional recurrent neural network}
  \nomenclature{MDRNN}{Multidimensional Recurrent Neural Network}

\begin{figure}[!ht]
\centering
  \includegraphics[width=0.75\textwidth]{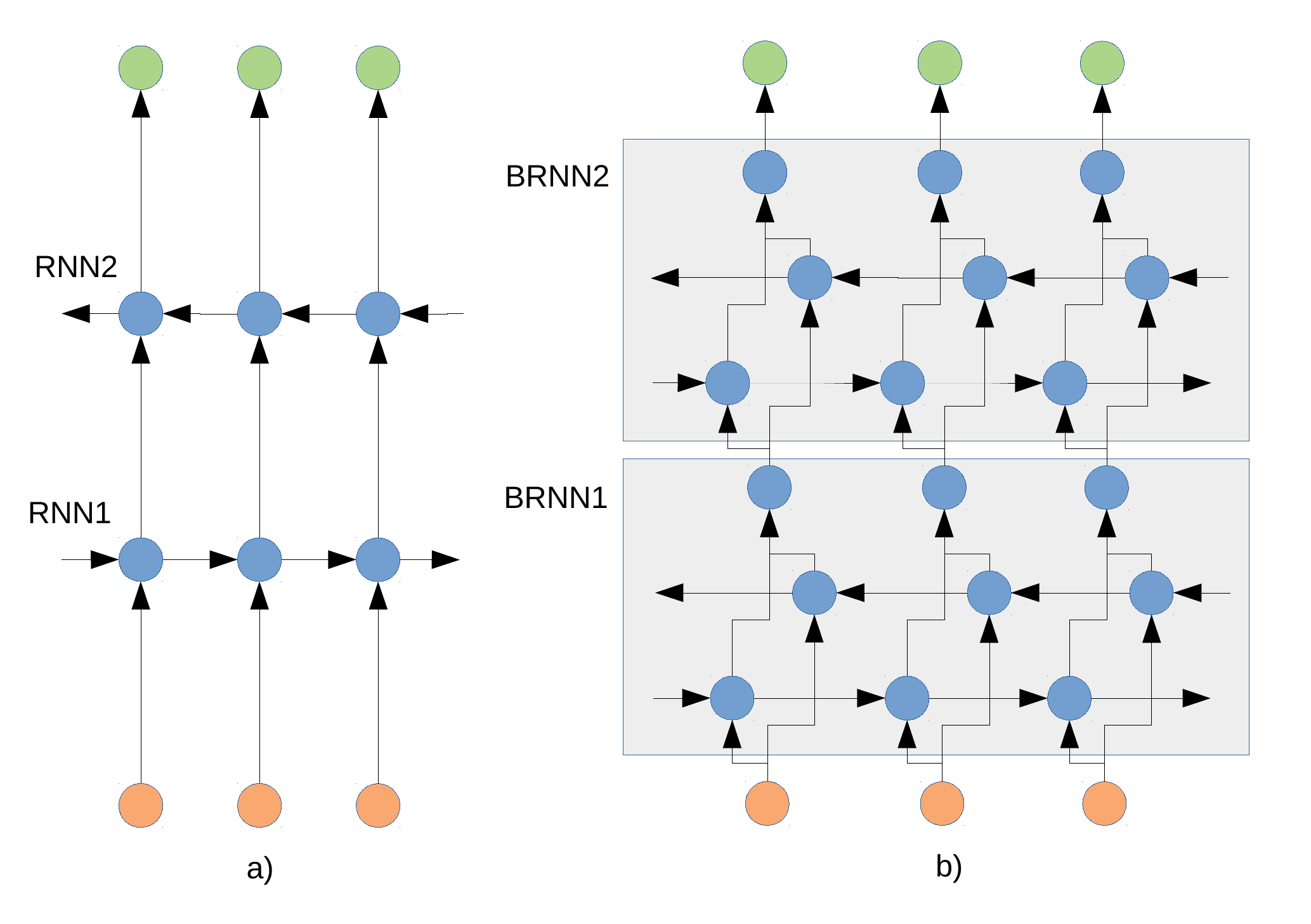}
\caption{Multi-layer RNN schemas: a) Stacked RNN. b) Bidirectional stacked RNN.}
\label{fig:Stacked_RNN_schema}
\end{figure}

Similarly to what it happens with the MLP networks, where multiple dense layers are connected to create a multi-layer architecture, it is also common to connect multiple recurrent layers so the layer output sequence is passed as an input sequence to the next layer in the stacked RNN architecture, shown in the left side of Fig. \ref{fig:Stacked_RNN_schema}.

The sequential combination of multiple Bidirectional RNN interconnected, is also common and it generates a Bidirectional stacked RNN architecture as the one shown in the right side of Fig. \ref{fig:Stacked_RNN_schema}. In Chapter \ref{chapter_word_models} of this Thesis we explore the application of these previous architectures to the HTR problem.

In this background chapter, basic RNNs with a vector $h$ as a recurrent unit have been explained. However, previous studies \cite{279181} confirm that these types of networks have a vanishing gradient problem and gradients propagated through many layers tend to vanish. This makes it impossible for the network to model long dependencies along time. The most common solution to this problem is the use of gated RNNs as these networks are based on creating paths through time which avoid the problem of vanishing gradient. The most commonly used ones are the \textit{Long Short-Term Memory} (LSTM) networks and the \textit{Gated Recurrent Units} (GRU) networks. Several experiments with these two types of gated RNN to the HTR problem were performed in Chapter \ref{chapter_word_models} for specific seq2seq models.
  \nomenclature{LSTM}{Long Short-Term Memory}

\subsubsection{LSTM}
\label{LSTM}
  \index{Long short-term memory network}

\begin{figure}[!ht]
\centering
  \includegraphics[width=0.75\textwidth]{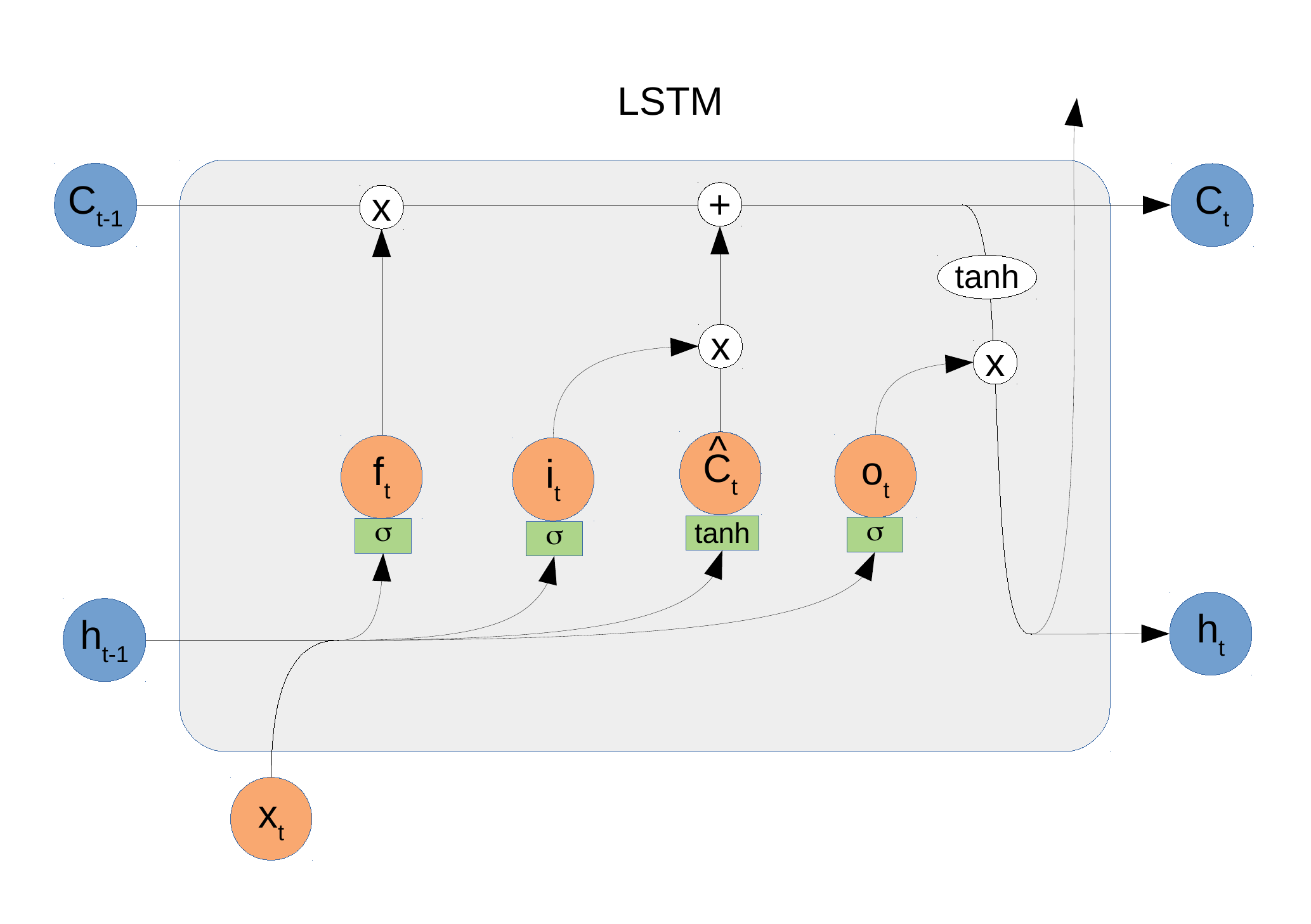}
\caption{Long Short-Term Memory (LSTM) cell.}
\label{fig:lstm}
\end{figure}

\textit{Long Short-Term Memory} (LSTM) networks are a special type of Recurrent Neural Networks (RNN) which are able to learn long-term dependencies. LSTM networks were introduced by Hochreiter and Schmidhuber \cite{hochreiter1997long} and include two ways to translate the previous information across the net: the output vector $h$ and the state vector $c$, that combined using three gates, are explicitly designed to store and propagate long-term dependencies. Fig. \ref{fig:lstm}, includes a LSTM cell schema. 
\nomenclature{LSTM}{Long Short-Term Memory}

The gate $i$, named the input gate, learns which values will be updated in the state vector. The gate $f$, named the forget gate, learns the information from the previous state that can be thrown away. With the output of these two layers the network creates a vector of new candidate state values $c$. Finally, the gate $o$, named the output gate, learns what information will go to the output $h$. The following equations model in detail the LSTM gate.

\begin{equation}
\begin{array}{lcl} 
f_{t} = \sigma( W_f[h_{t-1}, x_t] + b_f) \\ 
i_{t} = \sigma( W_i[h_{t-1}, x_t] + b_i) \\ 
o_{t} = \sigma( W_o[h_{t-1}, x_t] + b_o) \\ 
\hat{C}_{t} = \tanh( W_c[h_{t-1}, x_t] + b_c) \\ 
C_{t} = f_t \odot C_{t-1} + i_t \odot \hat{C}_t \\ 
h_{t} = o_t \odot \tanh(C_t)
\end{array}
\end{equation}

where $W$ and $b$ are trainable parameters, $\sigma$ is the sigmoid function, $x$ is the input sequence, $i$ is the input gate, $f$ is the forget gate, $o$ is the output gate, $C$ is the state vector and $h$ is the output vector and $\odot$ denotes the element-wise product.

\subsubsection{GRU}
\label{GRU}
  \index{Gated recurrent unit}

The \textit{Gated Recurrent Units} (GRU) \nomenclature{GRU}{Gated Recurrent Units} \cite{cho-etal-2014-learning} is an alternative gated RNN to the LSTM architecture with less parameters. There is no clear consensus to determine the best type of architecture. Depending on the problem, sometimes the performance of the LSTM can be superior, whereas in other scenarios the GRU offers better results \cite{Greff2017LSTM}. However, in general, it is necessary to experiment both for each specific problem. As exceptions, some authors as \cite{Junyoung2014Empirical} and \cite{10.3389/frai.2020.00040}, indicate that GRUs exhibit better performance that LSTM networks on smaller datasets. In reference \cite{cho-etal-2014-learning} it is pointed out that the GRU have been demonstrated to be more efficient that LSTM networks for manage long-distance temporal correlations in different applications. 

\begin{figure}[!ht]
\centering
  \includegraphics[width=0.75\textwidth]{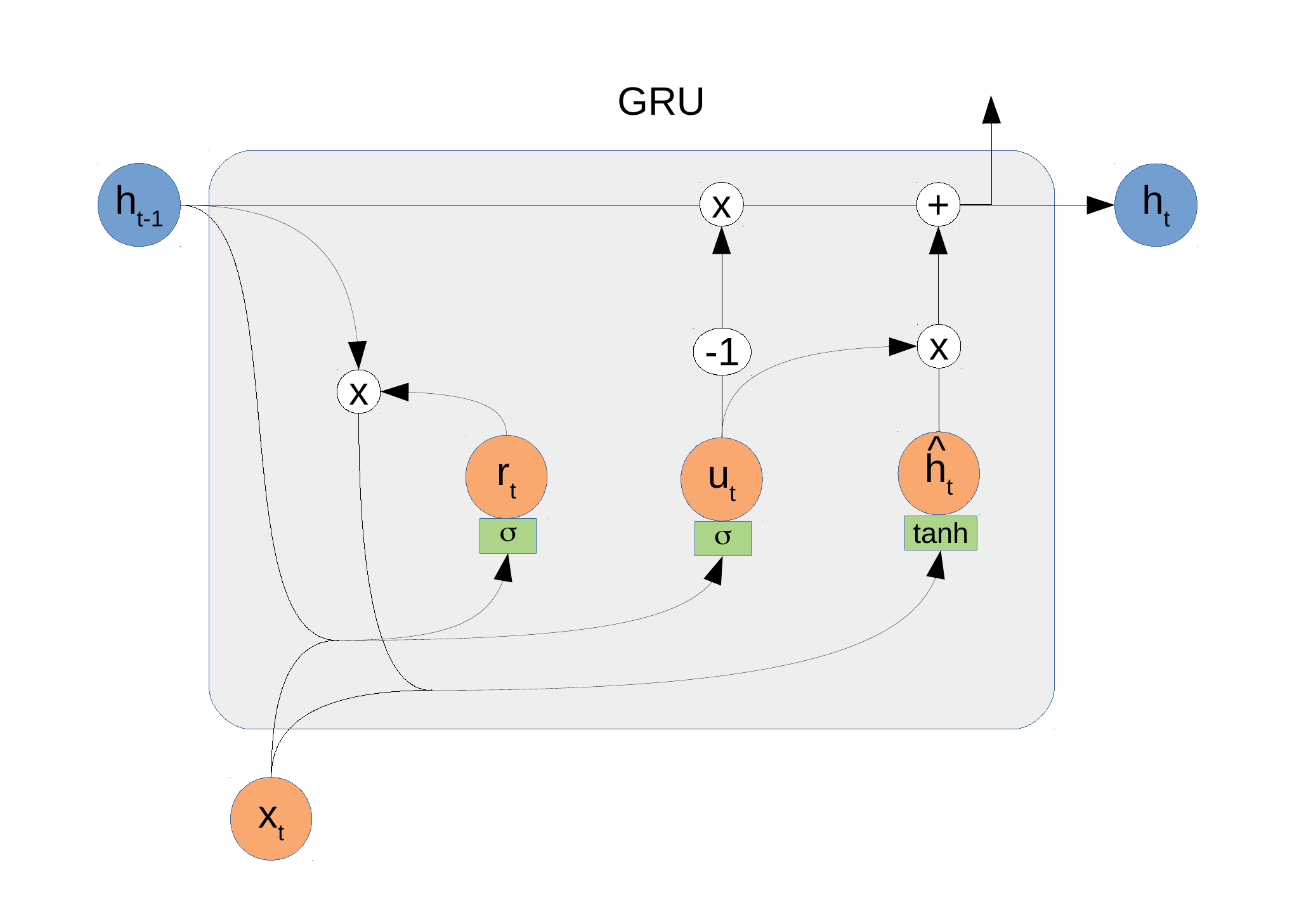}
\caption{Gated Recurrent Unit (GRU) cell.}
\label{fig:gru_schema}
\end{figure}

The main difference between a GRU network and a LSTM network is that in the GRU only one gated simultaneously controls the forgetting factor and the decision to update the state unit. Fig. \ref{fig:gru_schema} shows a scheme of one GRU network which uses two gates. The reset gate $r$ which controls the parts of the current state will be used to calculate the new state and the update gate $u$ which controls what information will be moved to the output $h$.

For the notation of Fig. \ref{fig:gru_schema}, the equations for the forward step are the following ones:

\begin{equation}
\begin{array}{lcl} 
u_{t} = \sigma( W_u[h_{t-1}, x_t] + b_u) \\ 
r_{t} = \sigma( W_r[h_{t-1}, x_t] + b_r) \\ 
\hat{h}_{t} = \tanh( W_h[r_t \odot h_{t-1}, x_t] + b_h) \\ 
h_{t} = (1-u_t) \odot h_{t-1} + u_t \odot \hat{h}_t
\end{array}
\end{equation}

where $W$ and $b$ are trainable parameters, $\sigma$ is the sigmoid function, $x$ is the input sequence, $r$ is the reset gate, $u$ is the update gate and $h$ is the output vector and $\odot$ denotes the element-wise product.

\subsubsection{RNN in handwriting}

The initial use of the neuronal networks in the HTR problem focused on combining them with HMM in the hybrid approach \cite{bengio1999markovian}. In the specific case of the RNN, multiple combined models with HMM have been proposed, for example \cite{senior1993line}, \cite{senior1998off} or \cite{schenk2006novel}.

The use of models with only RNN was restricted to the recognition of characters \cite{bourbakis1995handwriting}, mostly because the target function needs one training signal for each element of the entrance sequence, and it implies that data has to be pre-segmented. To our best knowledge, the work by Graves et al. \cite{graves2009novel} is the first paper where one full RNN architecture was proposed for the HTR problem. In order to solve the previous limitation, these authors used one \textit{Connectionist Temporal Classification} (CTC)\nomenclature{CTC}{Connectionist Temporal Classification} layer as an decoder. This layer performs the alignment between the output of the model and the target sequence of characters. CTC only imposes that the input sequence must have a larger or equal length to the output sequence. The proposed model is a one-layer Bidirectional LSTM with 100 memory cells. From that work, these RNN are going to be present in one way or another in most of the publications related to the HTR problem.

The popularity of RNN approaches to the HTR increased with the help of the CTC. Initially, the most common approach used was Multidimensional RNN, especially the MDLSTM models (which are discussed in detail in Subsection \ref{section:Multidimensional Recurrent neural netwoks}). MDLSTM seemed better suited to the bi-dimensional nature of the input data, that is the images of the handwritten texts.

In 2017, several publications such as \cite{bluche2017gated} and \cite{puigcerver2017multidimensional} criticized the high computational cost of the MDLSTM models. These publications proposed alternative models based on the CNN-RNN-CTC architecture which are discussed in depth in Subsection \ref{section:The CNN - LSTM - CTC Architecture}. In these models, the RNN component is usually composed by several bidirectional layers stacked (like in \cite{puigcerver2017multidimensional} or \cite{dutta2018improving}).

RNNs are also the core of current approaches to the HTR problem, based on sequence-to-sequence (seq2seq) models. In these models, both the encoder and the decoder include several types of RNNs. For example, Bluche et al. \cite{bluche2016scan} used MDLSTM, Sueiras et al. \cite{sueiras2018offline} used Bidirectional LSTM and Kang et al. \cite{kang2018convolve} employed GRU layers.

In recent years new approaches without using RNNs in their architectures have emerged. For example, Kang et al. \cite{kang2020pay} used transformer layers \cite{NIPS2017_3f5ee243} to build the HTR model. Coquenet et al. \cite{Coquenet2020} have proposed a model which uses only Convolutional Neural Networks (CNN) and the CTC as a decoding layer. Convolutional Neural Networks (CNN) and their detailed application to the HTR problem are discussed in the next section.

\subsection{Convolutional Neural Networks}
\label{section:Convolutional Neural Networks}
  \index{Convolutional neural network}

\textit{Convolutional Neural Networks} (CNN) \cite{lecun1989backpropagation} are a type of neural network specialized in modelling the information with a grid-like structure, such as images considered a two-dimensional pixel grid. These networks use the convolutional mathematical operation ($*$) instead of the standard matrix multiplication to calculate the network output. 

\begin{figure}[!ht]
\centering
\includegraphics[width=0.75\textwidth]{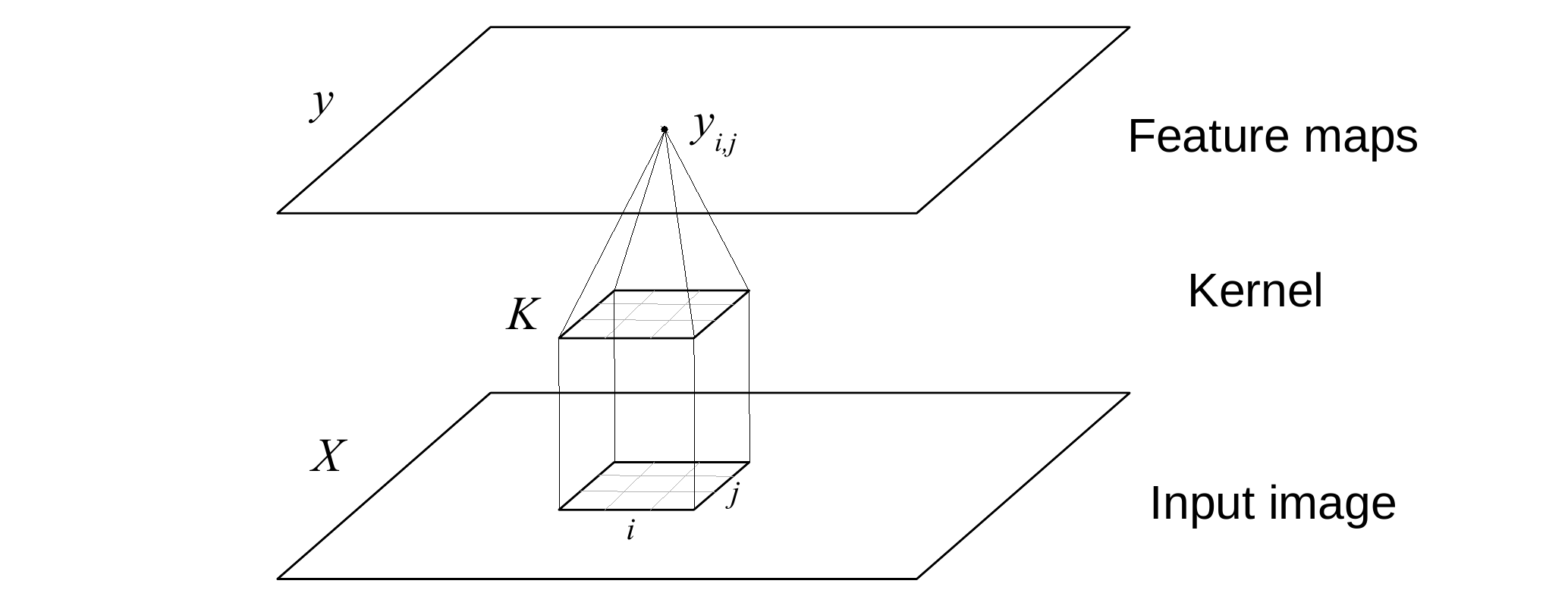}
\caption{CNN schema for a kernel of 3$\times$3 and one feature map.}
\label{fig:cnn_schema}
\end{figure}

One convolutional layer transforms an input array $X$ into an output $y$ commonly known as feature map, by convolutioning the input with a array of parameters $K$ known as kernel. This operation is performed locally so that the output $y_{i,j}$ is calculated based on the neighborhood of $x_{i,j}$ with the kernel convolutional operation. Fig. \ref{fig:cnn_schema} illustrates this operation with a two dimensional input and a 3$\times$3 dimension kernel which generates an output with a single feature map. 

On the notation of Fig. \ref{fig:cnn_schema}, the equation corresponding to the forward step for this network is as follows:

\begin{equation}
y(i,j) = (K * X)(i,j) = \sum_{n}\sum_{m} X(i-n, j-m)K(m,n)
\end{equation}

where $y$ is the the output feature map, $K$ is the kernel of trainable parameters and $X$ is a 2-D input array. 

Convolutional networks allow to identify local characteristics of the input. For this purpose, small kernels are used, generally with a dimension much smaller than the input dimension. These small kernels make the CNN very computationally efficient and robust to input transformations such as scale changes, translations or rotations.

A standard convolutional architecture usually contains three elements. First, a convolutional layer as described above. Secondly, a nonlinear activation function, usually a \textit{Rectiﬁed Linear Unit} (ReLU) function, which controls the range of variation of the output. The third stage consists of applying a \textit{pooling operation} on the output of the activation function. 

The pooling operation replaces the network output at a given position $(i,j)$ by a statistic summary of the output at the nearby positions. The most common pooling function is called \textit{max pooling} \cite{zhou1988computation} which assigns to the position $(i, j)$ a value calculated as the maximum of the outputs within a rectangular neighborhood. Pooling helps the convolutional architecture to be invariant to small translations of the input image. This is a very relevant property for the case of handwritten text images, for which a translation of the image does not modify the text written on it that is intended to be recognized.

Convolutional architectures play a fundamental role in all models working with data of spatial nature, especially 2-D images, but also 1-D time series or 3-D videos. Over the years, multiple convolutional architectures have been proposed for different problems related to image analysis. In general, these architectures include at least the three elements of convolution, activation and pooling indicated above. A complete review of these architectures is out of scope of this Thesis. For a complete reference of the different convolutional architectures proposed in recent years see Khan et al. \cite{Khan2020ASO}.

In this Thesis, we have worked with three types of convolutional architectures, which are: the LeNet \cite{lecun1995comparison}, VGG \cite{simonyan2014very}, and ResNet \cite{he2016deep}. These architectures have been used in both the character recognition experiments in Chapter \ref{chapter_character_models} and the word recognition models in Chapter \ref{chapter_word_models}.

LeNet architecture was the first proposed one for handwritten digit recognition in 1995. It is characterized by using 5$\times$5 kernels and having two layers: the first with 20 feature maps and the second with 50 feature maps. VGG architecture is characterized by proposing a convolution-convolution-pooling layer configuration with 3$\times$3 kernels. ResNet architecture is characterized by proposing direct connections (residuals) between the input and output of each convolutional component. Additionally, it employs the regularization strategy called \textit{batch normalization} \cite{ioffe2015batch}. In the following chapters, more information of these three architectures are provided, including a detailed definition of the architectures and diagrams of specific configurations used in the experiments of this Thesis.

\subsubsection{CNN in handwriting}

In the HTR problem domain, CNNs have been initially used for handwriting digits recognition \cite{lecun1989backpropagation}. Later, these networks were applied to handwritten text recognition in specific domains such as online handwriting \cite{bengio1995lerec} or bank checks transcription \cite{lecun1997reading} \cite{lecun1998gradient}.

One of the first applications of CNNs on the general handwriting recognition problem was proposed in 2013 by Bluche and collaborators \cite{bluche2013tandem}. In this work, a CNN-HMM model was implemented using the tandem approach. The convolutional architecture selected for feature extraction was inspired by LeNet-5 model. The results with CNN were superior than using handcrafted features and similar to those obtained with LSTM networks. 

In 2016, several related publications appeared \cite{krishnan2016deep} \cite{krishnan2016matching} \cite{poznanski2016CVPR}, which applied the nearest neighbor approach to the HTR problem. In these works, a CNN-type model is used to extract the features employed to identify the nearest neighbors and perform text recognition. In the same year, a full convolutional model was proposed for the identification of $n$-grams of rank 1, 2 and 3 in images of handwritten words that are then used to recognize the text by means of \textit{Canonical Correlation Analysis} \cite{poznanski2016CVPR}.
  \index{Canonical correlation analysis}

In 2017, CNNs take on a central role in an HTR recognition model \cite{bluche2017gated}. In particular, convolutional gates where used to enable context-sensitive and hierarchical feature extraction. A novel strategy of pre-training the convolutional part with multiple corpora of different languages was also proposed. Subsequently, the rest of the network is trained with each corpus separately.

Starting in 2017 with the publications of Puigcerver \cite{puigcerver2017multidimensional}, and Bluche and Messina \cite{bluche2017gated} popularizing the CNN-RNN-CTC architecture, CNNs are included in almost every approach to the HTR problem to nowadays. With the emergence of the seq-2seq-type models (discussed in Section \ref{section:seq2seq architectures}), CNNs remain an essential part of the solutions. In these cases, they are also always incorporated as the first component of the network, with the goal of extracting visual attributes from handwritten text images.

Finally, it should be noted that recently, some authors have proposed a full convolutional architecture with CTC to avoid the costs of running LSTM network architectures. Yousef et al. \cite{yousef2020accurate} proposes a Fully Convolutional Network (FCN)\nomenclature{FCN}{Fully Convolutional Network} architecture for the line recognition problem. The model is based on the concatenation of several blocks that use attention gates to control the information that is transferred between them. In this way, signals that are not relevant, such as background noise, are filtered out. The attention gates used are based on the highway network architecture proposed by Srivastava et al. \cite{srivastava2015training}. The same author presents, in a later publication \cite{yousef2020origaminet}, a fully convolutional architecture for HTR in pages with multiple lines. The paper proposes a strategy for adapting any fully convolutional model (at the line level) for text recognition (at the paragraph level). This adaptation is performed without needing the transcription to be segmented into lines, which is an advantage over other page-level or paragraph-level approaches, such as \cite{bluche2016scan} or \cite{wigington2018start}.

\subsection{Multidimensional Recurrent neural netwoks}
\label{section:Multidimensional Recurrent neural netwoks}
  \index{Multidimensional recurrent neural network}

In this subsection, we first introduce the \textit{Multidimensional Recurrent Neural Networks} (MDRNN), describing their main characteristics, and then detailing their application to the offline handwritten recognition problem. Next, we review the usage of MDRNN in the HTR problem.

MDRNN network architectures were introduced in 2007 by Graves et al. in \cite{graves2007multidimensionalrecurrent}, applying them to the image segmentation problem. MDRNNs generalize standard RNNs by defining recurrent connections along multiple dimensions, as opposed to RNNs that provide recurrence in a single dimension. This allows these types of networks to model spatial recurrences such as those that appear in handwriting images. Additionally, they are more robust to local distortions than the standard RNN along the different input dimensions. Distorsions include rotations or shears, which are very common in handwritten text.

\begin{figure}[!ht]
 \centering 
 \includegraphics[width=4cm]{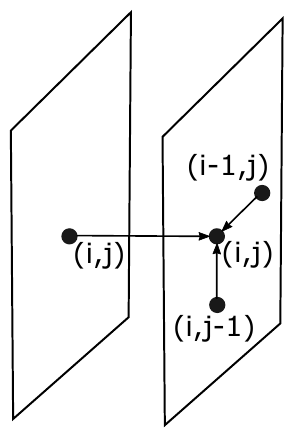} 
 \caption{MDRNN 2 layer context example (adapted from \cite{graves2007multidimensionalrecurrent}).}
 \label{fig:MDRNN_2layer}
\end{figure}

To illustrate how MDRNNs take into account the context in multiple dimensions, Fig. \ref{fig:MDRNN_2layer} shows how the calculation of the hidden layer at position $(i,j)$ use the values of the hidden layer at positions $(i-1, j)$ and $(i, j-1)$, respectively.

One-dimensional RNNs can define the context of a given point in the sequence in two different ways depending on whether the sequence is followed in the forward or backward direction. When the sequence is followed in both directions, a Bidirectional Recurrent Neural Networks (BRNN) \cite{schuster1997bidirectional} are obtained. In the same way, MDRNNs can define the context of a given point in multiple ways depending on how the different dimensions are followed. 
  \index{Bidirectional recurrent neural network}

\begin{figure}[!ht]
 \centering 
 \includegraphics[width=12cm]{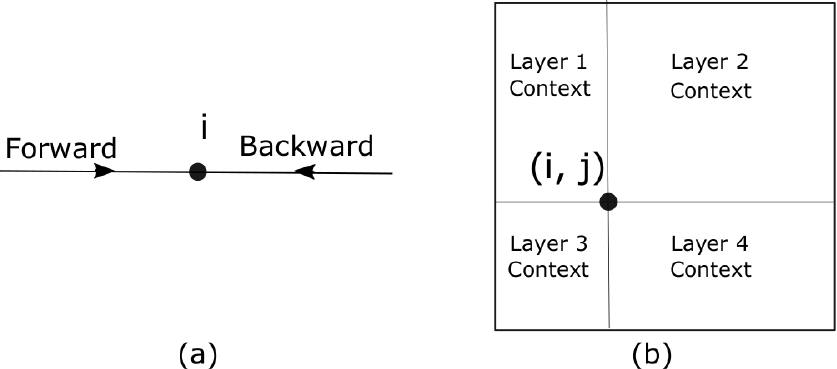} 
 \caption{MDRNN example of contexts. (a) 1-dimensional BRNN. (b) 2-dimensional MDRNN (adapted from \cite{graves2007multidimensionalrecurrent}).}
 \label{fig:MDRNN_context_sample}
\end{figure}

Fig. \ref{fig:MDRNN_context_sample} shows the two contexts associated to a one dimensional RNN and comparatively some possible contexts associated to a two dimension MDRNNN, which incorporates four different networks starting each one of them in a corner of the analysis surface and with a different directionality. The exact definition of the contexts in the multidimensional case can be quite varied, depending on the computation order strategy of the input elements. For example, in the case of a 2-D input, some possible orders can be: by rows first, by columns first or starting diagonally from each of the origin points of the corners.

MDRNNs are usually built with LSTM networks, and abbreviated as MDLSTM. LSTM networks avoid the vanishing gradient problem and ensure that the range of the context used in a given step is large enough, since they are able to model relationships across the entire context.
  \index{Long short-term memory network}

\subsubsection{MDRNN in handwriting}
\index{Multidimensional LSTM}

The first use of MDRNNs in the offline handwriting recognition problem was proposed in 2009 by Alex Graves and Jürgen Schmidhuber \cite{graves2009offline}, the same authors who two years earlier had introduced MDRNNs in \cite{graves2007multidimensionalrecurrent}. MDRNNs were proposed as a unified alternative to address a problem such as offline handwriting recognition, which requires both Computer Vision and Sequence Learning. Previously, the proposed systems addressed these two requirements separately, the former one with complex preprocessing and feature extraction techniques and the latter with sequence learning through models, such as HMMs that generated the transcriptions. This paper proposed the specific use of MDLSTM networks built with multidirectionality from the four corners of the text image to be processed.
  \index{Computer vision}
  \index{Sequence learning}

\begin{figure}[!ht]
 \centering 
 \includegraphics[width=16cm]{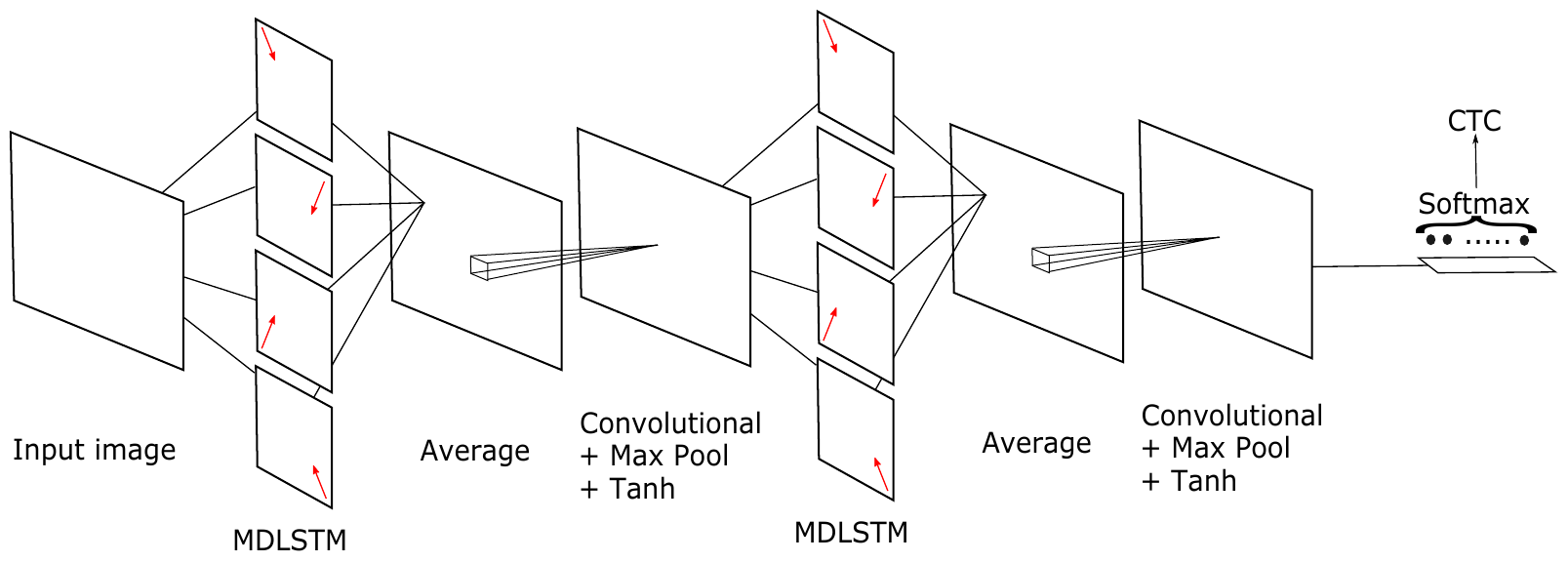} 
 \caption{MDLSTM standard architecture (adapted from \cite{voigtlaender2016handwriting}).}
 \label{fig:MDRNN_architecture}
\end{figure}

The architecture proposed in \cite{graves2009offline} included several MDLSTM type layers with interleaved convolutional and feedforward layers and a final CTC layer \cite{graves2006connectionist}, which is a type of decoding layer also introduced by Graves et al. three years before. Fig. \ref{fig:MDRNN_architecture} shows a schematic representation of the proposed architecture, which was later adopted by other authors with few variations. This architecture is fully trainable from raw pixel to text decoding, it does not require any alphabet-specific preprocessing like HMMs and it is directly applicable on any character set of different languages.
  \index{Connectionist temporal classification}

Even if the work by Graves and Schmidhuber \cite{graves2009offline} showed a significant improvement in recognition results, it was not until 2012 when other authors \cite{menasri2012a2ia} start embracing MDLSTM networks in general HTR problem, almost always with a few changes about the initial architecture proposed which is shown in Fig. \ref{fig:MDRNN_architecture}. 

In 2012, the French company A2IA in the paper \cite{menasri2012a2ia} explained their approach to the RIMES-ICDAR2011 competition. It stated the use of a MDLSTM architecture with CTC as part of an ensemble consisting of two models based on HMM and the MDLSTM model. To our best knowledge this is the second reference of MDLSTM use applied to HTR after the original 2009 reference. The paper does not detail the architecture used but indicates the results obtained by each model in the ensemble. The results of the MDLSTM model were clearly superior to the other two HMM-based models: 9.2\% of WER of the MDLSTM model versus 22.0\% of WER of the HMM-based models, in the isolated word recognition task on RIMES database. With the MDLSTM model alone, this paper already obtained  a clearly better result in the competition than the rest of the participants. 
  \index{Company!A2IA}

In 2013, the same company presented the same architecture in the competition of the Arabic Handwriting recognition OpenHaRT \cite{bluche2014a2ia}, also with good results. From here on, during the years from 2014 to 2017, the architecture based on MDLSTM and schematized in Fig. \ref{fig:MDRNN_architecture}, appears in several papers about offline handwritten text recognition, with very competitive results (see for example the works of \cite{pham2014dropout}, \cite{Leifert2016CITlabAF}, \cite{voigtlaender2016handwriting}, \cite{bluche2016scan}, and \cite{chen2017simultaneous}). 

Pham et al. \cite{pham2014dropout} proposed to improve the standard MDLSTM-based architecture with the application of the \textit{dropout regularization strategy} \cite{hinton2012improving}. This technique randomly removes a fixed percentage of hidden units from a neural network at each iteration of the training phase of the network, so that in the backpropagation error step, only some of the weights are updated at each training step. The dropout has a reducing overfitting effect and accelerates the convergence of the training process in deep models with multiple layers. Then, in the evaluation phase, all hidden units and weights are used. This paper \cite{pham2014dropout} proposes the application of dropout only in the feedforward layers of recurrent networks, without including it in the recurrent connections, to avoid affecting the ability of RNNs to model sequences. A detailed analysis of different dimensioning of the recurrent layers and, especially, on which layers to apply dropout is performed. The authors found that the optimal strategy for this architecture is the application of dropout on all recurrent layers as opposed to the usual practice at that time that was using it only on the last layers of the model.
  \index{Dropout}

Until 2016, the application of MDLSTM networks had the drawback that the existing implementations of MDLSTM networks were CPU-based and computationally expensive, which meant that training a network could take weeks \cite{Leifert2016CITlabAF}. The paper by Voigtlaender et al. \cite{voigtlaender2016handwriting} made publicly available an efficient implementation of GPU-based MDLSTM networks that allows the training of these architectures with more layers and with more hidden units in each layer than before. The paper proposed an implementation strategy for MDLSTM contexts with a parallelizable diagonal pattern instead of the usual hierarchical row and column implementation. Finally, it was shown that the increase in number and size of MDLSTM layers and the adoption of a different initialization strategy (proposed by Glorot et al. in \cite{glorot2010understanding}) provided substantial improvements in both the convergence speed of the train step and the accuracy in the evaluation step.

The MDLSTM architecture was proposed by Bluche et al. in \cite{bluche2016scan} and \cite{bluche2016joint}, as the basis for the implementation of a handwritten paragraph-level recognition systems, without the need of line segmentation. These are the first cases where an end-to-end handwritten paragraph recognition system is proposed. The solution is based on the combination of an MDLSTM-like network with the usual architecture, augmented with an attention mechanism \cite{bahdanau2014neural}.

In 2017, a variant of the MLSTM network called \textit{Separable MDLSTM} (SepMDLSTM) is proposed in the paper \cite{chen2017simultaneous} by Chen et al., where the rows and columns of the input image to the recurrent layer are traversed sequentially in two successive layers instead of in parallel layers as previously done. This new architecture is applied in a multi-task problem for simultaneous script identification with a classifier and handwriting recognition with a CTC layer.

Finally, in 2017 two publications by Bluche and Messina \cite{bluche2017gated} and Puigcerver \cite{puigcerver2017multidimensional}, respectively, criticized the high computational cost of MDLSTM networks and proposed the same alternative architecture. This is based on first employing a set of CNN-type layers, that capture visual attributes of the text image, followed by LSTM-type layers that capture the recurrent nature of the handwritten text. In both cases, the CTC is used as the final decoding layer and yields results that significantly improve those obtained with MDLSTM architectures.

After the above publications, MDLSTM-based architectures are no longer used in the handwritten recognition problem and CNN-LSTM-CTC based architectures become the new standard. These architectures are discussed in the next section.

\subsection{The CNN-LSTM-CTC Architecture}
\label{section:The CNN - LSTM - CTC Architecture}
  \index{Convolutional neural network}
  \index{Recurrent neural network}
  \index{Connectionist temporal classification}

\begin{figure}[!ht]
 \centering 
 \includegraphics[width=10cm]{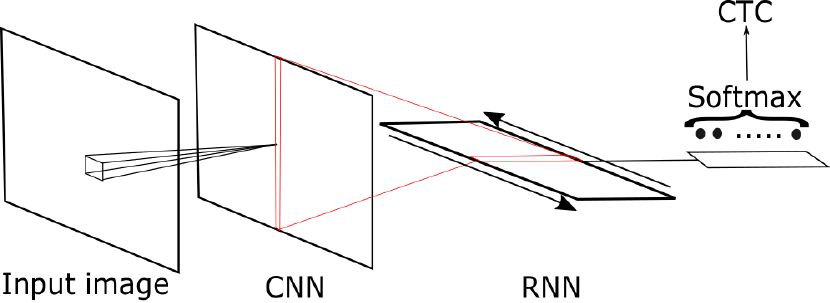} 
 \caption{Standard CNN-RNN-CTC general architecture.}
 \label{fig:CNN_RNN_CTC_general_architecture}
\end{figure}

As described in the previous subsections, CNNs and RNNs had been frequently used in the HTR problem although usually separately. In 2016, a paper by Suryani et al. \cite{suryani2016benefits} proposed an architecture for HTR based on CNNs followed by BILSTM-type of RNNs, using HMM for final transcription decoding. The model was applied to Chinese handwriting recognition with discrete results compared to those obtained at that time with MDLSTM type architectures. In any case, it is, to the best of our knowledge, the first reference of a CNN-RNN architecture applied to the HTR problem. In the same year, Shi et al. \cite{shi2016end} published, and also proposed a CNN-RNN architecture applied to the recognition of printed text in natural images. But it is in the publications by Bluche et al. \cite{bluche2017gated} and Puigcerver \cite{puigcerver2017multidimensional} where the CNN-LSTM-CTC based architectures become the reference to address the handwritten recognition problem. The general architecture proposed by the different authors is very similar, although each author introduces some variations that are described below. A general outline of the architecture is shown in Fig. \ref{fig:CNN_RNN_CTC_general_architecture}.

These architectures have proven to achieve good results in another text recognition tasks. For example, Breuel \cite{breuel2017high} shows how this architecture improves the obtained results with BLSTM-CTC architecture in optical characters recognition (OCR) task. Li et al. \cite{LI201814} used it to obtain the state-of-the-art in the license plate recognition problem. Shi et al. \cite{shi2016end} also used this same architecture for scene text recognition in natural images.

Bluche and Messina \cite{bluche2017gated} proposed a standard CNNs alternative, called gated-CNN. Basically, the idea is that the convolutional filter was followed by a sigmoig activation that controls the propagation or not of the feature to the next layer, depending on the values of the analyzed position and the neighboring positions. The paper employs a large multilingual corpus of 132,983 lines in a total of 7 languages with Latin characters. The convolutional layers are trained with this joint corpus and subsequently a fine tuning per language of the LSTM layers is performed.

In Puigcerver \cite{puigcerver2017multidimensional} the same previous architecture is proposed: CNN layers followed by LSTM layers with a CTC decoding. The author uses batch normalization \cite{ioffe2015batch} and dynamic train data augmentation (as it will be described in Section \ref{section:Data_normalization}), which are techniques that have been successfully employed in image classification problems with convolutional architectures at that time.

Dutta et al. \cite{dutta2018improving} also proposes the CNN+LSTM+CTC architecture but introducing at the beginning a \textit{spatial transformer network} (STN) layer \cite{jaderberg2015spatial}. It is a trainable layer that produces geometrical transformations on the input similar to those produced in handwriting due to random hand movements. In the published ablation study the authors indicate an improvement of the \textit{Word Error Rate} (WER) of 9\% for IAM database at word level by the introduction of the STN layer in the model. Unfortunately, comparisons with results from other authors are difficult since the paper uses a charset reduced to lowercase characters, test lexicon decoding and test-time data augmentation.
  \index{Metrics!Word error rate}
  \nomenclature{STN}{Spatial Transformer Network}
  \index{Spatial transformer network}
  \index{Databases!IAM}

Just as MDLSTM-based models have been used as a basis for building page-level HTR systems as in \cite{bluche2016scan}, CNN-LSTM-CTC architectures have also been used for the same purpose. In 2018, Wigington et al. \cite{wigington2018start} propose an end-to-end full-page handwriting recognition system composed with three elements. First, a Region Proposal Network to identify the start position of each text line in the page. Second, a novel line follower network that pre-process and convert text line into a normalized line image. Third, a CNN-LSTM-CTC type network that transcribes the normalized line.

Recently, some publications question the efficiency of CNN-RNN architectures for modeling complex spatial and temporal relationships like the ones in handwriting text. In particular, it is questioned the fact that both relationships could be extracted optimally with two sequentially connected components but not integrated. As alternative, variations of recurrent networks to introduce the spatial component \cite{huang2020end} or variations of convolutional networks to introduce the temporal or sequential component \cite{ingle2019scalable} are proposed.

Huang et al. \cite{huang2020end}, propose a new neuronal architecture that combines convolution, pooling and recurrence in a unified framework named \textit{Convolutional Recurent Neural Network} (CRNN)\nomenclature{CRNN}{Convolutional Recurent Neural Network}. The CRNN can be seen as an improved version of RNN where the direct connection between input and hidden layer is replaced by a local connection and shared weights like a convolutional kernel.

In Ingle et al. \cite{ingle2019scalable}, it is discussed the model scalability with a RNN component because its execution cannot be parallelized like with a CNN. As alternative to LSTM, it is proposed the use of \textit{Gated Recurrent Convolutional layers} (GRCL)\nomenclature{GRCL}{Gated Recurrent Convolutional Layers} \cite{wang2017gated}. In summary, these layers are conventional convolutional layers with added recurrent connections in the time direction. Additionally, they include a gatting mechanism that allows to control the information propagated by these recurrent connections like in GRU layers.

\section{Connectionist Temporal Classification (CTC)}
\label{section:Connectionist Temporal Classification - CTC}

The \textit{Connectionist Temporal Classification} (CTC) is an output layer defined to align sequences. The layer can directly transform an input sequence $S_i$ into an output sequence $S_o$ provided that the size of the input sequence is larger or equal than the size of the output one, i.e. $|S_i|>=|S_o|$. This layer performs the output decoding of the visual model by converting it into the sequence of characters that transcribe the text contained in the image. This allows training HTR models in a single step, since it does not need an additional decoding process like the one needed by HMM-based models. 
  \index{Connectionist temporal classification}

\begin{figure}[!ht]
\centering 
\includegraphics[width=8cm]{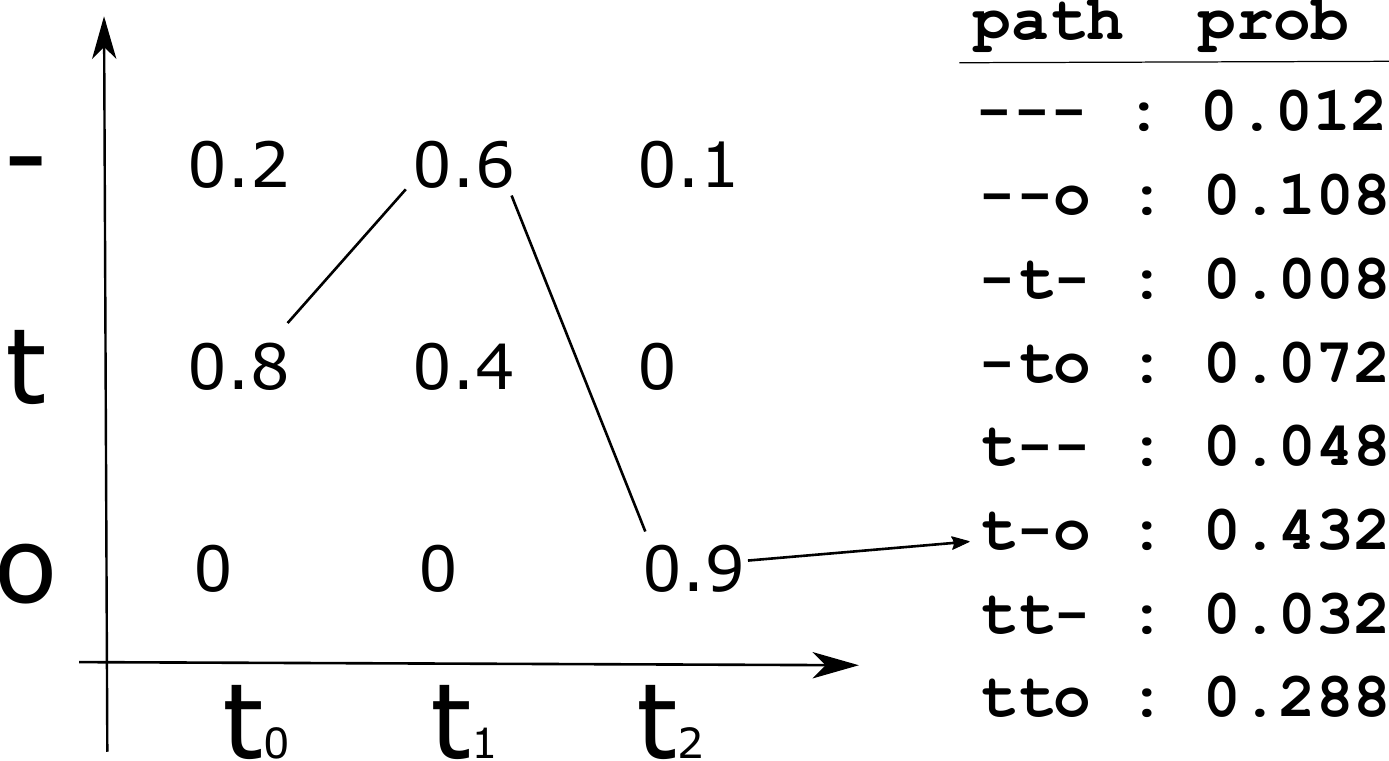}
\caption{CTC loss calculation: all possible paths to decode the output probabilities sequence are computed.}
\label{fig:CTC_loss}
\end{figure}

The CTC interprets the network output as a probability distribution over all possible character sequences that can happen, conditioned to an input sequence. From this probability distribution, it is possible to obtain a differentiable objective function that maximizes the probability of the correct output sequence from the input one, that can be trained by backpropagation in conjunction with the other layers of the network. Fig. \ref{fig:CTC_loss} shows all possible decoding paths in a particular example simplified to a 2-character dictionary. For this particular case, the probability of the correct sequence "to" is: 0.072 (-to) + 0.432(t-o) + 0.288(tto) = 0.792.

To better handle transitions between characters, CTC introduces a special character called "blank", usually represented by the symbol "-", which should not be confused with the whitespace character for hyphenation. This "-" character allows to properly model transitions between characters where there may be a certain separation, as well as possible spaces at the beginning or end of a handwritten text image. It also facilitates learning with images of the same text that may have different alignment and spacing between characters.

\begin{figure}[!ht]
\centering 
\includegraphics[width=6cm]{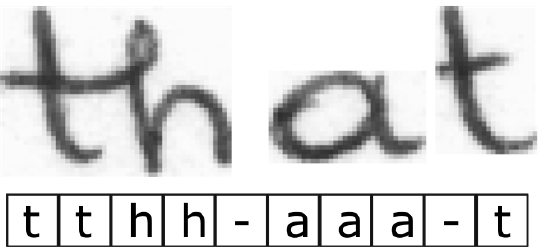}
\caption{CTC Decoding path for one word.}
\label{fig:CTC_decoding}
\end{figure}

For the prediction of new images in the evaluation phase, the CTC selects the sequence or decoding path with the highest marginal probability given the input and applies a operator ($B$) that firstly removes duplicate characters and secondly removes the special character 'blank' (-). An example of decoding a handwritten word can be seen in Fig. \ref{fig:CTC_decoding}.

The white character can optionally appear between the other characters and is only required to appear between two identical characters that are repeated in the input image.

Thus, for example, the decoding of the following sequences is as follows
\begin{itemize}

\item B(-aa-b) = ab
\item B(aa--bb) = ab
\item B(aa--b-bb-) = abb
\end{itemize}

The first two examples illustrate a case of decoding the same text due to differences in the alignment of the characters in the input image.

The use of the white character is optional. In Bluche et al. \cite{bluche2015framewise}, an analysis on the inclusion or not of the white character is provided. This analysis conclude that its inclusion favors the use of CTC in handwriting recognition.

There are different variants of the decoding process that allow to restrict the decodings to a prefixed dictionary or to introduce a language model \cite{graves2009novel}. A detailed efficiency analysis of the CTC applied to the handwritten text recognition problem, compared to the classical HMM-based approach, can be found in \cite{bluche2015framewise}. It also compares the performance of CTC with inputs from RNN and MLP, concluding that RNN provides better results because it can better handle dependencies in an arbitrary length context.

\subsubsection{CTC in handwriting}
\label{section:CTC usage in handwriting}

The CTC model was introduced in 2006 by Graves et al in \cite{graves2006connectionist} as a general method for temporal classification with RNNs applied to the phonetic labeling problem. Later, in 2009, the same authors applied it to the handwritten word recognition problem in \cite{graves2009novel} where an architecture based on bidirectional LSTM (BLSTM) with a final CTC layer for decoding was proposed. This model is applied on a set of nine graphical features obtained from the normalized image by means of a one-pixel sliding window. In addition, the initial definition of CTC is modified to allow the integration of CTC with a dictionary and a language model in a new algorithm called \textit{CTC token passing}. This allows to restrict and orient the decoding process and to obtain results directly comparable to those obtained in HMM-based architectures, where the language model, usually based on $n$-grams, supports the decoding process.

In the same year, Graves and Schmidhuber published a novel architecture for the handwritten recognition problem \cite{graves2009offline} that, starting directly from the image pixels, applied a Multidimensional LSTM (MDLSTM) architecture in several layers with a final CTC layer. To the best of our knowledge this is one of the first times that a unified probabilistic model is proposed for the unconstrained handwriting recognition problem. It trains character sequence recognition directly from the pixels of the text image. This MDLSTM+CTC architecture applied directly on the image and has become the reference for the handwriting problem between 2013 and 2017. 
  \index{Multidimensional LSTM} 

Although the first use of CTC in handwriting dates from 2009 in \cite{graves2009novel}, it is not until 2014 that the CTC has become popular in the field of handwritten text recognition. Since them it is a standard used in the vast majority of HTR publications between 2014 and 2018 \cite{bluche2014comparison} \cite{pham2014dropout} \cite{bluche2015framewise} \cite{voigtlaender2016handwriting} \cite{bluche2016joint} \cite{bluche2016scan} \cite{chen2017simultaneous} \cite{bluche2017gated} \cite{puigcerver2017multidimensional} \cite{wigington2018start} \cite{Chowdhury2018AnEE} \cite{dutta2018improving}.

Recently Scheid et al. in \cite{scheidl2018word}, have reviewed and compared different decoding strategies for the CTC algorithm. A new decoding algorithm called \textit{word beam search decoding} was proposed. It allows the integration of a dictionary and a language model, like the \textit{CTC token passing algorithm} proposed in \cite{graves2009novel}, but reducing its computational complexity and allowing the decoding of special characters between words in the dictionary, such as numbers or punctuation marks.

In recent years, it has become increasingly common to decode the final sequence of characters using a decoder layer of the seq2seq architecture, which is also described in this chapter. In any case, CTC is still used sometimes with encoder-decoder architectures like seq2seq. For example, in Bluche et al. \cite{bluche2016scan} the CTC is used as an auxiliary method to perform a pre-training at word or line level and these authors apply an architecture with attention and decoder to perform a prediction at paragraph level. Another use case of CTC in an encoder-decoder type architecture can be found in \cite{coquenet2020end}, where this architecture with hybrid attention for paragraph or page level handwriting recognition using CTC for the final decoding of the text, is proposed. Similarly, Yousef et al. \cite{yousef2020origaminet} describe a neural network that can convert any fully convolutional handwritten line recognition model trained with the CTC into a handwritten page recognition model without needing for the image and its transcription to be segmented by lines.

\section{Solutions based on Nearest Neighbors}
\label{section:Solutions based on Nearest neighbors}

Between 2014 and 2016 some authors have proposed an alternative to HMM or CTC-based approaches common at the time, that consists in representing the images and transcriptions of handwritten words in a shared \textit{n}-dimensional embedding space and then applying a \textit{k-Nearest Neighbors} (k-NN) algorithm to associate transcriptions and images. This approach was primarily aimed at solving the problem of word spotting, which consists of identifying whether a given word is present in a text image or not (usually for the purpose of indexing databases of historical manuscript text images). By extension, the application of k-NN to single word images is also a way of transcribing such images,

In 2014, Almazan et al. \cite{almazan2014word} address the problems of word spotting and word recognition with the above approach. They introduce a technique called \textit{Pyramidal Histogram of Characters} (PHOC)\nomenclature{PHOC}{Pyramidal Histogram of Characters} to encode transcripts in a way that considers whether a specific character appears at a particular position in the text string. Then, the word images are projected into another space, where each dimension encodes whether a word image contains a given character in a particular region. Finally, the method trains a projection of the previous spaces into a lower dimensional common subspace where it uses k-NN for both word spotting and word recognition problems. Later, some other authors, like \cite{krishnan2016deep} and \cite{poznanski2016CVPR}, employed a similar procedure, including the PHOC technique for transcription coding, but using attributes trained with a CNN for the generation of embedding associated to the word images. In reference \cite{krishnan2016deep}, a proprietary architecture called \textit{HWNet} is used, which is described in detail in \cite{krishnan2016matching}, and in the case of reference \cite{poznanski2016CVPR}, a VGG \cite{simonyan2014very} style network is used.

Due to the search solution approach in a shared \textit{n}-dimensional embedding space, these methods require a default lexicon to perform text recognition, which is a drawback because they cannot identify \textit{out of vocabulary} (OOV) words.
  \index{Out of vocabulary}
  \nomenclature{OOV}{Out Of Vocabulary}

\section{seq2seq architectures}
\label{section:seq2seq architectures}

In recent years a new type of architectures applied to the offline HTR problem has appeared, the so-called \textit{sequence-to-sequence (seq2seq)} architectures that follow the encoder-decoder framework \cite{sutskever2014sequence}, and that are oriented to model relationships between unaligned sequences, as it happens in the case of offline HTR.
  \index{Sequence-to-sequence}
  
This section introduces the seq2seq architectures and details their use in the HTR problem in recent years. These architectures are the ones used in Chapter \ref{chapter_word_models} of this Thesis about deep architectures to continuous offline handwriting recognition.

\subsection{Overview of seq2seq architectures}
\label{sect:Introduction to seq2seq architectures}

The seq2seq architectures were introduced in 2014 by Sutskever et al in \cite{sutskever2014sequence}, and Bandanau et al. \cite{bahdanau2014neural} proposed a key modification of it by including an \textit{attention mechanism} that allowed a good management of long sequences. These new kind of neural architectures are oriented to model sequence transduction problems, where it is necessary to decode an information sequence into another sequence not necessary aligned with the first one. The most common cases and these architectures are applied are: speech recognition \cite{chorowski2015attention} \cite{bahdanau2016end}, machine translation \cite{bahdanau2014neural} \cite{sutskever2014sequence} and image captioning \cite{xu2015show}. Their success in the above problems has also led to their use in other problems, such as text recognition in natural images \cite{lee2016recursive} \cite{shi2016robust} or even speech recognition from lip movement in videos \cite{chung2016lip}. 

\begin{figure}[!ht]
\centering 
\includegraphics[width=8cm]{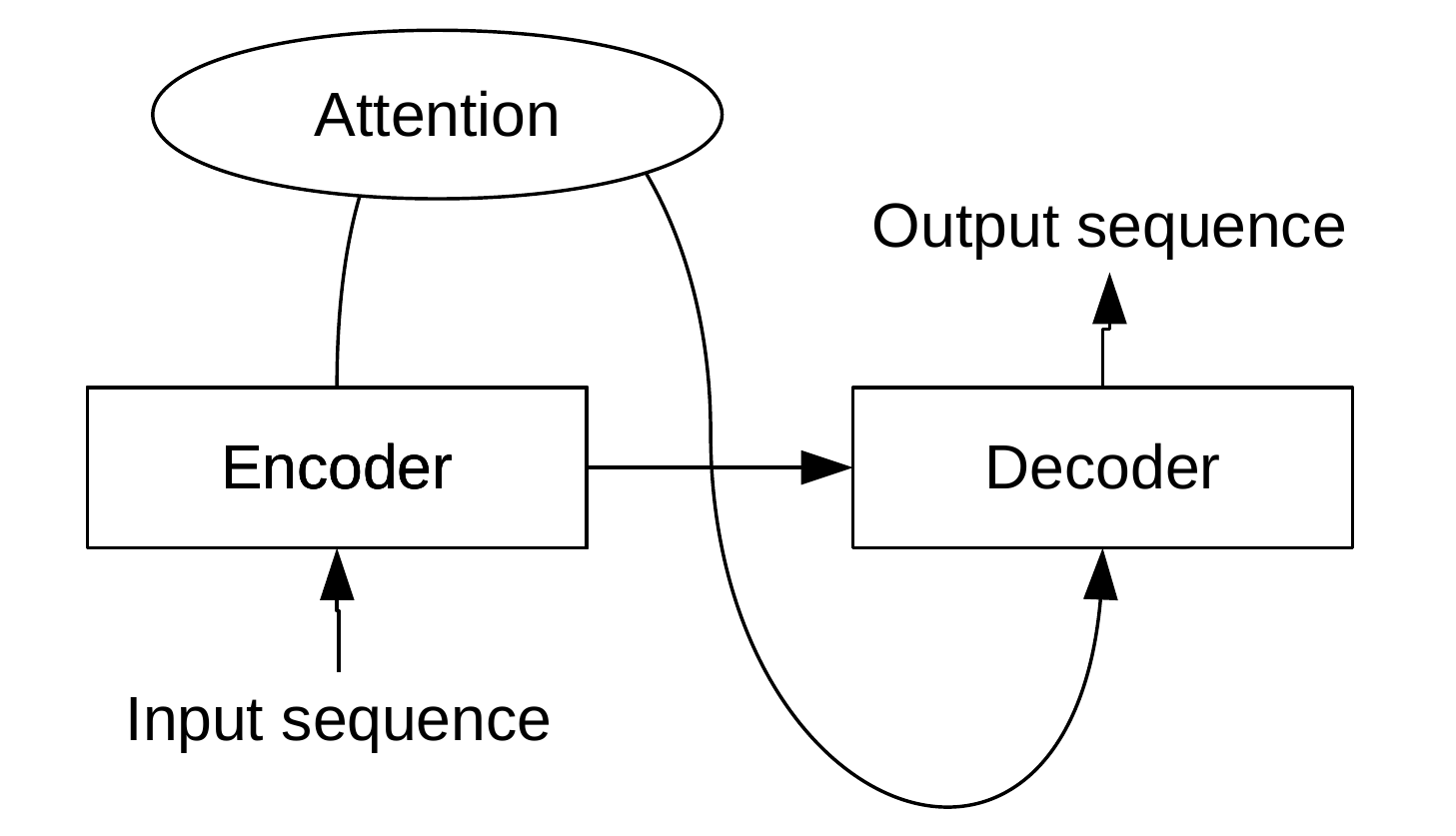} 
\caption{Basic schema of the Seq2seq architecture.}
\label{fig:seq2seq_basic_schema}
\end{figure}

In Fig. \ref{fig:seq2seq_basic_schema} is outlined the different principal components of seq2seq architecture, composed of the following elements:

\begin{itemize}
  \item An \textit{encoder} that allows coding variable length input sequence in a fixed subspace. It uses to be composed by recurrent layers and, in the case that the input is a image or a video, it usually also includes convolutional layers. 
  \item A \textit{decoder} that, from the decoder output, builds the output sequence element by element. It is usually composed by a unidirectional RNN.
  \item An \textit{attention mechanism}, that allows each decoder step to access to the most relevant encoder information for that step.
\end{itemize}

\subsection{seq2seq Encoder}

In seq2seq architecture the \textit{encoder} is in charge of coding each input sequence, that could have variable length, into a fixed length sequence that codifies such input in a feature subspace. 
In that way, given a variable length input elements sequence $\mathbf{x}=(x_1, ..., x_{n_x})$ it is transformed in a fixed length sequence $\mathbf{h}=(h_1, ..., h_T)$ by the encoder, usually through a input sequence previous transformation that depends of its nature and, after that, by a RNN $f$ in such a way that:

\begin{equation}
\begin{array}{lcl} 
(s_1, ..., s_{T}) = g(x_1, ..., x_{n_x}) \\
h_t=f(s_t, h_{t-1})
\end{array}
\end{equation}

where $g$ can be simply padding over the input sequence for length alignment, or a complex convolutional architecture in the case that one or several images (of one or more channels) being the input, or other options depending of the problem nature and the coding data. Additionally, $h_t \in {\rm I\!R}^n$ is a hidden state at time $t$. 

Finally, the output sequence $\mathbf{h}=(h_1, ..., h_T)$ is transformed in a context vector $\mathbf{c}$ that will be sent to the decoder. For example, in the original seq2seq model by Sutskever et al. \cite{sutskever2014sequence}, the transformation is $\mathbf{c} = q(h_1, ..., h_T) = h_T$ and the context vector $\mathbf{c}$, that is the encoder sequence last element, it is moved to first decoder element. In the case that the model include an attention mechanism, it will allow to each decoder step, access to the complete encoder output sequence $\mathbf{h}=(h_1, ..., h_T)$.

The encoder architecture in a seq2seq model is very variable and dependent on the problem nature and on the input data. In fact, most of the variability in the different approaches that use seq2seq models for HTR problem is in this component and, in occasions, in the attention component used.

Although the encoder architecture can be very diverse, the last layer is usually RNN type, in particular those kind that can capture well dependent relationships in long sequences, like LSTM or GRU. The decoder usually uses the same kind of RNN network with the same cell number in these cases, because it is usual that the final state of the encoder $\mathbf{c}$ will be the initial state of the decoder. If a bidirectional RNN is used as encoder, it is possible to select as decoder initialization the final state of the forward layer or the same of the backward layer or a combination of both.

Quite complex cases can be codified by the encoder with multiple input sequences of different length. For example, in reference \cite{chung2016lip} the aim is to recognize an audio sequence that came along with a video sequence with the speaker's facial movements to complement the audio signal.

In handwriting recognition problem, where a 1-channel grayscale image is usual, the encoder typical architecture is composed by convolutional and pooling layers followed by recurrent layers. The convolutional and pooling layer sequence is oriented to extract the image visual characteristics of the handwritten text. The final recurrent layer sequence learns the characteristics of the sequential nature of the text. It is the same underlying idea as in CNN-RNN-CTC model described in Subsection \ref{section:The CNN - LSTM - CTC Architecture}. Different encoder architectures used in HTR literature will be described in detail in Subsection \ref{section:seq2seq in handwriting}. In Chapter \ref{chapter_word_models} multiple experiments are detailed with different alternatives related to the encoder module of a seq2seq model applied to HTR in the main reference databases.

\subsection{seq2seq Decoder}
\label{section:seq2seq Decoder}

In the seq2seq architecture, the \textit{decoder} is in charge of generating the output sequence of the model, which is usually of different length than the input sequence. Unlike the CTC case, the output sequence may be longer than the input sequence.

The decoder predicts each of the elements of the output sequence $y_t$ as a function of the previous elements in the sequence $y=(y_1, y_{t-1})$ and of a context $c$ provided by the attention mechanism. Thereby, the decoder defines a probability on the output sequence $y$ as the product of the ordered conditional probabilities of each previous element of the sequence, as given in the equation:

\begin{equation}
p(y) = \prod_{t=1}^{T} p(y_t | \{y_1,...,y_{t-1}\}, c)
\label{eq:seq2seq_decoder_probability}
\end{equation}

The decoder is usually implemented as a unidirectional RNN \cite{sueiras2018offline} \cite{kang2018convolve}, generally speaking of LSTM or GRU types. The RNN output is additionally transformed by a dense layer $d$ to fit its dimension to the cardinal of possible set of values of $y_t$. It is finally applied an activation softmax function \cite{bridle1990probabilistic} to get the probability distribution estimation $p(y_t)$.
  \index{Softmax}

The initial state of the RNN decoder $h_0$ is normally initialized with the value of the final state of the encoder $h^e_{last}$. This is shown in Fig. \ref{fig:seq2seq_basic_schema} by the direct connection defined between the encoder and the decoder. Although it can also be randomly initialized \cite{kang2018convolve}.

\begin{figure}[!ht]
\centering 
\includegraphics[width=16cm]{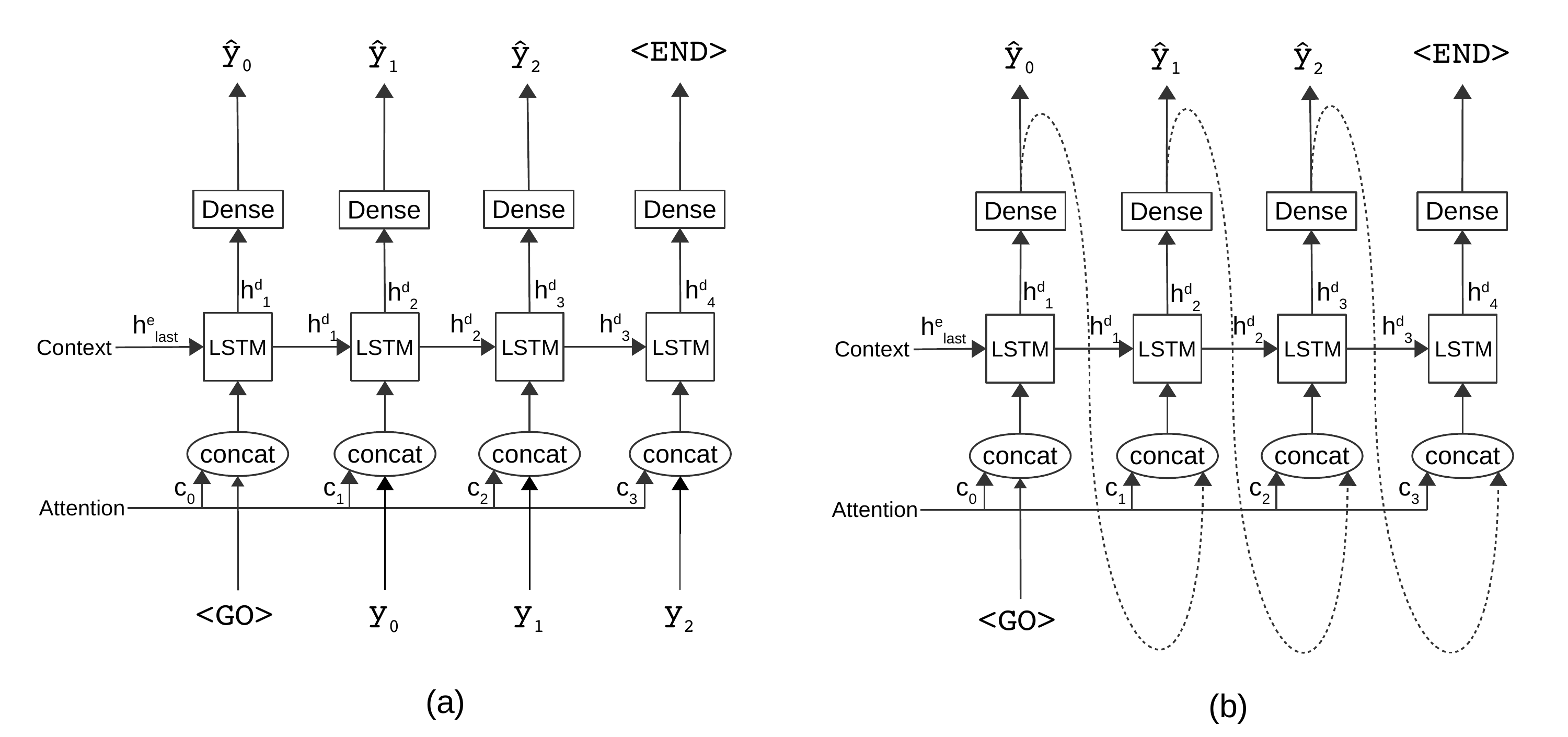} 
\caption{Schema of a LSTM based decoder: (a) train(teacher forcing) and (b) evaluation.}
\label{fig:seq2seq_decoder}
\end{figure}

The decoder RNN input during training phase is the target sequence of characters $\mathbf{y}$ but initiated with the special element $\textless GO\textgreater$. The output sequence finalization is indicated when the special element $\textless END\textgreater$ is decoded, like is illustrated in the left of Fig. \ref{fig:seq2seq_decoder}. This figure includes a decoder detail with all previous elements. As it can be observed, in the training stage the input is compound by the original values of the output sequence $\mathbf{y}$ (left sub-figure in Fig. \ref{fig:seq2seq_decoder}). In the evaluation stage the input at step $t$ is the step $t$-1 output estimation, denoted as $\hat{y}_{t-1}$.


The training of the seq2seq model by passing as input to the decoder the original values of the output sequence, but shifted to $t-1$ is called \textit{teacher forcing} \cite{williams1989learning}. This training strategy can lead to problems in the evaluation phase because an error in one of the generation steps of the output sequence is propagated through the conditioned probabilities. This causes a decrease in performance because the predicted sequence diverges from the real one from the point of error.

Teacher forcing also has a negative effect: the train overfit. This is because at each step $t$ of the decoding process the previous real value of the character sequence $y_{t-1}$ is used. To mitigate this, several alternatives have been proposed such as the approach of combining the actual value and the estimate of the previous step as input to the next step \cite{bengio2015scheduled}. It is also common to train with the same criterion used in the evaluation step, i.e., to use as input to step $t$ the estimate generated in the previous step $\hat{y}_{t-1}$ instead of the actual value $y_{t-1}$.

The values $y_t$, usually discrete, have to be coded so they can be used in the decoder input. They can be transformed by one-hot codification or a dense layer can be used that codifies the values of $y_t$ in low dimensionality embeddings.

In addition, Fig. \ref{fig:seq2seq_decoder} also shows that, in case that an attention mechanism is included in the seq2seq model (in the $t$ estimation step) the attention vector $c_t$ is introduced into the decoder by concatenating it with the RNN input. This concatenation is done directly with the previous element of the output sequence $y_{t-1}$ in the training step with the teacher forcing strategy. In the evaluation step, it is done with the estimation obtained in the previous step $\hat{y}_{t-1}$ (see sub-figure right in Fig. \ref{fig:seq2seq_decoder}).

The following equations describe the detailed calculation carried out by the RNN decoder:

\begin{equation}
 \begin{array}{lcl} 
  \hat{y_t} = argmax(p(y_t)) \\
  p(y_t) = softmax(dense(h_t)) \\
  h_t = LSTM(concat(e_t, a_t),h_{t-1})\\ 
  e_t = embbeding(y_{t-1}) \\
  h_0 = h^{e}_{last}\\
  e_0 = embedding(<GO>)
 \end{array}
 \label{eq:decoder_detail}
\end{equation}

The estimated output sequence $\mathbf{\hat{y}}=(y_1, ...,y_k)$ can be easily built by selecting the maximum probability values in $p(y_t)$ by an \textit{argmax} function for each $t$, as it is indicated by previous equations. But, it is also possible to use more complex decoding strategies like bean search decoding or even addressed to a language model \cite{Chorowski2017towards}. 

There are several variations on the above formulation. For example, one can try to obtain a more robust decoder that could explore another decoding sequences by selecting a random element based in \textit{softmax} probability distribution instead of selecting the higher probability element. This general technique is called \textit{multinomial decoding} \cite{cho-etal-2014-learning}.

Another common strategy to improve the stability in multi-category predictive models is the \textit{label smoothing} \cite{szegedy2016rethinking}, which consists of replacing the identificators 0, 1 of the selected category in each step of the decoder by a regularized version $\frac{\epsilon}{k}$ and $1-\epsilon\frac{k-1}{k}$. This procedure usually improves the generalization capacity of the model, but spending longer training periods. In the case of reference \cite{kang2018convolve} this technique has a significant positive impact when the model seq2seq is applied to the HTR problem. 

Finally, although the decoder is usually an unidirectional RNN, sometimes it uses another different architecture. For example, in \cite{kalchbrenner2013recurrent} a combination of RNN and de-convolutions is used.

\subsection{seq2seq Attention Mechanism}
\label{section:C3_seq2seq_attention}

This component, introduced by Bandanau et al. \cite{bahdanau2014neural}, confronts the original seq2seq problem as it connected the encoder and the decoder only with the state vector $c$, limiting the use of seq2seq models with long longitude sequences because the performance of the original model deteriorated rapidly when the length of the sequences increased \cite{bahdanau2014neural}.

The attention mechanism solves the previous problem by connecting each step of the decoder with all the elements of the encoder sequence that may be relevant for the decoding of that step. The model then predicts each element of the output sequence based on the predictions of the previous elements and this context provided by the attention mechanism.

More formally, and following the notation of the previous sections, to calculate the attention vector $c_t$ at each output time $t$ over a encoder output length $N_e$:

\begin{equation}
\begin{array}{lcl} 
e_{t,i} = w^T \tanh(W h^e_i + V h^d_{t-1} + b)\\
a_t = softmax(e_t)\\
c_t = \sum_{i=1}^{N_e}a_{t,i}h_i
\end{array}
\label{eq:content_attention}
\end{equation}

where $w$ and $b$ are trainable weight and bias vectors, $W$ and $V$ are trainable matrices, $h^e_i$ is the hidden state of the encoder at time $i\in{0,1, ..., N_e}$ and $h^d_{t-1}$ is the hidden state of the decoder at time $t-1$. The attention mechanism is implemented as a neural network layer that combines the states of the encoder and the previous states of the decoder with the hyperbolic tangent as activation function. The vector $e_t$ contains in the position $i$ a score of how much attention need to put the current decoder step $t$ on the \textit{i}-th encoder hidden state $h_i$. Then, we apply the $softmax$ function to normalize it and create the attention masks $a_t$ over all the encoder states. The final attention vector $c_t$ are concatenated with the input of the decoder layer as we show in Fig. \ref{fig:seq2seq_decoder}.

\subsubsection{Location Based Attention}
\label{suubsection:Location Based Attention}

The previous attention mechanism, also called \textit{content-based attention}, is one of the most common and is the one used in this Thesis. This algorithm has been seen to have scalability problems with very long input sequences \cite{chorowski2015attention} (e.g. of the order of thousands, as for example those that had the problem of speech recognition). This is motivated by the fact that, in this type of sequences, the previous model is not capable of distinguishing multiple representations of the same input fragment in different positions. In this Thesis, where we work at word images level, the length of the maximum input sequence corresponds to the width of the word image and it does not exceed a few hundred pixels, and the attention mechanism does not seem to face the those problems. However, in a line, or even paragraph or page level of the handwritten recognition approach, it can be relevant.

The modification of the previous attention mechanism which has been proposed to solve this problem \cite{chorowski2015attention} is called \textit{location-based attention}. It is based on the fact that the previous attention mechanism explicitly takes into account, both the identification of the position considering the previous step, and the characteristics of the input sequence. This is achieved by adding auxiliary convolutional characteristics as inputs to the attention mechanism $l_t$. They are extracted by convolving the attention weights from the previous step $a_{t-1}$ with trained filters $F$. 
In this way, the calculation of $e_{t,i}$ is replaced in the equation \ref{eq:content_attention} by:

\begin{equation}
\begin{array}{lcl} 
l_t=F*a_{t-1}\\
e_{t,i} = w^T \tanh(W h^e_i + V h^d_{t-1} + U l_{t,i} + b)\\
\end{array}
\label{eq:location_attention}
\end{equation}

\subsubsection{Attention smoothing}
\label{subsection:Attention smoothing}

Another problem that has the classical attention mechanism detailed in the equations \ref{eq:content_attention} 
is that the included \textit{softmax normalization}, calculated as: 

\begin{equation}
a_{t,j} = \frac{exp(e_{t,j})}{\sum_{i=1}^{L}exp(e_{t,i})}
\label{eq:attention_softmax}
\end{equation}

is likely to generate noisy information when the input sequence $h$ is long. This makes it difficult for the calculation equation of attention to focus on the relevant input elements $h_i$ for the decoding of each of the elements of the output sequence $y_j$. To solve this, Chorowski et al. \cite{chorowski2015attention} proposed to replace the normalization with softmax, with another one using the bounded logistic sigmoid $\sigma$ such that: 

\begin{equation}
a_{t,j} = \frac{\sigma(e_{t,j})}{\sum_{i=1}^{L}\sigma(e_{t,i})}
\label{eq:attention_sigma}
\end{equation}

which softens the focus found by the attention mechanism.

This smoothing technique has been used in the case of HTR (for example, in \cite{kang2018convolve}) with worse results than those obtained with the normalization. In this case, it has worked at the word level, so the length of the input sequences is relatively short and smoothing does not seem to be an advantage, but at line level it could be useful.

\subsubsection{Transformers}
\label{section:seq2seq_Transformers}

In recent years, the attention mechanisms have gained great relevance, since they are the basis of the networks called \textit{transformers} \cite{NIPS2017_3f5ee243}. These have produced important advances in modeling both textual information \cite{devlin-etal-2019-bert} as well as images \cite{chen2020generative}. The transformers are an encoder-decoder type architecture similar to the seq2seq architecture with the difference that the encoder and the decoder do not use CNN or RNN but are basically designed on attention mechanisms. They were first introduced by Vaswani et al. \cite{NIPS2017_3f5ee243} in 2017, and they were applied to machine translation and English constituency parsing.

Transformers have two advantages over RNN-based encoder-decoder models. The first is that their execution is highly parallelizable, and even though they generally have a much higher volume of parameters, their computational cost and training times are lower than using RNN architectures. 
The second is that they handle long-range dependencies better, since the length of the paths that forward or backward signals must travel to cross the network are constant in a transformer. On the other hand, in an RNN-based seq2seq architecture, the length of paths are of the order of $O(n)$, where \textit{n} is the length of the input sequence. The shorter these paths, the easier it is to learn long-range dependencies \cite{hochreiter2001gradient}.

\subsection{seq2seq in handwriting}
\label{section:seq2seq in handwriting}

From the success of the seq2seq models in other sequence transduction problems, its application to the offline HTR problem became popular. Additionally, these models tried to overcome certain limitations inherent to CTC-based architectures, such as the fact that the alignment between the input and the output had to be strictly monotonous and that the maximum length of the output sequence was conditioned to the length of the input sequence.

An additional advantage of seq2seq architecture is that the decoder module can potentially learn a language model from input character sequences. This model is limited because of the small size of the available handwriting text databases compared to the volume of text typically used in the construction of general-purpose language models (i.e. usually larger).

The first reference to such seq2seq architecture applied to HTR is probably Bluche et al. \cite{bluche2016scan} in 2016. The model is applied to full paragraph recognition, using a line level pre-trained MDLSTM-CTC auxiliary model as a base for the encoder. 

Later on, in early 2018, Sueiras et al. \cite{sueiras2018offline} proposes a standard seq2seq architecture to the problem of recognition handwritten words images. This architecture is analyzed in depth in Chapter \ref{chapter_word_models}.

In the paper by Kang et al. \cite{kang2018convolve}, the seq2seq architecture is also proposed with a CNN-GRU type encoder, an attention mechanism and a GRU decoder. The GRU components of the encoder and decoder are configured as 2 layers of size 512 after comparative analysis. As CNN component, a VGG-19-BN architecture \cite{simonyan2014very} started with pre-trained weights from ImageNet is selected. Additionally, it is experimented with both content-based and location-based attentions, as well as with attention smoothing techniques, multi-nominal decoding \cite{cho-etal-2014-learning} and label smoothing \cite{szegedy2016rethinking}. In conclusion, the selection of a location attention with the application of the label smoothing technique provided the best results. This same conclusion is obtained by the same authors in \cite{kang2021candidate}. 

Most of the works that apply seq2seq to the HTR problem perform the decoding with the standard greedy decoding algorithm that decodes in each step the class with the highest probability provided by the softmax transformation. As an improvement, Chowdhury and Vig \cite{Chowdhury2018AnEE} propose an alternative decoding algorithm called \textit{beam search} that maximizes the joint probability of the entire output sequence. This is the same algorithm as the one proposed in \cite{scheidl2018word} for the CTC. According to the experiments published in \cite{Chowdhury2018AnEE}, by applying the beam search algorithm to the decoding provides significant performance improvements to the seq2seq model of more than 20 \% in CER and WER on the same model using a greedy decoding.

Another contribution to the decoding process is provided by Kang et al. \cite{kang2021candidate}, who proposed a new way of integrating a language model in the decoder. This technique, called \textit{candidate fusion}. It is an alternative to other language model integration methods such as \textit{shallow fusion} and \textit{deep fusion} \cite{2015arXiv150303535G}.

In reference \cite{Chowdhury2018AnEE}, two additional improvements are proposed over the standard model. The first one is the use of an alternative loss function to the standard cross entropy (CE)\nomenclature{CE}{Cross Entropy} called \textit{focal-loss} \cite{lin2017focal} which solves the problems that the CE loss has with unbalanced samples (where the CE makes that the most frequent and easily classifiable cases tend to dominate the gradients). The second one is the application of batch normalization \cite{ioffe2015batch} in convolutional layers, which has also been used by other authors \cite{kang2018convolve}. In both cases, the modifications represent a significant improvement of the WER and the CER in the experiments carried out with the IAM and RIMES databases.

An extensive analysis of different alternatives for the attention model in the application of the seq2seq to HTR can be found in Michael et al. \cite{michael2019evaluating}. In particular, the following variants were reviewed: \textit{penalized attention} \cite{DBLP:journals/corr/SankaranMAI16} that penalizes feature vectors that have received high attention scores in the past; \textit{monotonic attention} \cite{raffel2017online} that processes the attention in a left-to-right manner: \textit{Monotonic Chunkwise Attention} (MoChA) \cite{chiu2018monotonic}, which relaxes the hard monotonicity condition that splits the input sequence into chunks within which a content based attention is applied. The paper concludes that monotonous and chunkwise attention mechanisms tend to provide better results.
  \index{Monotonic chunkwise attention}
  \nomenclature{MoChA}{Monotonic Chunkwise Attention}

Finally, in 2020, Kang et al. \cite{kang2020pay} proposed for the HTR problem an encoder-decoder type model based on transformers. As indicated in the previous section, the transformers emerge as an alternative to RNN-based seq2seq models to speed up the processing and better handle long dependencies on the input sequences. These models have obtained very good results in several areas, and for the HTR, they also achieve superior results than those obtained with seq2seq architectures. 

The \textit{transformer-type} models are currently a very active research area and, in our opinion, it is one of the most promising work lines of in HTR \cite{kang2020pay}. Another very promising line is the one described in the next section, which it is based on the use of synthetic data and transfer domain strategies. This is intended to compensate the problem caused by the limited size of existing annotated handwritten text image databases.

\section{Synthetic data and domain adaptation}
\label{section: synthetic data and transfer domain}

The use in HTR of increasingly complex and layered neural architectures is conditioned by the limited volume of existing annotated data. The most extensive database available in English is the IAM database \cite{marti2002iam} with only 13,353 lines annotated. This limits the complexity of the models that can be built, which quickly begin to overfit.

There are several known strategies to alleviate these limitations. The most common is data augmentation (discussed in the Section \ref{section: Data_normalization_background}). Briefly, this technique consists of constructing distorted versions of the input images that keep the strict invariant text in them, for example scale changes, rotations, slant or skew transformations or elastic distortions. The technique is simple to implement and quite effective, so it is very popular. The new transformed images are usually built dynamically in each epoch of the training which generates a theoretically infinite sample. This technique has the disadvantage that it cannot generate new texts, only modify existing ones in the train database.

\begin{figure}[!ht]
\centering 
\includegraphics[width=10cm]{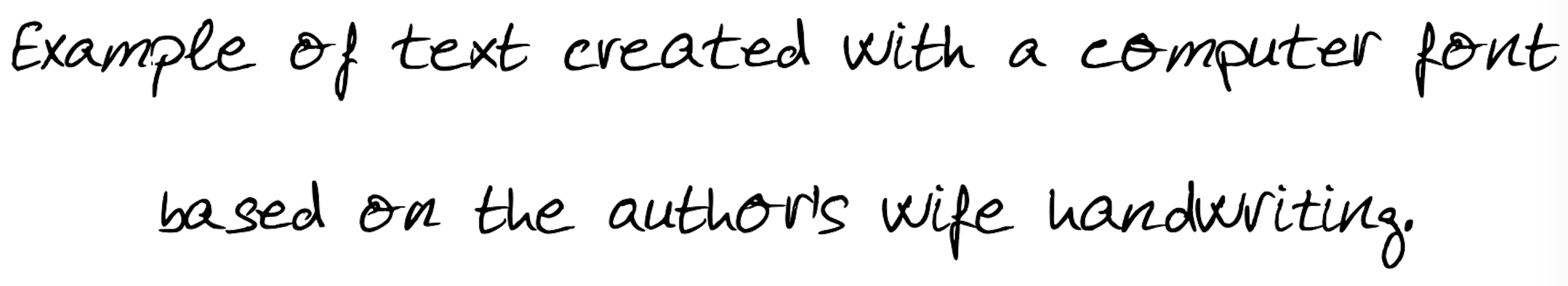} 
\caption{Sample of text written with a typographic font that emulate the handwritten text.}
\label{fig:sample_typograpyc_handwriting_font}
\end{figure}

A second more recent strategy to compensate for the scarcity of annotated data is the use of synthetically generated images, concretely using computer fonts that imitate the handwriting style. In 2014, in the paper by Jaderberg et al. \cite{DBLP:journals/corr/JaderbergSVZ14}, the use of standard typographic fonts is proposed to build a database of synthetic images used in the natural scene text recognition problem. This approach provides very good results in terms of error reduction. Inspired by the above previous papers, Krishnan and Jawahar \cite{Krishnan2016GeneratingSD} proposed the use of typographic fonts that emulate the handwritten text to generate synthetic images that can be used in the HTR problem. To do this, they generated a database of 90,000 word images from 750 handwritten-style fonts with a limited character set. This database has been used in paper \cite{dutta2018improving} to pre-train a CNN-LSTM-CTC architecture with the result of a relative improvement of more than 10 \% in CER and WER by introducing pre-training with synthetic data over a not pre-trained baseline of the same characteristics. Fig. \ref{fig:sample_typograpyc_handwriting_font} provide an example of text written with a typographic font that emulate the handwritten text.

The same pre-training strategy with synthetic data generated from freely available electronic sources that imitate cursive handwriting has been followed in \cite{kang2020pay} at line level, and in \cite{kang2021candidate} at word level. In the case of \cite{kang2020pay} relative improvements close to 40\% are cited in CER and WER at the line level for the IAM database produced by pre-training with synthetic data. 

The use of synthetic images generated from handwritten fonts provides improvements in the cases where it has been used \cite{zhang2019sequence} \cite{kang2020pay}. However, the resulting images have not the enormous variability that the handwritten text has. It also has an artificial regularity in the spatial distribution of the characters that makes a significant domain difference to the actual handwritten text. This generates a shift domain problem, whose solution is based on methodologies to adapt a model trained in an original domain (in this case of synthetic words) to a new domain such as a database of real handwritten text.

The \textit{shift domain problem} is well studied in images \cite{pmlr-v37-ganin15}, although usually the proposed solutions are focused on classification problems. Most use convolutional architectures to project the source and the target domains in a shared space in which the two domains are aligned. The function to be optimized is generally a combination of the error of the original problem and the error in the classification of each sample in its correct domain. This is relatively simple to implement for the case of a multinomial target classification problem, but it cannot be directly applicable to the HTR problem of transcribing an image into a sequence of multiple characters of variable length (precisely due to this characteristic variable dimension of the target).

To solve the above, Zhang et al. \cite{zhang2019sequence} proposed a seq2seq type architecture that included a module called \textit{Gated Attention Similarity}. This module allows aligning the distributions of the source and target domains in a common feature space. On this space, a loss function is defined at the character level that is used to make the model invariant to the domain transformations. 

Another work in the same line is that of Kang et al. \cite{kang2020unsupervised}, where a seq2seq type architecture is also proposed to perform domain adaptation in HTR. To do this, it incorporates a discriminator module of the input image domain into the model. This module includes a temporary pooling step that allows handle sequences of variable length. This paper shows the efficacy of the proposed model to the problem of unsupervised writer adaptation, training the recognition model exclusively with synthetic images generated from typographic sources and applying the domain adaptation module on the destination domain of unlabeled handwriting databases. 
In this way, the efficiency of the algorithm is shown in cases in which there is no annotated information in the destination domain.

\section{Language models background}
\label{c4 - section - Language models}
\index{n-gram language models}

As described in the Section \ref{SubSect:Decoding_predictions}, the \textit{language models} (LM) allow to introduce the context into the decoding process. The most commonly used types of LM in the area of handwriting recognition are the statistical \textit{n-gram language models} \cite{marti2001using} \cite{bertolami2008hidden} \cite{doetsch2014fast}. 

An \textit{n}-gram language model is a probabilistic model that predicts the next word $w_n$ in a sequence as a function of the previous $w_0, ...,w_{n-1}$ words regardless of the word position in the sequence. These models are generally built from a large corpus that allows sufficient repetition of the most common word sequences in a language. In addition, they include a smoothing technique to handle sequences of words not present in the trainig corpus, the most common is the \textit{Kneser-Ney} technique \cite{kneser1995improved}. The details of how these models are built are given in the Section \ref{SubSect:Decoding_predictions}. 

These LM models played an important role in the HMM-based approaches because it was the usual way to obtain the a priori probability value of each word to decode $P(w)$ required in the decoding process with HMM. The type of \textit{n}-gram LMs used is varied, for example in the case of \cite{bertolami2008hidden} a 2-gram statistical language model is used and, in \cite{kozielski2013improvements} and \cite{bluche2014comparison}, both 3-gram and 4-gram LMs are employed. The corpora used for training the \textit{n}-gram LMs are fairly standard and dependent on the handwriting database used.

The \textit{n}-gram LMs implemented by the different authors are at word level. Therefore, they only decode those words that are in the LM training corpus. In order to recognize also those words that are out of vocabulary (OOV), some authors like \cite{kozielski2013improvements} add a second LM at character level that complements the word level LM. Also, in some cases, LMs other than the \textit{n}-gram type have been used. In particular, in \cite{zamora2014neural} a language model based on neural networks instead of an \textit{n}-gram based LM is proposed.

In the approaches based on CTC, LMs are not strictly necessary. Therefore, many authors do not use them directly, providing direct results from the visual model or those obtained after performing a search into lexicons. This eliminates one of the sources of variability when comparing results from different publications, using different LMs by each author. In any case, LMs can be included as part of CTC-based models; for example, a modification of the CTC algorithm called \textit{CTC token passing}, which includes a bi-gram LM in the decoding process, is proposed in \cite{graves2009novel}.

In the case of approaches to the HTR problem based on the seq2seq architecture, LM is generally not used with some exceptions. For example, Kang et al. propose in \cite{kang2021candidate} a method called \textit{candidate fusion} to introduce an LM in an integrated way in the decoder of seq2seq type models. The LM integration results in a WER improvement of 14\% for RIMES database, however for IAM database, the improvement is less than 1\%. Subsequently, in \cite{kang2020pay}  the same authors propose a model for continuous HTR based on transformers and point out that the results with or without LM are almost the same, concluding that the transformer already learns a particular type of LM.





\chapter{Deep architectures for isolated offline character handwriting recognition}\label{c_character}
\label{chapter_character_models}

This chapter describes the models and experiments performed on the problem of recognizing isolated handwritten characters present in images. The main objective is to analyze the performance of different convolutional architectures in the simplest version of the HTR problem: recognizing isolated characters. The aim is to identify the more efficient character recognition models, general enough to be used as part of a word recognition system without segmenting the words into their component characters. The results of this chapter will be used in the next chapter, where a word recognition model is proposed, and the character-level models developed here are applied over a sliding window that runs along with the words to provide evidence of the occurrence of each character sequentially.

The first section introduces the handwritten character recognition problem. Then, in Section \ref{character models - Handwritten character database generation} the creation of a new handwritten character database named COUT is detailed. For this, firstly, in Subsection \ref{subsection: Extraction of characters from UNIPEN online} we describe the process of the database creation. Next, in Subsection \ref{subsection: Transformations applied to curated original images} the curation algorithm proposed to emulate a continuous handwriting context is described. Once the new COUT database has been introduced, the following sections describe the proposed models and the experimental results. Section \ref{character models - Character recognition model} introduces the three convolutional model architectures with which the experiments will be performed. In Section \ref{character models - Experimental results}, all the experimental setup and achieved results are detailed. This section is divided into the following subsections: in Subsection \ref{character models - Experimental results - Baseline results} the baseline results on the MNIST reference database are provided; then, in \ref{character models - Experimental results - Comparative results}, for NIST and \ref{character models - Experimental results - Comparative results TICH} for TICH datasets, respectively results are provided to make a comparison of the proposed architectures with the results of other authors; finally, in Subsection \ref{character models - Experimental results - Detailed results} the detailed results over the new COUT database are presented. This chapter concludes with Section \ref{character models - Results discussion} in which all the obtained results are discussed.

\section{Introduction}
\label{character models - introduction}

The main goal of this Thesis is to provide a general solution to the continuous offline HTR problem, particularly the word recognition problem. For this aim, we propose an architecture with a first component based on convolutional networks, which performs a first step of extraction of the most relevant characteristics of the text images related to the characters which they contain. This first step is considered essential to obtain a solution to the continuous problem. To identify the most suitable convolutional architecture, an analysis of different types of network models for the most basic problem of recognizing isolated characters is carried out.

Building a character-based recognition model that is effective for characters included in a word without segmenting it is complicated since most handwritten character databases contain only isolated characters \cite{deng2012mnist} \cite{guyon1994unipen}. Additionally, existing character databases are limited to upper and lower case characters and digits. They do not include punctuation marks or special characters such as parentheses or brackets present in general handwritten text databases at the word or line level. These isolated character image databases have also been obtained from templates in which the characters have been written isolately. These characters may contain different strokes or stroke shapes than appear when they are written as part of a word. In these cases, ligatures are added, and the characters appear surrounded by artifacts from the strokes of the preceding and the following characters.

To avoid this, we have created a new offline handwritten character database using the UNIPEN online database \cite{guyon1994unipen} that includes all visible ASCII characters, from ASCII 33 (corresponding to the symbol '!') to ASCII 126 (corresponding to the symbol '\textasciitilde{}'). In addition, an algorithm has been defined to include, in a synthetic way, artifacts for each character image. We aim to approximate the effect of the character shape when it is placed inside a word, basically by moving and truncating it. We also have added some artifacts to the left and right of each character. We named this new database as COUT (Characters Offline from Unipen Trajectories).

In order to build our model, we tested new architectures based on CNNs due to the recent impressive results produced by these networks in image classification problems, both in general \cite{krizhevsky2012imagenet} and also in handwriting digit recognition \cite{deng2012mnist}, and character classification problems \cite{zhang2015character}  \cite{yuan2012offline}.

For the experiments, we have selected three architectures that have shown good results in the handwritten character recognition problem in the past \cite{deng2012mnist} \cite{cirecsan2012multi}, and that have been proposed as the first convolutional component in general HTR models \cite{puigcerver2017multidimensional} \cite{kang2018convolve}. These architectures are LeNet \cite{lecun1995comparison}, VGG \cite{simonyan2014very} and ResNet \cite{he2016deep}, respectively. These have also been selected because they have an increasing level of complexity in terms of layers and number of parameters that allow a varied range of experiments to be conducted.

Different handwritten character databases were used for the problem of isolated character handwriting recognition by several authors. For example, Deng \cite{deng2012mnist} uses the MNIST database, Ciresan and collaborators \cite{cirecsan2012multi} built ensembles of CNNs over the NIST database, and Van der Maaten \cite{van2009new} used neural networks from 2 to 5 layers over the TICH database. In order to evaluate the proposed architectures against the results of other authors, experiments have been performed on the same public databases MNIST, TICH, and NIST, respectively.

Finally, systematic experiments have also been performed on the proposed COUT database, including models on different character sets. Extensive error analysis has also been carried out to understand well the limitations of the selected models.

\section{COUT character database generation}
\label{character models - Handwritten character database generation}

A new character recognition model is built, and its generalization capacity using different character sets is validated. Our objective is to use it to develop a new approach to handwritten word identification that avoids the character segmentation problem. To develop it, we need a handwriting character database where images of characters resemble as closest to the corresponding images obtained with a sliding window over a word image (and that is why the current databases are not helpful at all). Thus, we decided to create a new offline handwritten character database to overcome these limitations.

For the construction of this new image database of handwritten characters, we used UNIPEN online handwriting database. This online database contains the information of the $(x,y)$ coordinates of the handwritten text points defining the strokes captured with a digitizing device.

Several reasons to create this new database are enumerated next. First, the UNIPEN database includes a character set extended to symbols and punctuation marks that are not included in other character databases but do appear in continuous handwritten text databases. Second, the UNIPEN database is an online handwritten database of words, lines, and paragraphs segmented at character level. It means that the characters were written as a part of a word. In other character databases, these were written isolatedly in a template. Finally, when building this database, we can control the resolution and the stroke size to get character images as closest as possible to the word-level databases. 

In this section, we describe the steps followed to create the new dataset of handwritten characters. It is specially oriented to test model architectures that work with characters but also as a component of a more general system to automatically recognize handwritten words. 

The COUT database was built in two steps:

\begin{itemize}
  \item First, we extracted isolated characters from the UNIPEN online handwritten database.
  \item Next, we perform several transformations on the initial character images to simulate how these characters can appear inside real words. 
\end{itemize}

\begin{figure}
\centering 
\includegraphics[width=10cm]{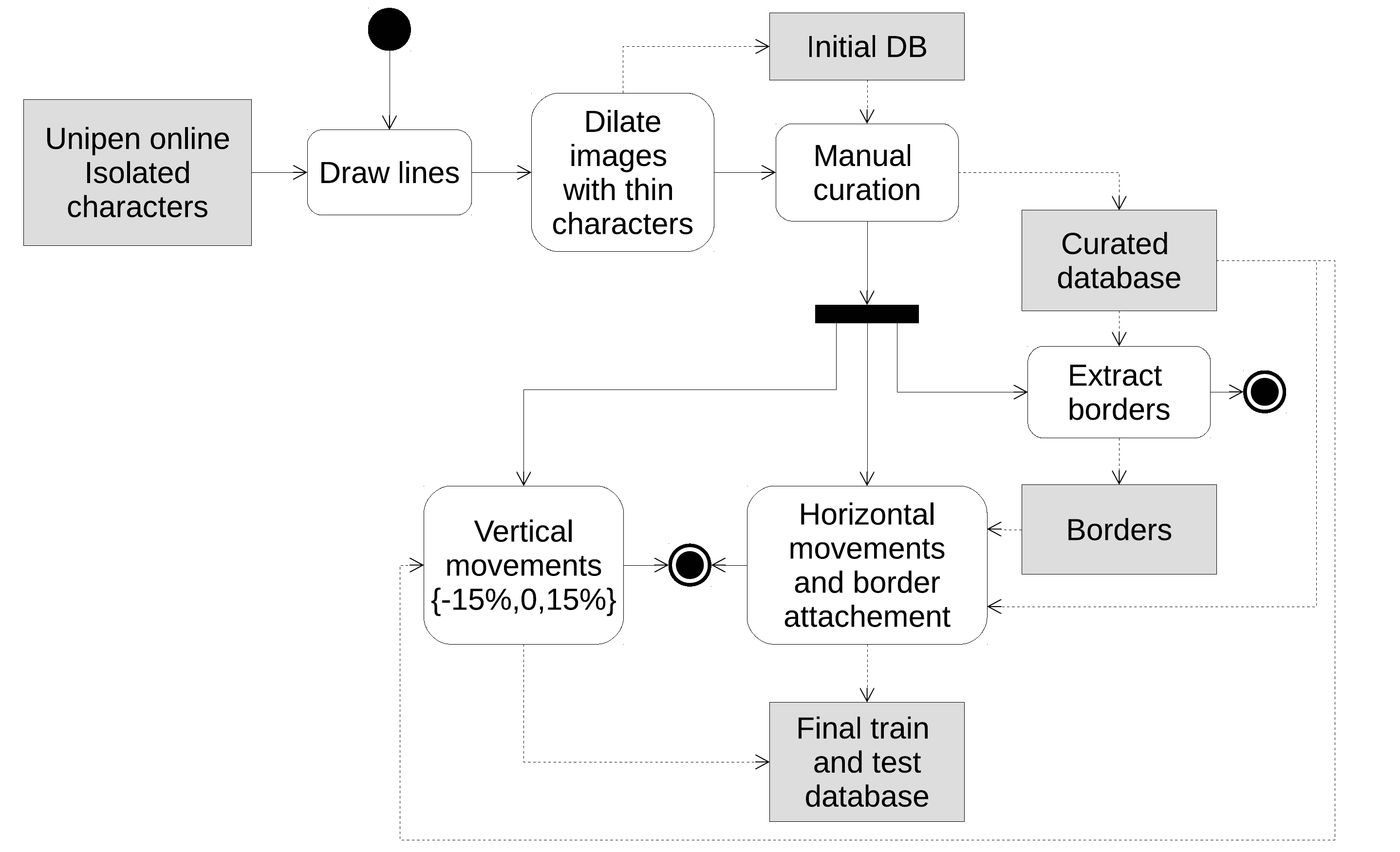} 
\caption{UML activity diagram of creation and transformation process of the character database. Processes are presented as white boxes and outputs as gray boxes.}
\label{fig:charModelDatabaseCreation}
\end{figure}

The UML activity diagram shown in Fig. \ref{fig:charModelDatabaseCreation} illustrates the stages involved for producing the new character database. The next two subsections describe in detail these stages.

\subsection{Extraction of characters from UNIPEN online}
\label{subsection: Extraction of characters from UNIPEN online}

Using this online database to extract the "base" characters, we avoid the character segmentation problem in offline handwritten text, and we also obtain perfect isolated characters. The isolated annotated characters are first identified. Next, these characters are hand-retouched so that the $x$ and $y$ coordinates of pixels are connected in each stroke which shapes the character. Variable-thickness strokes are used depending on the original resolution of the characters to ensure that all final characters have a similar thickness. 

The images from each available category are generated: upper case, lower case, digits, and punctuation marks, respectively. These generated images are resized to 64$\times$64 pixels without changing their aspect ratio or relative size in paragraphs. Finally, the generated images are curated manually, one by one, to make sure that they are assigned to their correct category and that are human-legible. Our database contains 93 categories and 68,382 image characters, where the average number of samples per category is 735. Table \ref{tab:charModelCharactersList} illustrates the types of characters present in our dataset. Our database of curated characters can be downloaded from \url{https://github.com/sueiras/handwritting_characters_database} to facilitate the reproducibility of the experiments. We also provide here a Python code that implements the transformations described in the next subsection.

\begin{center}
\begin{table}[!ht]
\begin{centering}
\begin{tabular}{cccccccccccccccccccc}
! & " &\# &\$ &\% &\& & ' & ( & ) & * & + & , & - & . & / & 0 & 1 & 2 & 3 & 4 \tabularnewline
5 & 6 & 7 & 8 & 9 & : & ; & \textless & = & \textgreater & ? & @ & A & B & C & D & E & F & G & H \tabularnewline
I & J & K & L & M & N & O & P & Q & R & S & T & U & V & W & X & Y & Z & [ & ] \tabularnewline
\textasciicircum{} & \_ & ` & a & b & c & d & e & f & g & h & i & j & k & l & m & n & o & p & q \tabularnewline
r & s & t & u & v & w & x & y & z &\{ & \textbar &\} & \textasciitilde{}\tabularnewline
\end{tabular}
\caption{Characters contained in the COUT database. \label{tab:charModelCharactersList}}
\par\end{centering}
\end{table}
\end{center}

A detailed description of this new database, including sample images and character frequency, was included in Chapter \ref{chapter_Handwriting_Databases}.

\subsection{Transformations applied to original images}
\label{subsection: Transformations applied to curated original images}

\begin{figure}[!ht]
\centering 
\includegraphics[width=10cm]{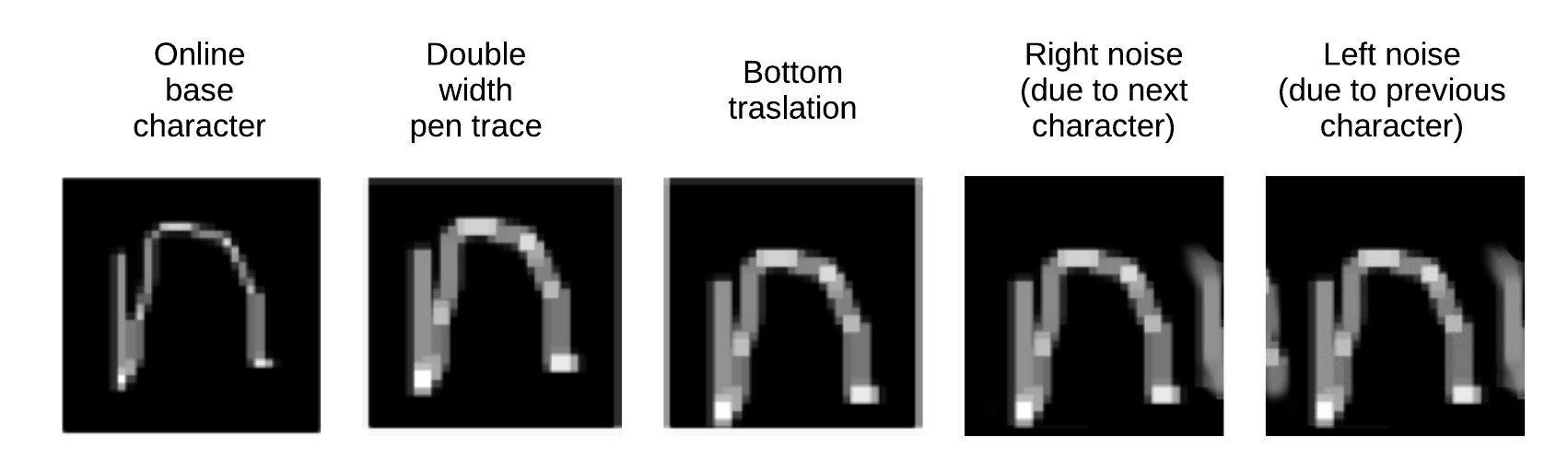}
\caption{Example of transformations chain applied to build a training sample.}
\label{fig:charModelTransformationsExample}
\end{figure}

A set of transformations were applied to the original character images with two goals: augmenting the training set size in experiments and ensuring good results with the models when they are applied on parts of word images (i.e., at the word level where the characters are not isolated). Fig. \ref{fig:charModelTransformationsExample} illustrates some examples of transformations applied to Unipen characters to build the training sample.

The applied transformations are the following ones. First, we resize characters and translate them up or down. Specifically, we reduce the characters by $25\%$ of their original size, and the new image is moved up or down for each character. The main idea is that some characters only can appear up or centered in a word, while others can only appear down and centered, and finally, other ones can appear in any place. This location depends on whether the characters usually cover the ascenders or descenders areas. In this way, the relative vertical positions of the characters match with those when extracted from a continuous line of text. The characters included in each case are:

\begin{itemize}
\item The letters: 'a', 'c', 'e', 'i', 'm', 'n', 'o', 'r', 's', 'u', 'v', 'w', 'x', 'z' are placed up centered and down because they do not cover the ascenders or descenders areas. 
\item The letters: 'g', 'j', 'p', 'q', 'y' are placed centered and down because they cover the descenders area.
\item Finally the letters: 'b', 'd', 'f', 'h', 'k', 'l', 't', 'A', 'B', 'C', 'D', 'E', 'F', 'G', 'H', 'I', 'J', 'K', 'L', 'M', 'N', 'O', 'P', 'Q', 'R', 'S', 'T', 'U', 'V', 'W', 'X', 'Y', 'Z' are placed up and centered because they cover the ascenders area.
\end{itemize}
        
Second, we move left or right the letter image and include in its sides (left, right or both) the beginning or end of another letter. In this way, similar training examples will be available to those that will be found when applied the model to a fragment of a word image. The borders of the characters (with 3 pixels of thickness) are extracted from the available sample to achieve this. These borders are added to the extremes of the existing characters in order to generate a new sample. In this process, the non-overlapping between the characters and the added borders is guaranteed.

\section{Convolutional architectures}
\label{character models - Character recognition model}

\textit{Convolutional Neural Networks} (CNN) have been used for many years for the offline handwritten character recognition problem \cite{lecun1998gradient}. In our experiments, we apply three convolutional architectures with incremental complexity to identify those that may be the most suitable ones as the first component of the word recognition models described in Chapter \ref{chapter_word_models}.
  \index{Convolutional neural network}

\begin{itemize}
  \item The first one is \textit{LeNet} \cite{lecun1995comparison}. It was the first proposed for handwritten digit recognition in 1995. It is characterized by using 5$\times$5 kernels, and having two sets of convolution and pooling layers, with an incremental number of feature maps.
  \index{LeNet}

  \item The second one is \textit{VGG} \cite{simonyan2014very}, which is characterized by proposing a convolution-convolution-pooling layers with 3$\times$3 kernels, and by duplicating the number of feature maps in each module.
  \index{VGG}
  
  \item The third one is the \textit{ResNet} architecture \cite{he2016deep}, which is characterized by proposing direct connections (residuals) between the input and output of each convolutional component. Additionally, it employs the regularization strategy called \textit{batch normalization}\cite{ioffe2015batch}.
  \index{ResNet}
\end{itemize}

In every case, we use zero-padding to preserve the spatial size and a stride of ones in all the convolutional layers. All hidden layers include the \textit{non-linear rectification units} (ReLU) \cite{krizhevsky2012imagenet}. Finally, we use the dropout regularization \cite{hinton2012improving} in the dense layers with a ratio of 0.5. The configuration of each of these architectures used in the handwritten character recognition experiments is described below.

\subsection{LeNet architecture}
\label{c5 - section - Lenet architecture}

The LeNet architecture was originally proposed in 1995 by Yann LeCun and collaborators \cite{lecun1995comparison}. This model alternates convolutional layers and pooling layers, and ends with dense layers. The strategy of alternating convolutional layers and downsampling layers was initially proposed by Fukushima in 1975 in \cite{fukushima1979neural}, and in \cite{lecun1995comparison} is adapted to the problem of recognizing isolated handwritten digits.

\begin{figure}[!ht]
\centering 
\includegraphics[width=12cm]{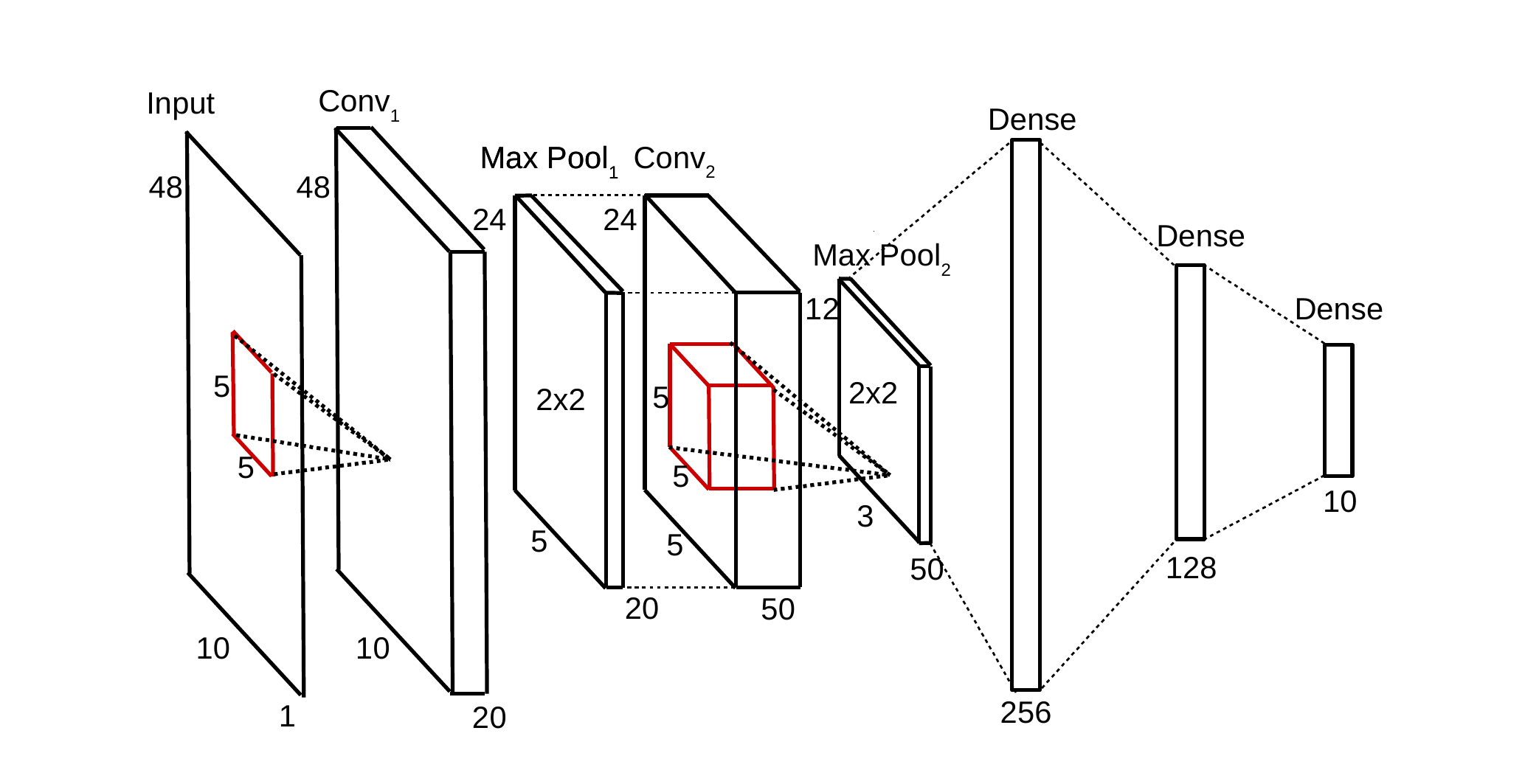} 
\caption{Proposed LeNet architecture schema for the MNIST database.}
\label{fig:charModelLenetArchitecture}
\end{figure}

The original LeNet architecture was composed of four layers organized in two groups of a convolutional layer followed by a pooling layer. Then, in some cases, it was completed with dense layers at the end. Since this architecture was proposed precisely for the handwritten digit recognition problem, we decided not to modify it substantially for the present experiments. It has been taken into account that the number of categories in the current general character recognition problem is much larger. Additionally, the computational limitations of the computers at that time were conditioned by the total number of network parameters that could be handled. Therefore, it has been decided for the current experiments to increase the dense layers dimension and the number of feature maps. Fig. \ref{fig:charModelLenetArchitecture} shows the LeNet architecture selected in the present character recognition experiments. The architecture is characterized by employing two groups of convolutional layers with kernels of size 5$\times$5 and pooling layers with kernels of size 2$\times$2 that reduce the size of the output feature maps by half in each case. The number of feature maps in the first group is set to 20, and in the second group is set to 50. Three dense layers are added at the end. The first two ones are oriented to identify interactions between the output elements of the feature maps. They are sized so that the second is half the first and uses a ReLU type activation. The third one is a classification layer with an output size equal to the number of categories of the target, including a \textit{softmax} activation that allows the network to return as output a probability distribution of the possible targets.

For the different databases used, the size of the dense layers have been modified to adapt them to the different number of categories in the target set. The simplest case is the MNIST database which has only 10 categories, since it is a digit database. For this case, we have set the size of the two dense layers to 256 and 128, respectively. The rest of the databases already have 26 or more categories as they include the letters of the alphabet in upper case (TICH) or upper and lower case (NIST and UNIPEN). In this case, the two dense layers have been configured with sizes of 512 and 256, respectively. Table \ref{table:lenetConfigurationsSummary} includes a detail of the configurations used of this architecture for the different databases and character sets used in the experiments and compares them with the original LeNet-1 and LeNet-5 architectures.

\begin{table}[!ht]
\centering 
\begin{tabular}{l m{1.5cm} l m{1.3cm} m{1.3cm} m{1.2cm} m{1.2cm}}
  \toprule 
  Database & Charset size & Model & Feature maps 1 & Feature maps 2 & Dense layer 1 & Dense layer 2 \\
  \midrule
  MNIST   &          10& LeNet-1 \cite{lecun1995comparison} &   4& 12&   0&  0  \\
  MNIST   &          10& LeNet-5 \cite{lecun1995comparison} &   6& 16& 120& 84  \\
  MNIST    &         10& ours                               &  20& 50& 256& 128 \\
  TICH-NIST-COUT &  +28& ours                               &  20& 50& 512& 256 \\
  \bottomrule 
\end{tabular}
\caption{Configurations of the LeNet architecture used in the different databases.}
\label{table:lenetConfigurationsSummary}
\end{table}

Table \ref{table:lenetLayersSummary} includes a detailed information of the layers for the MNIST database with an input image size of 28$\times$28 and 10 categories. The layer type, the output tensor dimension, and the number of trainable parameters of each layer are indicated. The input tensor size of each layer is the one indicated as output in the previous layer. An input layer has been incorporated in the table indicating the size of model input data to obtain this information in all the layers. The total number of parameters is 687,142 for this case.

\begin{table}[!ht]
\centering 
\begin{tabular}{lrr}
\toprule 
    Layer type&          Output Shape&   Parameters \\
    \midrule 
    input&                (28, 28, 1)&   0 \\
    \arrayrulecolor{gray}  
    \midrule 
    Convolution2D(ReLU)&  (28, 28, 20)&  500 \\
    MaxPooling2D&         (14, 14, 20)&  0 \\
    \midrule 
    Convolution2D(ReLU)&  (14, 14, 50)&  25,000 \\
    MaxPooling2D&         (7, 7, 50)&    0 \\
    \midrule 
    Flatten&              (2,450)&        0 \\
    Dense(ReLU)&          (256)&         627,456 \\
    Dense(ReLU)&          (128)&         32,896 \\
    Dense(Softmax)&       (10)&          1,290 \\
    \arrayrulecolor{black}
    \bottomrule 
\end{tabular}
\caption{Summary of layers, output shape and number of parameters for the proposed LeNet architecture applied over MNIST database.}
\label{table:lenetLayersSummary}
\end{table}

\subsection{VGG architecture}
\label{c5 - section - VGG architecture}

This second architecture is based in the VGG network \cite{simonyan2014very} that is characterized by proposing a convolution-convolution-pooling layers module with 3$\times$3 kernels and by duplicating the feature map number in each module. Thus, the main difference with the previous LeNet architecture is that the convolutional and pooling layer blocks have the following configuration:

\begin{itemize}
  \item A first convolutional layer with a kernel of size 3$\times$3.
  \item A second convolutional layer of kernel 3$\times$3. 
  \item A third pooling layer with a kernel of size 2$\times$2 that reduces the output dimension to half the input dimension.
\end{itemize}

\begin{figure}[!ht]
\centering 
\includegraphics[width=16cm]{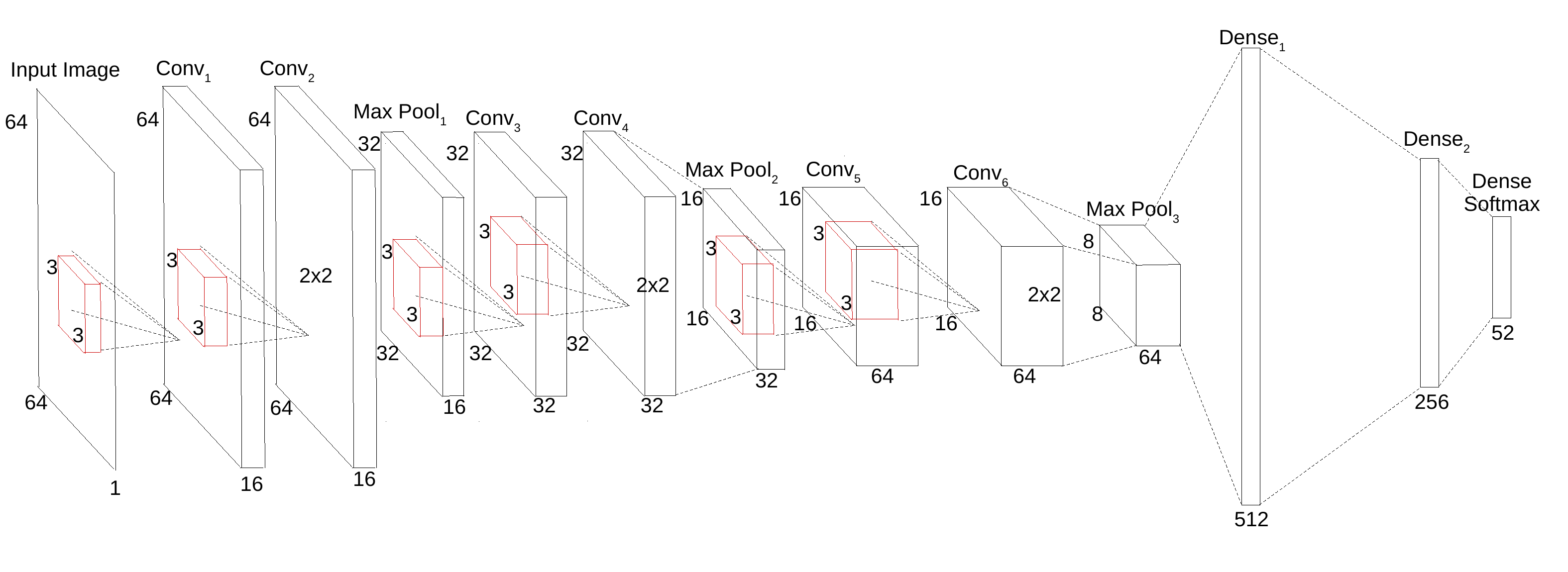} 
\caption{VGG architecture model with the sizes of the layers and the sizes of the convolutional masks (in red). This model is instantiated for an input size of 64$\times$64 and a final dense layer of 52 categories corresponding to the model applied on upper and lower case letters of the COUT database.}
\label{fig:charModelVGGArchitecture}
\end{figure}

The original VGG architecture has between 16 and 19 layers and it is oriented to the more complex task of the ImageNet challenge \cite{deng2009imagenet} that is to classify a color image into 1,000 general categories. Our problem is more limited, and we do not need an architecture so complex. The proposed model has 9 trainable layers, grouped in 3 stacks of convolutional and subsampling layers, and 3 final dense layers similar to those proposed in the previous LeNet-type architecture. The detailed architecture is shown in Fig. \ref{fig:charModelVGGArchitecture}.

The options of 2, 3 and 4 convolutional blocks have been evaluated using the subset of images of upper and lower case letters of the new COUT database to determine the number of convolutional blocks with which to perform the experiments. Table \ref{table:vggBlocksExperiments} summarizes the results of accuracy in the validation and test partitions of the experiments performed, from which the configuration with 3 convolutional blocks is selected as the one that obtains the highest accuracy in the tests.

\begin{table}[!ht]
\centering 
\begin{tabular}{lcc}
  \toprule 
  Num. VGG blocks& \multicolumn{2}{c}{Accuracy} \\
 \cmidrule(r){2-3} 
  & Validation(\%) & Test(\%) \\
  \midrule 
  2 &  84.77&  85.62 \\
  3 &  86.83&  \textbf{87.41} \\
  4 &  \textbf{87.12}&  87.20 \\
  \bottomrule 
\end{tabular}
\caption{Accuracy of the VGG architecture over the COUT upper and lower case database. Comparative over the number of blocks.}
\label{table:vggBlocksExperiments}
\end{table}

Finally, the Table \ref{table:vggLayersSummary} includes some detail of the layers for the case of the subset of upper and lower case letters of the COUT database with an input image size of 64$\times$64 and 52 classes. The layer type, the output tensor size and the number of trainable parameters of each layer are detailed. The input tensor size of each layer is the one indicated as output in the previous layer. An input layer has been incorporated to the table indicating the size of the model input data to obtain this information in all layers. The total number of parameters is 4,625,492 for this case.

\begin{table}[!ht]\footnotesize
\centering 
\begin{tabular}{lrr}
    \toprule 
    Layer type&          Output Shape&   Parameters \\
    \midrule 
    input&                (64, 64, 1)&     0 \\
    \arrayrulecolor{gray}  
    \midrule 
    Convolution2D(ReLU)&  (64, 64, 32)&    288 \\
    Convolution2D(ReLU)&  (64, 64, 32)&    9,216 \\
    MaxPooling2D&         (32, 32, 32)&    0 \\
    \midrule 
    Convolution2D(ReLU)&  (32, 32, 64)&    18,432 \\
    Convolution2D(ReLU)&  (32, 32, 64)&    36,864 \\
    MaxPooling2D&         (16, 16, 64)&    0 \\
    \midrule 
    Convolution2D(ReLU)&  (16, 16, 128)&   73,728 \\
    Convolution2D(ReLU)&  (16, 16, 128)&   147,456 \\
    MaxPooling2D&         (8, 8, 128)&     0 \\
    \midrule 
    Dense(ReLU)&          (512)&           4,194,816 \\
    Dense(ReLU)&          (256)&           131,328 \\
    Dense(Softmax)&       (52)&            13,364 \\
    \arrayrulecolor{black}
    \bottomrule 
\end{tabular}
\caption{Summary of layers, output shape and number of parameters for the selected VGG architecture applied on upper and lower case letters of the COUT database.}
\label{table:vggLayersSummary}
\end{table}

\subsection{ResNet architecture}
\label{c5 - section - Resnet architecture}

The \textit{Residual Networks} (ResNet)\nomenclature{ResNet}{Residual Networks} was proposed in 2015 by He el al. \cite{he2016deep} as a strategy to overcome the limitations when trying to train models with a large number of layers. The strategy used is to define convolutional modules that have a built-in direct connection between the module input and output. The output of the convolutional module is combined with its input, and this combination is passed as input to the next convolutional module. In this way, in the mentioned paper 150-layer models are trained. 

These networks are a particular case of the \textit{Highway Networks} previously proposed by Srivastava et al. in \cite{srivastava2015training}. The Highway Networks include in each convolutional module a gate that decides when the input to the module must be concatenated with its output. Thus, a ResNet is a Highway Net whose gates are always open.

\begin{figure}[!ht]
\centering 
\includegraphics[width=14cm]{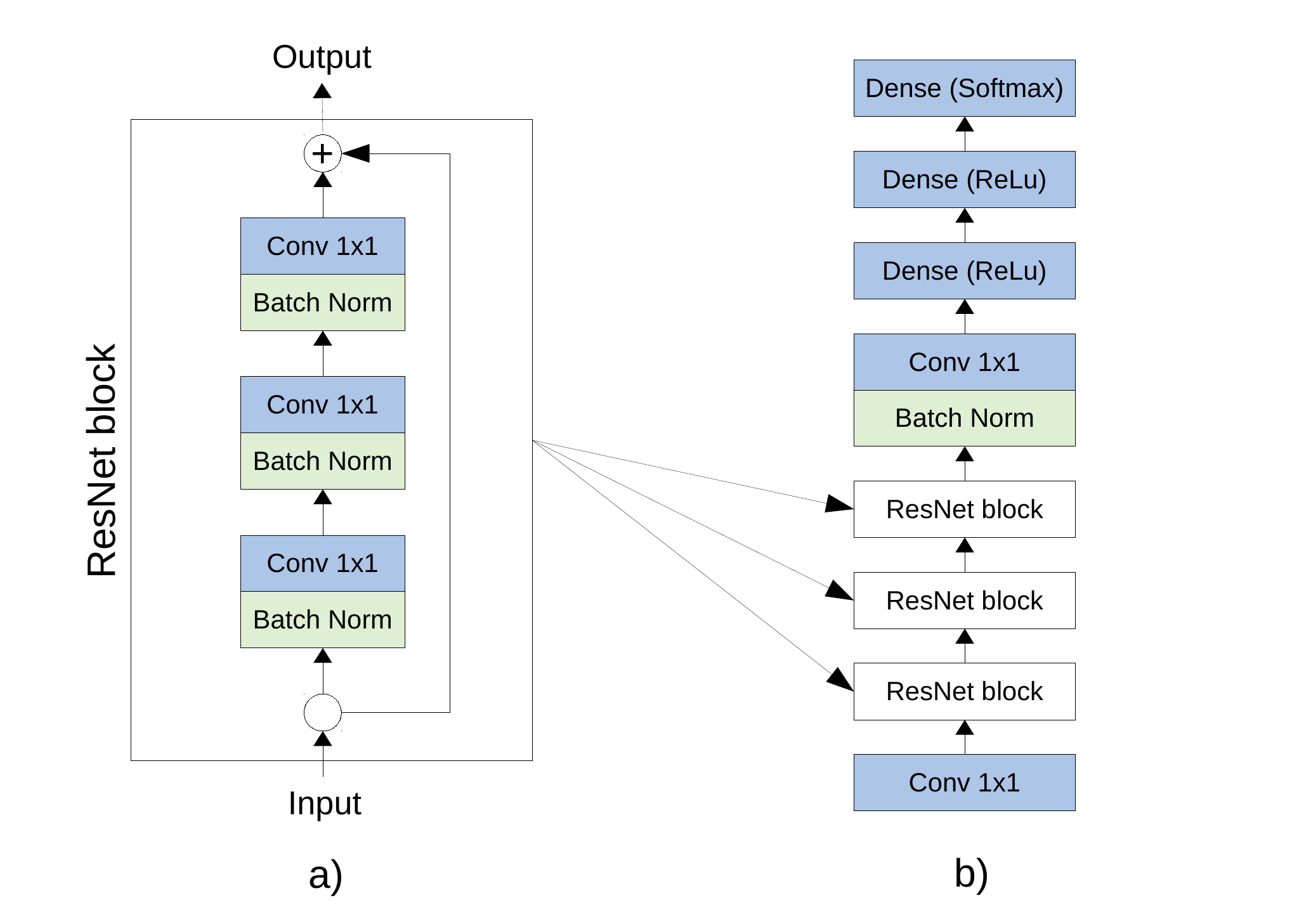} 
\caption{ResNet architecture: a) schema of a ResNet module and b) schema of the complete architecture with 3 modules.}
\label{fig:charModelResnetArchitecture}
\end{figure}

The proposed architecture scheme is similar to the two previous cases. Several convolutional layer modules are followed by 3 dense layers. The difference is that in the case of the ResNet architecture, each convolutional module is more complex than those proposed before. Fig. \ref{fig:charModelResnetArchitecture} shows the detail of components of a ResNet module. The module is composed of the following layers:

\begin{itemize}
  \item A first convolutional layer with a kernel of size 1$\times$1 with 16 output feature maps.
  \item A second convolutional layer of kernel size 3$\times$3 with 16 output feature maps.
  \item A third convolutional layer with a kernel of size 1$\times$1 with 64 output feature maps.
\end{itemize}

All of the above convolutional layers are preceded by a batch normalization layer and employ ReLU activations. On top of the above, each block adds a direct connection between the input and output of the module, which is called \textit{residual} and from which the name of the architecture derives.

In the original proposal, ResNet was oriented to the classification of natural images into 1,000 categories, and architectures of up to 150 layers were proposed. However, in our experiments with character databases, which are less complex, we decided to use only 3 blocks. The number of blocks has been fixed after performing experiments with a different number of blocks on the NIST database for classification of upper case letters. Table \ref{table:resnetBlocksExperiments} summarizes the accuracy results in the validation and test partitions of the experiments performed.

\begin{table}[h!]
\centering 
\begin{tabular}{l c c}
 \toprule 
 Num. ResNet blocks & \multicolumn{2}{c}{Accuracy} \\
 \cmidrule(r){2-3} 
  & Validation(\%) & Test(\%) \\
 \midrule 
 1 &  95.72&  92.74 \\
 2 &  95.36&  91.85 \\
 3 &  95.82&  \textbf{95.97} \\
 4 &  \textbf{96.25}&  93.67 \\
 5 &  96.09&  93.78 \\
 \bottomrule 
\end{tabular}
\caption{Accuracy with the ResNet model over the NIST upper case database. Comparative over the number of blocks.}
\label{table:resnetBlocksExperiments}
\end{table}

Fig. \ref{fig:charModelResnetArchitecture} shows the ResNet architecture used in the experiments for the handwritten character databases. On the left, the detail of each of the convolutional blocks is included, composed of convolutional layers and batch normalization layers \cite{ioffe2015batch} as indicated above. Additionally, the residual connection linking the input to the output is observed. The complete architecture is shown on the right, which includes 3 convolutional blocks.

Finally, Table \ref{table:resnetLayersSummary} includes a detail of the layers for the case of the upper case letters image subset of the NIST database, with 64$\times$64 input image size and 26 classes. The layer type, the output tensor dimension, and the number of trainable parameters of each layer are detailed. The size of the input tensor of each layer is the one indicated as output in the previous layer. The model is sequential except for the add-type layers that configure the residual connection. In these cases, the input layers to be added are indicated. The total number of model parameters is 8,531,242.

\begin{table}[!ht]\footnotesize
\centering 
\begin{tabular}{llrr}
\toprule 
Layes name&  Layer type&   Output Shape&   Parameters \\
\midrule 
input&  InputLayer&            (64, 64, 1)&    0 \\
    \arrayrulecolor{gray}  
    \midrule 
    conv0&  Convolution2D(ReLU)&   (64, 64, 16)&   144 \\
    \midrule 
    bn0&    BatchNormalization&    (64, 64, 16)&   64 \\
    conv1a& Convolution2D(ReLU)&   (32, 32, 8)&    128 \\
    bn1b&   BatchNormalization&    (32, 32, 8)&    32 \\
    conv1b& Convolution2D(ReLU)&   (32, 32, 8)&    576 \\
    bn1c&   BatchNormalization&    (32, 32, 8)&    32 \\
    conv1c& Convolution2D(ReLU)&   (32, 32, 16)&   144 \\
    add1&   Add conv1c and conv0&  (32, 32, 16)&   0  \\
    \midrule 
    bn2a&   BatchNormalization&    (32, 32, 16)&   64   \\
    conv2a& Convolution2D(ReLU)&   (32, 32, 8)&    128   \\
    bn2b&   BatchNormalization&    (32, 32, 8)&    32   \\
    conv2b& Convolution2D(ReLU)&   (32, 32, 8)&    576   \\
    bn2c&   BatchNormalization&    (32, 32, 8)&    32   \\
    conv2c& Convolution2D(ReLU)&   (32, 32, 16)&   144   \\
    add2&   Add conv2c and add1&   (32, 32, 16)&   0   \\
    \midrule 
    bn3a&   BatchNormalization&    (32, 32, 16)&   64   \\
    conv3a& Convolution2D(ReLU)&   (32, 32, 8)&    128   \\
    bn3b&   BatchNormalizationV1)& (32, 32, 8)&    32   \\
    conv3b& Convolution2D(ReLU)&   (32, 32, 8)&    576   \\
    bn3c&   BatchNormalization)&   (32, 32, 8)&    32   \\
    conv3c& Convolution2D(ReLU)&   (32, 32, 16)&   144   \\
    add3&   Add conv3c and add2&   (32, 32, 16)&   0   \\
    \midrule 
    bn4a&   BatchNormalization&    (32, 32, 16)&   64   \\
    conv4a& Convolution2D(ReLU)&   (32, 32, 8)&    128   \\
    bn4b&   BatchNormalization&    (32, 32, 8)&    32   \\
    conv4b& Convolution2D(ReLU)&   (32, 32, 8)&    576   \\
    bn4c&   BatchNormalization&    (32, 32, 8)&    32   \\
    conv4c& Convolution2D(ReLU)&   (32, 32, 16)&   144   \\
    add4&   Add conv4c and add3&   (32, 32, 16)&   0   \\
    \midrule 
    bnF& (BatchNormalization&      (32, 32, 16)&   64   \\
    \midrule 
    Flatten& Flatten&              (16384)&        0       \\
    dense1& Dense(ReLU)&           (512) &         8,389,120 \\
    dense2& Dense(ReLU)&           (256)&          131,328 \\
    output& Dense(Softmax)&        (26)&           6,682 \\
    \arrayrulecolor{black}
    \bottomrule 
\end{tabular}
\caption{Summary of layers, output shape and number of parameters for the ResNet architecture applied over the NIST upper case database.}
\label{table:resnetLayersSummary}
\end{table}

\section{Experimental results}
\label{character models - Experimental results}

This section details the experimental results obtained by applying the selected convolutional architectures on the considered databases. The results obtained are detailed next:

\begin{itemize}
  \item Baseline results on the MNIST database, which allows having a first validation of the models on a well-known basic database.

  \item Comparative results. It includes the experiments carried out with the TICH and NIST databases. These databases are more complex, and published results are available to evaluate the models compared to models of other authors.

  \item Detailed results over the new COUT database to evaluate the behavior of the models with images of all types of characters, including special characters and punctuation marks. It allows us to evaluate the models in a scenario similar to that found in a sliding window strategy applied to the word recognition problem.

\end{itemize}

The proposed models have been applied to different character dictionaries. In particular, for the MNIST database, the 10 available digits have been used. We have also selected the available characters for the TICH database, which are digits and capital letters. In the case of the NIST database, separate models have been built for upper case, lower case and the combination of both types of letters to compare us with other authors. Finally, for the new COUT database, models have been built with the previous sets, plus one with the entire set of 93 characters, including letters, digits, punctuation marks, and other characters. Although the characters used in all cases are Latin, the methodology presented can be used on other alphabets with minimal changes and starting with a character corpus for the considered language.

The training has been performed using a learning rate of 0.0001 and the optimization algorithm RMSProp \cite{mukkamala2017variants}. A primary data augmentation strategy consisting of the following random transformations has also been employed: direct and inverse rotations up to 15 degrees, horizontal and vertical translations up to 20\%, shear up to 15\%, and zoom in and out up to 20\%. A basic data augmentation strategy has been chosen because the objective of these experiments is to compare different architectures, not to obtain the best possible result. As indicated in \cite{ciresan2011convolutional}, a more elaborate data augmentation strategy, including for example elastic deformations, would improve the overall accuracy of all models evaluated. 
\index{RMSProp algorithm}

The models are trained with an \textit{early stopping} strategy until the validation error does not improve in 20 epochs. The best validating model is selected as the final model and it is used to evaluate the test data.

The main metric used to evaluate the results of character recognition models, given that they are multi-category classification models, is accuracy. Accuracy is defined as the percentage of cases correctly classified by the model over the total number of cases in an evaluation partition. A case is classified by a probabilistic model in the class in which it presents a higher output probability. Additionally, to better understand the errors made by each model, several confusion matrices described in Section \ref{section:Metrics to evaluate character recognition models} have been included. Since the number of target categories is large, only the rows and columns of the confusion matrix with significant errors will be shown.
\index{Metrics!Accuracy}

\subsection{Baseline results on MNIST database}
\label{character models - Experimental results - Baseline results}
  \index{Databases!MNIST}

The MNIST database is one of the most widely used in the analysis of Computer Vision models, and many publications use it in their benchmarks of new algorithms or methodologies (see for example \cite{Mazzia2021EfficientCapsNetCN}, \cite{wan2013regularization}, \cite{kowsari2018rmdl}, or \cite{cirecsan2012multi}).

As indicated above, the particular configurations of the architectures for the experiments with this database have been adjusted, taking into consideration the reduced number of categories of the database. In particular, the first two dense layers have been set to 256 for the first one, and 128 for the second one. 

The results obtained by the three architectures in terms of accuracy are included in the Table \ref{table:mnistAccuracyResults}. It can be seen that the VGG architecture obtains the best result with an accuracy of 99.57\%. In any case, the three models obtain good results, especially considering that no hyperparameter optimization has been performed to obtain them, and it can be considered that all these models are valid for this problem.

\begin{table}[!ht]
\centering 
\begin{tabular}{lcc}
 \toprule 
 Architecture & \multicolumn{2}{c}{Accuracy} \\
 \cmidrule(r){2-3} 
  & Validation(\%) & Test(\%) \\
 \midrule 
 LeNet&   99.15&  99.28 \\
 VGG&     \textbf{99.51}&  \textbf{99.57} \\
 ResNet&  98.49&  98.69 \\
 \bottomrule 
\end{tabular}
\caption{Accuracy by architecture over the MNIST database.}
\label{table:mnistAccuracyResults}
\end{table}

The best results on the MNIST database with a convolutional architecture are reported in \cite{wan2013regularization}, which reports an accuracy of 99.79\% using a new optimized dropout strategy called \textit{dropconnect}. State-of-the-art is obtained by Mazzia et al. in \cite{Mazzia2021EfficientCapsNetCN} with an accuracy of 99,84\% employing a type of model called \textit{capsule networks}.

\begin{figure}[!ht]
\centering 
\includegraphics[width=10cm]{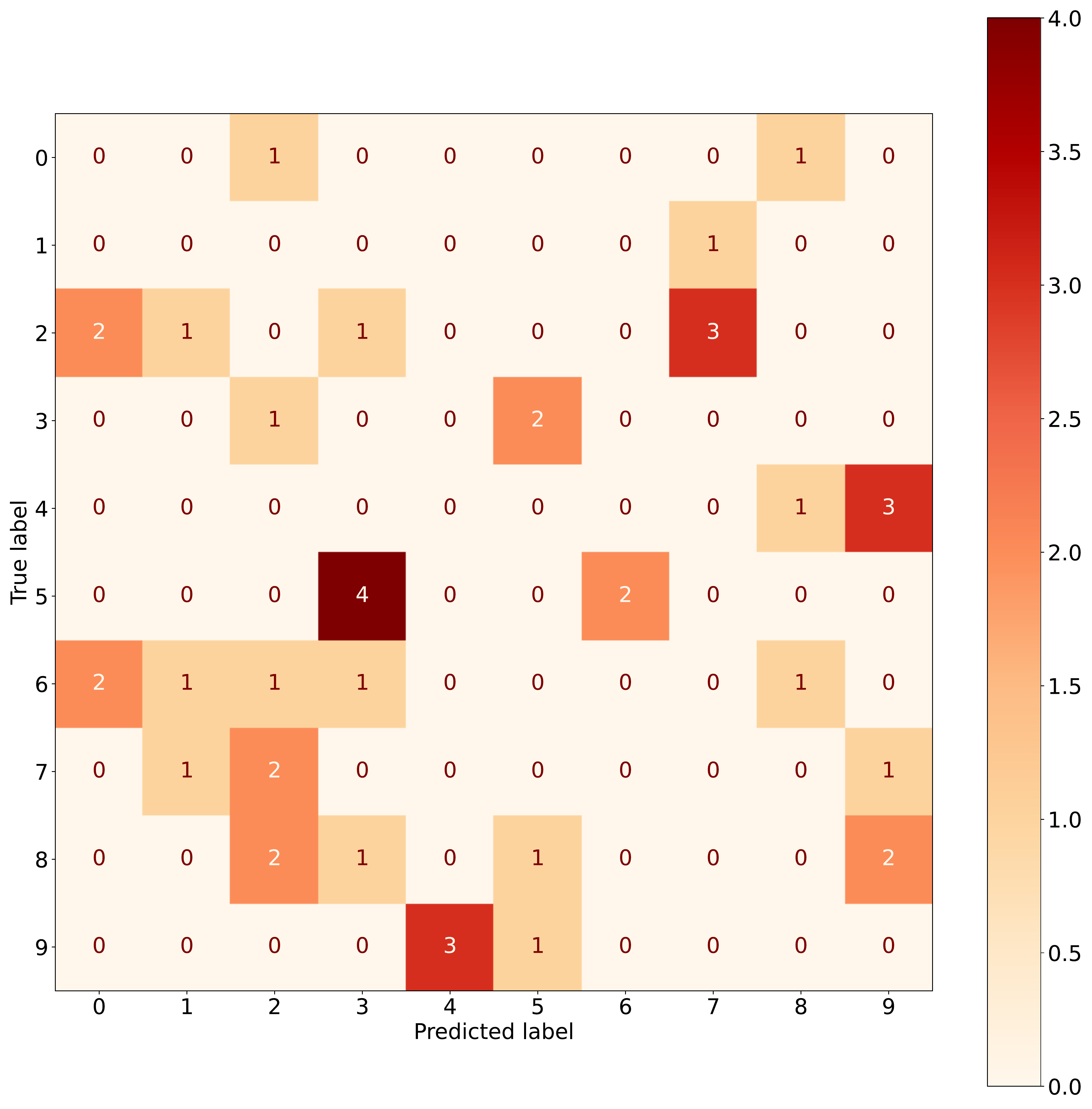} 
\caption{Confusion matrix (only errors) of the VGG model over the test partition of MNIST database.}
\label{fig:cm_mnist_vgg}
\end{figure}

As shown in Fig. \ref{fig:cm_mnist_vgg}, the most frequent error is when confusing characters '9' and '4', and characters '3' and '5'. In the VGG model, these errors account each one for 14\% of the total errors (6 out of 43 in both cases) in the test set. In the rest of the models, this error is also the most frequent one.

\subsection{Comparative results on NIST database}
\label{character models - Experimental results - Comparative results}
\index{Databases!NIST}

To determine whether the proposed models are competitive for character recognition over the whole vocabulary, we use the specific databases NIST and TICH. These two databases are different since NIST has the particularity that the images are binary, with only zero or one pixel values. Moreover, the TICH database contains grayscale images.

In the NIST database, experiments have been carried out with the three proposed architectures by setting the size of the first two dense layers to 512 and 256, respectively. The size has been doubled with respect to the experiments performed with the MNIST database, since the target size increases from 10 categories to a minimum of 26 categories. Results are provided for three subproblems summarized: 
\begin{itemize}
  \item Upper case characters recognition with 26 categories.
  \item Lower case characters recognition with 26 categories.
  \item Upper and lower case characters recognition with 52 categories.
\end{itemize}

The upper and lower case model that we have built to classify the character images has 52 letter categories (i.e., 26 upper case and 26 lower case ones). These categories are specially difficult because several letters like: 'c', 'f', 'i', 'j', 'k', 'o', 'p', 'u', 'v', 'w', 'x' or 'z' have similar shape for upper and lower case. 

There exist several benchmarks for using the NIST database (2nd edition) \cite{grother2016nist}. One of the most accurate ones that use deep networks is \cite{cirecsan2012multi}, which provided recognition rates of $98.17\%$ for upper case and $92.53\%$ for lower case, respectively.

Other results are provided by Ciresan et al. \cite{ciresan2011convolutional}, where a convolutional network architecture model with two convolution-pooling groups and a dense layer very similar to those proposed in this Thesis is used. In this work multiple trainings of the same architecture are executed, and committees of five trainings are used to analyze the error stability. Tables \ref{table:NISTUppertAccuracyResults} and \ref{table:NISTLowertAccuracyResults} included the results of this publication obtained on average by the committees and also the results of the best individual trainings.

In this database, the test sample does not follow the same distribution as the train and valid partitions. In the case of train and validation, there are certain characters that have a higher frequency, up to 10 times more than the less frequent ones, as is the case of the characters 'O' and 'S'. On the other hand, in the test partition, all characters have a similar frequency, about 400 cases per category. It makes the accuracy calculated for tests significantly lower than that obtained for the validation. In any case, the relative comparison between different test accuracy values is correct.

\subsubsection{Results on NIST dataset for upper case letters}
\label{subsubsection:Results over the NIST upper case}

Table \ref{table:NISTUppertAccuracyResults} includes the results of the upper case letter classification models from the NIST database. In this case, the best results are obtained by the VGG architecture, although the LeNet model obtains similar results. The ResNet model shows lower accuracy results.

\begin{table}[!ht]
\centering 
\begin{tabular}{llcc}
 \toprule 
 Publication & Model &\multicolumn{2}{c}{Accuracy} \\
 \cmidrule(r){3-4} 
 & & Validation(\%) & Test(\%) \\
 \midrule 
 Ciresan 2011 \cite{ciresan2011convolutional}& Committee &  &  98,09 \\
 Ciresan 2011 \cite{ciresan2011convolutional}& Best model&  &  97.49 \\
 Ciresan 2012 \cite{cirecsan2012multi}       &           &  &  \textbf{98.17} \\
 Ours&  LeNet&   97.48&  96.13 \\
 Ours&  VGG&     \textbf{98.49}&  97.35 \\
 Ours&  ResNet&  95.37&  91.85 \\
 \bottomrule 
\end{tabular}
\caption{Accuracy by architecture over the NIST upper case dataset.}
\label{table:NISTUppertAccuracyResults}
\end{table}

\begin{figure}[!ht]
\centering 
\includegraphics[width=10cm]{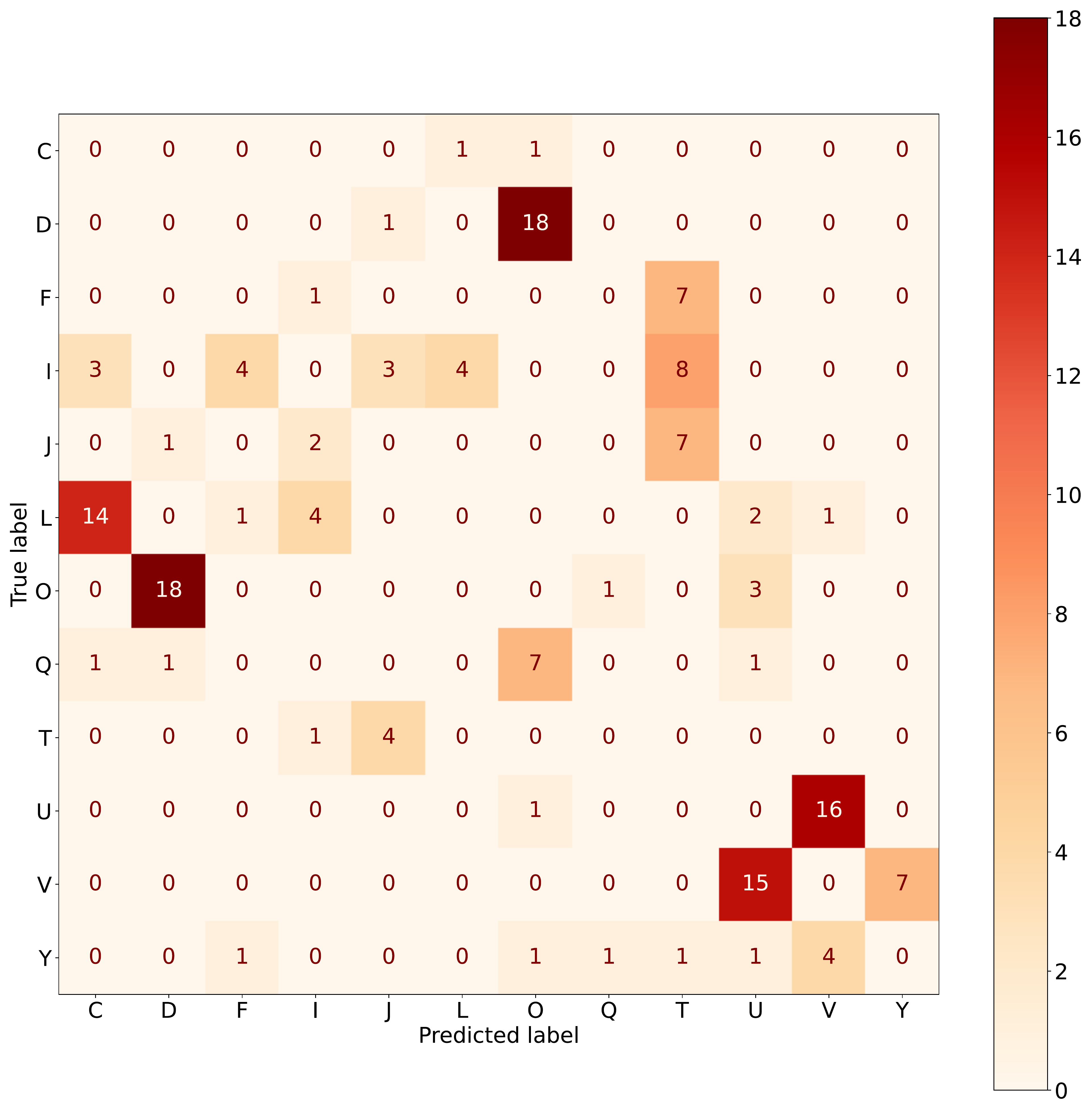} 
\caption{Truncated confusion matrix (only errors) of the VGG model over the test partition of NIST upper case letters.}
\label{fig:cm_NIST_upper_vgg}
\end{figure}

Fig. \ref{fig:cm_NIST_upper_vgg} shows a part of the confusion matrix (including only errors) of the model in which the most frequent errors of the model occur. It can be seen that the most common errors fall in the case of the upper case letters for VGG model occur when confusing the characters 'D' and 'O' (36 test cases), when confusing the characters 'U' and 'V' (31 test cases), and when assigning as 'C' 14 characters that are really 'L' ones.

In the upper case, the result of the VGG model is quite close to the published results \cite{ciresan2011convolutional} \cite{cirecsan2012multi}, even though the two works cited above seek to minimize the error. For this purpose, these works optimize the preprocessing of the database, apply a complex data augmentation, and adjust both the model design and the training algorithm. All these steps are applied in order to minimize the error. The objective of the present study is not to optimize the above parameters to obtain the best possible result but to evaluate three different architectures to determine their suitability for using them in continuous HTR models. It partly explains why the results obtained are a bit worse than those published.

Upper case letters are easier to identify than lower case letters, and upper case letter shapes have less variation because they are well defined and have smaller intra-class variability (because there is less stroke variations in the different writing styles).

\subsubsection{Results on NIST dataset for lower case letters}
\label{subsubsection:Results over the NIST lower case}

Table \ref{table:NISTLowertAccuracyResults} presents the results of the lower case letter classification models from the NIST database are included. The results are similar to the previous ones, with the VGG model obtaining the best result and the LeNet and ResNet models obtaining a worse results, especially ResNet.

\begin{table}[!ht]
\centering 
\begin{tabular}{llcc}
 \toprule 
 Publication & Model &\multicolumn{2}{c}{Accuracy} \\
 \cmidrule(r){3-4} 
 & & Validation(\%) & Test(\%) \\
 \midrule 
 Ciresan 2011 \cite{ciresan2011convolutional}& committee&     &  92,29 \\
 Ciresan 2011 \cite{ciresan2011convolutional}& best model&    &  91.16 \\
 Ciresan 2012 \cite{cirecsan2012multi}       & Convolutional& &  \textbf{92.53} \\
 Ours& LeNet&   95,05&  87,30 \\
 Ours& VGG&     \textbf{96.60}&  90.58 \\
 Ours& ResNet&  92.99&  83.12 \\
 \bottomrule 
\end{tabular}
\caption{Accuracy by architecture over the NIST lower case dataset.}
\label{table:NISTLowertAccuracyResults}
\end{table}

\begin{figure}[!ht]
\centering 
\includegraphics[width=10cm]{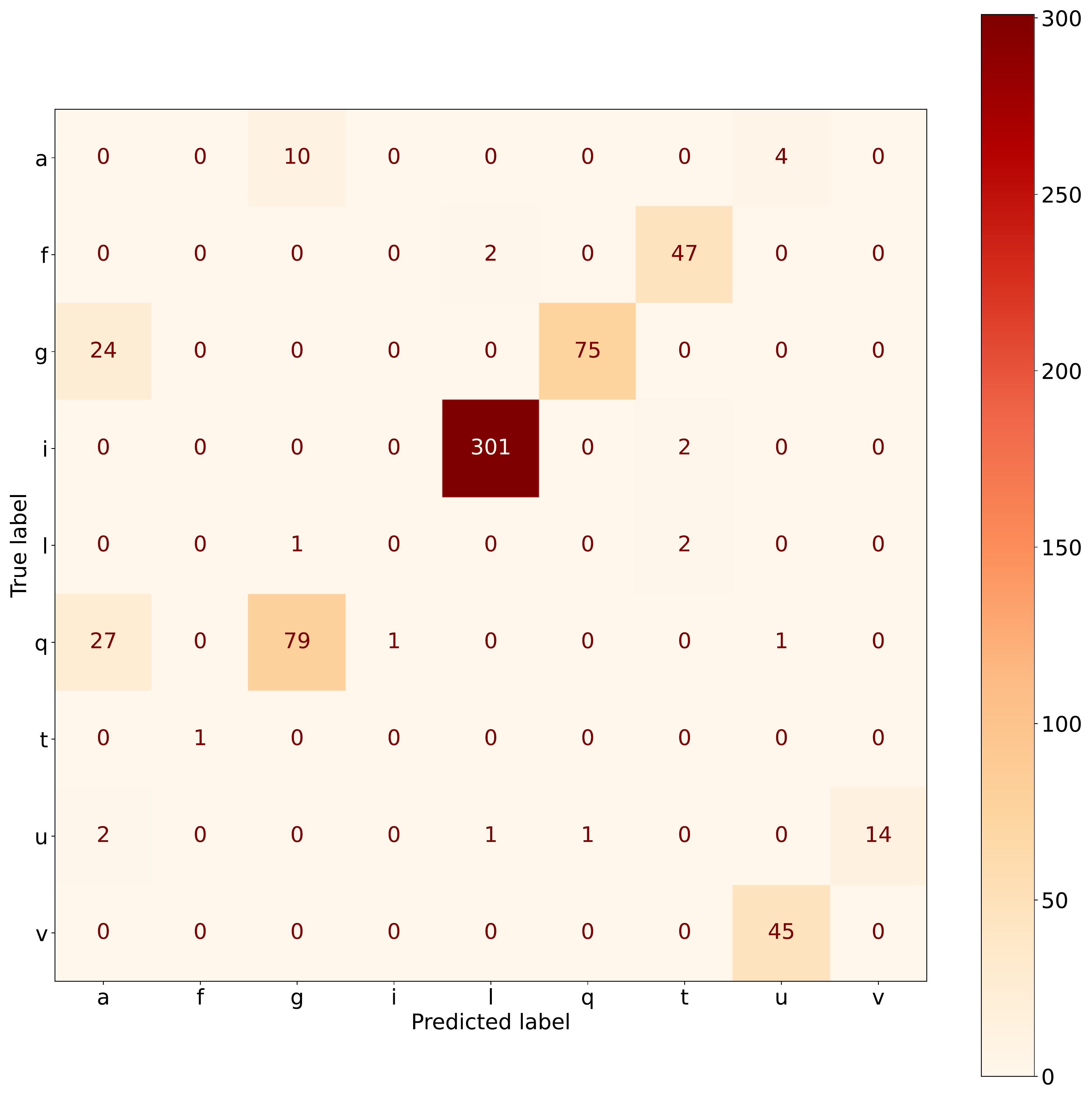} 
\caption{Truncated confusion matrix (only errors) of the VGG model over the test partition of NIST lower case letters.}
\label{fig:cm_NIST_lower_vgg}
\end{figure}

The best performing model (i.e. VGG) also achieves an accuracy relatively close to the results of the individual models of other authors. Fig. \ref{fig:cm_NIST_lower_vgg} shows the confusion matrix of this model. Again, only the rows or columns with relevant errors are shown.

The main source of error comes from classifying most of the images of the character 'i' as the character 'l' (el). It is also important to remark that there are the main errors when confusing 'g' and 'q', 'f' and 't', and 'u' and 'v'.

The full confusion matrix shows that practically all 'i' characters are classified as 'l'. This is because these characters are very similar and that in the train partition, there are 15,213 cases of the 'l' character and only 2,517 of the 'i' character. This difference in frequency is reflected in the biases associated with each target class of the last classification layer. This means that any image of the 'i' character on which the model does not achieve a precise identification ends up by classifying it as 'l'.

\subsubsection{Results on NIST dataset for upper and lower case letters}
\label{subsubsection:Results over the NIST upper and lower case}

Finally, Table \ref{table:NISTUpperLowertAccuracyResults} includes the results on NIST database for upper and lower case letters. This is the most complex case since there are several characters with similar upper and lower case spellings. In particular, the characters: 'C', 'F', 'I', 'J', 'K', 'L', 'M', 'O', 'P', 'S', 'U', 'V', 'W', 'X', 'Y', and 'Z' have that particularity and are usually distinguished by their relative size compared to other character.

\begin{table}[!ht]
\centering 
\begin{tabular}{llcc}
 \toprule 
 Publication & Model &\multicolumn{2}{c}{Accuracy} \\
 \cmidrule(r){3-4} 
 & & Validation(\%) & Test(\%) \\
 \midrule 
 Ciresan 2011 \cite{ciresan2011convolutional}& committee&   &  \textbf{78,59} \\
 Ciresan 2011 \cite{ciresan2011convolutional}& best model&  &  76.99 \\
 Ours& LeNet&   85.70&  68.13 \\
 Ours& VGG&     \textbf{87.56}&  70.86 \\
 Ours& ResNet&  82.16&  64.53 \\
 \bottomrule 
\end{tabular}
\caption{Accuracy by architecture over the NIST upper and lower case letters dataset.}
\label{table:NISTUpperLowertAccuracyResults}
\end{table}

The VGG architecture once again obtained the best results. In this case, the accuracy of models is further away from those obtained by other authors. 

\begin{figure}[!ht]
\centering 
\includegraphics[width=12cm]{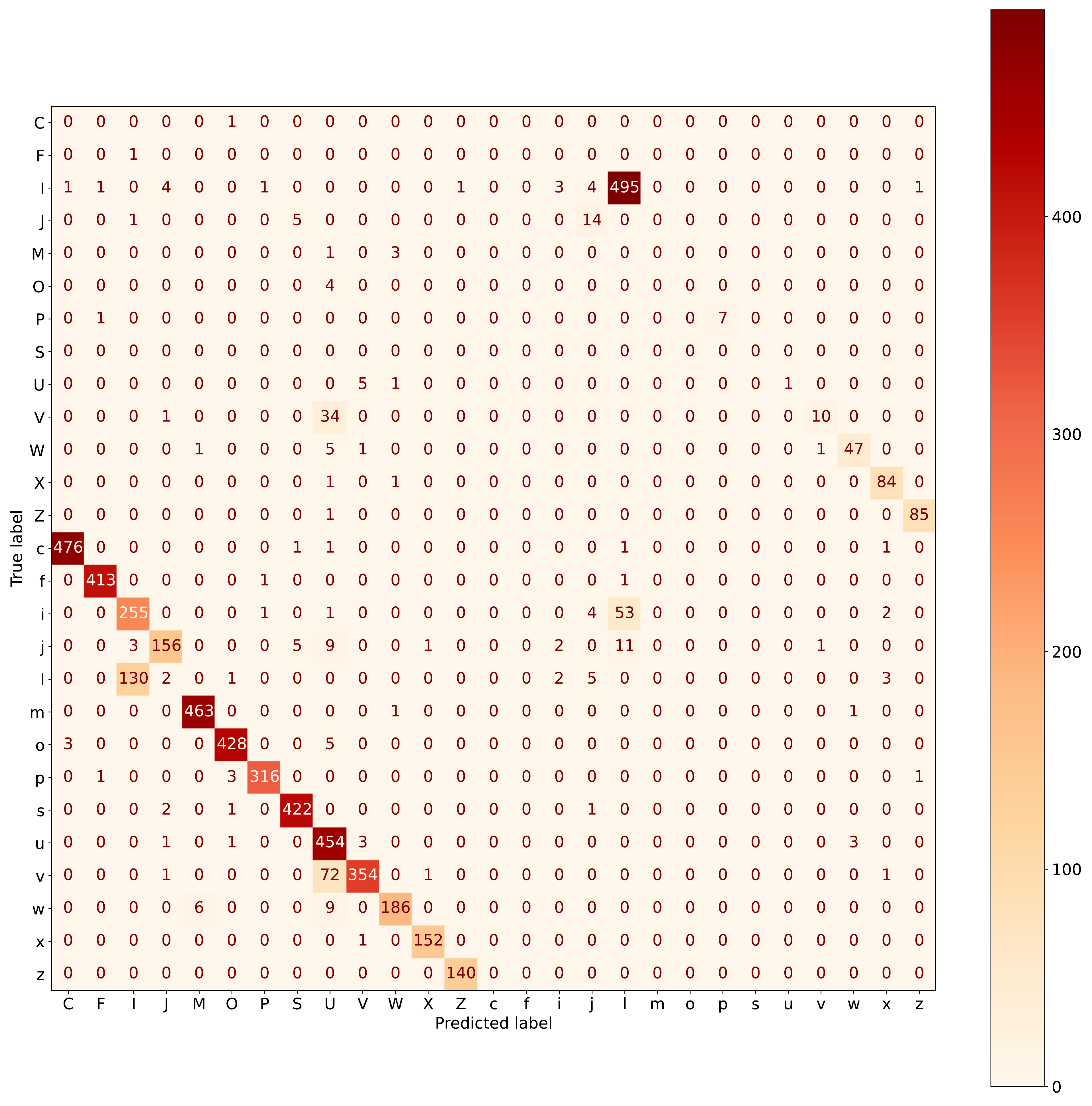} 
\caption{Truncated confusion matrix (only errors) of the VGG model over the test partition of NIST upper and lower case letters.}
\label{fig:cm_NIST_all_vgg}
\end{figure}

Fig. \ref{fig:cm_NIST_all_vgg} shows that most frequent errors occur precisely when confusing the same character in upper and lower case. In particular, and by error volume, with the characters: 'I', 'C', 'M', 'U', 'O', 'S', 'V', 'P', 'W', 'J', 'X', and 'Z' respectively.

Where creating the NIST database, all characters are resized to the same size, and then a blank border is added to them. It eliminates the possibility of differentiating between characters of the same size in the upper and lower cases, and it explains the high number of errors in this confusion matrix.

\subsection{Comparative results on TICH database}
\label{character models - Experimental results - Comparative results TICH}
\index{Databases!TICH}

The TICH database contains grayscale images of the upper case letter and digits characters with 36 categories. In Table \ref{table:TICHAccuracyBenchmark} the results obtained by the three proposed architectures are detailed. A baseline proposed by the database author is also included in reference \cite{van2009new}.

\begin{table}[!ht]
\centering 
\begin{tabular}{llcc}
 \toprule 
 Publication & Model &\multicolumn{2}{c}{Accuracy} \\
 \cmidrule(r){3-4} 
 & & Validation(\%) & Test(\%) \\
 \midrule 
 Van der Maaten \cite{van2009new} & k-NN  &  & 82.77$\pm$0.82 \\
 Ours &              LeNet  &   94.91 &  95.35 \\
 Ours &              VGG    &   \textbf{95.94} &  \textbf{96.54} \\
 Ours &              ResNet &   95.04 &  94.91 \\
 \bottomrule 
\end{tabular}
\caption{Accuracy by architecture over the TICH database.}
\label{table:TICHAccuracyBenchmark}
\end{table}

Using the TICH database \cite{van2009new}, Van der Maaten provided a best error recognition result of $82.77\% \pm 0.82\%$ in recognition using 10-$fold$ cross-validation procedure and a $k$-NN classifier. In this case, the comparative result corresponds to a baseline provided by the database author based on $k$-neighbors which is not optimized, and which the convolutional models outperform by far.

\begin{figure}[!ht]
\centering 
\includegraphics[width=10cm]{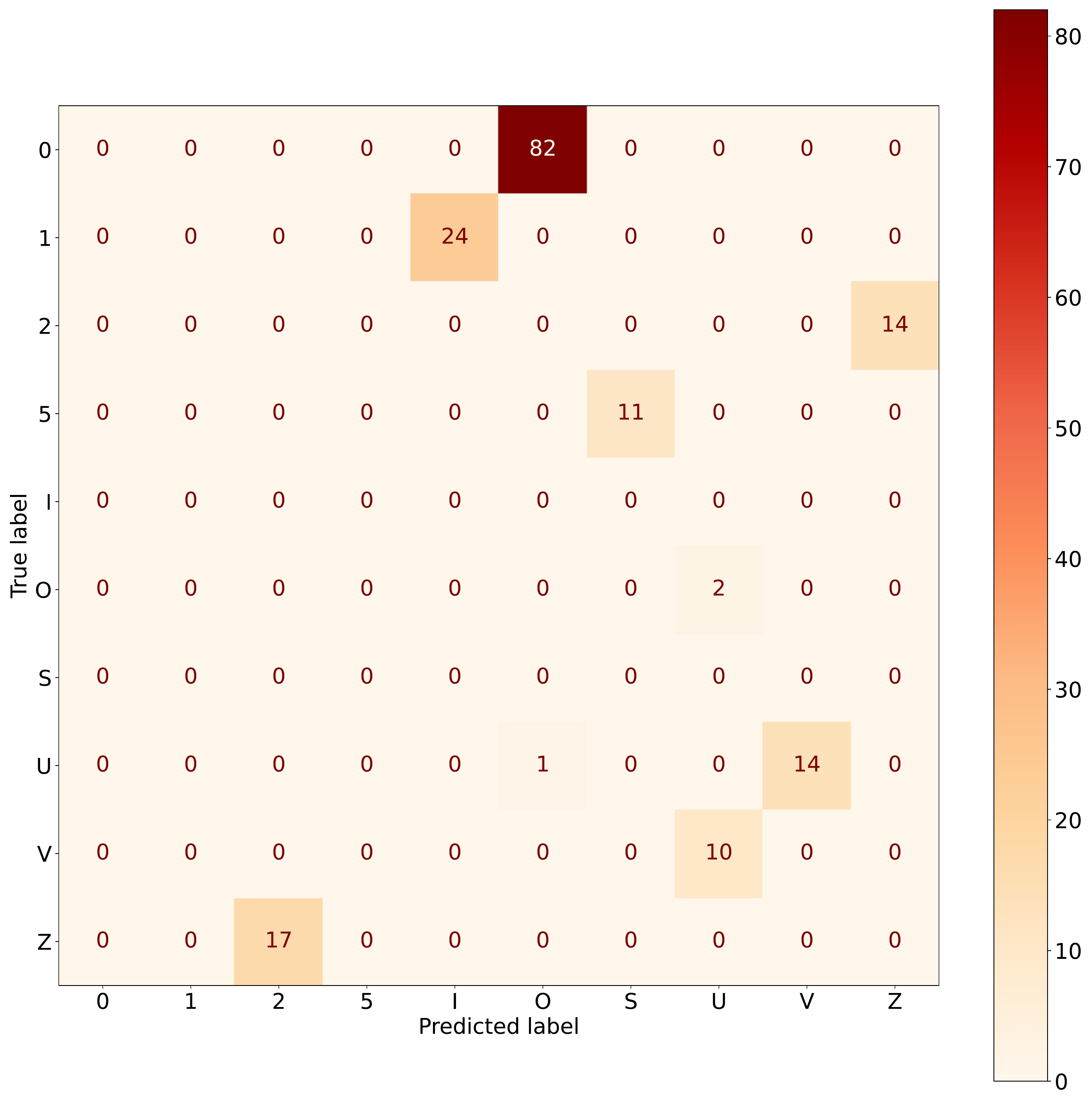} 
\caption{Truncated confusion matrix (only errors) of the VGG model over the test partition of TICH database.}
\label{fig:cm_tich_vgg}
\end{figure}

In the case of TICH dataset, the architecture with the lowest test error is the VGG with an accuracy of 95.54\%. The results obtained with this database, which includes the 10 digits and 26 upper case letters, are very similar to those obtained for upper case letters in the NIST database detailed in Table \ref{table:NISTUppertAccuracyResults}. The fact that the results, in this case, are lower comes from the fact that classifying the digits attached to upper case letters introduces more sources of error. In particular, the confusion between the characters 'O' and '0' (i.e., letter O and digit zero) and between the characters 'I' and '1' (letter I and digit one) are the two primary sources of error, as seen in the confusion matrix in Fig. \ref{fig:cm_tich_vgg}. This figure shows the section of the confusion matrix that includes the main errors of the model.

\subsection{Results on the new COUT database}
\label{character models - Experimental results - Detailed results}
  \index{Databases!COUT}

In this section, the experimental results on several subsets of the new COUT database (Characters Offline from UNIPEN Trajectories database) are presented. This database is built by us from character stroke trajectory data extracted from the UNIPEN online database.

In particular, classification experiments have been performed on the following subsets of characters:

\begin{itemize}
  \item Upper case letters: with 26 categories (results presented in Table \ref{table:UNIPENUpperAccuracyResults}).
  \item Lower case letters: with 26 categories (results presented in Table \ref{table:UNIPENLowerAccuracyResults}).
  \item Upper and lower case letters: with 52 categories (results presented in Table \ref{table:UNIPENUpperLowerAccuracyResults}). 
  \item All characters in the database: including upper and lower case letters, digit, punctuation and other symbols: with 93 categories (results presented in Table \ref{table:UNIPENAllAccuracyResults}).
\end{itemize}

The model that contains all the characters include a total of 93 classes: All the upper and lower case letters, the 10 digits, and the characters: '!', '\#', '\$', '\%', '\&', '(', ')', '*', '+', ',', '-', '.', '/', ':', ';', '\textless', '=', '\textgreater', '?', '@', '{]}', '\textasciicircum{}', '\_', '`', '\{', '|', '\}', '\textasciitilde{}'.

For each of the experiments, two runs of each experiment have been carried out. The first one (i.e. original data) uses the original images from the COUT database. In the second one (i.e. altered data), the training partition is transformed following the algorithm presented in Subsection \ref{subsection: Transformations applied to curated original images}, which modifies the images of each character by adding artifacts that simulate traces of previous and subsequent characters. This process simulates the effect of having been extracted them by a sliding window that runs through a handwritten word. In this way, it will be tested whether the proposed architectures are robust to the presence of these artifacts. In this second case, the algorithm is applied with a multiplicative factor of 8 that increases the training sample by this factor. That is, for each image of a character, 8 distorted versions of that image are generated. The validation and test partitions are not modified to obtain comparable results.

\subsubsection{Results on COUT for upper caseletters}

As shown in the Table \ref{table:UNIPENUpperAccuracyResults}, the test accuracy for upper case characters is similar in the models built on the original images and on the altered ones. It indicates that the ability to discriminate upper case letters is not reduced by the artifacts added to the left and right character images. The accuracy is also similar to that obtained in the experiments with the TICH and NIST databases.

\begin{table}[!ht]
\centering 
\begin{tabular}{llcc}
 \toprule 
Model& Altered &\multicolumn{2}{c}{Accuracy} \\
 \cmidrule(r){3-4} 
 & & Validation(\%) & Test(\%) \\
 \midrule 
  LeNet&   Yes &   94.66 &  95.44 \\
       &    No &   96.87 &  96.67 \\
 \midrule 
  VGG&     Yes &   96.55 &  97.41 \\
     &      No &   \textbf{98.31} &  \textbf{98.34} \\
 \midrule 
  ResNet&  Yes &  85.94 &  86.07 \\
       &    No &  96.55 &  97.23 \\
 \bottomrule 
\end{tabular}
\caption{Accuracy by architecture over the new COUT upper case dataset.}
\label{table:UNIPENUpperAccuracyResults}
\end{table}

The VGG model obtains the best results on the two datasets. The ResNet model performs well on the original image database. However, on the altered database, the performance drops significantly. The most frequent errors occur between the letters 'U' and 'V', and between the letters 'M' and 'N', respectively.

\subsubsection{Results on COUT for lower case letters}
The results corresponding to the lower case character set are detailed in the Table \ref{table:UNIPENLowerAccuracyResults}.

\begin{table}[H]
\centering 
\begin{tabular}{llcc}
 \toprule 
Model& Altered &\multicolumn{2}{c}{Accuracy} \\
 \cmidrule(r){3-4} 
 & & Validation(\%) & Test(\%) \\
 \midrule 
  LeNet&   Yes &   89.83 &  89.48 \\
       &    No &   95.20 &  95.09 \\
 \midrule 
  VGG&     Yes &   93.03 &  92.82 \\
     &      No &   \textbf{96.04} &  \textbf{96.09} \\
 \midrule 
  ResNet&  Yes &  68.70 &  69.58 \\
       &    No &  92.12 &  92.52 \\
 \bottomrule 
\end{tabular}
\caption{Accuracy by architecture over the new COUT lower case dataset.}
\label{table:UNIPENLowerAccuracyResults}
\end{table}

The model with the highest accuracy is again the VGG. Moreover, the difference with the other two models is quite significant in the modified database. The ResNet architecture model obtains worse results than the other two ones, as it was the case in the experiments performed on the lower case subset of the NIST database.

In the lower case letters, test results are superior to those obtained with the NIST database for the same character set. However, this happens because in the NIST database the test results were biased due to the difference of target distribution in that partition. Comparing with the results in the validation partition of the NIST database, which has a target distribution equal to that of the train, it is observed that the results for the two databases are very similar. For example, the VGG model in NIST obtains an accuracy of 96.6\% in validation, and in the original COUT database, it obtains 96.04\%. 

There are more significant differences between the original and the modified database results in this lower case instance than in the upper case. It is especially important in the ResNet model, which drops its accuracy from 92.52\% to 69.58\%.

\begin{figure}[!ht]
\centering 
\includegraphics[width=10cm]{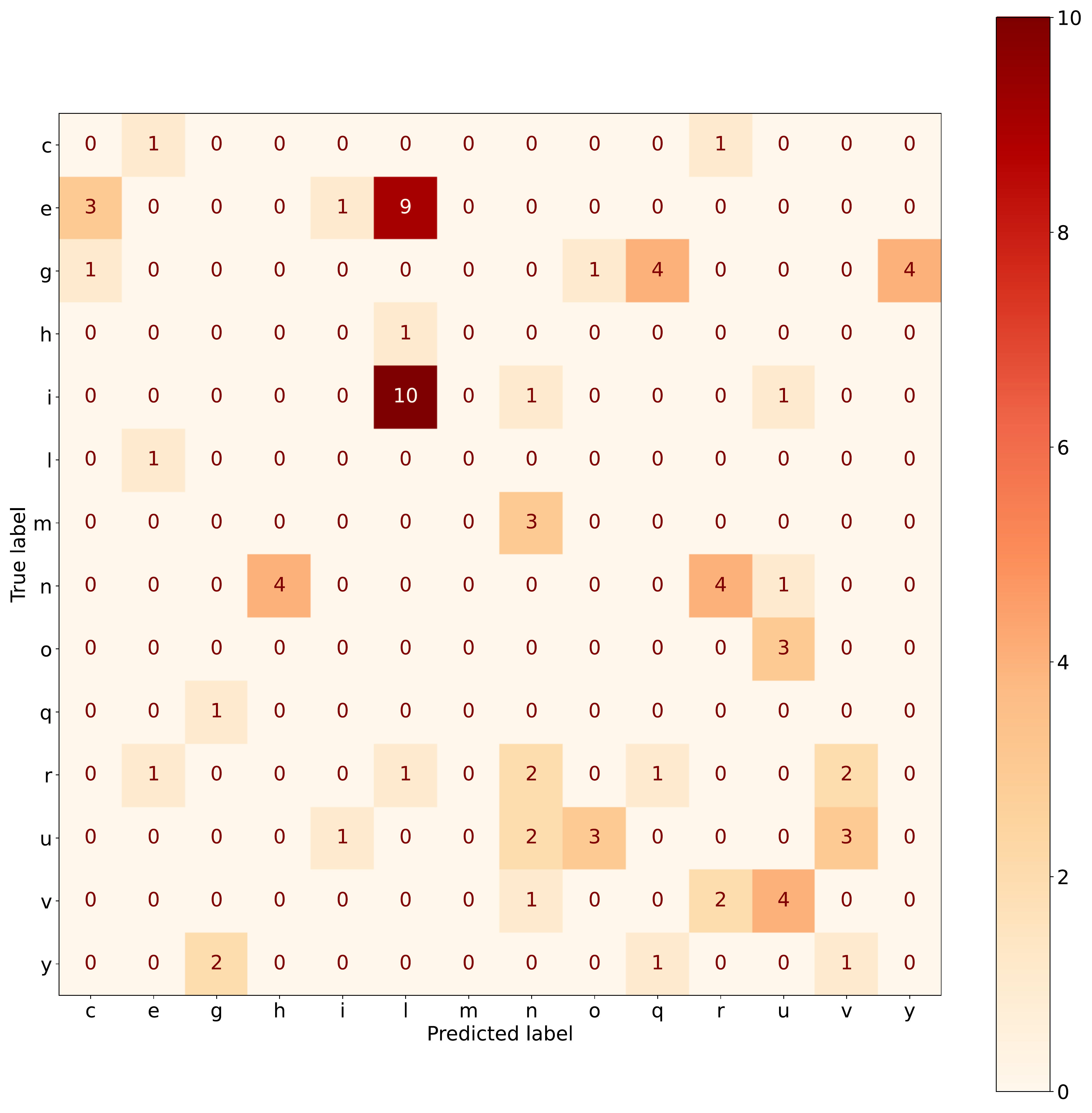} 
\caption{Truncated confusion matrix (only errors) of the VGG model over the test partition of COUT lower case letters.}
\label{fig:cm_COUT_lower_vgg}
\end{figure}

As it can be seen in the confusion matrix shown in Fig. \ref{fig:cm_COUT_lower_vgg} the main errors come from confusing the characters 'e' and 'i' with the character 'l', and confusing the characters 'u' and 'v'.

\subsubsection{Results on COUT for upper and lower case letters}
The results corresponding to the charset of upper and lower case letters are detailed in the Table \ref{table:UNIPENUpperLowerAccuracyResults}. 

\begin{table}[!ht]
\centering 
\begin{tabular}{llcc}
 \toprule 
Model& Altered &\multicolumn{2}{c}{Accuracy} \\
 \cmidrule(r){3-4} 
 & & Validation(\%) & Test(\%) \\
 \midrule 
  LeNet&   Yes &   76.74 &  77.47 \\
       &    No &   84.50 &  85.66 \\
 \midrule 
  VGG&     Yes &   81.48 &  82.37 \\
     &      No &   \textbf{87.03} &  \textbf{87.20} \\
 \midrule 
  ResNet&  Yes &  55.91 &  56.51 \\
       &    No &  80.89 &  82.56 \\
 \bottomrule 
\end{tabular}
\caption{Accuracy by architecture over the new COUT upper and lower case dataset.}
\label{table:UNIPENUpperLowerAccuracyResults}
\end{table}

The conclusions are very similar to those obtained with the lower case model above. The VGG model is the one that obtains the best results. It also happens that the results in the validation partition are very similar to those obtained by each of the architectures on the same NIST database charset. Finally, also ResNet model is the one that shows more differences when trained with the distorted database.

As in the NIST database, the main errors come from confusing the characters with similar graphism for upper and lower case letters. However, in this case, the errors are less frequent, probably because in COUT database construction the relative differences in the size of the characters have not been altered, as it was the case in the NIST database. It allows the model to better differentiate characters with identical spelling in upper and lower cases.

\subsubsection{Results on COUT for all the character set}
To conclude this subsection, the experimental results with the images of the 93 characters included in the database are summarized. Only results from the original database are included since the implemented data augmentation algorithm only applies to upper and lower case letters (i.e. the altered data is not available for this character set). The results are collected in the Table \ref{table:UNIPENAllAccuracyResults}.

\begin{table}[!ht]
\centering 
\begin{tabular}{lcc}
 \toprule 
Model &\multicolumn{2}{c}{Accuracy} \\
 \cmidrule(r){2-3} 
 & Validation(\%) & Test(\%) \\ \midrule 
  LeNet&   83.86 &   83.62 \\
 \midrule 
  VGG&     \textbf{86.22} &   \textbf{86.17} \\
 \midrule 
  ResNet&  78.52 &   78.81 \\
 \bottomrule 
\end{tabular}
\caption{Accuracy by architecture over the new COUT database}
\label{table:UNIPENAllAccuracyResults}
\end{table}

The differentiation of the 93 characters of the COUT database is probably the most complex problem among those addressed in this chapter. In addition to the similarities already analyzed between certain upper and lower case letters, there are also similarities between punctuation marks and other special characters such as: '(', '[', '\{', ')', ']' or '\}'.

The VGG model obtains the best outcome, and the results of LeNet and VGG models are very similar to those obtained in the previous experiment with the character set of upper and lower case letters. It shows that the recognition of special characters does not present significant errors.

\begin{figure}[!ht]
\centering 
\includegraphics[width=12cm]{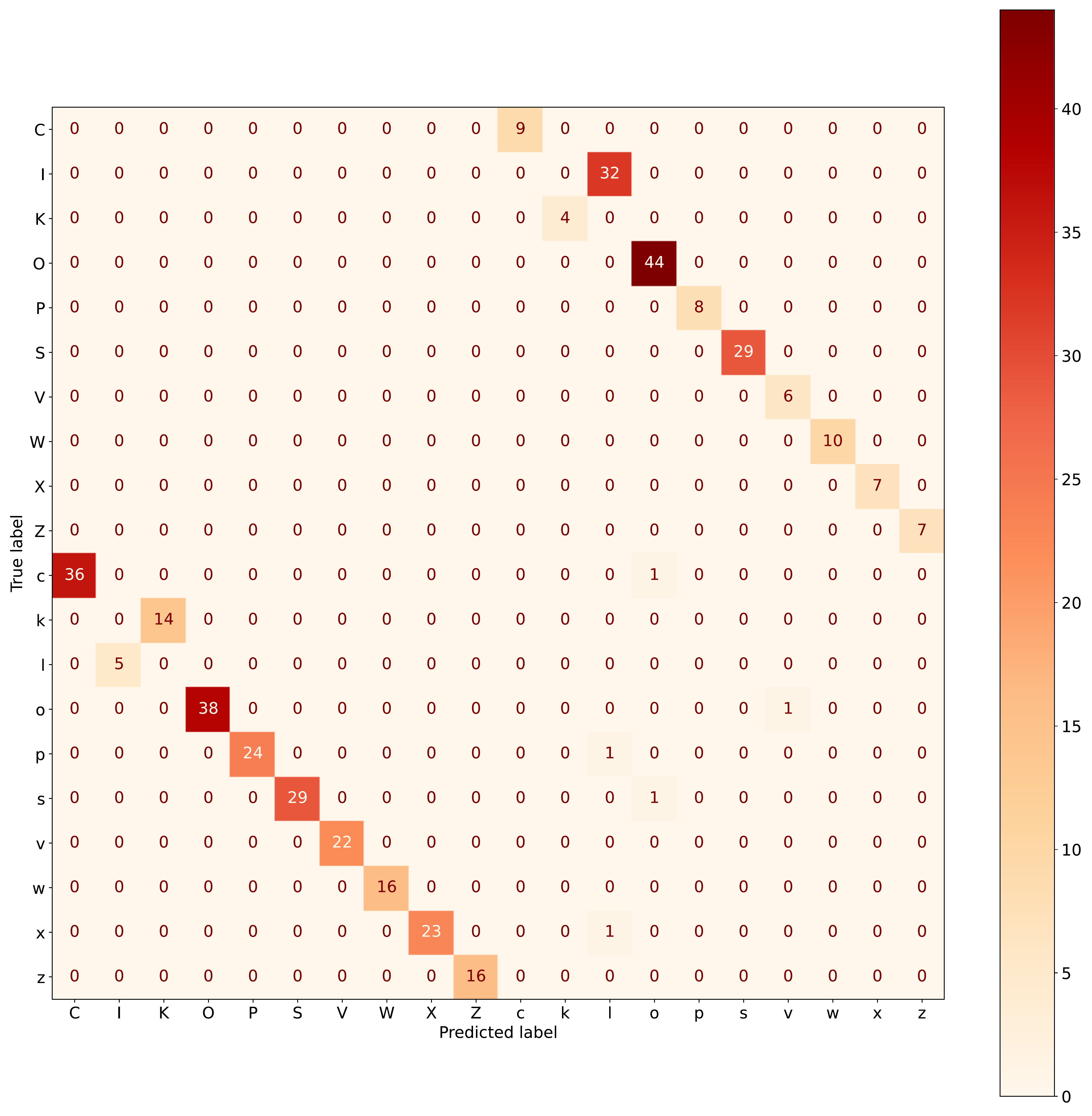} 
\caption{Truncated confusion matrix (only errors) of the VGG model over the test partition of COUT database.}
\label{fig:cm_COUT_all_vgg}
\end{figure}

A section of the confusion matrix of the VGG model for this problem is included in Fig. \ref{fig:cm_COUT_all_vgg}. It can be seen that the main errors are those appearing in the previous model on the upper and lower case letter charset when confusing characters with similar graphics in the two cases. In the case of special symbols, the most frequent error consists in confusing the addition symbol '+' with the letter 't'.

As a summary of the different experiments performed with this COUT database, it is indicated that the VGG architecture obtains the best results regardless of the subset of characters analyzed. It is also the architecture that presents the most similar results when comparing the augmented database versus the original one.

\subsubsection{Input resolution influence on the COUT database}
To determine the importance of the resolution of the input images, we built the model of the lower case charset with all architectures and the same training parameters but only changing the resolution of initial images to 32$\times$32 pixels. The lower case model is selected because it deals with the most frequent characters in continuous writing, and the errors presented by this model will be the most common ones when dealing with the continuous HTR problem. A recognition accuracy comparative over the test images was included in Table \ref{table:UNIPENAccuracyResultsSize} including accuracy for this new image resolution.

\begin{table}[!ht]
\centering 
\begin{tabular}{llccc}
 \toprule 
 Model& Resolution & Parameters &\multicolumn{2}{c}{Accuracy} \\
 \cmidrule(r){4-5} 
 & & & Validation(\%) & Test(\%) \\
 \midrule 
 LeNet &  64$\times$64&  0&  95.20 &  95.09 \\
       &  32$\times$32&  0&  94.40 &  94.59 \\
 \midrule 
 VGG &    64$\times$64&  0&  96.04 &  96.09 \\
     &    32$\times$32&  0&  \textbf{96.08} &  \textbf{96.43} \\
 \midrule 
 ResNet & 64$\times$64&  0&  92.12 &  92.52 \\
        & 32$\times$32&  0&  93.72 &  94.56 \\
 \bottomrule 
\end{tabular}
\caption{Accuracy by input resolution on the COUT test partition.}
\label{table:UNIPENAccuracyResultsSize}
\end{table}

The models obtain very similar accuracy when applied to a 32$\times$32 pixel image instead of a 64$\times$64 pixel image. In the sense that it is closest in accuracy at the two resolutions, being the most stable model the VGG one.

The biggest difference is observed in the \textit{ResNet} model, which obtains better results at the lower resolution of 32$\times$32. It may be because the ResNet architecture needs detailed fine-tuning to obtain good results in each particular case. For example, when changing the size of the input image. This need for fine-tuning may also explain the very different behavior of this model in the tests performed with the different databases. It is recalled that for selecting the \textit{ResNet} architecture for these experiments, basic fine-tuning of the architecture was done to determine the most appropriate number of \textit{ResNet blocks}. This fine-tuning was done on the upper case letters of the NIST database. It is precisely in the upper case classification models, built for the different databases, where this model has a more similar result to the other two ones. It can be seen in Table \ref{table:NISTUppertAccuracyResults} for the NIST database, Table \ref{table:TICHAccuracyBenchmark} for the TICH database and Table \ref{table:UNIPENLowerAccuracyResults} for the COUT database without data augmentation.

\section{Discussion of results}
\label{character models - Results discussion}

The main results and conclusions obtained from these experiments are the following ones.

\begin{itemize}

    \item The VGG architecture has proved to be the most efficient and the most versatile. It obtained the best results in all the experiments. It has not lost efficiency, always providing high accuracy results and close to those published in the literature, even though all the experiments have been performed with the same parameterization, and without any specific fine-tuning. 

    \item LeNet architecture, despite being the simplest in terms of layer depth, has shown to work correctly in most of the experiments performed. Moreover, the number of parameters of this architecture is lower than the other two ones. This makes it a good candidate to obtain a suitable baseline for this type of HTR problem. It has only presented a decrease in accuracy with respect to the VGG architecture in the more complex models, with a higher number of classes. We assess that it is likely to be necessary to increase the number of network parameters by increasing the dimension of the layers in these cases.

    \item The ResNet architecture, which was the most complex one in terms of the number of layers, presents a lower accuracy than the other, in some cases much lower. It has been found that for the upper case dataset, on which the initial fine-tuning of this architecture was done, good results were obtained. However, for other cases, the performance dropped significantly. We conclude that this architecture may need specific fine-tuning to adapt it to each of the different experiments performed.

    \item In the experiments performed with the upper and lower case character set, it has been observed that the results improve if the database maintains the relative sizes between letters. This is because the model makes fewer errors when distinguishing between characters with the same spelling in the upper and lower case. This insight has been used in the normalization strategy of the word databases used in the next chapter.

    \item The results on extended character sets, combining digits, upper and lower case letters, and symbols, are the most relevant ones to evaluate the performance of each architecture in the field of free text recognition. But precisely these results on isolated characters present obvious errors because several characters have very similar strokes and that people can only distinguish them by context. For example, the letter 'O' (upper case) and the digit '0'. As it is already known, these experiments do not allow to evaluate the ability of a model to differentiate characters by context, and only by spelling. When these architectures are used as part of a more general model of continuous text recognition, there will be such an answer. At that point, it will be possible to evaluate the ability of these models to take context into account when recognizing text.

\end{itemize}


\chapter{Deep architectures applied to continuous offline handwriting recognition}
\label{chapter_word_models}

This chapter describes the models and experiments performed for recognizing continuous handwritten text present in images. The main objective is to develop a model capable of identifying the sequence of characters appearing in a continuous text image. For this task, a new model based on the \textit{seq2seq} architecture, introduced by Sutskever et al. in \cite{sutskever2014sequence}, is proposed. Systematic experiments have been performed to evaluate the new model on multiple corpora, and the results have been compared to those of other authors.

In the next sections, we explain a novel architecture proposal for the offline handwriting recognition problem. This architecture combines a Convolutional Neural Network (CNN) with a seq2seq model. The CNN models the visual attributes of the handwritten words and provides a sequence of visual features from each part of the word image. This feature sequence is used as input to the seq2seq model. Across these inputs, the seq2seq network identifies the characters of the word using its encoder-decoder functionality. This functionality is designed to solve transcoding problems that need to transform one input sequence signal into another output sequence signal of different length. These architectures have been proven to successfully solve complex problems, such as machine language translation \cite{sutskever2014sequence}, speech recognition \cite{chorowski2015attention} or lip-reading \cite{chung2016lip}. 

For the evaluation of the proposed model, an experimental setup is defined, which includes a novel text image normalization strategy and a specified data augmentation algorithm. A model training strategy and a set of lexicons for dictionary-based decoding results are also proposed.

In addition, we analyze the effect of different parameterizations in the proposed architecture over the accuracy of the model. This analysis aims to check the stability of this model over changes in the parameters values, and conduct ablation analysis that measure the impact of the several components of the model on the error. It has been done using three different corpora in different languages to guarantee the generality of conclusions. Although the three selected databases are from languages with Latin characters, we think that the proposed solution is general enough and could be used on other character sets as Arabic or Chinese. 

Finally, our results are compared to other related approaches to evaluate the quality of the results concerning the state-of-the-art of the continuous HTR problem. Comparing results with different authors is complex due to the tremendous variability in the literature for existing data, dictionaries, and experimental strategies. These difficulties are analyzed, and solutions are proposed in specific sections.

The following section details the architecture of the model proposed. The necessary details of the experimental setup are provided in Section \ref{c6 - section - Experimental setup}, and the results of the experiments performed in Section \ref{C6 - section - Experiments}. In Section \ref{section:On the difficulties of comparing results} the difficulties encountered in the comparison with other authors are analyzed. Based on the previous analysis, in Section \ref{section:Results comparative}, a framework for standardizing the results is proposed to solve these difficulties, and the best results obtained by the proposed model are compared with other approaches over the same databases and with the same lexicons. Finally, the results obtained in Section \ref{C6 - section - Results discussion} are discussed.

\section{Model architecture}
\label{c6 - section - Model architecture}

\begin{figure}[!ht]
\centering
  \includegraphics[width=0.5\textwidth]{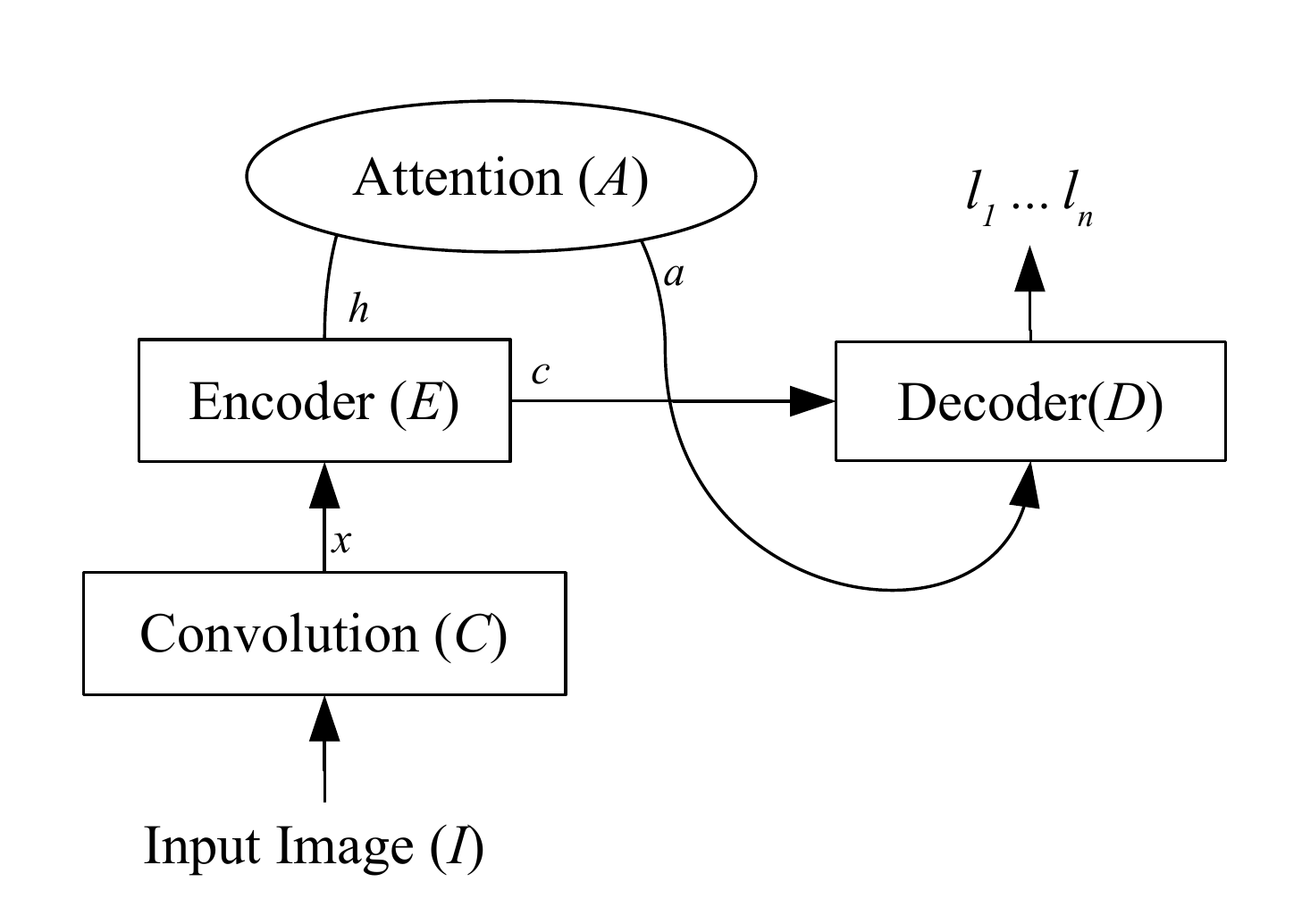} 
  \caption{Components of the proposed high-level model architecture for word recognition.}
  \label{fig:seq2seq_architecturehigh_level}
\end{figure}

This section describes the architecture of the new model proposed. The model transforms a variable-length handwritten word image into the variable-length sequence of the characters that conform the previous word. The model has three main components: a convolutional reader, an encoder, and a decoder with an attention mechanism. These components are combined into a seq2seq architecture as shown in Fig. \ref{fig:seq2seq_architecturehigh_level}.

Following the notation of Fig. \ref{fig:seq2seq_architecturehigh_level}, it is possible to summarize the previous model by the sequence of equations \ref{eq:c6_seq2seq_1}:

\begin{equation}
  \begin{array}{lcl} 
  x = C(I) \\ 
  (c, h) = E(x) \\ 
  a = A(h) \\
  l_t = D(a, c, l_{t-1})
  \end{array}
\label{eq:c6_seq2seq_1}
\end{equation}

The convolutional reader function $C$ is a deep CNN that converts the input image $I$ into a vector of visual features $x$. The input to this convolutional reader consists of the bi-dimensional array of pixel values of the word image. The convolutional model extracts feature vectors of this image related to the letters that appear in it. Two types of convolutional readers have been defined. The first one employs a single convolutional model on the input image and uses a concatenation strategy of feature maps by horizontal positions to obtain an ordered sequence of output features. The second one divides the image into overlapping patches that run through the image from left to right, and applies the convolutional model on each of the defined patches. This second strategy naturally generates an ordered sequence of convolutional features. 

Next, the encoder $E$ function, a RNN of type LSTM or GRU, reads the sequence of convolutional features $x$ and extracts the sequence relationships among these features. So, the encoder transforms $x$ into a sequence of encoded output features $h$ that model the relationships among the word letters. In addition, the final state $c$ of the encoder RNN was preserved to use as an initial state in the RNN decoder.

Finally, a decoder $D$ gets the previous encoder output transformed by the attention mechanism $A$ to predict, letter by letter, the word. The decoder consists in a LSTM that generates each current letter $l_t$ combining the state vector $c$, the encoder outputs transformed by the attention mechanism $a$ and the previous letter $l_{t-1}$.

The attention mechanism $A$ helps to focus on specific parts of the encoder output features $h$ that are relevant to identify specific letters. For example, the first letters need to focus on the first elements of the encoder output features that represent the first part of the word image. On the other side, the last letters need to focus on the final part of the encoder features associated with the last part of the word image.

The following sections describe this architecture in detail. Subsection \ref{section:C6-model-Detailed_schema} comprises a detailed model schema in which the architecture of each of its components is included. The following subsections detail each of the components of the model: he convolutional reader is presented in Subsection \ref{section:C6-model-Convolutional_reader}; the encoder, in Subsection \ref{section:C6-model-Encoder}; the decoder, in Subsection \ref{section:C6-model-Decoder}; and the attention mechanism in Subsection \ref{section:C6-model-Attention_mechanism}.

\subsection{Detailed schema of the model}
\label{section:C6-model-Detailed_schema}

\begin{figure*}[!ht]
  \includegraphics[width=1\textwidth]{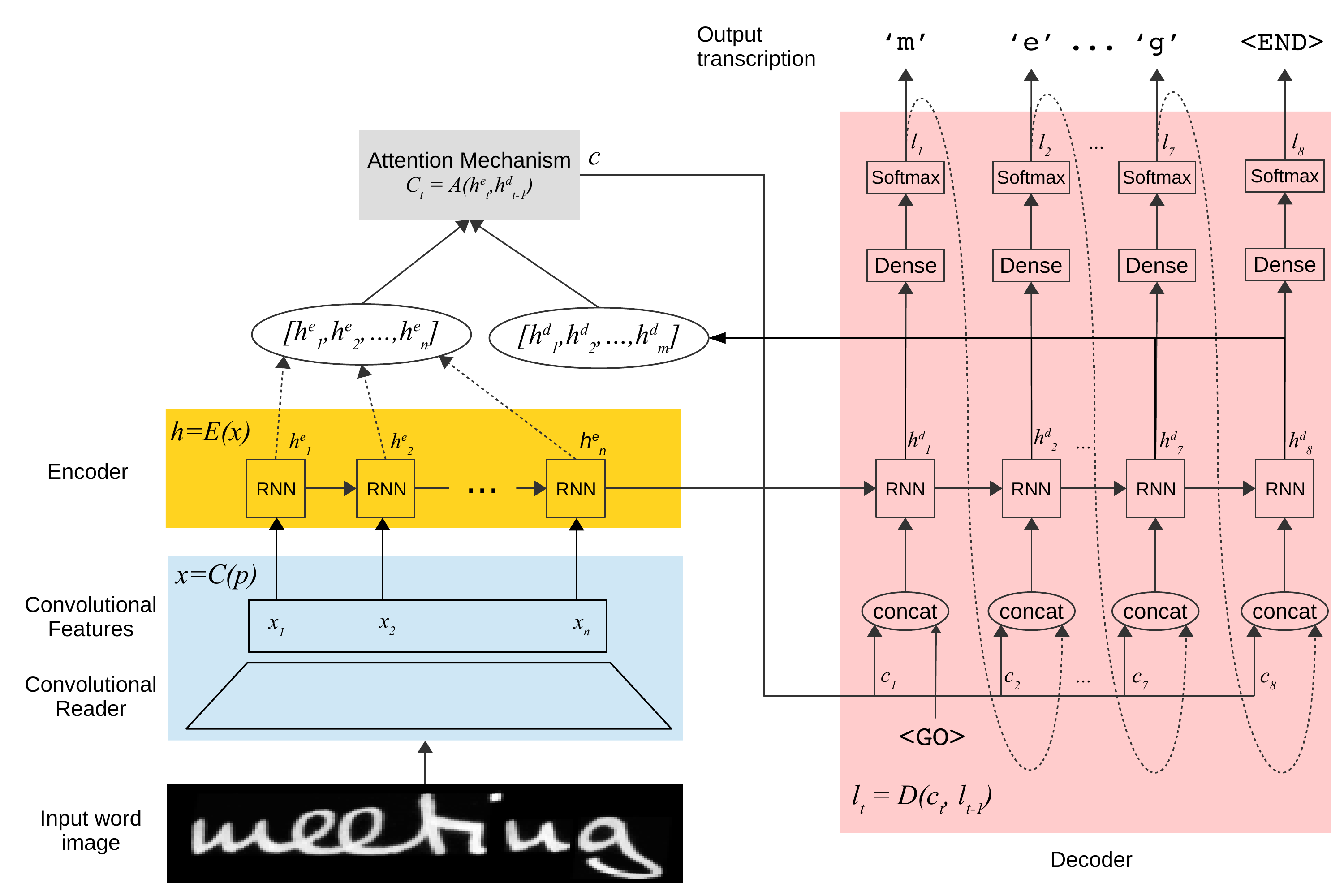} 
\caption{Detailed model architecture.}
\label{fig:Seq2Seq_extend}
\end{figure*}

In this subsection, we include the detailed schema of the proposed model. In the schema, we include the details of the different model components: convolutional reader, encoder, decoder, and attention mechanism. Additionally, we also include the details of how these components are connected. Fig. \ref{fig:Seq2Seq_extend} shows the detailed schema of the complete model architecture. This figure follows the notation introduced previously.

The proposed solution is designed to model each letter $l_{i}$ in the word $w = (l_{1}, l_{2},...,l_{n})$ as a conditional probability of the input image $I$ and the probability distribution of the previous letters of the word $l_{<i}$.

\begin{equation}
\mathbf{P}\left ( l_{i}|I, l_{<i} \right )
\end{equation}

Therefore we model the output probability distribution of a word given the sequence of patches, as follows:

\begin{equation}
\mathbf{P}\left (w |I \right ) = \prod_{i}\mathbf{P}\left ( l_{i}|I, l_{<i} \right )
\end{equation}

In the following subsections, all the elements of the architecture that implements this model are detailed.

\subsection{Convolutional reader}
\label{section:C6-model-Convolutional_reader}

The objective of the convolutional reader is to extract the visual features which are relevant to identify the characters appearing in the input word image. The selection of the convolutional architecture is one of the most relevant aspects in the model design. It is also the component of the model that presents more variability in its configuration. In the previous chapter, on models for isolated characters identification, three different convolutional architectures were analyzed: LeNet, VGG, and ResNet, respectively. In that chapter, it was concluded that LeNet and VGG architectures obtained good results in character recognition, especially the second one, and that ResNet architecture only offered good results if it was properly configured for each specific target. Based on this, the VGG and LeNet architectures are the ones on which the current word recognition model will be evaluated. The details of the architectures and a diagram of them can be seen in Subsection \ref{c5 - section - Lenet architecture} for LeNet and in Subsection \ref{c5 - section - VGG architecture} for VGG.

Since the output of the convolutional component is fed as input to it, the encoder must consist of a sequence of convolutional feature vectors. Furthermore, the features entering the encoder are expected to encode aspects of the input image sequentially. The first features encode characteristics related to the initial part of the image, and the last features encode the corresponding ones of the final part of the image of the handwritten text. The encoder will describe the sequential relationships in the image in the same way that humans read the text sequentially. Following the notation of Fig. \ref{fig:Seq2Seq_extend}, the output of this convolutional component is a sequence of vector features $x=(x_{1}, x_{2},...,x_{n})$ that encapsulate the visual features of the handwritten word image.

One way to obtain the above representation is to extract from the image a sequence of vertical patches of fixed width in pixels, using a given pixel step size. From each patch, the convolutional component computes a vector that contains the visual features related to the character or characters included in it. The image patches are the inputs, and the features obtained from the last dense layer are the outputs. 
Alternatively, the convolutional component can be applied directly to the complete input image. In this case, a specific transformation must be applied to the output of the convolutional component to convert it into the input of the decoder RNN. The transformation must construct an ordered sequence of feature vectors that sequentially encode the image as indicated above. 

Two types of convolutional components are used: the ones based on image patches and the ones using directly on the complete image, are described in detail below.

\subsubsection{Convolutions over image patches}
\label{section: Image patches architectures}

\begin{figure}[!ht]
\centering
  \includegraphics[width=0.6\textwidth]{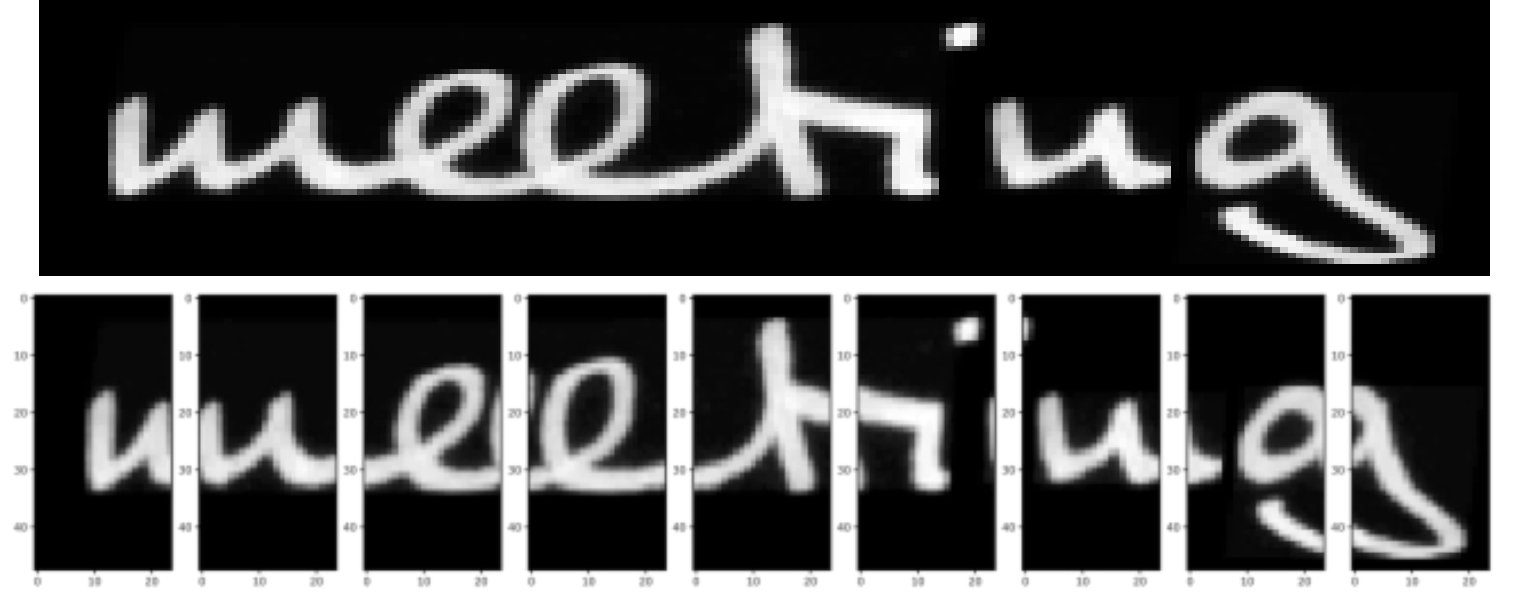}
\caption{Word patch extraction: (up) original image word, and (down) extracted input model patches.}
\label{fig:patches:4} 
\end{figure}

The first strategy for applying the convolutional architecture consists of using the same convolutional model to an ordered sequence of patches extracted from the original image with a sliding window that runs horizontally from left to right. In this way, we extract a sequence of horizontal overlapped slides for each original word image. These slides can include a part of a letter, one letter, or more than one letter. The slides are the inputs to our model, and they enable it to view the word image not globally but as a sequence of patches. These are characterized by two parameters: the patch size (i.e., its horizontal size) and the patch step (i.e., the number of pixels that we move to the right to extract the next patch). Fig. \ref{fig:patches:4} shows an example of an original word image and its corresponding extracted patches.

The same convolutional model is applied on the previous slide images using a sharing parameters technique \cite{zhang2018learning}. In this way, the model focuses on extracting the convolutional characteristics of the image region that appears in each patch, generating an ordered sequence of such feature vectors. The RNN of the decoder will model the sequential nature of these features.

The two convolutional architectures used are VGG and LeNet, that were selected in the previous Chapter \ref{chapter_character_models}. The only modification made is the removal of the last two dense layers. These layers are not included because they were oriented to build the character classifier. In this case, what we seek to obtain as output are the convolutional characteristics of each patch.

\subsubsection{Convolutions over complete images}

In the case of convolutional architectures used directly on the input image without creating patches, an operation must be applied in order to adapt the output of the convolutional network to the input required by the RNN of the decoder. In addition, this operation must generate a sequence of vectors ordered concerning the columns of the input image; that is to say, the first vectors must collect features from the left side of the image and the last vectors, the ones from the right side. 

\begin{figure*}[!ht]
\centering
  \includegraphics[width=0.7\textwidth]{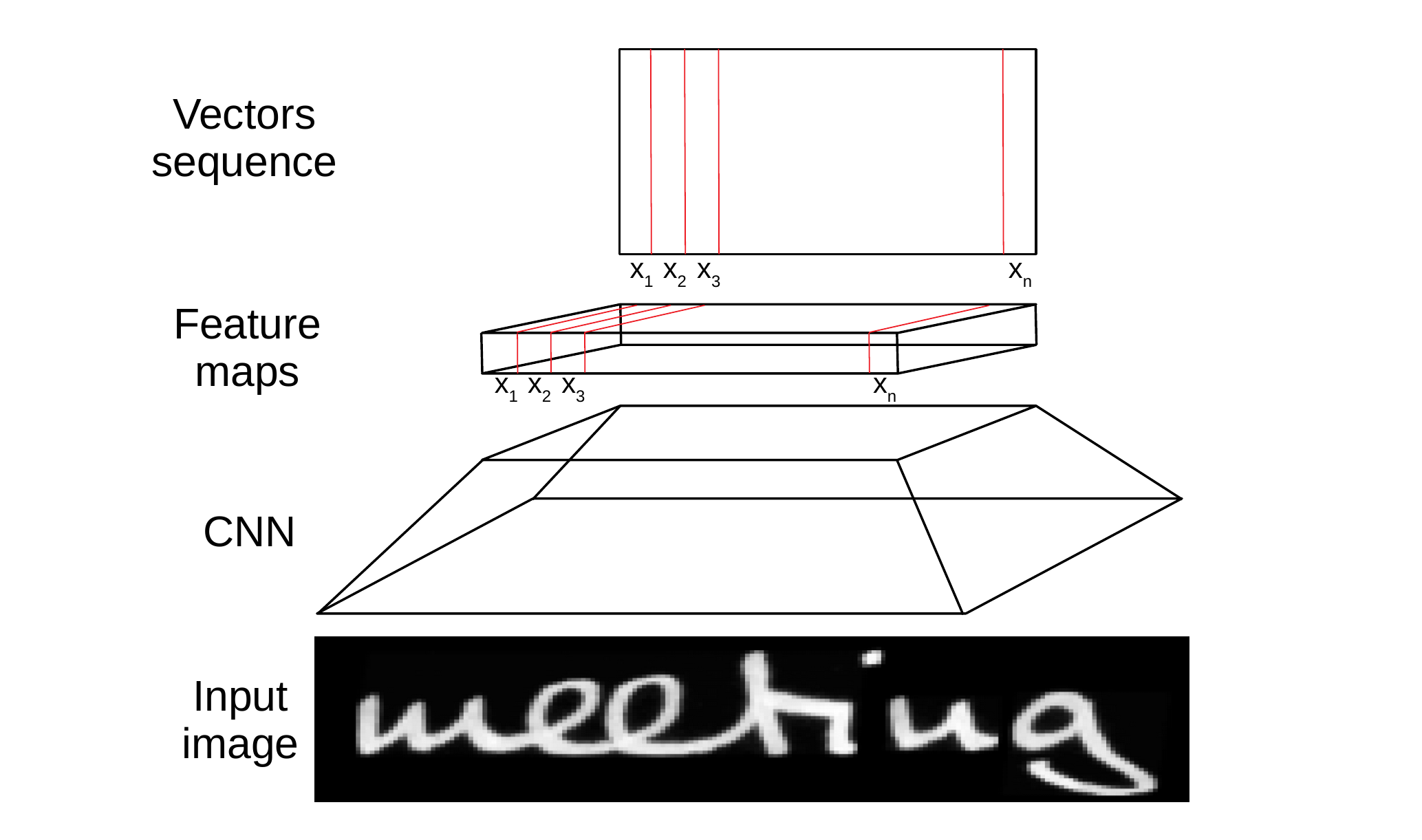} 
\caption{Residual connection to convert the convolutional output into the feature vectors sequence.}
\label{fig:residual_connection}
\end{figure*}

For this purpose, the convolutional network output feature maps are resized into a two-dimensional array used as input to the RNN component of the encoder. The resizing is done by concatenating the output feature maps of the convolutional component by columns. It ensures that the input sequence to the encoder RNN has the information ordered in the same way as in the input image but transformed by the convolutional component. Fig. \ref{fig:residual_connection} illustrates how columns traverse the feature maps to generate each vector of the output sequence by concatenation.

The convolutional architectures applied in this case are the same as in the previous patch application, LeNet and VGG. The only difference is that the previous transformation can only be applied to the output of a convolutional or pooling layer. Therefore, in this case, the dense layers have been removed from the convolutional component.

\subsection{Encoder}
\label{section:C6-model-Encoder}

The purpose of the encoder is to capture the input image characteristics corresponding to the sequential nature of the text. To do that, the encoder is implemented using RNN layers. The RNN encoder receives the sequence of vector features $x_t$ generated by the convolutional reader and generates as output a sequence of output vectors $h^e_t$, each one for each encoder step $t$. The first element $h^e_0$ of the RNN of the encoder is initialized randomly.

\begin{equation}
 \begin{array}{lcl} 
 h^e_0 = random() \\
 h^e_t = RNN(x^e_t, h^e_{t-1})
 \end{array}
\end{equation}

The encoder architecture is another essential component to be analyzed along with the convolutional component described above, since it must model the elements related to the sequential nature of the text. In our case, we chose to perform several experiments to determine if there is an encoder configuration that consistently provides better results in the different databases analyzed. Specifically, we have analyzed the type of RNN to be used, which can be LSTM or GRU, the number of RNN layers to be used, as well as their dimensions, and we have also experimented with simple forward RNNs or bidirectional RNNs. In Subsection \ref{C6 - Experiments - Encoder architecture} we detail the results of these experiments.

The output sequence of the RNN $h^e$ will be passed as input to the attention mechanism. In bidirectional RNNs, where there are two output sequences, forward and backward, these are concatenated to generate the final input sequence to the attention mechanism. In addition, the encoder RNN returns a final state vector $h^e_{last}$, which is used as the initial state of the decoder RNN. In bidirectional RNNs, the state of the forward network is selected as the final state.

\subsection{Decoder}
\label{section:C6-model-Decoder}

The decoder is responsible for generating the output sequence formed by the ordered characters that make up the text of the input image, i.e., the transcription of the handwritten text image. 

Like the encoder, the decoder is also based on an RNN network. In addition, in the decoder architecture proposed in this Thesis, it has been chosen to use the same type of RNN as the encoder. Thus, in the experiments in which the encoder uses an LSTM type network, the decoder will use an RNN of the same type. The same happens if the encoder uses a GRU type network. 

\begin{figure*}[!ht]
  \includegraphics[width=1\textwidth]{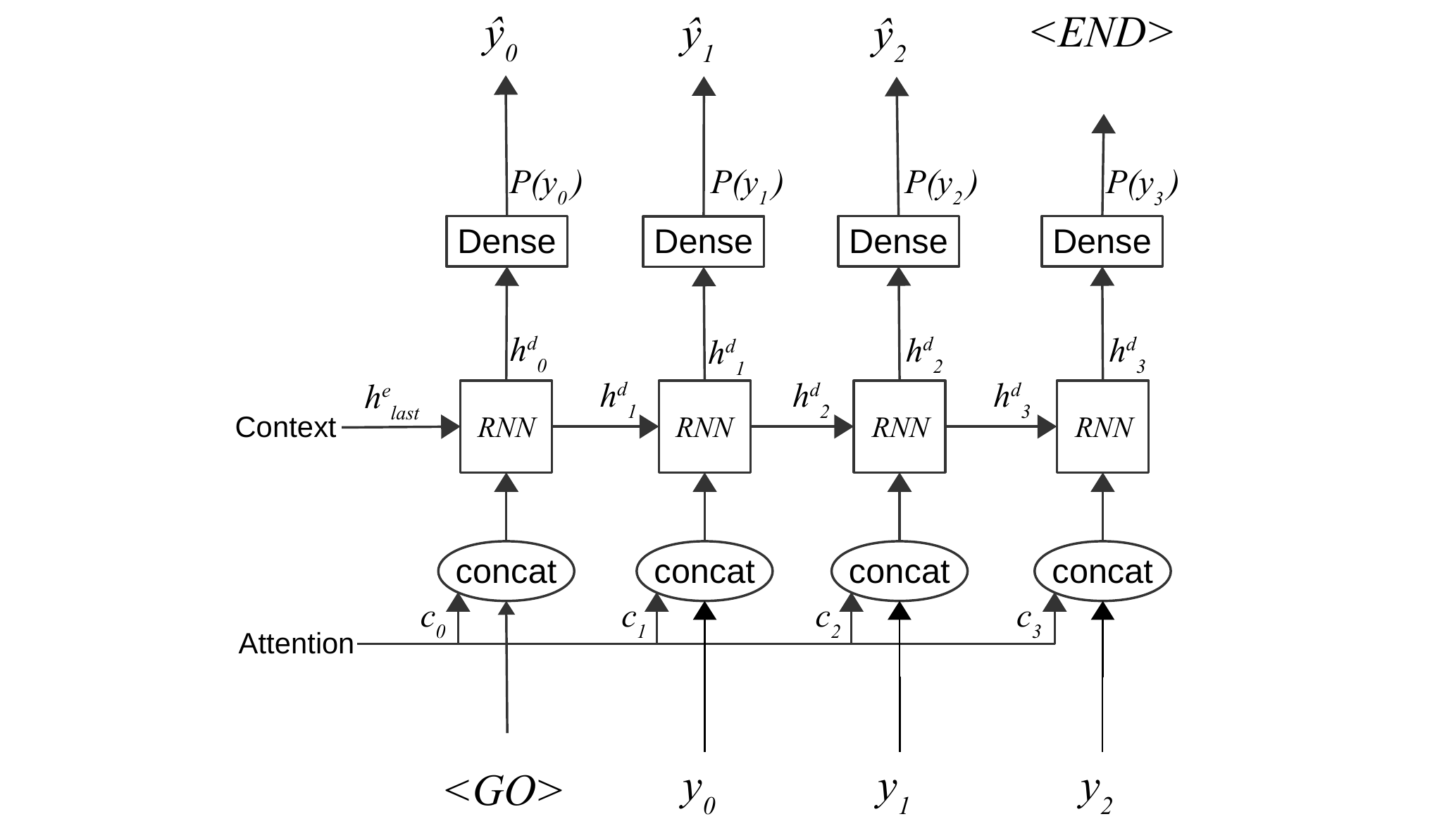} 
\caption{Decoder training diagram.}
\label{fig:seq2seq_decoder_train_c6}
\end{figure*}

The decoder operates so that each step generates one by one the characters of the transcription. During the training stage of the model, the decoder receives as input in each step the previous character of the transcript $y_{t-1}$ and the context information provided by the attention mechanism $a_t$. With this information, it generates as output an estimate of the next character of the transcript $\hat{y_t}$. The decoder starts with the special signal $\textless GO\textgreater$ as input to decode the first character of the word image. It ends when at the output appears the special character $\textless END\textgreater$, by selecting the sequence of previous output characters as the decoded word. The initial state of the decoder RNN $h_0$ is initialized with the last state of the encoder $h^{e}_{last}$. This way of initializing the decoder means that the dimension of the decoder RNN must be the same as the dimension of the encoder RNN. Fig. \ref{fig:seq2seq_decoder_train_c6} schematizes the operation of the decoder in the train stage. 

Performing model training using the real $y_t$ values as input to the RNN is called \textit{teacher forcing} and was introduced by Williams et al. in \cite{williams1989learning}. This technique is described in detail in Subsection \ref{section:seq2seq Decoder} of Chapter \ref{chapter_background}.

The probability distribution of the output character for each step is generated by a $dense$ layer with the $softmax$ activation function \cite{bridle1990probabilistic} applied over the output sequence of the RNN. Eq. \ref{eq:c6_decoder_detail} details the behavior corresponding to the decoder train process following the notation of Fig. \ref{fig:seq2seq_decoder_train_c6}.
\index{Softmax}

\begin{equation}
 \begin{array}{lcl} 
  y_0 = \textless GO\textgreater \\
  h^d_0 = h^e_{last}\\
  h^d_t = RNN(concat(y_{t-1}, c_t), h^d_{t-1})\\ 
  p(y_t) = softmax(dense(h_t)) \\
  \hat{y_t} = argmax(p(y_t)) \\
 \end{array}
 \label{eq:c6_decoder_detail}
\end{equation}

The estimated output sequence $\mathbf{\hat{y}}=(\hat{y_1}, ...,\hat{y_k})$ is built selecting the maximum probability values in $p(y_t)$ by argmax function for each $t$, as it is indicated in the equation \ref{eq:decoder_detail}. 

\begin{figure}[!ht]
  \includegraphics[width=1\textwidth]{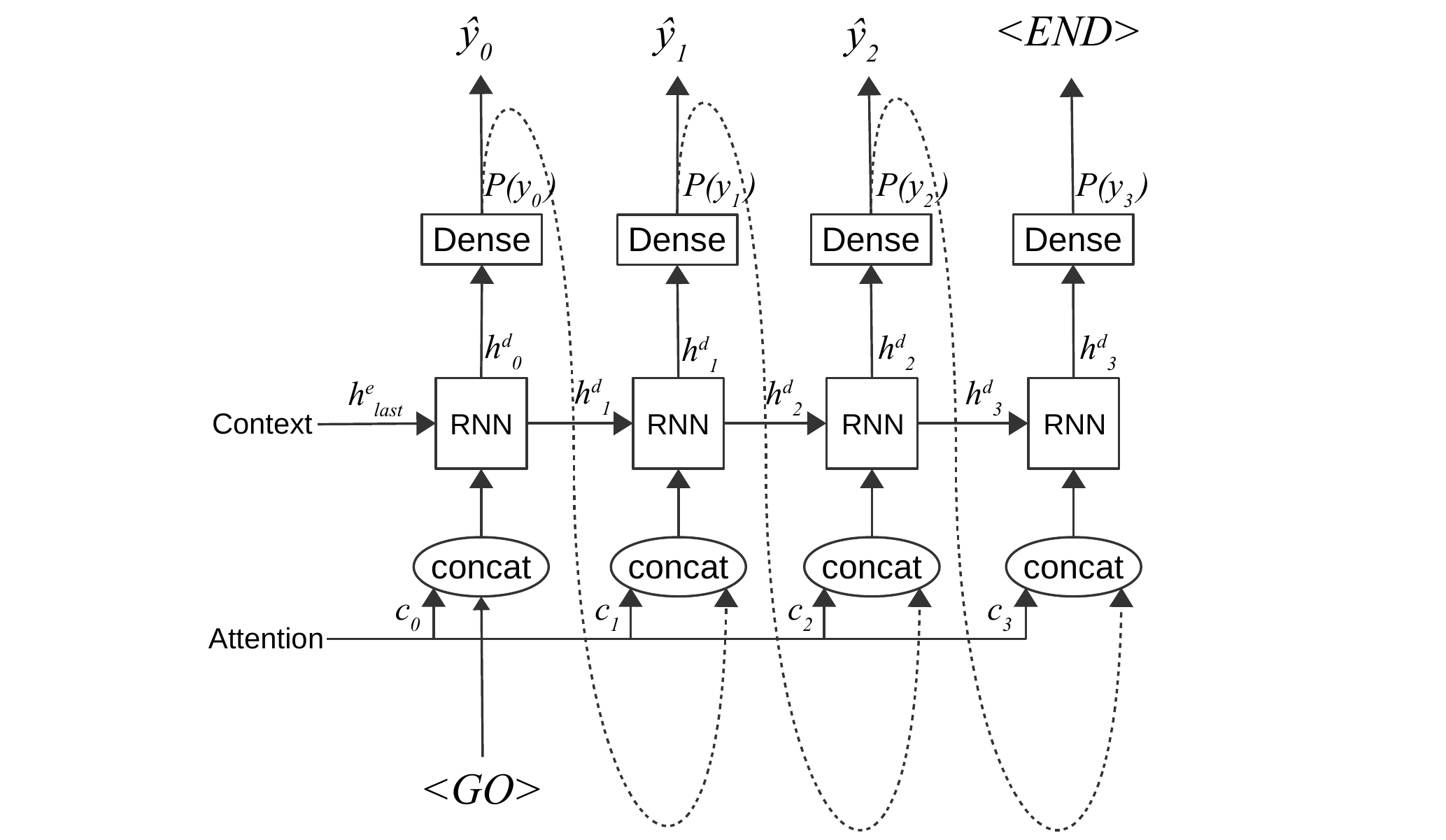} 
\caption{Decoder score diagram.}
\label{fig:seq2seq_decoder_score_c6}
\end{figure}

Unlike the rest of the model, the decoder has a different behavior over the model train phase and the model evaluation or score phase. During the scoring phase, it is not possible to use the output character sequence $y_t$ as input to the decoder, as is done during the training phase. Instead, as input for step $t$, the character estimated by the decoder for step $t-1$ is used, i.e., $\hat{y_{t-1}}$. Details of the scoring process are given in Fig. \ref{fig:seq2seq_decoder_score_c6}.

In the score or validation stage, the equation of the previous RNN is modified as shown in Eq. \ref{eq:c6_decoder_detail_score}. Both real and estimated characters are used encoded with one-hot encoding.

\begin{equation}
  h_t = RNN(concat(\hat{y}_{t-1}, c_t), h_{t-1})
 \label{eq:c6_decoder_detail_score}
\end{equation}

The training process can also be performed following the same scoring principle of using the prediction of the previous step as input to the next step. It is an alternative to the teacher forcing strategy mentioned above.

The objective of the decoder is to learn how to generate the output character sequence. For identifying the next character in the sequence, the RNN of the decoder receives as input the context information of the input image produced by the attention mechanism in addition to the previous character or the estimate of that character in the previous step. The estimation of the next character in a sequence from the previous characters through an RNN is precisely one of the types of language models analyzed in Section \ref{c4 - section - Language models} of Chapter \ref{chapter_background}. It means that the decoder could be acting as a character-level language model. However, our model is trained at word level only with words present in the handwriting training database. This fact considerably limits the capacity of the decoder to act as a true language model. A true character language model typically needs characters of the previous words to capture the grammar and several orders of magnitude more text than is present in the handwriting databases.

\subsection{Attention mechanism}
\label{section:C6-model-Attention_mechanism}

The attention mechanism connects the sequence of feature vectors generated by the encoder $h^e_1,...,h^e_n$ with each step of the decoder. The encoder output encodes the visual features obtained by the convolutional module and the sequential features obtained by the encoder RNN. It allows each decoder step to focus on a specific part of the relevant encoder output for this step. It is very important on the architecture because it allows the decoder to pay attention to the input data part that is more relevant to decode the current character. For example, if the decoder estimates the first letter of the word, it can focus on the features generated by the first part of the word image. And so on, with all the characters to be predicted.

Following the notation of the previous sections, to calculate the attention vector $c_t$ at each output time $t$ over a encoder output length $N_e$, the detailed equations in \ref{eq:c6_content_attention} are used:

\begin{equation}
\begin{array}{lcl} 
e_{t,i} = w^T \tanh(W h^e_i + V h^d_{t-1} + b)\\
a_t = softmax(e_t)\\
c_t = \sum_{i=1}^{N_e}a_{t,i}h^e_i
\end{array}
\label{eq:c6_content_attention}
\end{equation}

where $w$ and $b$ are trainable weight and bias vectors, $W$ and $V$ are trainable matrices, $h^e_i$ is the hidden state of the encoder at time $i\in{0,1, ..., N_e}$ and $h^d_{t-1}$ is the hidden state of the decoder at time $t-1$. The attention mechanism is implemented as a neural network layer that combines the states of the encoder and the previous states of the decoder with the hyperbolic tangent as an activation function. The vector $e_t$ contains in the position $i$ a score of how much attention we need to put the current decoder step $t$ on the \textit{i}-th encoder hidden state $h_i$. Then, we apply the $softmax$ function to normalize it and create the attention masks $a_t$ over all the encoder states. The final attention vector $c_t$ is concatenated with the input of the decoder layer as we show in Fig. \ref{fig:Seq2Seq_extend}.

As indicated in Subsection \ref{section:C3_seq2seq_attention} of Chapter \ref{chapter_background} there are different types of attention mechanisms. In the case of the implemented model, a non-monotonic content base attention network characterized by the above equations \ref{eq:c6_content_attention} has been selected.

The selected attention mechanism is a content-based type, which does not encode positional information. This information is necessary to distinguish the same character at different locations in the input image. This is done because it is assumed that the RNN of the encoder already encodes this positional information, and therefore it is not necessary for the attention mechanism to do so as well.

The choice of a non-monotonic attention mechanism may be controversial since reading a handwritten word could be considered a monotonous task because it is needed to read the left letters before the right ones. However, this fact is not true for algorithms. Reading letters of a word, written using a Latin script, from left to right, is conventionalism, and it is possible reading them from right to left without loss of information. So, to allow the model to consider all the neighborhood information, the selected attention mechanism has no monotonicity restrictions.

\begin{figure}[!ht]
\centering
  \includegraphics[width=0.8\textwidth]{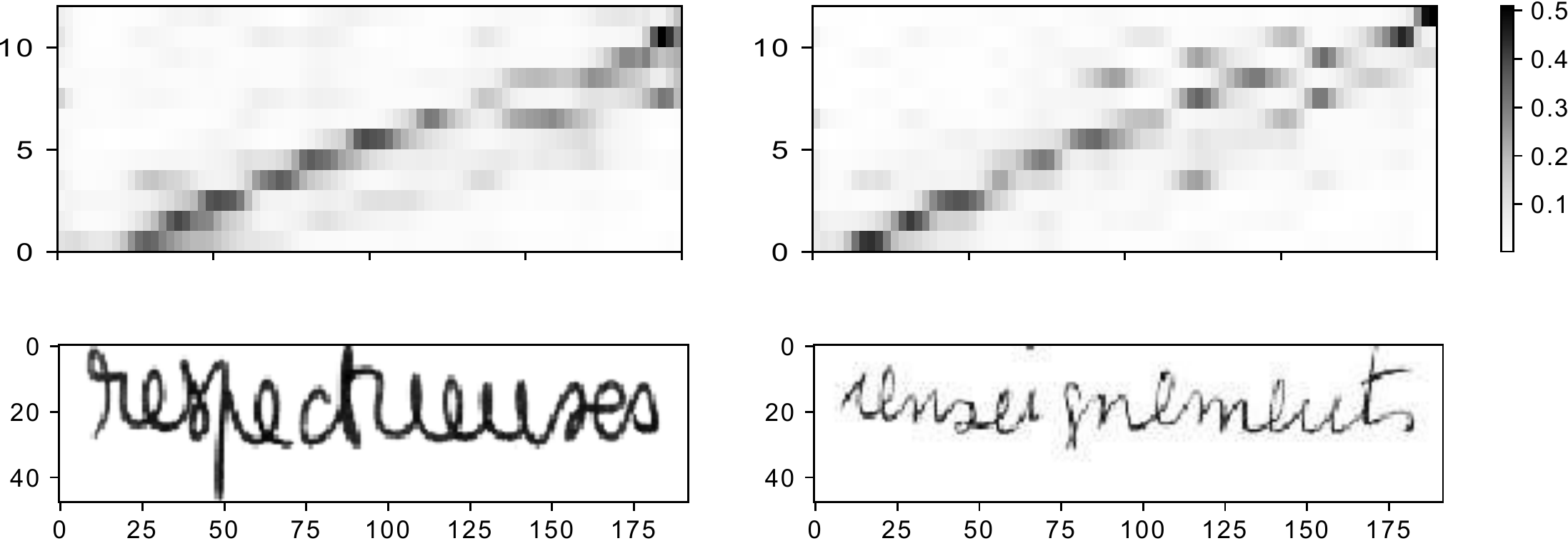}
\caption{Example of attention mechanism outputs for two words and their respective original handwritten word images.}
\label{fig:attention}
\end{figure}

To illustrate this assumption, we plot in Fig. \ref{fig:attention} the outputs $a_{t,l}$ of the attention mechanism for two words of the validation partition. These figures present in the horizontal axis the image patch number and in the vertical axis the predicted character position. The figures show that, in general, the attention mechanism is working monotonically, but for some character positions, it uses information from patches at the left of what will be a monotonous behavior.

\section{Experimental setup}
\label{c6 - section - Experimental setup}

A set of experiments has been designed to validate the usefulness of the proposed new model in the continuous HTR problem. In this section, the common elements of these experiments are defined. These elements are defined to ensure that the results of the experiments provide adequate information on the effectiveness of the proposed model. They also ensure that the results allow a correct comparison with results of other models from different authors.

In Subsection \ref{section:Source_code_and_software} we include a brief description of the software tools employed to build the different experiments. Next, Subsection \ref{c6 - section - Implementation details} details the configuration defined for the training of each model. In Subsection \ref{C6 - section - Image Preprocesing} describes in detail the image preprocessing algorithm used to normalize the handwritten text images before inserting them into the model. Subsection \ref{C6 - section - Data augmentation} describes the data augmentation algorithms implemented and used in some of the experiments. Then, in Subsection \ref{C6 - section - Lexicons} we provide details about the different lexicons used for word decoding.

\subsection{Source code and software}
\label{section:Source_code_and_software}

Different software tools are used to develop the experiments and to perform the different steps of the recognition process. Table \ref{table:Experiments_sofware} includes a version summary, download links, and application areas. 

In order to facilitate that the results can be reproduced independently, all the software selected has open licenses and is freely available to download and use.

\begin{table}[!ht]
 \centering
 \begin{tabular}{lrrr}
  \toprule
  Name &  Version & Link & Usage \\
  \midrule
  Tensorflow &  1.8 &  \url{www.tensorflow.org} & Modelling \\
  OpenCV &      3.0 &  \url{www.opencv.org}   & Image preprocesing \\
  imgtxtenh &   1.0 &  \url{www.github.com}   & Image Preprocesing \\
  \bottomrule
 \end{tabular}
 \caption{Software employed in the experiments.}
 \label{table:Experiments_sofware}
\end{table}

\textbf{Tensorflow} \cite{abadi2016tensorflow} is a definition and execution of machine learning algorithms software. It has much flexibility to implement complex algorithms and is capable of scaling the execution of the same on large volumes of data.
It is available for download from \url{http:\\www.tensorflow.org}.
\index{Software!Tensorflow}

\textbf{Open CV} (Open Source Computer Vision Library) is an open-source Computer Vision and machine learning software library. It is available for download from \url{http:\\www.opencv.org}. It has a BSD-license that allows, among other things, its use for research purposes. The library provides C++, Python, Java, and MATLAB interfaces and supports the most common operating systems. This Thesis uses the Open CV Python interface over the Ubuntu operating system to develop some image transformations in the image preprocessing step.
\index{Software!Open CV}

\textbf{imgtxtenh} \cite{villegas2015modification} is a tool to clean and enhance noisy scanned text images, which could be printed or handwritten text. It is available for download from \url{https://github.com/mauvilsa/imgtxtenh} (last access September 2020)
\index{Software!Imgtxtenh}

\subsection{Implementation details}
\label{c6 - section - Implementation details}
\index{Dropout}
\index{Adam algorithm}

As trainer, we used the Adam stochastic optimization algorithm, with a batch size of 128 and with a dropout of 0.5 \cite{pham2014dropout} in the dense layers of the convolutional part and in the cells of the encoder RNN layers. An initial learning rate of 0.001 was used, and it was decreased by 2\% in each epoch. We applied stopping criteria of 20 epochs without any improvement in the Word Error Rate (WER) on the validation set. In general, the experiment configuration that has obtained a lower WER in the validation data is considered the most appropriate. In any case, the simple parsimony principle is applied \cite{vandekerckhove2014model}, and when two configurations with similar results are presented, the one that produces a model with a smaller number of parameters is selected. The loss function used is the \textit{weighted cross-entropy} loss for a sequence of logits. On this loss, it is applied on all the weights of the model a type L2 regularization with a parameter of 0.0001.

In all the experiments, the model weights are randomly initialized through the technique called \textit{Glorot normal} \cite{glorot2010understanding} that initializes the weight of each node using a random value obtained from a Gaussian distribution of zero mean and standard deviation proportional to the number of connections of the node. 
\index{Glorot normal}

In each epoch, we randomized the order of the training data. The different experiments took between 60 and 280 epochs. The duration of the experiments varies greatly depending on the database, the model parameters volume, and whether or not data augmentation is used. In general terms, each experiment with the IAM and RIMES databases took between 10 and 20 hours, and the experiments with the Osborne database between 2 and 4 hours.

Once the model has been trained, it is applied to the evaluation and test datasets to obtain the accuracy metrics, producing three different groups of results: the raw accuracy of the visual model without any lexicon, the improvement obtained when using search over the standard lexicon of each database, and finally, the error obtained when eliminating the out-of-vocabulary (OOV) words using the test lexicon.

\subsection{Image preprocesing}
\label{C6 - section - Image Preprocesing}

\begin{figure}[!ht]
 \centering 
 \includegraphics[width=7cm]{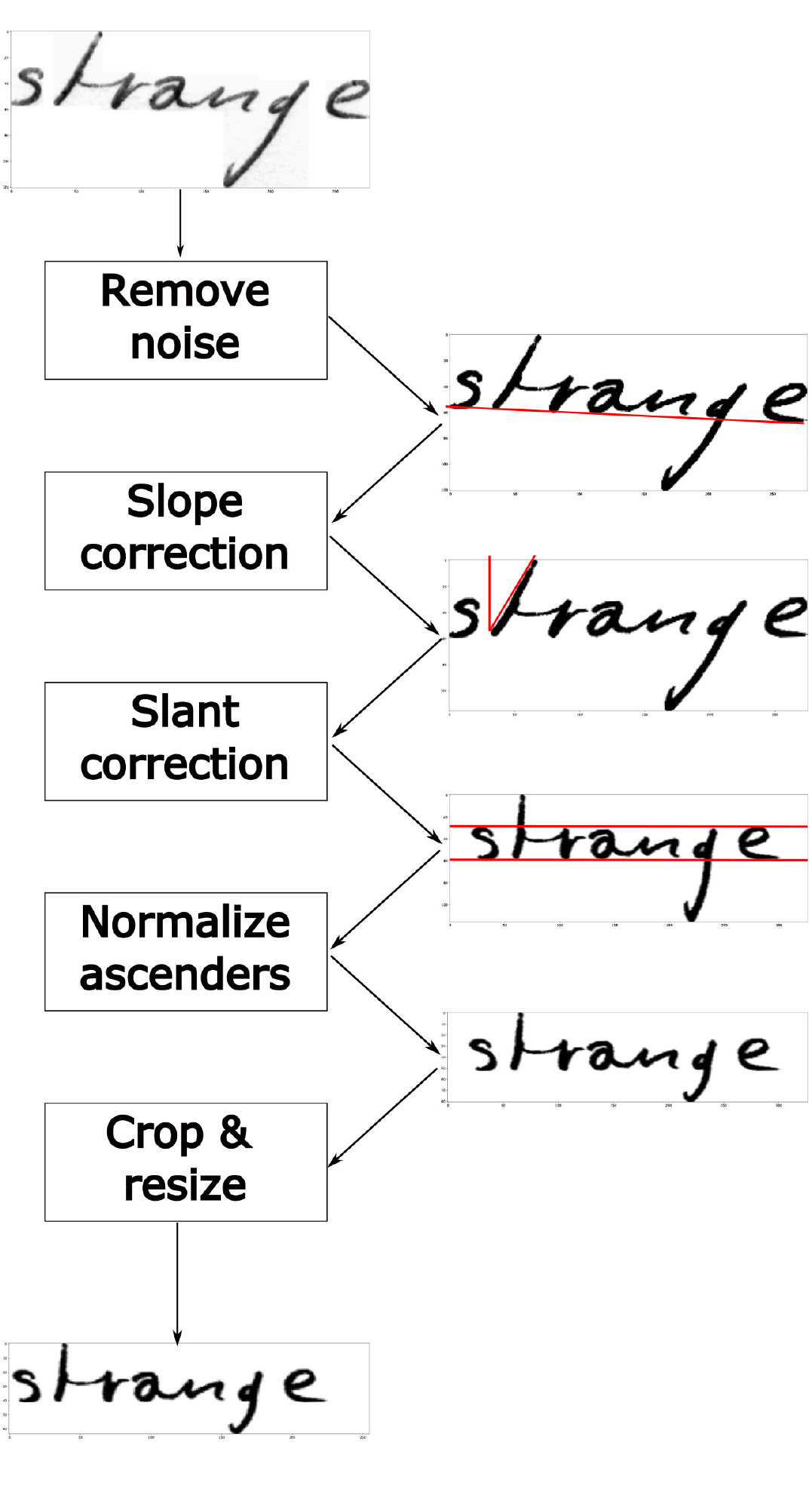} 
 \caption{Pre-processing algorithm steps.}
 \label{fig:preprocessing_algorithm}
\end{figure}

The data used in our experiments were preprocessed to correct some typical characteristics of the handwriting text that difficult the recognition task. For this purpose, a specific algorithm has been developed which, in summary, has the following steps. First, the image is cleaned, improving the image contrast and eliminating noise. Next, baseline and corpus line are identified and used to correct the line skew. Next, the slant is corrected, and the height of the characters is normalized based on baseline and corpus line. Finally, the input image size is adjusted to the model using a fixed height and width without modifying the aspect ratio of the text. Fig. \ref{fig:preprocessing_algorithm} shows a schema of this algorithm's main steps.

This specific image pre-processing algorithm, which includes all the previous steps, has been applied to all the experiments performed, except for a specific one that measures the impact of the normalization algorithm itself on the error. In addition, as in \cite{Chowdhury2018AnEE} we invert the images so that the traces are composed with the pixels of higher intensity on a black background of zero-valued pixels. It makes slightly easier to learn the activations of the CNN component. This inversion has been applied in all the models and results included in this Thesis. 

\begin{figure}[H]
 \centering 
 \includegraphics[width=0.6\textwidth]{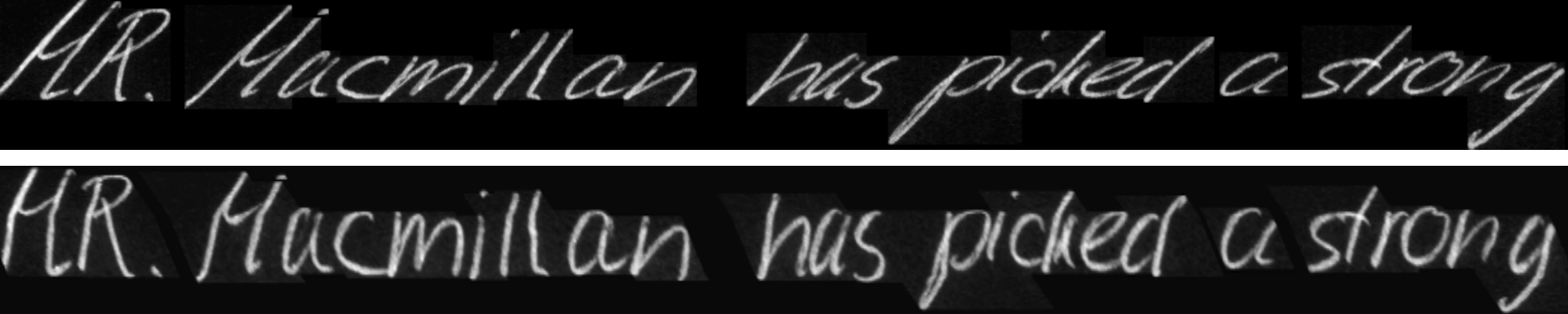}
 \caption{Line preprocessing example: (upper) original line and (lower) preprocessing results.}
 \label{fig:preprocessing_vf:3} 
\end{figure}

The preprocessing performed is the same for the three databases used in the experiments. However, in the IAM database, it was applied at line-of-text level, and for the RIMES and Osborne databases, it is performed at word level. Fig. \ref{fig:preprocessing_vf:3} shows one example of an original line of the IAM database and the result produced after the described normalization process.

In the following paragraphs, we review this new normalization algorithm, detailing its use in the processing carried out in this Thesis and outlined in Fig. \ref{fig:preprocessing_algorithm}.

\subsubsection{Remove the noise of the image}
  \index{Image denoising}
  
The first step of the algorithm is to clean the image. The method proposed by Villegas et al. \cite{villegas2015modification} has been used, which combines the advantages of local binarization methods and preservation of the grayscale. To do this, these authors modified Saviola algorithm \cite{sauvola2000adaptive} so that instead of binarizing each pixel based on the neighborhood, they convert it to a gray value through a linear transformation. An implementation of the algorithm is available at \url{https://github.com/mauvilsa/imgtxtenh}.

\begin{figure}[!ht]
 \centering 
 \includegraphics[width=0.9\textwidth]{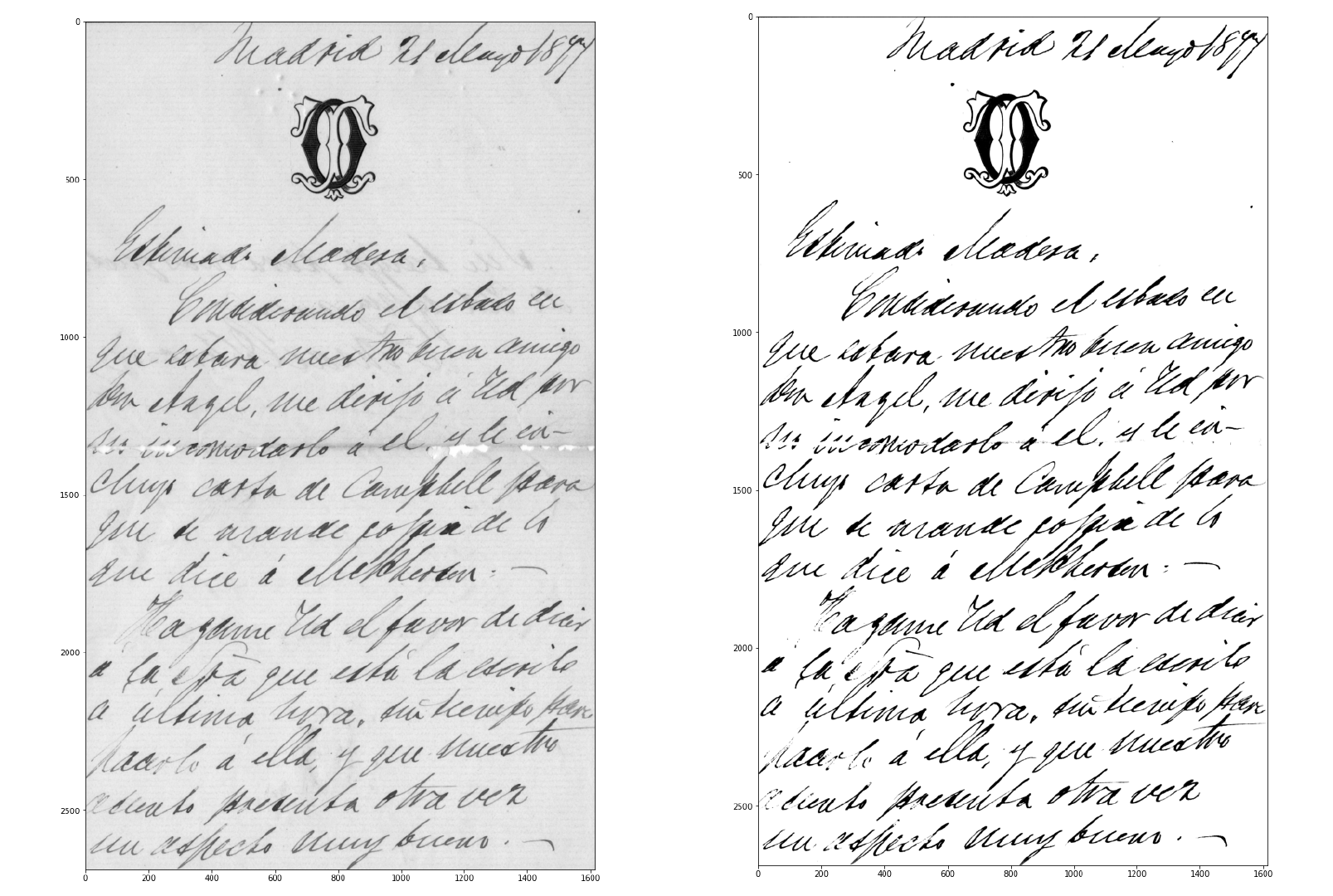} 
 \caption{Example of result of the enhanced contrast algorithm used in this Thesis applied on a Osborne database page.}
 \label{fig:remove_noise_samples}
\end{figure}

Fig. \ref{fig:remove_noise_samples} shows two examples of applying this algorithm on one page of the IAM database and another page for the Osborne database.

\subsubsection{Slope correction}
\label{c6_subsubsection_slope correction}
\index{Slope correction}

\begin{figure}[!ht]
 \centering 
 \includegraphics[width=0.9\textwidth]{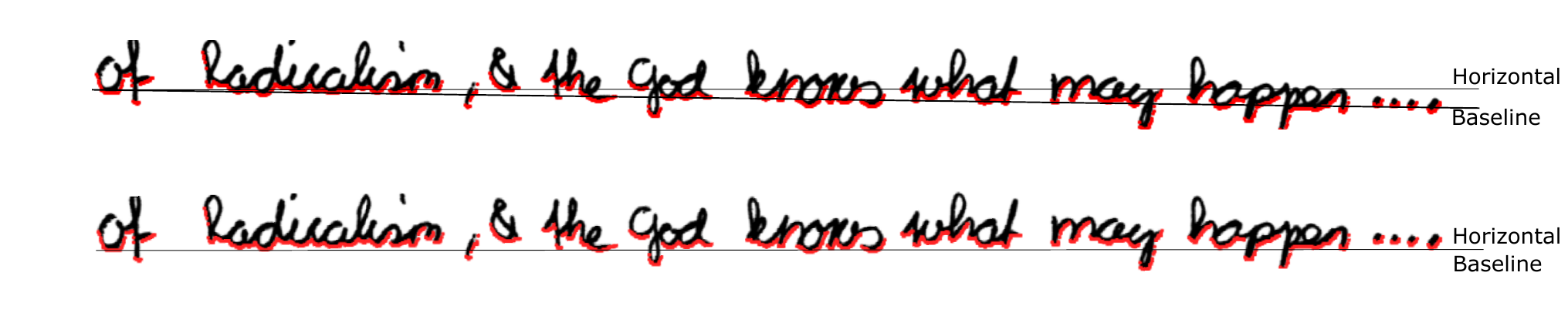} 
 \caption{Example of slope angle estimation used in this Thesis.}
 \label{fig:slope_example}
\end{figure}

The slope correction is carried out separately and before the slant correction described in the next section. For the experiments in this Thesis, we develop a new method similar to Gupta and Chanda \cite{gupta2014efficient}, that uses a linear regression of the {\textit{x, y}} positions of the pixels in the core region. We identify the slope angle from the estimate of the baseline obtained by applying the robust regression \textit{RANdom SAmple Consensus} (RANSAC) \cite{fischler1981random} method on the vertical pixels that exceed a certain gray threshold and that are located in the lowest position of each pixel column. In Fig. \ref{fig:slope_example} these pixels can be seen marked in red in a line of handwritten text from the RIMES database. The robust regression performs an elimination of outliers that mostly correspond to the extreme positions of the descenders.
\index{Random sample consensus}

\subsubsection{Slant correction}
\index{Slant correction}

This Thesis develops a variant of the direct estimation methods provided in papers \cite{you2002slant}, \cite{Kim1997} and \cite{kimura1993improvements}, based on the analysis of 8-directional chain code of the thinned image. The text shear angle is estimated from the analysis of pixels larger than all right boundary pixels, i.e., black pixels with a white pixel on their right. For these pixels, the top three pixels, the upper-left, the upper-center, and the upper-right are checked so that evidence about the local inclination of the line to the left or the right is accumulated. The ratio between the difference of evidence to the right and the normalized evidence to the left provides an estimate of the slant angle tangent. The code is detailed in the algorithm \ref{Alg:slant_correction_algorithm}.

\begin{algorithm}[!ht]\footnotesize
\begin{lstlisting}[language=Python]
def slant_angle(img, th_up=100, th_down=100):
    C = 0
    L = 0
    R = 0
    for w in range(1,img.shape[1]-1):
        for h in range(2,img.shape[0]-1):
            if img[h,w] > th_up and ( img[h, w+1] < th_down):
                if img[h-1, w-1] > th_up:
                    if img[h+1, w] > th_up:
                        L += 1
                        C += 1
                    if img[h+1, w+1] > th_up:
                        L += 2
                elif img[h-1, w] > th_up:
                    if img[h+1, w-1] > th_up:
                        R += 1
                        C += 1
                    if img[h+1, w] > th_up:
                        C += 2
                    if img[h+1, w+1] > th_up:
                        L += 1
                        C += 1
                elif img[h-1, w+1] > th_up:
                    if img[h+1, w-1] > th_up:
                        R += 2
                    if img[h+1, w] > th_up:
                        C += 1
                        R += 1
    return np.arctan2((R-L),(L+C+R))
\end{lstlisting}
 \caption{Slant angle identification algorithm proposed.}
 \label{Alg:slant_correction_algorithm}
\end{algorithm}

Next, by knowing the slant angle $\alpha$ and the image width $w$, we corrected the slant using the affine transformation matrix defined by Eq. (\ref{eq:c6_slant}).

\begin{equation}
\left(\begin{array}{ccc}
1 & -\alpha & 0.5\cdot w\cdot\alpha\\
0 & 1 & 0
\end{array}\right)
\label{eq:c6_slant} 
\end{equation}

\begin{figure}[!ht]
 \centering 
 \includegraphics[width=12cm]{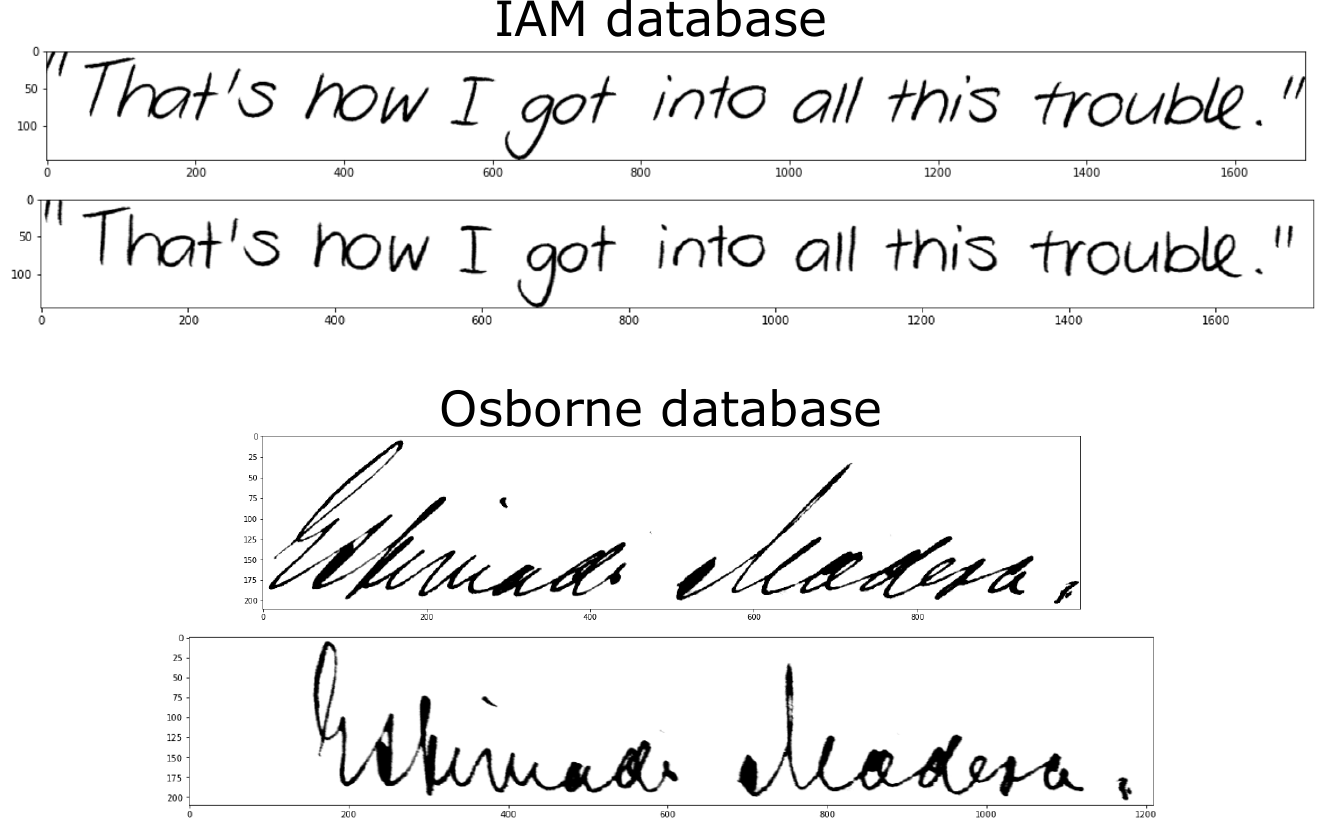} 
 \caption{Slant correction samples over lines of IAM and Osborne databases}
 \label{fig:slant_correction_samples}
\end{figure}

The algorithm application implemented on two examples of line images can be seen in Fig. \ref{fig:slant_correction_samples}. The first one corresponding to the IAM database with a small slant, and the second one corresponding to the Osborne database with a very pronounced slant that makes it difficult to read the text even for humans. The transcription is "Estimado Madera".

\subsubsection{Normalize ascenders and descenders regions}

To normalize the size of the ascender and descender zones is necessary to identify them from the baseline and upperline lines. These lines have already been identified in the previous slope correction step \ref{c6_subsubsection_slope correction}, and the image has been transformed so that these lines are completely horizontal. However, the subsequent slant correction step may have modified them. Therefore, they are recalculated following the robust regression procedure described above but restricting that it must be a horizontal line with no slope angle. This constraint is easy to meet since the text slope has already been eliminated.

Once the horizontal lines of the baseline and upperline have been identified, the image is divided into three horizontal regions as shown in Fig. \ref{fig:calligraphy_zones}. A height rescaling is applied to the ascending and descending regions with a factor proportional to the ratio of the height of the region with respect to the height of the core region, as indicated in the equation \ref{eq:ratio_ascenders}.

\begin{equation}
    r= 
\begin{cases}
    \frac{h_{cr}}{h_r},& \text{if } h_r > h_{cr}\\
    1,              & \text{otherwise}
\end{cases}
\label{eq:ratio_ascenders}
\end{equation}

\begin{figure}[!ht]
\centering 
\includegraphics[width=15cm]{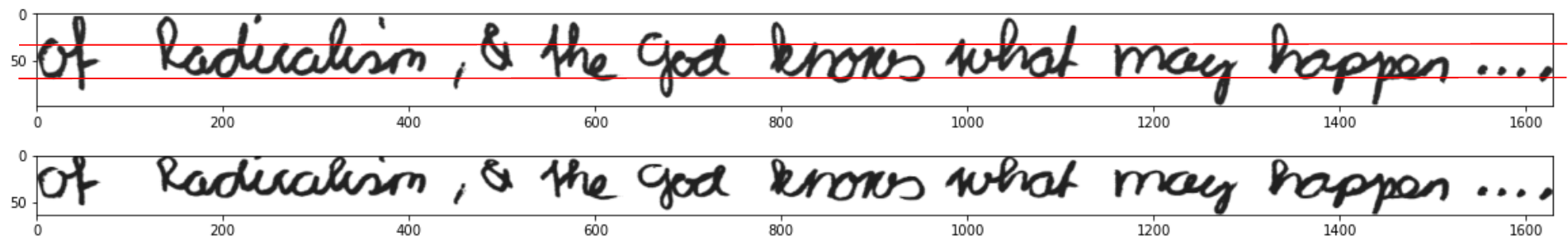} 
\caption{Example of normalization of the ascenders and descenders regions.}
\label{fig:normalize_ascenders}
\end{figure}

Fig. \ref{fig:normalize_ascenders} shows an example of a line from the IAM database on which the previous algorithm has been applied to normalize the ascending and descending regions.

\subsubsection{Crop and resize input images}

Finally, an image centering and resizing process is performed as follows. First, the left and right empty edges of each image are cut out, but the lower and upper edges are not changed. It is done to keep the relative vertical position of the baseline and upperline of the different images when this is possible. It happens when they are not already cropped in the original source, as is the case with Osborne's database. When the words top and bottom edges that have only ascenders or only descenders are removed, the relative positions of the baseline and the upperline are changed, sometimes significantly. Thus, a word-centered image that has only ascenders will have its core region at the bottom of the word, and, conversely, another word that has only descenders will have its core region at the top. Since we want to avoid this and keep the core region in the center of the image in all cases, no trimming of the top and bottom edges is done.

Secondly, the image is resized to a fixed height of 48 pixels, which has been visually checked to ensure that it does not alter text recognition by humans. In addition, empty columns up to a width of 192 pixels are added on the left. With this size of 48$\times$192 pixels, it has been proven that it is possible to correctly transform more than 99.9\% of the images in the databases used. For the few cases in which the word image is longer than 192 pixels after resizing, its size is adjusted to 192 by changing the aspect ratio of the word. In addition, we experiment with three different resolutions, 32$\times$128, 48$\times$192, and 64$\times$256, to evaluate the impact of this selection on the final result.  

\begin{figure}[!ht]
\centering 
\includegraphics[width=8cm]{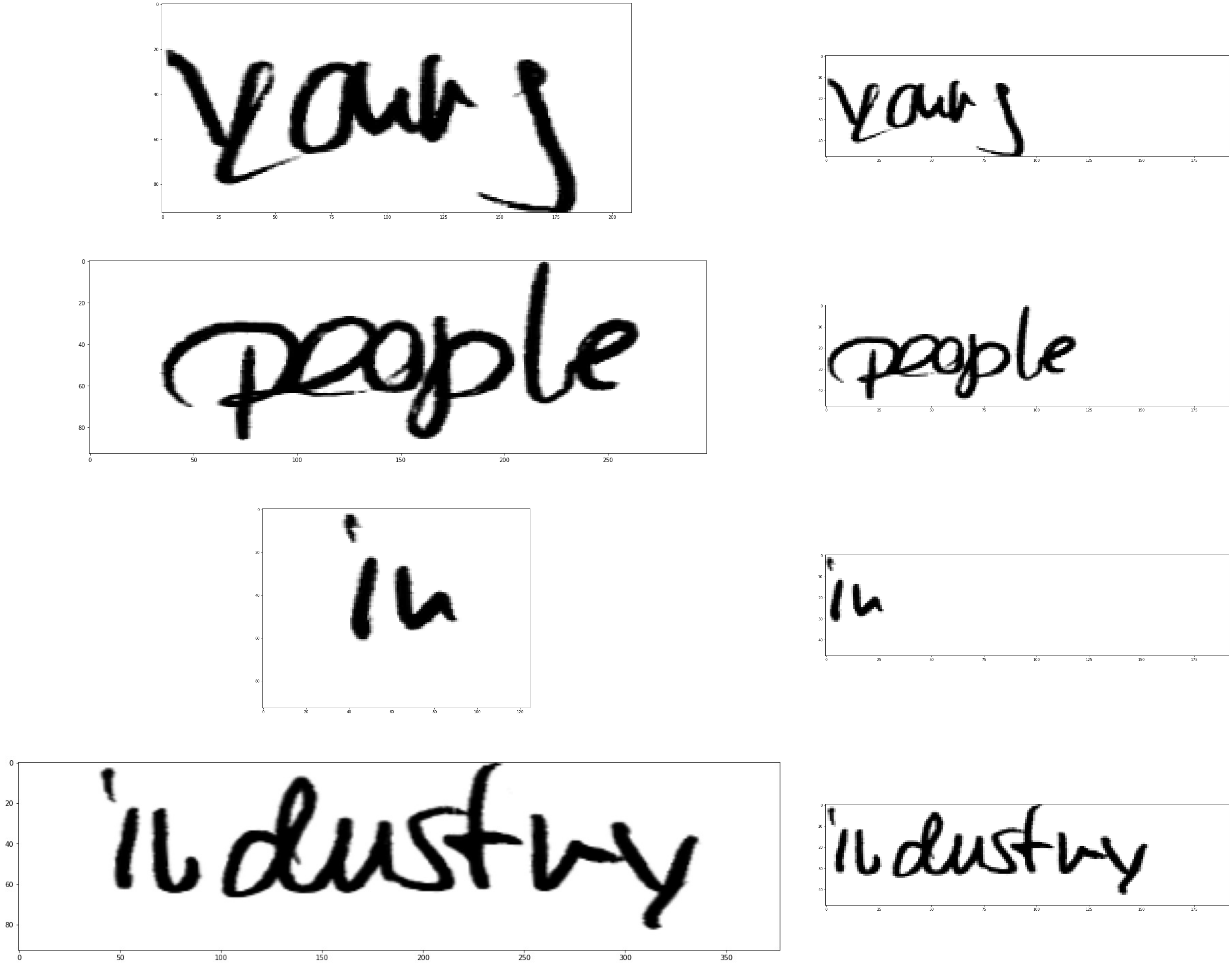} 
\caption{Example of crop an resize word images: original at the left, and resized ones to fixed size at right.}
\label{fig:crop_and_resize}
\end{figure}

Fig. \ref{fig:crop_and_resize} shows several examples of images before and after the cropping and resizing described above. The images are printed, keeping the relative sizes between them.

\subsection{Data augmentation}
\label{C6 - section - Data augmentation}
\index{Data augmentation}

The high variability of HTR data, and the fact that existing databases for model development are limited in size, make difficult the training process of deep architecture models such as the one proposed. Therefore, we have experimented with applying data augmentation in training following the strategy of introducing the data augmentation process in each iteration. This integration makes it possible that a set of input data different from those of the previous epoch is obtained in each epoch of the model training algorithm. It allows performing the training with a theoretically infinite sample of data. Thus the training algorithm is defined next. It is the identical implementation of the data augmentation done in other HTR publications as, for example, in \cite{puigcerver2017multidimensional}.

\begin{algorithm}[!ht]\footnotesize
\begin{lstlisting}[language=Python]
def train(train_data, model, lr=learning_rate):
    model.w = initialization()
    for epoch in range(num_epochs):
        img_augmented = augmentation(train_data)
        y_pred = model.forward_pass(img_augmented)
        w_gradients = model.backward_pass(y, y_pred)
        model.w = model.w + (lr * w_gradients)
    return model.w
\end{lstlisting}
 \caption{Train algorithm with data augmentation integrated.}
 \label{Alg:train_data_augmentation}
\end{algorithm}

The proposed algorithm performs random transformations of data that are invariant to the text content present in the image. The transformations used are: translations, resizings, slant modifications, elastic distortions, random projective transforms, and erode and dilate transformations.

The elastic distortion transformation is a technique based on the realization of small local distortions of the image. For this purpose it is defined a grid of points in the image, on which tiny random modifications of the points position are applied. The elastic distortion transformation is obtained by adapting the original image to the distorted grid of points. This transformation imitates the small oscillations in the stroke present in many handwritten calligraphies.

Random projective transform, also called homography, changes an image as if the point of view from which the image is viewed has been modified. Projective geometry is used to do this. 

The erode and dilate transformations are another type of morphological transformations that allow to increase or decrease the size of the elements in the foreground of the image. The dilation operation adds points to the edges of the foreground image. If the foreground image is composed of handwritten text strokes, the size of these strokes is increased. In the case of dilation transfer, the thickness of the strokes will be reduced.

\begin{figure}[!ht]
 \centering 
 \includegraphics[width=10cm]{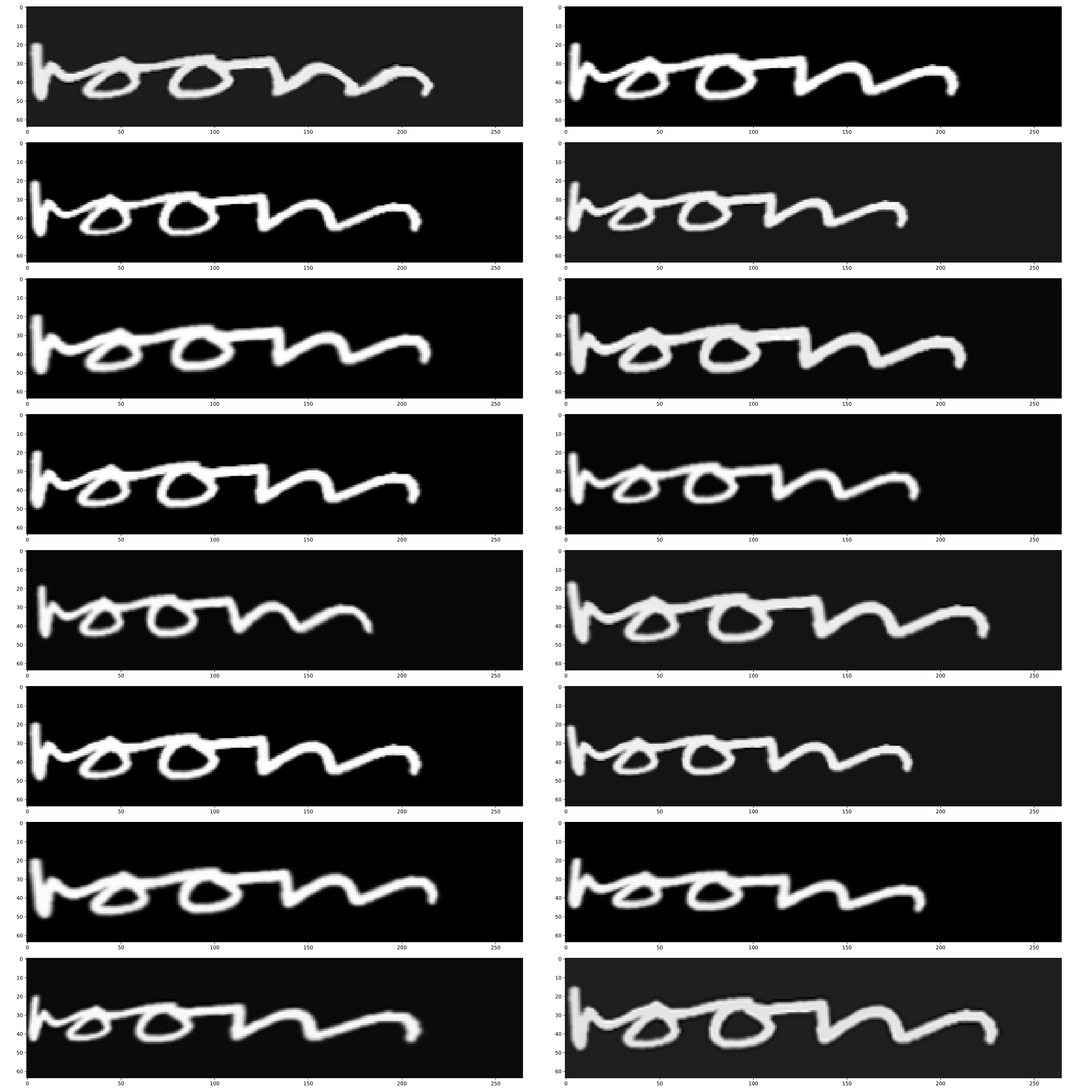} 
 \caption{Examples of images generated by the data augmentation algorithm used in this Thesis.}
 \label{fig:data_augmentation_example}
\end{figure}

Fig. \ref{fig:data_augmentation_example} show some examples of the word "room" augmented by the previous transformations. 

The detailed algorithm is included in \ref{Alg:data_augmentation_thesis}. The algorithm defines a level of randomness and an order of application of the above transformations. It places the smoothest transformations first, by assigning them a higher probability of application, and leaves for the last the most aggressive transformations, which are the erosion and dilation transformations with a lower probability of occurrence. For a given image, several transformations can be applied consecutively, producing a final composite transformation.

\begin{algorithm}[!ht]\footnotesize
\begin{lstlisting}[language=Python]
def augmentation(img):
    if random_uniform(0, 1) < 0.5:
        img = ElasticDistortion(img)
    if random_uniform(0, 1) < 0.5:
        img = RandomTransform(img)

    # Skew
    if random_uniform(0, 1) < 0.8 :
        angle = random_uniform(-0.15, 0.15)
        img = affineTransformation(img, angle)

    #Resize
    if random_uniform(0, 1) < 0.4:
        img = resize_down(img)
    elif random_uniform(0, 1) < 0.4:
        img = resize_up(img)

    #Erode - dilate
    if random_uniform(0, 1) < 0.3:
        img = erode(img, iterations=1)
    elif random_uniform(0, 1) < 0.3:
        img = dilate(img, iterations=1)

    return img
\end{lstlisting}
 \caption{Data augmentation algorithm proposed.}
 \label{Alg:data_augmentation_thesis}
\end{algorithm}

\subsection{Lexicons}
\label{C6 - section - Lexicons}
\index{Lexicon}

For each considered handwriting word database (IAM, RIMES and Osborne, respectively), a standard lexicon is used in this chapter. It is selected considering that it is a lexicon mainly accepted in the literature or in its absence that it is easy to obtain or construct to facilitate the reproducibility and comparison of the results.

The application of each lexicon to decode the results is made by a systematic search of the lexicon word closest to the resulting string of characters in the visual model, in order to select the one that minimizes Levenshtein's distance \cite{stuner2020handwriting}.
\index{Levenshtein distance}

In addition to the results obtained with the standard lexicons, validation metrics have been computed, by considering as a result the direct output of the visual model without any kind of subsequent decoding. It facilitates the direct comparison of different visual models, each one with each other. 

Finally, results have also been provided using as lexicons the word sets present in the test partitions of the different corpora. Obviously, this does not provide a realistic result of the model's ability to recognize the text, but it allows the error impact produced by the OOV words to be measured when the standard lexicon is used. In addition, it provides an easy way to compare different models in the case of optimal decoding conditions.

\subsubsection{Lexicons applied with the IAM database}

The standard validation lexicon used was the set of 50,000 most frequent words in the union of Lancaster-Oslo/Bergen (LOB) \cite{johansson1986tagged} and the Brown \cite{francis1989manual} corpora when previously the IAM paragraphs from the Brown corpus were removed. Table \ref{table:IAM_LOB_pages} contains the complete list of excluded pages of the LOB corpus used in the IAM database.
\index{Corpus!LOB}
\index{Corpus!Brown}

\begin{table}[!ht]\footnotesize
\centering 
\begin{tabular}{c c c c c c c } 
 \toprule
 A1 & A2 & A3 & A4 & A5 & A6 & -  \\
 B1 & B2 & B3 & B4 & B5 & B6 & -  \\
 C1 & C2 & C3 & C4 & -  & C6 & -  \\
 D1 & -  & D3 & D4 & D5 & D6 & D7 \\
 E1 & E2 & -  & E4 & -  & E6 & E7 \\
 F1 & F2 & F3 & F4 & -  & -  & F7 \\
 G1 & G2 & G3 & G4 & G5 & G6 & G7 \\
 H1 & H2 & -  & H4 & H5 & H6 & H7 \\
 J1 & -  & -  & J4 & -  & J6 & J7 \\
 K1 & K2 & K3 & K4 & -  & -  & K7 \\
 L1 & -  & L3 & L4 & -  & -  & L7 \\
 M1 & M2 & M3 & M4 & -  & M6 & -  \\
 N1 & N2 & N3 & N4 & -  & N6 & -  \\
 P1 & P2 & P3 & -  & -  & P6 & -  \\
 -  & R2 & R3 & -  & -  & R6 & -  \\
 \bottomrule
\end{tabular}
\caption{List of excluded pages of the LOB corpus used in the IAM database.}
\label{table:IAM_LOB_pages}
\end{table}

To identify the words in the previous corpora, we use a blank space tokenizer, and next, we isolate the punctuation marks because in the IAM database these characters are considered as single words.

\subsubsection{Lexicons applied with the RIMES database}

All the present words in the training and validation partitions are used as a standard lexicon in the RIMES database. It includes a total of 5,334 different words. Using the previous lexicon, the number of OOV words in the test partition is 410. 

In this Thesis, for the RIMES database we use exactly the transcriptions of words provided with that database, differentiating between capital and small letters. However, other authors such Stuner et al. \cite{stuner2020handwriting}, or Menasri et al. \cite{menasri2012a2ia} use only lowercase letters because they indicate that the labeling of certain uppercase characters, especially the letter 'j' (in the words 'je' or 'j') has errors.

\subsubsection{Lexicons applied with the Osborne database}

For the Osborne database, we also used all the words present in the train and validation partitions as a standard lexicon. In this case, we have a total of 2,708 different words, and the OOV words in the test set are 185. 

\section{Experiments}
\label{C6 - section - Experiments}

This section details the experimental results achieved with the proposed architecture for the offline continuous HTR problem. The experiments aim to explore this architecture capabilities and identify the components and parameterizations that provide the best results in terms of recognition capability. 

The experiments have been repeated on three handwritten text databases of three different languages to validate the consistency of the experiment results. First, two well-known handwriting datasets were selected: IAM offline handwriting \cite{marti2002iam} in English, and RIMES databases \cite{augustin2006rimes} in French. These two databases have been extensively used by multiple authors and allow us to evaluate the proposed model compared to the results of other authors. In addition, we also provide results for the new Osborne database in Spanish. The Osborne database is especially challenging because it is a dataset of historical documents with very accentuated cursive handwriting and this dataset is smaller in size than the other two ones.

The handwritten text recognition ability in each experiment is measured using the metrics of Character Error Rate (CER) and Word Error Rate (WER) \cite{frinken2014continuous} metrics described in Section \ref{SubSect:Evaluation_metrics}. In summary, CER measures the Levenshtein distance \cite{levenshtein1966binary} between the predicted and the real character sequence of the word. WER represents the percentage of words correctly identified by the model and, therefore, defines the model accuracy. Since the goal of our experiments is to obtain conclusions about the performance of the visual model, WER and CER values provided in the experimental results are obtained by the direct application of the model, without using any lexicon in the decoding.
\index{Metrics!Character error rate}
\index{Metrics!Word error rate}

First, in Subsection \ref{C6 - Experiments - Baseline}, baseline results obtained using an initial configuration of the model selected, based on the results of the experiments carried out in the previous chapter (i.e., isolated character recognition and the most common configurations that this architecture usually has when applied to different problems) are presented \cite{bahdanau2016end} \cite{DBLP:journals/corr/JaderbergSVZ14} \cite{Chorowski2017towards}. Then, new experiments are carried out to explore the capabilities and limitations of the proposed architecture. The type of experiments performed are categorized as follows:

\begin{itemize}

    \item Experiments about image segmentation strategy. To identify the impact of the patch width and the step size parameters of the image segmentation into the error. The results are detailed in subsections \ref{c6_experiment_segmentation} and \ref{c6_experiment_segmentation_patch}.
    
    \item Experiments for the different convolutional architectures analyzed in Chapter \ref{chapter_character_models} regarding isolated character recognition. The results are detailed in Subsection \ref{C6 - experiments - Convolutional_architectures}.

    \item Experiments to select the most suitable encoder architecture in terms of the RNN type to be used, its dimension, and the number of layers. The results of these experiments are included in Subsection \ref{C6 - Experiments - Encoder architecture}.

    \item Experiments on the impact of the input image resolution on the error, with detailed results in Subsection \ref{C6 - Experiments - Image resolution}.

    \item Experiments on the importance of data normalization and data augmentation, detailed in Subsection \ref{C6 - Experiments - Normalization and data augmentation}.

    \item Experiment on the application of the teacher forcing strategy during training, with results in Subsection \ref{C6 - Experiments - Teacher forcing}.

\end{itemize}

From the results of the different previous experiments, a more suitable model configuration is selected. The last experiment is performed with this configuration to obtain a final result of the proposed model that can be compared with the results of other authors on the same databases. These results are detailed in Subsection \ref{C6 - Experiments - Final model}

In the case of the runs corresponding to the baseline and final models, these provide additional results using the selected lexicons in the decoding process. In particular, we offer three main results for each experiment:
\begin{itemize}

    \item First, the results obtained by the direct application of the visual model (without any lexicon usage).
    \item Second, we process the outputs of the visual model by finding the closest word in terms of Levenshtein distance \cite{levenshtein1966binary} over standard lexicons for IAM and RIMES databases, extensively used in the handwriting literature and described in Subsection \ref{C6 - section - Lexicons}.
    \item Third, we apply the same search using the test set lexicon.
 \end{itemize}

Finally, we dedicate the final Subsection \ref{C6 - Experiments - Error analysis} to analyze the nature of the error by investigating the error types that we got in order to propose some possible improvements to the current model.

\subsection{Baseline experiments}
\label{C6 - Experiments - Baseline}

An initial base configuration of the model has been defined to obtain a reference result that will be used to contrast the subsequent results of the different experiments performed. The selection of the configuration is based on the criteria that this is not excessively complex and that, based on our experience, can obtain good a priori results on the proposed problem. This configuration is as follows:

\begin{itemize}
  \item The strategy of converting the input image into a sequence of patches is selected as indicated in Subsection \ref{section: Image patches architectures}. The patch configuration is set up with an initial patch width of 10 pixels and a step size of 2 pixels.
  \item As convolutional architecture, LeNet applied on the patch sequence has been selected. \index{LeNet}
  \item The encoder is configured with two 256 LSTM layers.
  \item The height of the input size is set to 48
  \item The experiments are performed on the normalized data without any data augmentation in training and using the teacher forcing strategy.
\end{itemize}

With this configuration, experiments have been performed with the IAM, RIMES, and Osborne databases, and the following tables detail the results obtained for each of the decodings by lexicons used in all the experiments performed.

Table \ref{table:Baseline results. No lexicon.} includes CER and WER data for the validation and test partitions corresponding to the results obtained directly by the model, without using any lexicon to decode them.

\begin{table}[H]
\centering
 \begin{tabular}{l r r r r r}
 \toprule
 Dataset &  \multicolumn{2}{l}{Validation} & & \multicolumn{2}{l}{Test} \\
 \cmidrule(r){2-3} \cmidrule(r){5-6}
 & CER & WER & & CER & WER \\
 \midrule
 IAM     &  8.5 & 19.3 && 12.6 & 27.4 \\
 RIMES   &  6.0 & 18.0 &&  6.1 & 17.8 \\
 Osborne & 16.3 & 33.5 && 20.9 & 44.0 \\
 \bottomrule
 \end{tabular}
\caption{Baseline results with no lexicon.}
\label{table:Baseline results. No lexicon.}
\end{table}

If the standard lexicons to decode the results of the visual model are used, we obtain CER and WER data detailed in Table \ref{table:Baseline results. Standard lexicon.}.

\begin{table}[!ht]
\centering
 \begin{tabular}{lrrrrr}
 \toprule
 Dataset &  \multicolumn{2}{l}{Validation} & & \multicolumn{2}{l}{Test} \\
 \cmidrule(r){2-3} \cmidrule(r){5-6}
 & CER & WER & & CER & WER \\
 \midrule
 IAM     & 10.6 & 17.9 && 14.3 & 24.5 \\
 RIMES   &  7.0 & 14.2 &&  6.9 & 14.3 \\
 Osborne & 13.1 & 21.9 && 28.2 & 42.8 \\
 \bottomrule
 \end{tabular}
\caption{Baseline results using standard lexicon.}
\label{table:Baseline results. Standard lexicon.}
\end{table}

Finally, the results in Table \ref{table:Baseline results. Test lexicon.} are obtained by using the test lexicon to decode the model predictions. In this case, it only makes sense to apply this lexicon to the test data. 

\begin{table}[!ht]
\centering
 \begin{tabular}{lrrrrr}
 \toprule
 Dataset &  \multicolumn{2}{l}{Validation} & & \multicolumn{2}{l}{Test} \\
 \cmidrule(r){2-3} \cmidrule(r){5-6}
 & CER & WER & & CER & WER \\
 \midrule
 IAM     &  NA &  NA & & 10.3 & 17.0 \\
 RIMES   &  NA &  NA & &  3.6 &  7.2 \\
 Osborne &  NA &  NA & &  8.2 & 13.9 \\
 \bottomrule
 \end{tabular}
\caption{Baseline results using the test lexicon.}
\label{table:Baseline results. Test lexicon.}
\end{table}

\subsection{Image segmentation: step size}
\label{c6_experiment_segmentation}
\index{Image segmentation}

In this first set of experiments, we analyzed the effect over the metrics of CER and WER of the different parametrizations corresponding to the initial process of image patches creation described in Subsection \ref{section: Image patches architectures}. We considered two main parameters: the patch width and the step size to create the patches. For example, a patch width of 5 and a step size of 2 means that the first patch is created with the columns from 1 to 5, the second patch with the columns from 3 to 7, and so on.

We started analyzing the step size. Training is performed on the baseline model by modifying only the step size parameter with the following values: 1, 2 (baseline), 4, 6, and 8. It is recalled that in the baseline model, the patch width is 10.

Table \ref{table:step_size_results} shows the results of the experiments with the step size. It is observed that the results for step sizes 1, 2, and 4 are pretty similar in each database. From step sizes greater than 4, the error levels measured in terms of WER and CER increase with increasing the step size.

\begin{table}[!ht]
\centering
 \begin{tabular}{llrrrrr}
 \toprule
 Dataset &   & \multicolumn{2}{l}{Validation} & & \multicolumn{2}{l}{Test} \\
 \cmidrule(r){3-4} \cmidrule(r){6-7} & Step size& CER & WER & & CER & WER \\
 \midrule
       & 1          &                   8.6 &           20.0 &&         12.6 &          27.4 \\
       & 2 & \textbf{8.5}& \textbf{19.3} && \textbf{12.6}&  \textbf{27.4}\\
 IAM   & 4          &                   8.8 &          20.3 &&          12.9 &          28.0 \\
       & 6          &                   9.3 &          22.1 &&          13.6 &          29.5 \\
       & 8          &                   10.2&          23.2 &&          14.9 &          31.1 \\
 \midrule
       & 1          &                   6.1 &   \textbf{17.7}&&        6.1 &          17.9 \\
       & 2&\textbf{6.0}&           18.0 &&\textbf{6.1}&  \textbf{17,8}\\
RIMES  & 4          &                   6.4 &           18.2 &&        6.3 &          18.0 \\
       & 6          &                   6.6 &           19.0 &&        6,4 &          18.4 \\
       & 8          &                   7.5 &           20.3 &&        7.4 &          20.0 \\
 \midrule
       & 1          &                   17.6 &         36.1 &&        23.2 &         46.4 \\
       & 2& \textbf{16.3}& \textbf{33.5}&&\textbf{20.9}&         44.0 \\
 Osborne & 4        &                   16.4 &         35.1 &&        21.0 & \textbf{43.5}\\
       & 6          &                   18.7 &         37.6 &&        23.6 &         46.0 \\
       & 8          &                   21.3 &         41.0 &&        27.4 &         51.7 \\
 \bottomrule
 \end{tabular}
\caption{Step size experiments results (baseline in step size of 2). Best result, by database, is presented in bold.}
\label{table:step_size_results}
\end{table}

\begin{figure}[!ht]
\centering
  \includegraphics[width=0.9\textwidth]{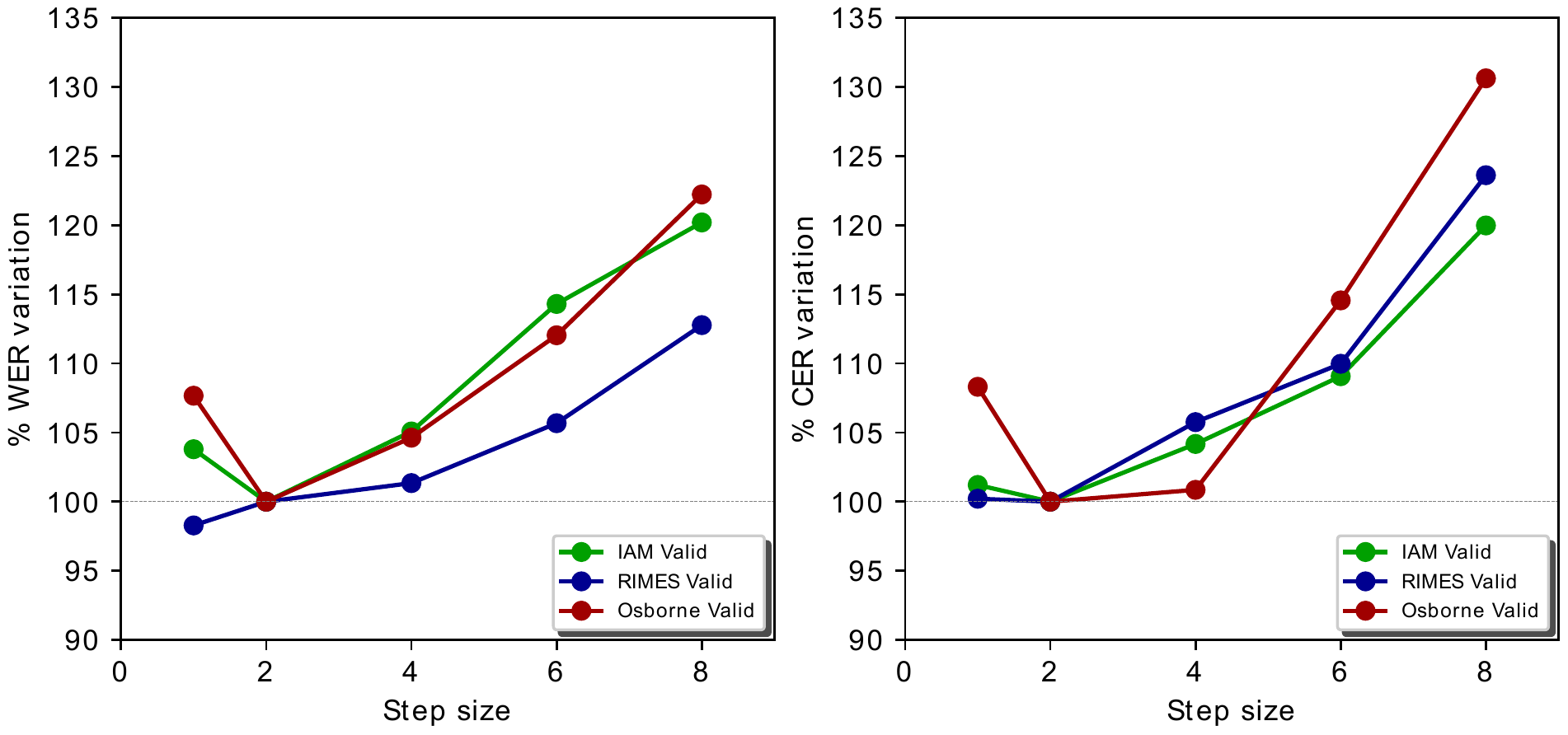}
\caption{Variation percentage of WER and CER in validation and test partitions for different step sizes over the baseline.}
\label{fig:WERVal_vs_stepize} 
\end{figure}

The relative difference of using different step sizes can be seen in Fig. \ref{fig:WERVal_vs_stepize}. It shows the relative increase or decrease of WER and CER concerning the baseline model, which in the graph shows the value 100. 

We observe the increasing trend of the error as a function of the step size, with the only exception of the step size of 1 applied to the Osborne database.

\subsection{Image segmentation: patch width}
\label{c6_experiment_segmentation_patch}
\index{Image segmentation}

The patch width, i.e., the width of each of the patches extracted from the image, is now analyzed. The patches are overlapped unless the step size and patch width coincide. Because we used a convolutional architecture over the patches, the patch width must have a minimal size to apply the convolutions. This minimal width is 5 because in LeNet architecture the initial convolutional layer has a filter of 5$\times$5. The baseline model uses a patch width of 10 with a step size of 2 applied on a LeNet architecture. The experiments performed analyze patch widths between 5 and 25 in intervals of 5. 

Table \ref{table:Patch_size_results} shows the results obtained in the experiments performed to analyze the effect of patch width on the error.

\begin{table}[!ht]
\centering
 \begin{tabular}{llrrrrr}
 \toprule
 Dataset &   & \multicolumn{2}{l}{Validation} & & \multicolumn{2}{l}{Test} \\
 \cmidrule(r){3-4} \cmidrule(r){6-7} & Patch width& CER & WER & & CER & WER \\
 \midrule
     & 5            &       \textbf{8.3}&	        19.8 &&	        12.7 &	        27.7 \\
     & 10           &      8.5&    \textbf{19.3}&&         12.6 &          27.4 \\
 IAM & 15           &               8.5&	        19.6 &&	\textbf{12.5}&	\textbf{27.3}\\
     & 20           &               8.9&	        20.4 &&	        12.9 &	        28.0 \\
     & 25           &               9.3&	        21.2 &&	        13.4 &	        28.7 \\
 \midrule
       & 5          &               6.3&	        18.5 &&	        6.2 &	        18.1  \\
       & 10         &    6.0&            18.0 &&         6.1 &           17,8 \\
 RIMES & 15         &       \textbf{5.9}&	\textbf{17.3}&& \textbf{5.7}&	\textbf{16.8}\\
       & 20         &               6.4&	        18.8 &&	        6.2 &	        18.4 \\
       & 25         &               6.3&	        18.5 &&	        6.2 &	        18.3 \\
 \midrule
       & 5          &               17.7 &            35.1 &&           24.2 &            46.0 \\
       &         10 &   16.3 &            33.5 &&           20.9 &            44.0 \\
 Osborne & 15       &       \textbf{15.9}&    \textbf{32.0}&&           19.9 &            43.8 \\
       & 20         &               16.5 &            34.2 &&   \textbf{19.5}&    \textbf{42.6}\\
       & 25         &               19.2 &            38.2 &&           23.6 &            48.3 \\
 \bottomrule
 \end{tabular}
\caption{Patch width experiments results (baseline in patch width of 10). Best result by database in bold.}
\label{table:Patch_size_results}
\end{table}

Table \ref{table:Patch_size_results} shows that the use of different patch widths affects the model results relatively little. In terms of WER metric in validation, which is used in the training stop criterion, the minimum value is obtained for a patch width of 15. In any case, the results with patch widths between 5 and 20 are similar, and only with a patch width of 25, it starts to be seen an increase in the error metrics that is consistent between databases.

\begin{figure}[!ht]
\centering
  \includegraphics[width=0.9\textwidth]{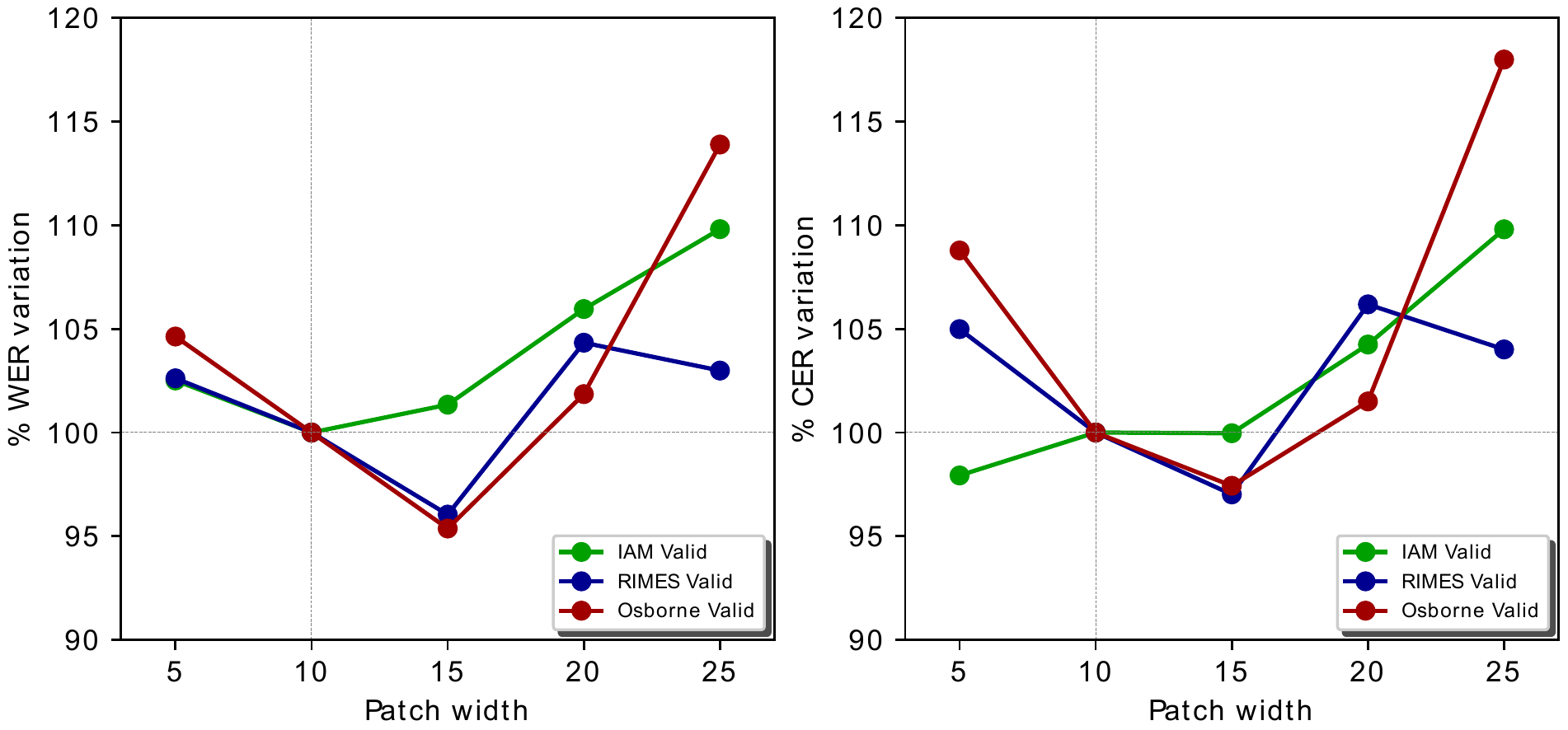}
\caption{Variation percentage of WER and CER in validation and test partitions with different patch widths over the baseline.}
\label{fig:WERVal_vs_patchSize} 
\end{figure}

Fig. \ref{fig:WERVal_vs_patchSize} shows the relative difference of using different patch widths. It presents the relative percentage increase or decrease of WER and CER concerning the baseline model, which corresponds to a patch width of 10, and which in the graph takes the reference value of 100.

It is consistently observed in the three databases that the patch width of 15 gives the best results and that smaller or larger patch widths increase the error. In any case, the relative differences are small with patch widths between 5 and 20.

\subsection{Convolutional architectures}
\label{C6 - experiments - Convolutional_architectures}
\index{Convolutional neural network}

The different convolutional architectures proposed are discussed below. The two selected architectures are LeNet and VGG. In addition, each convolutional architecture can be used with an image patching strategy or directly on the model input image. In this way, we experiment with four possible configurations regarding convolutional architecture:

\begin{itemize}
  \item LeNet architecture on image patches.
  \item VGG architecture on image patches.
  \item LeNet architecture on full image.
  \item VGG architecture on full image.
\end{itemize}

The results of these four experiments for each of the three considered databases are presented in Table \ref{table:convolutional_experiments}.

\begin{table}[!ht]
\centering
 \begin{tabular}{lllrrrrr}
 \toprule
 Dataset &   & & \multicolumn{2}{l}{Validation} & & \multicolumn{2}{l}{Test} \\
 \cmidrule(r){4-5} \cmidrule(r){7-8} & Patches & Architecture & CER & WER & & CER & WER \\
 \midrule
        &Yes                &LeNet  &           8.5&    \textbf{19.3}&&         12.6 &          27.4 \\
  IAM   &Yes                &VGG    &           9.1&            20.6 &&         13.3&           28.6 \\
        &No                 &LeNet  &           8.5&            19.9 &&         12.4&           27.1 \\
        &No                 &VGG    &   \textbf{8.3}&           19.6 && \textbf{12.0}&  \textbf{26.6}\\
 \midrule 
        &Yes                &LeNet  &           6.0&            18.0 &&         6.1&            17,8 \\
  RIMES&Yes                 &VGG    &           6.1&            18.2 &&         5.7&            17.7 \\
  &No                       &LeNet  &           5.7&    \textbf{16.5}&&         5.5&    \textbf{16.2}\\
  &No                       &VGG    &    \textbf{5.4}&          16.6 && \textbf{5.4}&           16.3 \\
 \midrule
            &Yes            &LeNet  &   \textbf{16.3}&  \textbf{33.5}&&         20.9 &          44.0 \\
  Osborne   &Yes            &VGG    &           36.2 &          47.2 &&         38.3 &          54.8 \\
            &No             &LeNet  &           17.1 &          35.2 && \textbf{19.9}&  \textbf{42.1}\\
            &No             &VGG    &           NA   &          NA   &&         NA   &          NA   \\
 \bottomrule
 \end{tabular}
\caption{Convolutional architectures experiments results (baseline in Lenet and patches). Best result by database in bold.}
\label{table:convolutional_experiments}
\end{table}

The behavior of the model depending on the convolutional architectures used is quite different for each database. With IAM database, the results are similar using patches and full images, obtaining a lower WER in validation with LeNet architecture applied on patches. In the case of RIMES database, it is clearly observed that the error decreases when using the convolutional component directly on the image instead of on the patches. In the case of the Osborne database, the VGG architecture does not work correctly. In the case of using it on patches, the error increases notably, and when trying to apply it to the whole image, the training of the model does not converge to an optimal solution.

\begin{figure}[!ht]
\centering
  \includegraphics[width=0.9\textwidth]{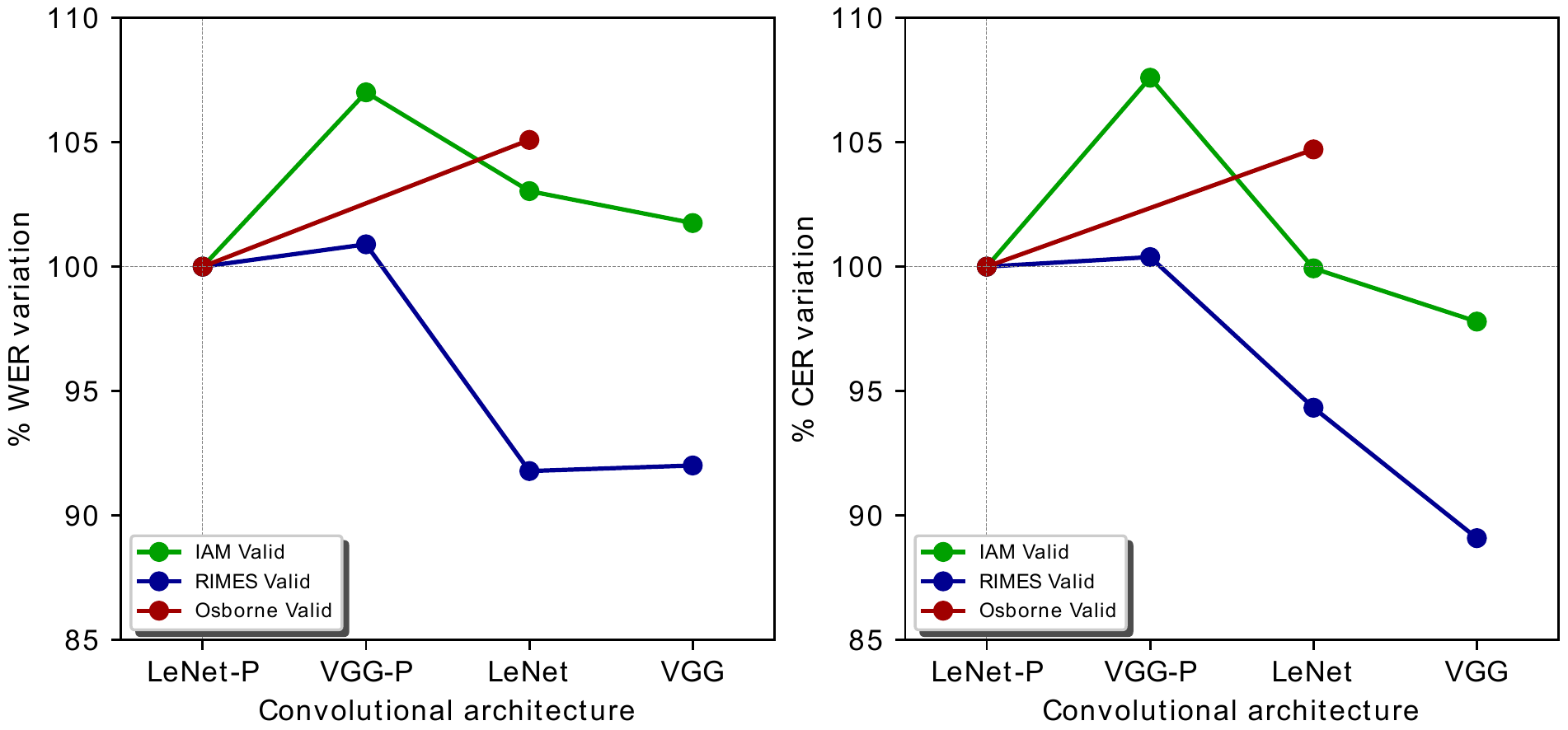}
\caption{Variation percentage of WER and CER in validation and test partitions with different convolutional architectures over the baseline.}
\label{fig:WERVal_vs_convolutional} 
\end{figure}

The graph in Fig. \ref{fig:WERVal_vs_convolutional} shows the relative differences between the convolutional architectures proposed. The figure shows the relative increase or decrease percentage of WER and CER concerning the baseline model that corresponds to a LeNet architecture applied on patches that in the graph takes the reference value 100.

It is observed that, in general, the relative differences are less than 5\%, except for the results of the full-image experiments performed on RIMES database. It is not easy to theorize on a possible cause of this behavior. It is proposed as a hypothesis that it could be related to the fact that the average word length in RIMES database is shorter than in the other two databases, and that may favor models applied on the whole image rather than on patches. This hypothesis should be tested in future work.

\subsection{Encoder architecture}
\label{C6 - Experiments - Encoder architecture}

Next, we analyzed the effect of the encoder architecture on the final error. Four encoder characteristics were analyzed and hypothesized to be relevant for the model performance. The following characteristics have been analyzed by taking into account that in the baseline model, the encoder is defined as a bidirectional LSTM-type RNN with two 256 layers:

\begin{itemize}
  \item First, the size of the RNN layers is analyzed, experimenting with values of 64, 118, 256, and 512.
  \item Second, the number of RNN layers is analyzed, considering both the number of layers and whether or not they are bidirectional. We experiment with values of 1, 2, and 3 layers for both the bidirectional and unidirectional cases.
  \item Finally, the RNN type is analyzed, experimenting with LSTM and GRU types.
\end{itemize}

\subsubsection{RNN dimension}

In the first experiment, we used a fixed baseline model with a two-layer bidirectional LSTM-type RNN encoder and built different experiments varying the size of the RNN layers. The results of the experiments for each of the three databases are included in Table \ref{table:encoder_dimension_results}.

\begin{table}[!ht]
\centering
 \begin{tabular}{llrrrrr}
 \toprule
 Dataset &   & \multicolumn{2}{l}{Validation} & & \multicolumn{2}{l}{Test} \\
 \cmidrule(r){3-4} \cmidrule(r){6-7} & Encoder size& CER & WER & & CER & WER \\
 \midrule
    &                       64 &        9.7 &	        22.8 &&	        13.9 &	        30.3 \\
 IAM&                       128&        8.3 &	        19.6 &&	        12.8 &	        27.9 \\
    &                       256&        8.5 &           19.3 &&         12.6 &          27.4 \\
    &                       512&\textbf{8.0}&   \textbf{18.6}&& \textbf{11.7}&  \textbf{26.0}\\
 \midrule
      &                     64 &        7.6 &	        22.3 &&	        7.3 &	        21.8 \\
 RIMES&                     128&        6.3 &	        18.1 && \textbf{6.0}&	        17.6 \\
      &                     256&\textbf{6.0}&           18.0 &&         6.1 &           17,8 \\
      &                     512&        7.8 &	\textbf{17.6}&&	        8.0 &   \textbf{17.4}\\
 \midrule
  &                         64 &        20.0 &	        38.0 &&	        25.0 &	        51.1 \\
 Osborne&                   128&\textbf{15.4}&	        33.7 &&	        21.5 &	        46.7 \\
  &                         256&        16.3 &  \textbf{33.5}&& \textbf{20.9}&  \textbf{44.0}\\
  &                         512&        51.2 &	        56.1 &&	        54.1 &	        63.3 \\
 \bottomrule
 \end{tabular}
\caption{Encoder size experiments results (baseline in encoder size of 256). Best result by database in bold.}
\label{table:encoder_dimension_results}
\end{table}

It is observed that in the cases of IAM and RIMES databases, the experiments that set a large encoder size obtain better results. For Osborne database, the model with encoder size 512 does not converge properly and obtains a very high error. It is probably because IAM and RIMES databases are much more complex in size than Osborne database, and therefore the larger encoder sizes work better for these databases and worse for Osborne database.

\begin{figure}[!ht]
\centering
  \includegraphics[width=0.9\textwidth]{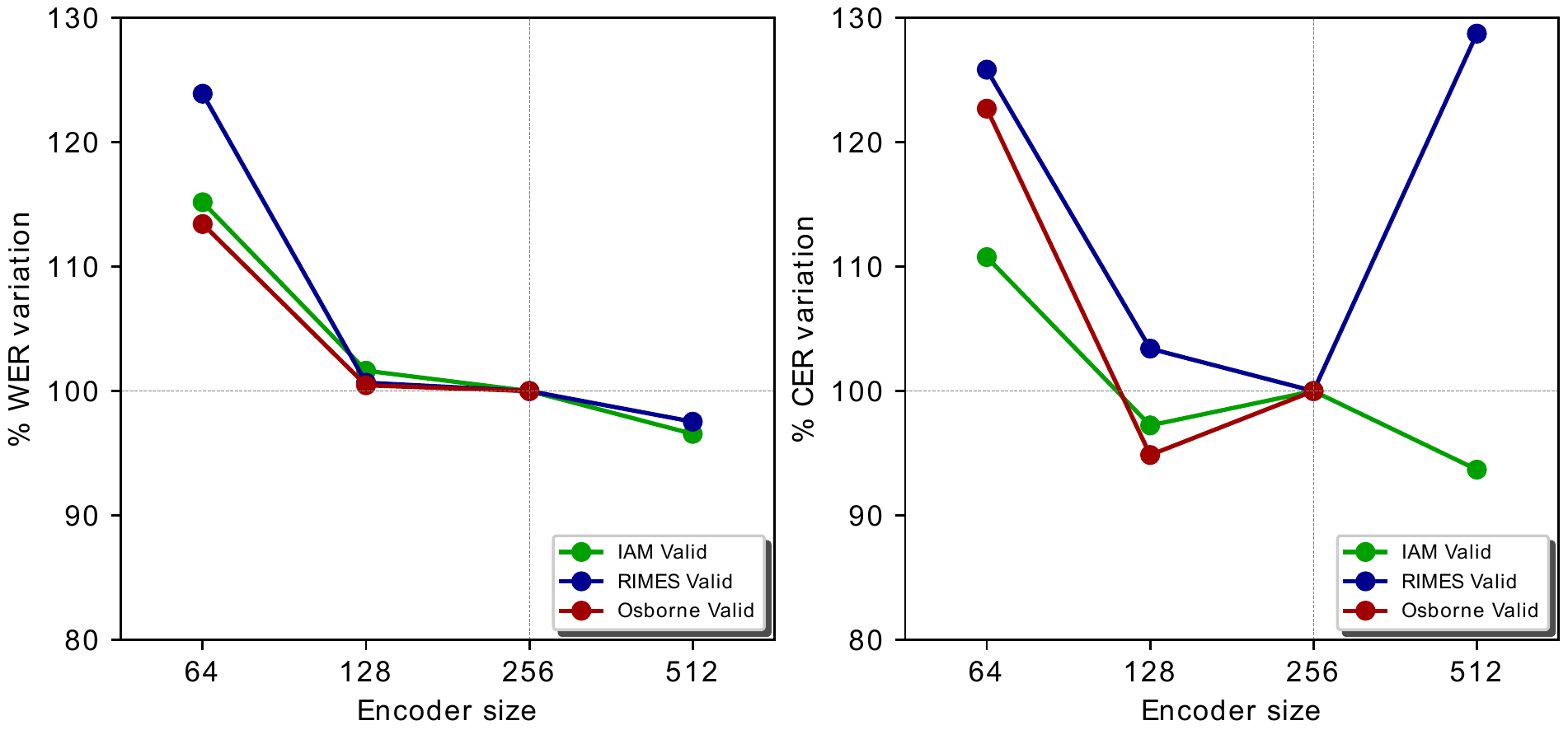}
\caption{Variation percentage of WER and CER in validation and test partitions with different encoder RNN sizes over the baseline.}
\label{fig:WERVal_vs_encoder_size} 
\end{figure}

In Fig. \ref{fig:WERVal_vs_encoder_size} we measure the percentage increase or decrease concerning the baseline of the validation CER and WER values. It allows to better observe the effect of the dimension change without being affected by differences in the complexity of the different databases.

From Fig. \ref{fig:WERVal_vs_encoder_size} it is clear that the 64 size is not suitable for any of the databases. In terms of WER in the validation data, the best results are obtained with size of 512 for IAM and RIMES, and size of the 256 for Osborne. The results obtained with a size of 128 are relatively close to these optimal with a much smaller number of parameters. As a conclusion, it is proposed that a number of parameters between 128 and 512 are suitable for the large databases (IAM and RIMES), and between 128 and 256 for the smaller Osborne database.

\subsubsection{Number of layers and bidirectionally}
  \index{Bidirectional recurrent neural network}

In the second set of experiments, we used a fixed baseline model with an RNN encoder of type bidirectional LSTM with dimension 256, and defined different experiments modifying the number of layers of the RNN and the fact that these layers are bidirectional or not. The results of the experiments for each of the three databases are included in Table \ref{table:encoder_layers_results}. This Table shows that the best validation results are obtained with three bidirectional layers for IAM and RIMES databases and two bidirectional layers for Osborne database. As before, it is hypothesized that the models for the larger and more complex databases improve by increasing the decoder complexity, in this case by increasing the number of layers. It is also observed that models with bidirectional layers generally perform better than models with unidirectional layers. This is true regardless of the database and the number of layers.

\begin{table}[!ht]
\centering
 \begin{tabular}{lllrrrrr}
 \toprule
 Dataset &   & & \multicolumn{2}{l}{Validation} & & \multicolumn{2}{l}{Test} \\
 \cmidrule(r){4-5} \cmidrule(r){7-8} & Bidirectional & num layers& CER & WER & & CER & WER \\
 \midrule
  &         No &            1&          9.6&	        22.3 &&	        14.1 &	        30.0 \\
  &         No &            2&          9.1&	        20.7 &&	        13.2 &	        28.3 \\
  IAM&      No &            3&          9.3&            20.7 &&         13.2 &          28.0 \\
  &         Yes&            1&          8.8&	        21.1 &&	        13.1 &	        28.9 \\
  &         Yes&            2&         8.5&            19.3 &&         12.6 &          27.4 \\
  &         Yes&            3&  \textbf{8.0}&   \textbf{18.8}&& \textbf{11.6}&  \textbf{25.8}\\
 \midrule
  &         No &            1&          7.7&	       21.1 &&	        7.5 &	       21.2 \\
  &         No &            2&          6.7&	       18.4 &&	        6.7 &	       18.3 \\
  RIMES&    No &            3&          6.4&           17.3 &&          6.2 &          17.3 \\
  &         Yes&            1&          7.1&	       20.0 &&	        6.7 &	       19.9 \\
  &         Yes&            2&         6.0&           18.0 &&          6.1 &          17,8 \\
  &         Yes&            3&  \textbf{5.1}&  \textbf{15.4}&&  \textbf{5.1}&  \textbf{15.5}\\
 \midrule
  &         No &            1&          23.8&	         43.8 &&	    27.9 &	        51.1 \\
  &         No &            2&          25.0&	         43.0 &&	    32.8 &	        50.0 \\
  Osborne & No &            3&          27.5&            41.9 &&        31.3 &          52.5 \\
  &         Yes&            1&          21.2&	         41.1 &&	    28.9 &	        54.5 \\
  &         Yes&            2&  \textbf{16.3}&   \textbf{33.5}&&        20.9 &          44.0 \\
  &         Yes&            3&          16.6&            34.3 &&\textbf{18.9}&  \textbf{40.7}\\
 \bottomrule
 \end{tabular}
\caption{Number of layers and bidirectionally experiment results (baseline in 2 bidirectional layers). Best result for each database in bold.}
\label{table:encoder_layers_results}
\end{table}

\begin{figure}[!ht]
\centering
  \includegraphics[width=0.9\textwidth]{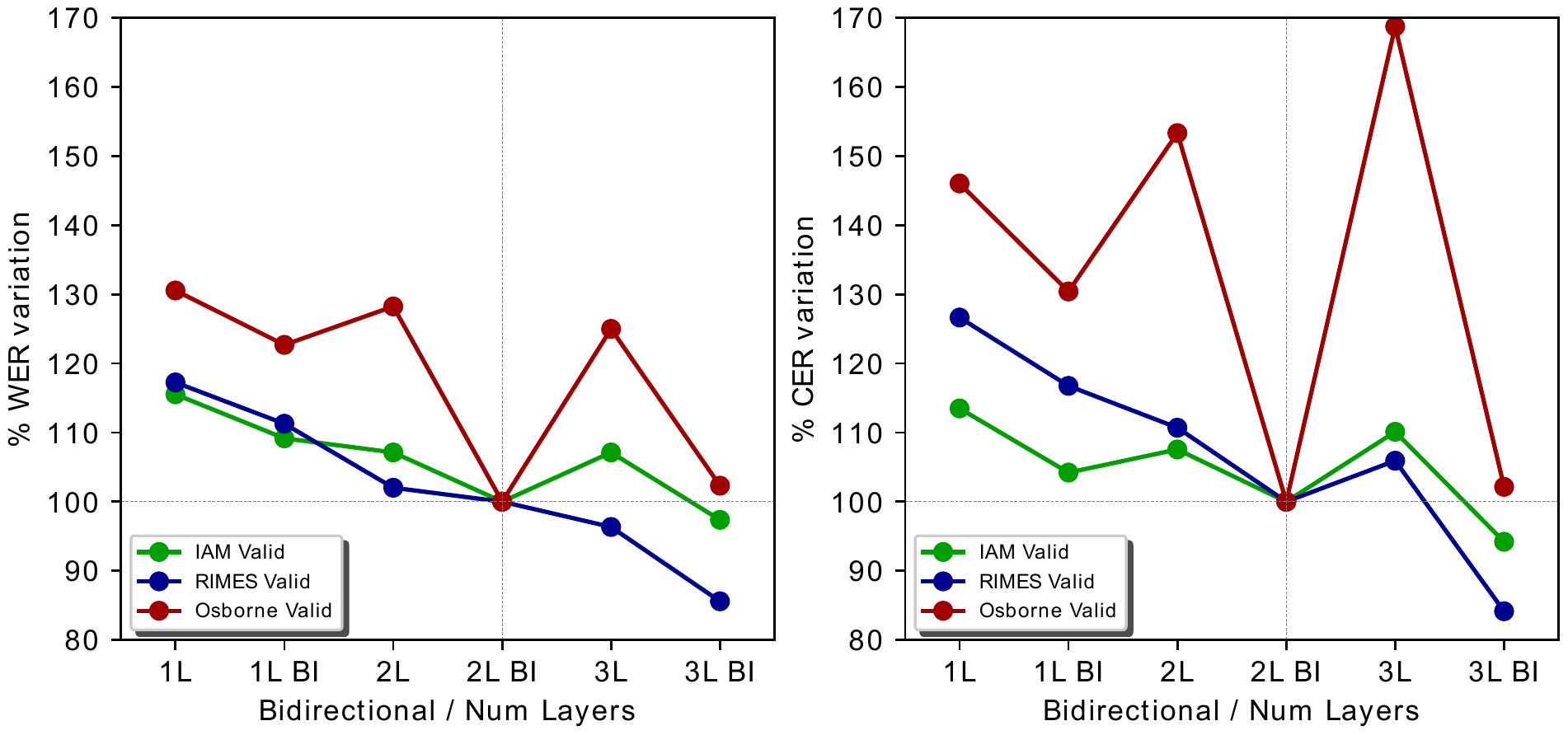}
\caption{Variation percentage of WER and CER in validation and test partitions with different encoder layers over the baseline.}
\label{fig:WERVal_vs_encoder_layers} 
\end{figure}

Fig. \ref{fig:WERVal_vs_encoder_layers} presents the percentage increases or decreases concerning the baseline of the validation CER and WER values. It allows better observing the effect of changing the number and type of encoder layers without being affected by differences in the complexity of the different databases.

It is observed that, in general, WER tends to decrease as the number of layers increases, except that for Osborne database. The results with three layers are not good, and the optimum is achieved with two bidirectional layers. The conclusion is similar to the one obtained when analyzing each layer size: that the models of the more complex databases benefit from an increase in the number of encoder parameters. It can be seen that increasing the encoder complexity by adding layers is more beneficial than increasing the encoder complexity by increasing the size of each layer when these data are compared with those obtained previously by experimenting with the size of the encoder layers, shown in Fig. \ref{fig:WERVal_vs_encoder_size}. In RIMES database, by increasing the layer size from 256 to 512 means a WER improvement of less than 5\%. However, for the same database, going from two bidirectional layers to three layers means an improvement of more than 15\%.

\subsubsection{Type of layers: LSTM vs GRU}
\index{Long short-term memory network}
\index{Gated recurrent unit}

The third set of experiments analyzes the impact on the error when changing the encoder RNN type, experimenting with LSTM and GRU types. In order not to draw conclusions about the type of architecture based on a single experiment per database, each type has been analyzed in the bidirectional and non-bidirectional cases. Table \ref{table:encoder_type_results} compares the results of this experiment for each of the three databases. Finally, it is recalled that, as indicated in Subsection \ref{section:C6-model-Decoder}, for the model architecture proposed for this Thesis, the decoder has been chosen to use the same type of RNN as the encoder.

\begin{table}[!ht]
\centering
 \begin{tabular}{lllrrrrr}
 \toprule
 Dataset &   & & \multicolumn{2}{l}{Validation} & & \multicolumn{2}{l}{Test} \\
 \cmidrule(r){4-5} \cmidrule(r){7-8} & Bidirectional & Type& CER & WER & & CER & WER \\
 \midrule
  &         No&             LSTM&           9.1 &	        20.7 &&	        13.2 &	        28.3 \\
  IAM &     No&             GRU &           9.0 &           20.6 &&         12.9 &          27.8 \\
  &         Yes&            LSTM&           8.5 &           19.3 &&         12.6 &          27.4 \\
  &         Yes&            GRU &   \textbf{8.0}&   \textbf{18.9}&& \textbf{11.6}&  \textbf{26.0}\\
 \midrule
  &         No&             LSTM&           6.7 &	        18.4 &&	        6.7 &	        18.3 \\
  RIMES &   No&             GRU&    \textbf{5.1}&   \textbf{15.5}&& \textbf{5.1}&   \textbf{15.6}\\
  &         Yes&            LSTM&           6.0 &           18.0 &&         6.1 &           17,8 \\
  &         Yes&            GRU &           5.3 &           16.5 &&         5.3 &           16.3 \\
 \midrule
  &         No&             LSTM&           25.0 &	        43.0 &&	        32.8 &	        50.0 \\
  Osborne & No&             GRU &           27.1 &          43.1 &&         32.2 &          51.9 \\
  &         Yes&            LSTM&   \textbf{16.3}&  \textbf{33.5}&& \textbf{20.9}&  \textbf{44.0}\\
  &         Yes&            GRU &           41.2 &          52.5 &&         44.9 &          59.3 \\
 \bottomrule
 \end{tabular}
\caption{RNN encoder type experiments results (baseline in LSTM bidirectional). Best result by database in bold.}
\label{table:encoder_type_results}
\end{table}

Table \ref{table:encoder_type_results} shows that the best results for IAM and RIMES databases are obtained with the GRU type. In Osborne database, the results with GRU type networks are inferior to those obtained with bidirectional LTSM type networks. In the case of RIMES database, the improvement of results by using the GRU type appears with both unidirectional and bidirectional layers.

\begin{figure}[!ht]
\centering
  \includegraphics[width=0.9\textwidth]{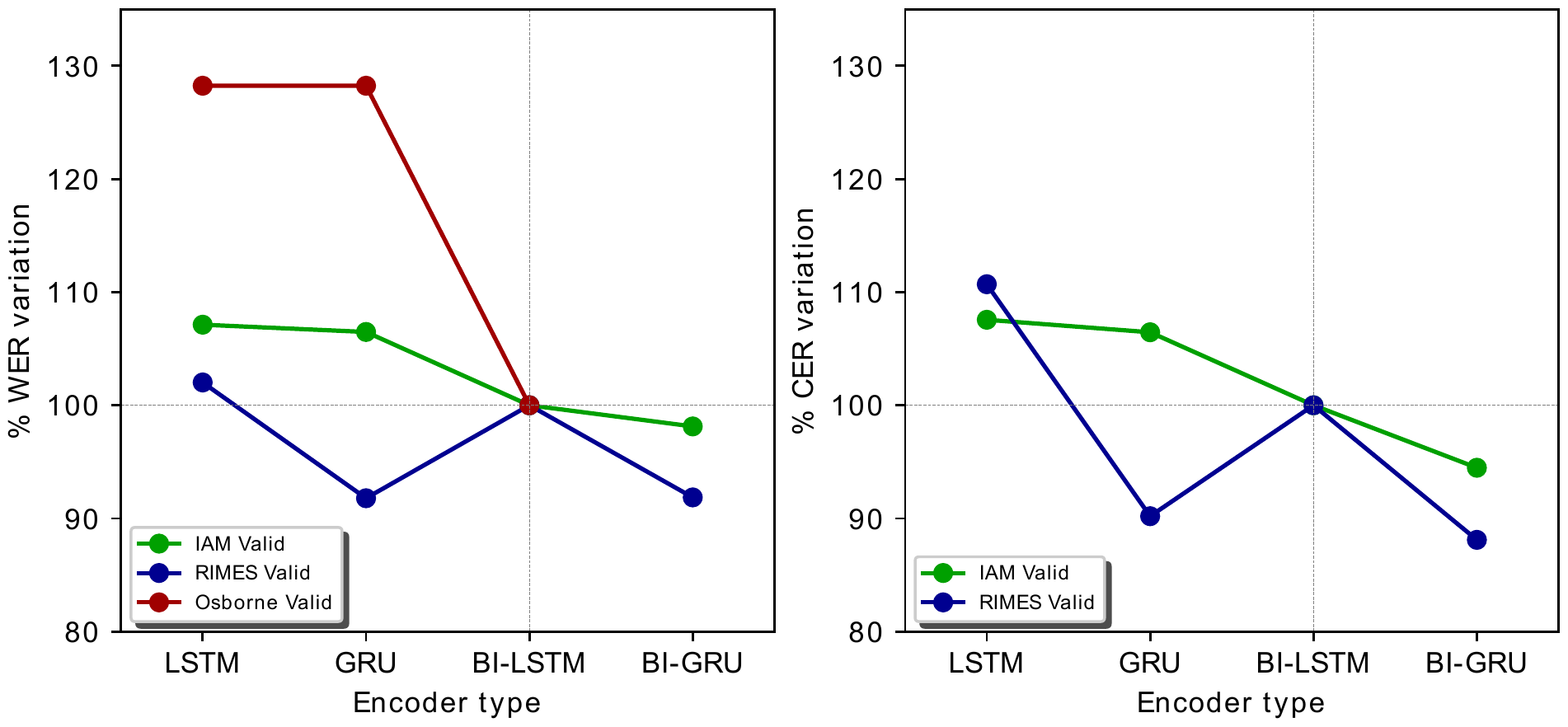}
\caption{Variation percentage of WER and CER in validation and test partitions with different encoder types over the baseline.}
\label{fig:WERVal_vs_encoder_type} 
\end{figure}

Fig. \ref{fig:WERVal_vs_encoder_type} measures the percentage increases or decreases for the baseline of the validation CER and WER values when changing the RNN type of the encoder and decoder. It allows to better observe the effect of the dimension change without being affected by the differences in the complexity of the different databases. This figure also shows how RIMES database is the most affected by the RNN type change. We have omitted the data from Osborne database, which had a huge percentage of change not to distort the graph.

In summary, IAM and RIMES databases benefit from an increase in encoder complexity greater than that established in the baseline. In addition, among the aspects analyzed, the most influential one in obtaining better WER and CER results in validation are increasing the number of layers from two to three and using GRU networks instead of LSTM type. In the case of Osborne database, the results indicate that the architecture proposed in the baseline would be the most appropriate one.

\subsection{Image resolution}
\label{C6 - Experiments - Image resolution}

Experiments were also conducted to determine if the input image resolution affected the model performance. For this purpose, three different resolutions (height$\times$width) were selected for the input images of 32$\times$128, 48$\times$196 (baseline), and 64$\times$256 pixels. The changes in image sizes have been made proportionally to keep the aspect ratio of images. The results of these experiments for each of the three databases are included in Table \ref{table:image_resolution_experiments}.

\begin{table}[!ht]
\centering
 \begin{tabular}{llrrrrr}
 \toprule
 Dataset &   & \multicolumn{2}{l}{Validation} & & \multicolumn{2}{l}{Test} \\
 \cmidrule(r){3-4} \cmidrule(r){6-7} & Image Resolution& CER & WER & & CER & WER \\
 \midrule
    &                    32x128&            9.1 &           20.9 &&         13.5 &          28.8 \\
 IAM&                    48x192&            8.5&    \textbf{19.3}&&         12.6 &          27.4 \\
    &                    64x256&    \textbf{8.4}&	        20.0 &&	\textbf{12.5}&	\textbf{27.3}\\
 \midrule
    &                    32x128&            6.5 &           18.3 &&         6.4 &           18.1 \\
 RIMES&                  48x192&  \textbf{6.0}&   \textbf{18.0}&& \textbf{6.1}&   \textbf{17,8}\\
    &                    64x256&            6.3 &           18.2 &&         6.2 &           18.1 \\
 \midrule
    &                    32x128&          19.9  &	         38.5 &&	      25.4  &	        50.0 \\
 Osborne&                48x192& \textbf{16.3}&  \textbf{33.5}&&  \textbf{20.9} &   \textbf{44.0}\\
    &                    64x256&          18.3  &            37.1 &&          22.4 &            45.4 \\
 \bottomrule
 \end{tabular}
\caption{Input image resolution experiments results (baseline in 48x192). Best result by database in bold.}
\label{table:image_resolution_experiments}
\end{table}

It is observed that in all cases, the lowest WER in validation is provided by the 48$\times$192 resolution. The database least affected by resolution changes is RIMES, and the most affected is Osborne. In general, by reducing the resolution to 32$\times$128 increases the error more than increasing it with 64$\times$256 resolution.

\begin{figure}[!ht]
\centering
  \includegraphics[width=0.9\textwidth]{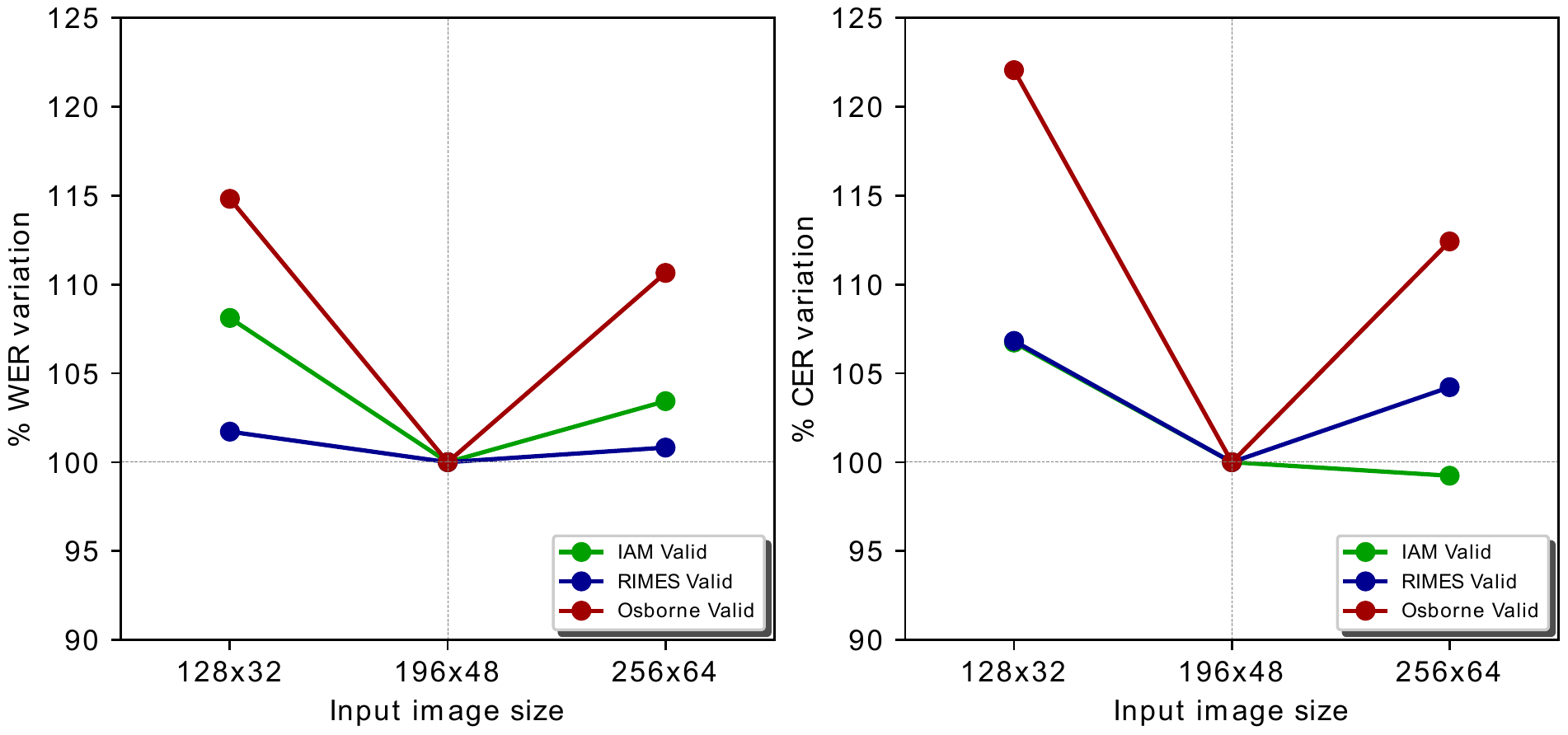}
\caption{Variation percentage of WER and CER in validation and test partitions with different input image resolutions over the baseline.}
\label{fig:WERVal_vs_imageSize} 
\end{figure}

The graph in Fig. \ref{fig:WERVal_vs_imageSize} shows the relative differences with respect to the baseline which defined an input resolution of 48$\times$192.

As shown in Fig. \ref{fig:WERVal_vs_imageSize}, in IAM and RIMES databases the relative differences are less than 10\%, especially when increasing the resolution with respect to the baseline. Therefore, it is concluded that working with a resolution greater than or equal to 48$\times$192 is a factor that has a relatively small impact on the final results.

\subsection{Normalization and data augmentation}
\label{C6 - Experiments - Normalization and data augmentation}
\index{Data augmentation}

The following experiments analyze the impact of normalization and train data augmentation strategy on WER and CER metrics. Experiments are performed with original and normalized data. The normalization strategy is the one selected in the baseline. The normalization process performed with the images of the three databases is described in the previous Subsection \ref{C6 - section - Image Preprocesing} corresponding to Image Preprocessing. For each of the above cases, two experiments are performed, the first one with the data augmentation strategy activated and the second one without train data augmentation (baseline). The train data augmentation strategy used is described in the previous Subsection \ref{C6 - section - Data augmentation}. The results of these four experiments for each of the three databases are included in Table \ref{table:normalization_experiments}. These results clearly show that both data normalization and data augmentation strategies provide a significant reduction of WER and CER in all three databases. In particular, it is the combination of the two activated strategies that provides a lower WER and CER in validation and test. It is also observed that the data augmentation strategy provides a greater improvement than the normalization strategy. For example, activating data augmentation on non-normalized data reduces the CER by 1.7 points and the WER by 4.2 points for IAM database. On the other hand, by applying normalization without data augmentation improves the CER by 1.5 and the WER by 3.8.

\begin{table}[!ht]
\centering
 \begin{tabular}{lllrrrrr}
 \toprule
 Dataset &   & & \multicolumn{2}{l}{Validation} & & \multicolumn{2}{l}{Test} \\
 \cmidrule(r){4-5} \cmidrule(r){7-8} & Normalization & Data augmentation & CER & WER & & CER & WER \\
 \midrule
  &             Yes&            Yes&    \textbf{7.3}&   \textbf{17.0}&& \textbf{10.1}&  \textbf{22.6} \\
  IAM&          Yes&            No &            8.5 &           19.3 &&         12.6 &          27.4 \\
  &             No&             Yes&            8.3 &           18.9 &&         12.1 &          26.0 \\
  &             No&             No &            10.0 &          23.1 &&         15.2 &          31.3 \\
 \midrule
  &             Yes&            Yes&    \textbf{4.7}&   \textbf{13.7}&& \textbf{4.6}&   \textbf{13.5}\\
  RIMES&        Yes&            No &            6.0 &           18.0 &&         6.1 &           17,8 \\
  &             No&             Yes&            5.4 &           14.9 &&         5.1 &           14.5 \\
  &             No&             No &            6.7 &           18.3 &&         6.5 &           17.8 \\
 \midrule
  &             Yes&            Yes&\textbf{13.6}&  \textbf{29.3}&& \textbf{15.1}&  \textbf{35.9}\\
  Osborne&      Yes&            No &        16.3 &          33.5 &&         20.9 &          44.0 \\
  &             No&             Yes&        21.1 &          39.2 &&         20.7 &          44.7 \\
  &             No&             No &        29.3 &          48.2 &&         28.4 &          51.2 \\
 \bottomrule
 \end{tabular}
\caption{Normalization and data augmentation experiments results (baseline with normalization and no data augmentation). Best result by database in bold.}
\label{table:normalization_experiments}
\end{table}

\begin{figure}[!ht]
\centering
  \includegraphics[width=0.9\textwidth]{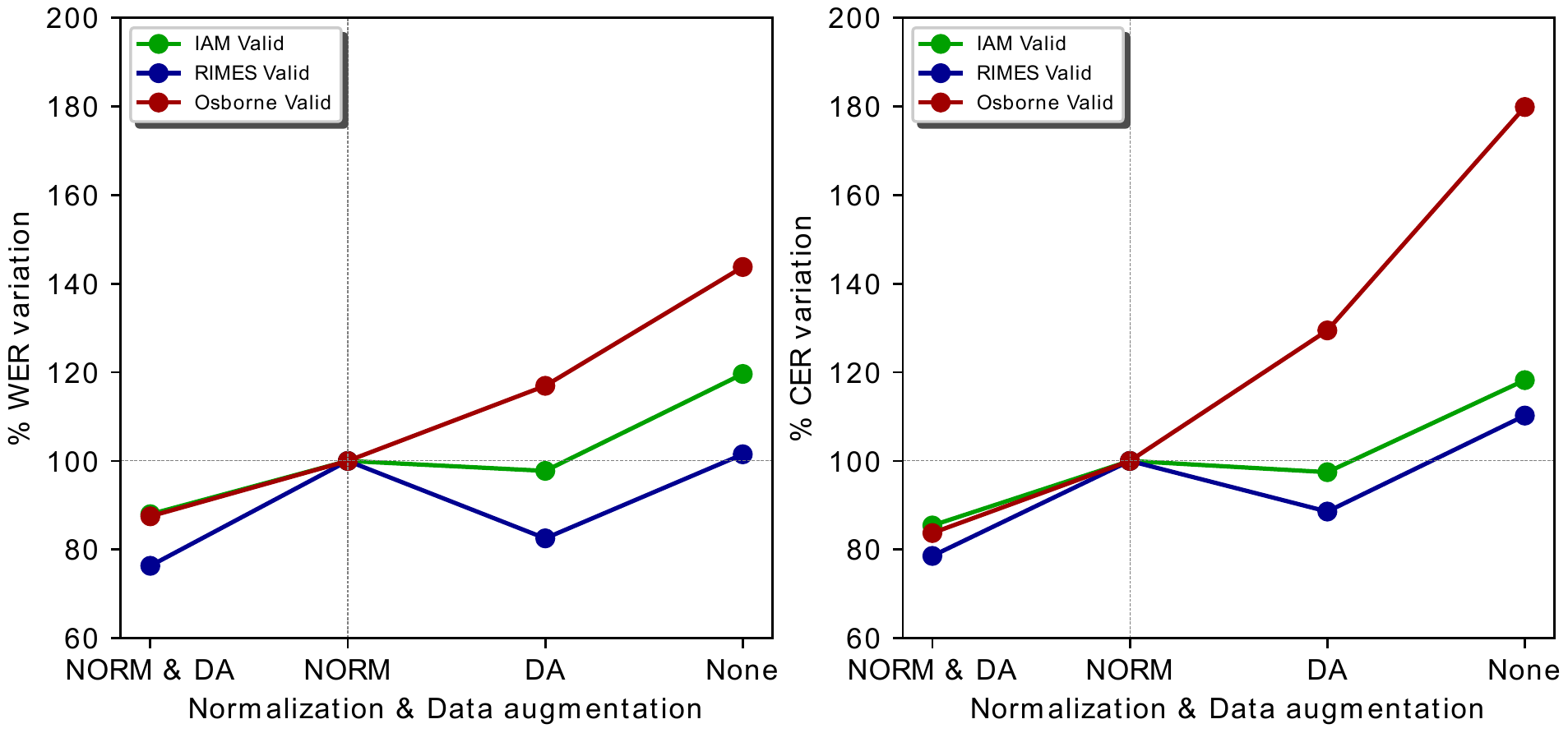}
\caption{Variation percentage of WER and CER in validation and test partitions with normalization and data augmentation over the baseline.}
\label{fig:normalization_wer_cer} 
\end{figure}

The graph in Fig. \ref{fig:normalization_wer_cer} shows the relative differences with respect to the baseline configuration that used normalization and did not use data augmentation.

When analyzing the relative differences in Fig.\ref{fig:normalization_wer_cer}, it is observed that these differences are important, especially for Osborne database. With a more limited size, this database takes more advantage of applying data augmentation in training. In the case of Osborne database, the effect of normalization also is greater. This is because this historical database has a very pronounced slant which, if not corrected by normalization, is likely to make text recognition very difficult.

\subsection{Teacher forcing}
\label{C6 - Experiments - Teacher forcing}

Finally, experiments have been conducted to determine the importance of using the \textit{teacher forcing} strategy during model training. As indicated in Subsection \ref{section:seq2seq Decoder}, teacher forcing \cite{williams1989learning} consists of train the seq2seq model by passing as input to the decoder the original values of the output sequence, but shifted to $t$-1. This training strategy can lead to problems in the evaluation phase because an error of one of the generation steps of the output sequence is propagated through the conditional probabilities. It causes a decrease in performance because the predicted sequence diverges from the real one from the point of error. Teacher forcing also has as negative effect the train overfit because, at each step $t$ of the decoding process, the previous real value of the character sequence $y_{t-1}$ is used. Table \ref{table:teacher_forcing_experiments} shows the results of these experiments for each of the three databases.

\begin{table}[!ht]
\centering
 \begin{tabular}{llrrrrr}
 \toprule
 Dataset &   & \multicolumn{2}{l}{Validation} & & \multicolumn{2}{l}{Test} \\
 \cmidrule(r){3-4} \cmidrule(r){6-7} & Teacher Forcing& CER & WER & & CER & WER \\
 \midrule
 IAM&                           Yes&        8.5 &   \textbf{19.3}&&         12.6 &         27.4 \\
      &                         No& \textbf{8.2}&           19.5 && \textbf{12.3} &\textbf{27.2}\\
 \midrule
 RIMES&                         Yes&\textbf{6.0} &  \textbf{18.0}&&         6.1 &          17,8 \\
      &                         No& \textbf{6.0} &          18.3 && \textbf{5.7}&  \textbf{17.5}\\
 \midrule
 Osborne&                       Yes&\textbf{16.3}&  \textbf{33.5}&& \textbf{20.9}&  \textbf{44.0}\\
        &                       No &        21.8&           41.8 &&         25.1&           50.5 \\
 \bottomrule
 \end{tabular}
\caption{Input image resolution experiments results (baseline with teacher forcing). Best result by database in bold.}
\label{table:teacher_forcing_experiments}
\end{table}

\begin{figure}[!ht]
\centering
  \includegraphics[width=0.9\textwidth]{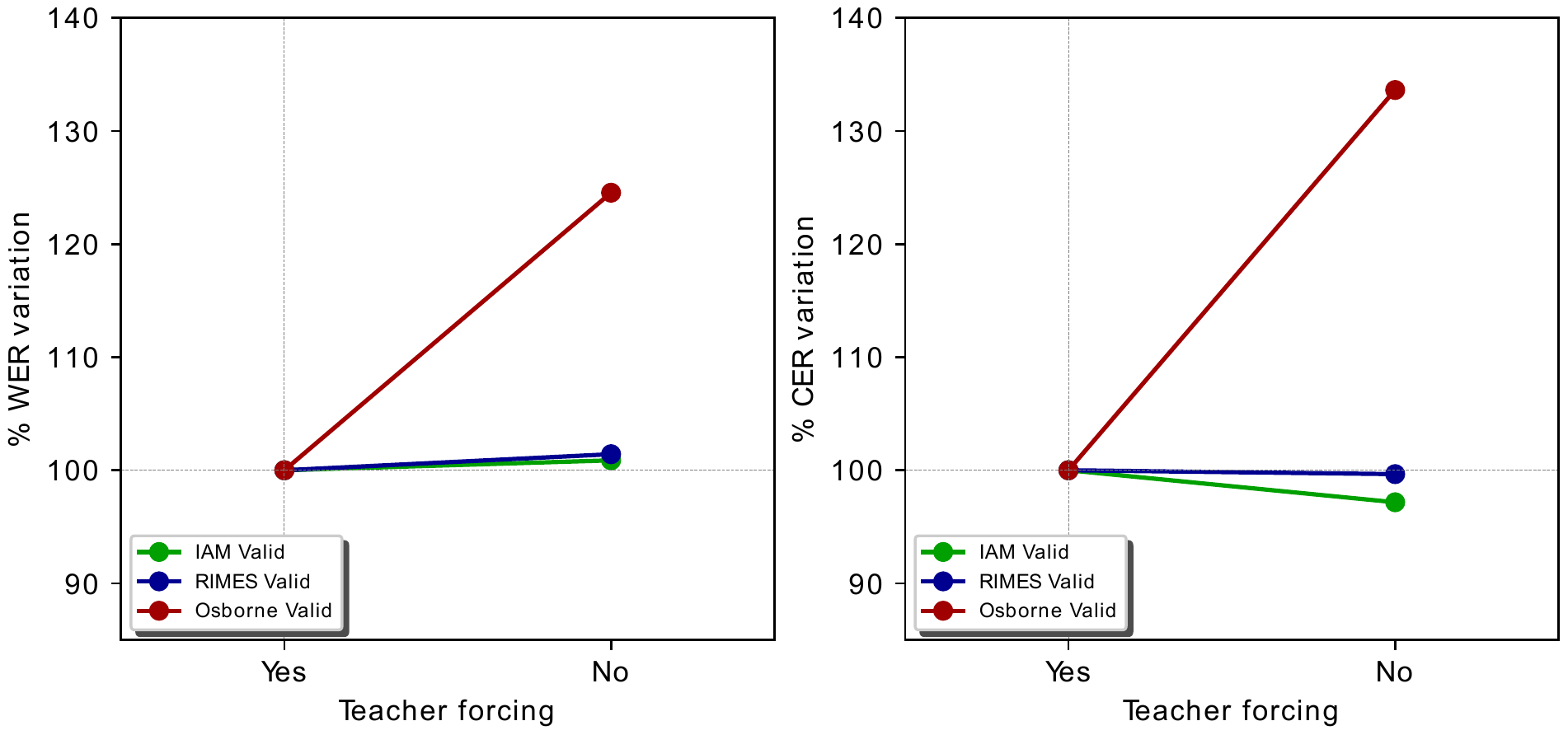}
\caption{Variation percentage of WER and CER in validation and test partitions with different training strategy over the baseline.}
\label{fig:WERVal_vs_teacher} 
\end{figure}

It can be seen that the best WER validation results are obtained with the teacher forcing strategy activated, although the results are quite similar, especially in the cases of IAM and RIMES databases. Fig. \ref{fig:WERVal_vs_teacher} shows the relative differences per database.

In IAM and RIMES databases, the relative differences in WER and CER are less than 2\%, so using or not this training strategy is not relevant in terms of accuracy. It has been observed that not using teacher forcing increases by approximately 50\% the number of epochs that the algorithm needs to reach the defined stopping criterion. It indicates that this strategy significantly accelerates the convergence speed of the model training.

\subsection{Final model}
\label{C6 - Experiments - Final model}

It has been observed in the experiments performed that the variations in model performance with the different configurations analyzed are relatively consistent between databases, especially between the IAM and RIMES databases. These experiments also allow identifying modifications on the baseline configuration that provide better results in terms of accuracy. From here, guidelines can be obtained to propose an optimized parameterization of the proposed model. The selected configuration does not necessarily include the settings that have provided the lowest validation error. It was decided to choose the configuration by defining a model with the smallest number of parameters, following the principle of parsimony \cite{vandekerckhove2014model}, since we have two configurations that provide similar validation results. These configurations are the following ones:

\begin{itemize}
  \item As a convolutional architecture, LeNet applied on the patch sequence is selected (as indicated in Subsection \ref{section: Image patches architectures}. The patch configuration is set with an initial patch size of 10 pixels and a step size of 2 pixels. In this case, the selection matches the one made in the baseline.
  \item The encoder is configured with three bidirectional layers of type GRU of size 256 for IAM and RIMES databases. For Osborne database, the encoder is configured with two bidirectional layers of type LSTM of size 256. The RNN of the decoder is set to the same type and size as used in the encoder.
  \item The size of the input image is set to 48$\times$192.
  \item The trainings are performed on the normalized data, with train data augmentation and teacher forcing strategy.
\end{itemize}

With this optimized configuration, experiments have been performed with IAM, RIMES, and Osborne databases, and the following tables detail the results obtained for each of the lexicon decodings used.

Table \ref{table:Optimal results. No lexicon.} includes the CER and WER values for the validation and test partitions, corresponding to the results obtained directly by the model, without using any lexicon to decode them.

\begin{table}[!ht]
\centering
 \begin{tabular}{lrrrrr}
 \toprule
 Dataset &  \multicolumn{2}{l}{Validation} & & \multicolumn{2}{l}{Test} \\
 \cmidrule(r){2-3} \cmidrule(r){5-6} & CER & WER & & CER & WER \\
 \midrule
 IAM     &  6.7 & 15.1 && 9.3 & 21.0 \\
 RIMES   &  4.6 & 13.1 &&  4.5 & 13.2 \\
 Osborne & 13.6 & 29.3 && 15.1 & 35.9 \\
 \bottomrule
 \end{tabular}
\caption{Optimal results without lexicon.}
\label{table:Optimal results. No lexicon.}
\end{table}

In the case of using the standard lexicons to decode the results of the visual model, we obtain the CER and WER values detailed in Table \ref{table:Optimal results. Standard lexicon.}.

\begin{table}[!ht]
\centering
 \begin{tabular}{lrrrrr}
 \toprule
 Dataset &  \multicolumn{2}{l}{Validation} & & \multicolumn{2}{l}{Test} \\
 \cmidrule(r){2-3} \cmidrule(r){5-6} & CER & WER & & CER & WER \\
 \midrule
 IAM     &  8.8 & 14.9 && 11.2 & 19.6 \\
 RIMES   &  6.2 & 12.3 && 5.9 & 12.2 \\
 Osborne &  8.7 & 14.7 && 24.0 & 34.5 \\
 \bottomrule
 \end{tabular}
\caption{Optimal results using a standard lexicon.}
\label{table:Optimal results. Standard lexicon.}
\end{table}

Finally, by using the test lexicon to decode the model predictions, the results are obtained from Table \ref{table:Optimal results. Test lexicon.}. It only makes sense to apply this lexicon to test data. It is possible to evaluate the error weight of the words of the test set not present in the standard lexicon (i.e., the OOV words) by comparing the results of this table with the previous one. 

\begin{table}[!ht]
\centering
 \begin{tabular}{lrrrrr}
 \toprule
 Dataset &  \multicolumn{2}{l}{Validation} & & \multicolumn{2}{l}{Test} \\
 \cmidrule(r){2-3} \cmidrule(r){5-6} & CER & WER & & CER & WER \\
 \midrule
 IAM     &  NA &  NA & & 7.6 & 13.0 \\
 RIMES   &  NA &  NA & & 2.7 & 5.4 \\
 Osborne &  NA &  NA & & 9.4 & 14.4 \\
 \bottomrule
 \end{tabular}
\caption{Optimal results using test lexicon.}
\label{table:Optimal results. Test lexicon.}
\end{table}

\subsection{Error analysis}
\label{C6 - Experiments - Error analysis}

This subsection analyzes the errors of the models presented to identify possible lines of work for improvement. The first analysis was performed in the previous subsection by studying the influence of the OOV words in the error. These errors can be reduced by improving the lexicons and, especially, by adding to the system some language models at character level that can identify OOV words. 

\begin{figure}[!ht]
\centering
  \includegraphics[width=0.7\textwidth]{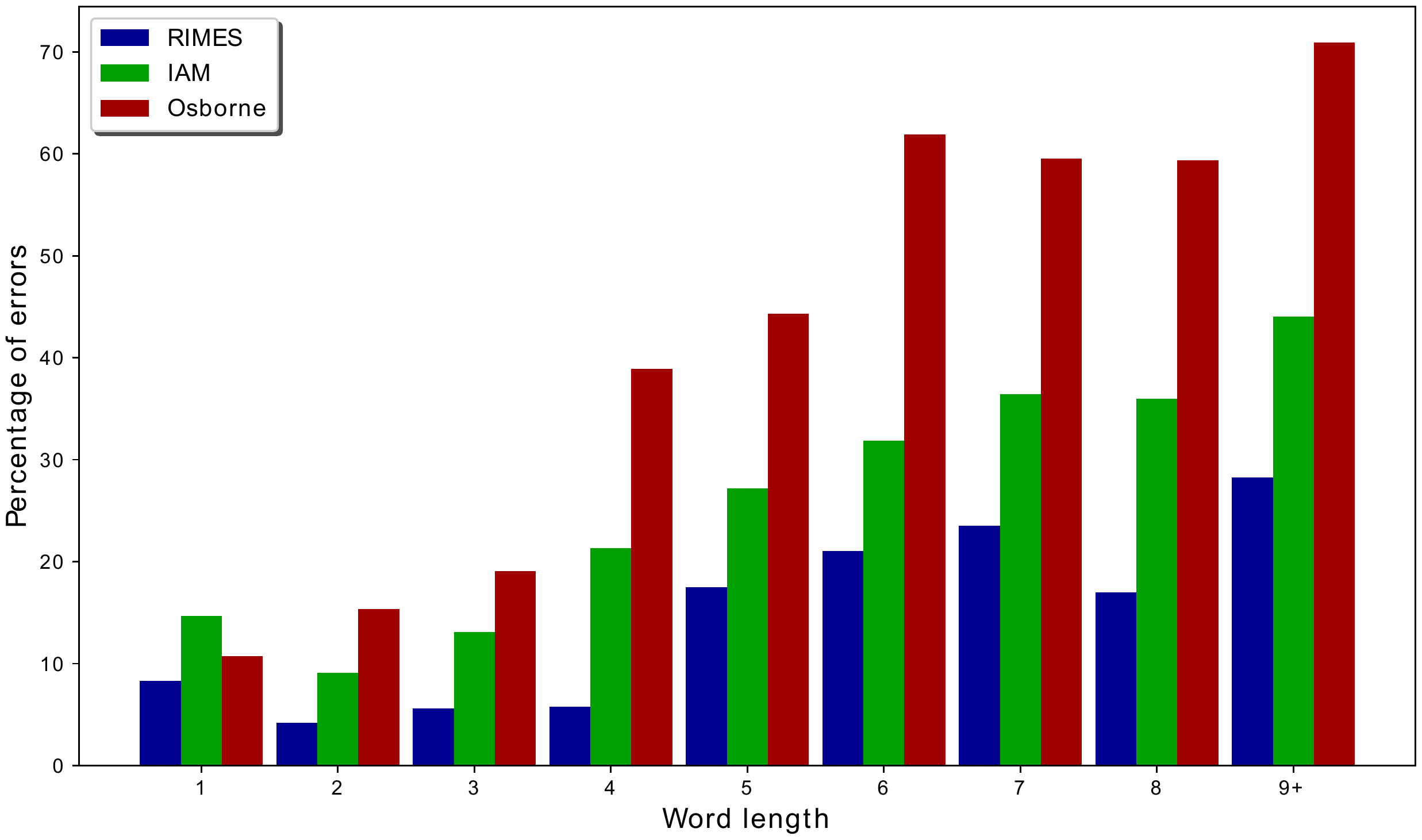}
\caption{Error analysis: WER in test vs word length.}
\label{fig:4_6_1} 
\end{figure}

First, we analyze how the errors change in terms of word length. Fig. \ref{fig:4_6_1} shows the WER with the standard lexicons in IAM and RIMES databases regarding word length. We observe that the error increases with the word length. It is related to the previous analysis on the OOV influence in the errors because the words with fewer letters are articles, prepositions, and similar words that typically appear in the lexicons. Regarding errors in identifying words of length one (isolated characters), in the RIMES and Osborne databases, the number of cases is residual (5 errors in RIMES and 6 in Osborne). There is a large volume of isolated characters in the IAM database due to the separation into words of punctuation marks, including apostrophes, and 641 errors appear, which represents 11.8\% of the total errors. The most common errors consist of confusing the characters '.' and ',' and the characters ''' and '"'.  

\begin{figure}[!ht]
\centering
  \includegraphics[width=0.7\textwidth]{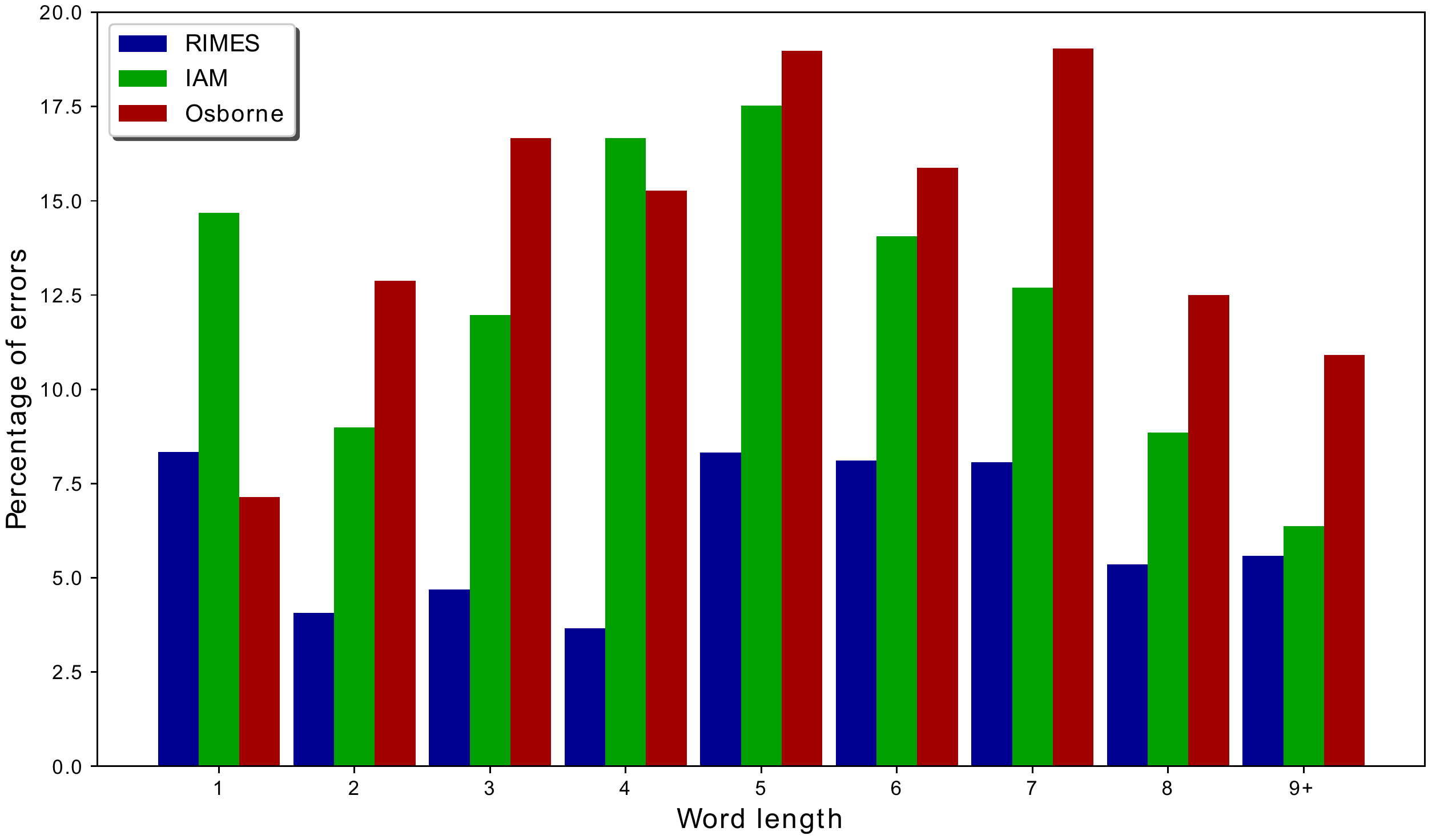}
\caption{Error analysis: WER in test vs word length with the test lexicon.}
\label{fig:4_6_2} 
\end{figure}

To eliminate the influence of the OOV words in the analysis of the error versus the word length, we show in Fig. \ref{fig:4_6_2} the WER test error using the test lexicon. We observed that most of the errors in long words originate from the absence of the correct word in the lexicon. Now, the highest percent of errors appears for word length of 4 in the IAM database and word length of 5 in the RIMES database.

Next, we analyze the amount of errors originated by replacing only one character in a word. The real and predicted words have the same characters and the same length, but one character has been replaced by another in the predicted word. In this case, we observed that the $57.4\%$ of errors of the visual model in the IAM database, the $59.1\%$ in the RIMES database, and the $56.9\%$ in the Osborne database are errors corresponding to only one character replaced. Probably, these errors could be solved in most cases using a character-level language model.

Finally, we analyze the errors originated by the confusion between the same character, one in lowercase and the other in uppercase. In this situation, we observe that a $3.7\%$ in the IAM database, a $5.6\%$ in the RIMES database and, a $0.9\%$ in the Osborne database are errors originated by the confusion of some uppercase character and the same one in lowercase, or vice versa. As we expected, the characters where it this confusion appears are those with a similar appearance in uppercase and lowercase. For example: 'f', 'c', 'v', 'l', 'k' or 'w'. These errors happen almost always at the beginning of words because they correspond to the beginning of a paragraph. They can be solved by finding a way to detect the end of a paragraph; for example, by identifying that the previous character is a '.' (dot).

\section{On the difficulties of comparing results}
\label{section:On the difficulties of comparing results}

While there is a reasonably widespread consensus about datasets and reference metrics for comparing different approaches to the general handwriting recognition problem, several factors influence the final results and which the authors address in different ways. So it is very difficult to compare the different approaches with each other.

These factors are fundamentally related to the choice of data for training, validation, and testing, the set of valid characters considered, the use of training and assessment improvement strategies, and the use of different lexicons and language models in decoding.

Some of these factors directly prevent the comparison of results. It is not possible to compare the metrics of two outcomes that have been calculated on different evaluation datasets. It is also impossible to compare results that have eliminated certain characters, such as punctuation marks or capital letters in the decoding process.

Other factors, such as the use of data augmentation in training or testing or the use of different language models in decoding, do not prevent the direct comparison of the final results of different publications. However, in these cases, it is essential to provide an ablation study that allows us to detail how much the different improvement strategies employed contribute to the accuracy level. It allows a more exact comparison with other authors who have not used these improvement strategies. 

Next, these factors are reviewed in detail, indicating how they influence the final result and proposing recommendations that allow a comparison between different publications results that provide the maximum information about the efficiency of the different components existing in a handwritten text recognition algorithm.

\subsection{Choice of train, validation and test partitions}
\label{subsection: c6-choice_train_test_partitions}

The use of the same training, validation, and evaluation or test partitions of the reference datasets in the different investigations on any modeling problem is a fundamental aspect where there is a general consensus. The training partition is used in the parameter setting of the different experiments, the validation partition allows to compare the different experiments in terms of accuracy or error metrics to select the optimal parameterization. The evaluation or test partition allows to calculate the metrics values on the selected optimal parameterization experiment. In this sense, the test partition must contain precisely the same data along the different lines of research so that the comparison of metrics between different authors can be reliable.

However, in the case of the general handwritten text recognition problem, there are exceptions to the above, in one of the most widely used databases to develop and compare experimental results. In the IAM database \cite{marti2002iam} coexist, over time, two different partitions of the train, validation, and evaluation datasets, which are widely used by different authors that obtain results with one of the partition but compare them with other publications results obtained for the other partition. 

The first partition (hereafter official) is provided on the official site of the database \footnote{\url{http://www.fki.inf.unibe.ch/databases/iam-handwriting-database} Last access: April 2021} and separates the total of 13,353 database lines into 6,161 for train, 940 for validation (named validation 2 in the site), and 1,861 for test. Additionally, it provides a second validation partition with 900 lines (named validation 1 in the site) and excludes from the partitions a total of 3,491 lines.

Several authors have used the second partition (usually referred to in the literature as Aachen partition) from 2013 to the present. To the best of our knowledge, it was first used by Kozielski et al. \cite{kozielski2013improvements} and include a total of 6,482 lines to train, 976 for validation and 2,915 for test. This partition excludes a total of 2,980 lines.

Table \ref{table:crosstab_IAM_partitions}, we provide a cross table that includes the number of common lines in both partitions. It can be seen that all training lines of the official partition are included in the training set of the Aachen partition. However, in the validation and test partitions, there is no match between the two partitions and the data are mixed. It implies that it is not a good practice the comparison of results obtained with the two partitions. Regarding the excluded lines, there is overlap except that the Aachen partition has incorporated lines excluded by the official partition into its partitions. In particular, it has included 321 lines to its train partition, 56 to validation, and 134 to eval or test.

\begin{table}[!ht]
\centering
\begin{tabular}{lrrrr} 
 \toprule
  & \multicolumn{4}{l}{\textbf{Aachen}} \\
 \cmidrule(r){2-5}
 \textbf{official} & excluded & Train & Test & Validation \\
 \hline
    excluded&   2,980 &   321 &   134 &  56 \\
    train &         0 & 6,161 &     0 &   0 \\
    test &          0 &     0 & 1,313 & 548 \\
    validation &    0 &     0 &   709 & 231 \\
    validation1&    0 &     0 &   759 & 141 \\
 \bottomrule
\end{tabular}
\caption{Cross table of lines count between Official and Aachen partitions of IAM database.}
\label{table:crosstab_IAM_partitions}
\end{table}

The main difficulty lies in the fact that it is not evident which partition is used by which author, as it is rarely explicitly stated in the publication. The principal indicator to identify it is the line or word counts per partition that are usually included in the description sections of the IAM datasets of each of the publications. These counts almost always coincide with the counts of one or another partition. Analyzing the data for this publication, it can be identified, in most cases, which partition has been used. 

Two additional particular cases can be observed analyzing this data. Firstly, it can be seen that in the publications before 2011, that claim to use the official partition that accompanies the IAM database, such as \cite{graves2009novel} or \cite{espana2010improving}, declare different line counts in the validation and test partitions from those currently available. In particular, declare 920 lines for validation instead of 940, and 2,781 lines for test instead of 1,861. It is probably due to a modification of the official partition after 2010. Secondly, several authors like Puigcerver \cite{puigcerver2017multidimensional} and Ingle et al. \cite{ingle2019scalable} declare to use a number of lines for the train partition that matches the official partition count, but in validation and test they declare a number of lines corresponding to the Aachen partition. In this case, and since the matching of the test partition is the key factor for comparing results, the outcome of these publications can establish valid comparisons with other results using the Aachen partition, even if they use the official training partition.

The use of one or the other partition impacts the results in terms of CER and WER in the validation and test partitions. Table \ref{table:IAM_partitions_results_comparative} includes a comparison of the results obtained using the two partitions on the model proposed in this Thesis, with the parameterization defined in the previous section, corresponding to the final model selected. CER and WER results are provided for the cases of not using any lexicon, using the standard lexicon, and using the test lexicon of the IAM database.

\begin{table}[!ht]
\centering
 \begin{tabular}{llrrrrr}
 \toprule
 Lexicon &   & \multicolumn{2}{l}{Validation} & & \multicolumn{2}{l}{Test} \\
 \cmidrule(r){3-4} \cmidrule(r){6-7} & IAM Partition& CER & WER & & CER & WER \\
 \midrule
 No lexicon     & Official&     7.4 &   17.8 &&     9.4&   19.8 \\
                &   Aachen&     6.7 &   15.1 &&     9.3&   21.0\\
 \midrule
 Standard lexicon&Official&     9.4 &   16.8 &&     11.3 &  18.7 \\
                &   Aachen&     8.8 &   14.9 &&     11.2 &  19.6\\
 \midrule
 Test Lexicon   & Official&     NA&   NA &&     7.4&  12.0 \\
                &   Aachen&     NA&   NA &&     7.6&  13.0 \\
 \bottomrule
 \end{tabular}
\caption{Results comparative between the Official and Aachen IAM partitions.}
\label{table:IAM_partitions_results_comparative}
\end{table}

It is observed that in the test partition, the results at the CER level are pretty similar. However, the results at the WER level, the Official partition obtains approximately one point lower WER than the Aachen partition consistently to use different decoding lexicons. It is also observed that differences in results between validation and test of the official partition are much smaller than those of the Aachen partition.

Since the test data of the two partitions are not the same, and as shown in the Table \ref{table:IAM_partitions_results_comparative} the results differ, it is recommended that a publication using one of the partitions to be compared with publications using the same partition. It has not always been the case; there are several cases in which a publication has performed the experiments with one partition and compared their results with publications that have used the other partition. Below are some cases that have been identified in the literature:

\begin{itemize}

  \item Publications \cite{kozielski2013improvements} of 2013, and \cite{pham2014dropout}, both use the Aachen partition (with 2915 lines in test) and compares they results with \cite{graves2009novel} of 2009 and \cite{espana2010improving} of 2010 that use the older version of the Official partition (with 2781 lines in test). To our understand, the first usage of the Aachen partition in a publication is in \cite{kozielski2013improvements}.

  \item Publication \cite{doetsch2014fast} use the Aachen partition and compares its results with, among others, \cite{espana2010improving} and \cite{dreuw2011hierarchical}, which use the Official partition.

  \item Publication \cite{poznanski2016CVPR} use the standard split that comes with the database (Official partition) and compares its results with, among others, \cite{kozielski2013improvements}, \cite{pham2014dropout}, \cite{doetsch2014fast} and \cite{bluche2014comparison}, which use the Aachen partition.

  \item Publication \cite{krishnan2016deep} use the Official partition and compares its results with, among others, \cite{doetsch2014fast} and \cite{bluche2014comparison}, which use the Aachen partition.

  \item Publication \cite{dutta2018improving} use the Official partition and compares its results with, among others, \cite{puigcerver2017multidimensional} and \cite{pham2014dropout}, which use the test Aachen partition.

\end{itemize}

The above cases do not modify the valuable contributions of each paper, but they do make it difficult to interpret the tables of comparative results provided in these publications.

In the case of the RIMES database, there are also two partitions used by different authors. The partition corresponding to the ICDAR2009 competition (with 51,739 words for train and 7,464 words for test), and the one corresponding to the ICDAR2011 competition (the most common). It uses the same train set as the previous partition (51,739 words), converts the previous test set to validation (7,464 words), and adds a new test set with 7,776 new words. The test set is different in both partitions, so it is considered that the comparison of results obtained with these two partitions is not valid. In this case, the two partitions have different names and are well differentiated in the literature.

\subsection{Filter transcription errors in IAM database}

The IAM database provides images and transcripts at different levels, including line and word level. Word level transcripts include a column indicating whether the transcript 'can be bad' or is correct. This transcription error mark is defined at the line level. If a line contains an inadequately segmented word, all words in the line are marked as incorrect. As explained in Subsection \ref{chapter:2 - subsect:IAM} this is because the segmentation and annotation of the IAM database at the word level were done with a procedure that manually validated the results at line level. There are approximately 15\% of words marked as wrong in the database, although only a part of them are really segmentation errors.

In most publications, there is no reference to this fact from the IAM database, we understand that in these cases the correct transcription mark is not considered, and all the images available in the selected partition are used, although we cannot be entirely sure. If all the images are not used, a part of the errors accounted in these publications comes from the segmentation error and not necessarily from an error in the proposed model. In any case, any two publications using all available data are perfectly comparable. Several publications explicitly eliminate these cases, such as \cite{sueiras2016using} or \cite{wigington2017data}. In the particular case of \cite{wigington2017data} it is indicated that the erroneous data have been excluded from the test partition but have been considered in the train and validation partitions, arguing that the additional cases included compensate for the actual transcription errors in terms of training improvement.

In terms of evaluation metrics, the elimination of cases marked as erroneous from all partitions has the advantage that transcription errors are eliminated from the test partition. However, it has the disadvantage that the training dataset is reduced, which makes the model difficult to train, especially in the case of deep architectures, for which having a larger volume of train data is essential. In any case, the results to be compared should be calculated on the same set of test images, whether they have these errors or not.

\subsection{Character set selected for decoding}

The IAM database has 79 characters that include the upper and lower case letters of the English language, the 10 digits, various punctuation marks, the character white space (ASCII 32), and a special character for garbage (\#). Most publications use the complete character set or ignore the garbage character and the white space character. The frequency of these last two characters in the database is very low (33 cases of each character), and there are no significant differences between considering them or not in the results.

Most authors perform the evaluation using the entire set of characters indicated. However, certain authors limit the characters used in the transcription. For example:

\begin{itemize}

  \item In reference \cite{krishnan2016deep}, it is indicated that the evaluation of the performance is done in a case-insensitive manner, and includes comparisons of that evaluation with references \cite{almazan2014word}, \cite{bluche2014comparison} and \cite{doetsch2014fast}, which in our understanding use the full charset with a distinction between upper and lower case characters.

  \item In reference \cite{poznanski2016CVPR} the author indicates that the character set employed contains the lower and upper case of the Latin alphabet. Digits were not included but are considered in the accounting of errors. Punctuation marks are ignored. The publication includes a table comparing test score with multiple authors (including \cite{almazan2014word}, \cite{bluche2014comparison} or \cite{doetsch2014fast}) who are understood to use the complete character set including punctuation marks.

  \item Another example appears in \cite{dutta2018improving} which uses a charset that includes only lowercase letters and compares it to reference works \cite{sueiras2018offline}, \cite{puigcerver2017multidimensional} or \cite{pham2014dropout} which use the full character set with capitalization and punctuation. 

  \item In the case of \cite{wigington2017data} the authors use different charsets, including or not capital letters and punctuation marks. It provides results for the different cases and compares them with other authors who have used the same character set. Thus, valid comparisons are obtained with multiple authors and for the different sets of characters used.

\end{itemize}

In the case of the RIMES database for the ICDAR2011 partition, it has a total of 81 characters. Most authors use the above character set, but some authors use a character set that ignores uppercase letters. 

\begin{itemize}

    \item In reference \cite{menasri2012a2ia}, the authors indicate that the evaluation is done over ICDAR2011 competition dataset, to which the publication refers, is case-insensitive. 
  
    \item In publication \cite{stuner2020handwriting}, the evaluation is carried out using lowercase letters for comparison purposes and indicating that there are transcription errors, especially for the letter 'j', in words 'je' and 'j'. This last publication establishes comparisons with \cite{menasri2012a2ia} which also provides results for the lower case character set, and with \cite{poznanski2016CVPR} which declares to use the full character set with upper and lower case.
    
    \item In reference \cite{wigington2017data} the authors use different char sets, and establish valid comparisons in different tables. However, the final results are obtained with a reduced character set and these are cited in comparative tables of other authors who use the full character set (for example \cite{kang2021candidate}).

\end{itemize}

The use of a reduced character set in decoding has a significant impact on WER and CER metrics. In Wigington et al. \cite{wigington2017data} a 23\% improvements are declared in WER (from 3.69 to 2.85) and a 20\% in CER (from 1.69 to 1.34) over the RIMES database.

\subsection{Data augmentation}
\label{section:comparing_results_data_augmentation}

There are two data augmentation strategies that can be applied separately or together: data augmentation during the training process and test-time data augmentation. 

Due to the variability of configuration strategies and the introduction of data augmentation in the training process, the different strategies and algorithms can be considered as part of the solution proposed by each author. Therefore, from the point of view of comparing the final results of two different approaches, it is perfectly correct to compare two approaches that use different data augmentation strategies in train or one that uses it with another that does not use it. There are references in the literature (such as \cite{mikolajczyk2018data}, or \cite{cui2015data}) that indicate that data augmentation techniques improve the training of deep network architectures, especially in areas where the dataset size is not very large, reducing overfitting and improving the generalizability of the model. However, in the case of HMM applied to handwritten text recognition, it is not a commonly performed practice. 

The test-time data augmentation tends to improve significantly the accuracy metrics by depending on the type of model used and on the size of the training dataset \cite{shanmugam2020and}. For example, Wigington et al. \cite{wigington2017data} report a 20\% of WER reductions for IAM word level (official partition) and a 25\% reductions for RIMES at word level based on a specific experiment to evaluate the test-time data augmentation strategy.

The test-time data augmentation is a procedure that can be applied to any model built, which will generally improve the evaluation metrics. In this sense, its use is commonly conditioned on whether such a procedure is a common practice in the problem literature. In handwritten text recognition, it is not a common practice to provide the evaluation metrics results using test-time data augmentation. In cases where it is used, as in \cite{poznanski2016CVPR}, \cite{dutta2018improving} or in \cite{wigington2017data}, direct comparison of the results with other authors who do not use test-time data augmentation makes it difficult to interpret the data since it is not possible to determine whether the differences of results between methods are due to larger or smaller efficiency, or to the use of test-time data augmentation by one of them and not by the other. Ideally, if this technique is used, we suggest the inclusion of an ablation study that provides comparative results of the algorithm with and without this technique.

\subsection{Data normalization}
\label{section:Data_normalization}

A common practice in the various approaches to offline handwritten text recognition problem is to perform an image normalization pre-processing to reduce the inherent variability of the handwritten text. In fact, in older publications, the text normalization step is a practice almost always present (see, for example, references \cite{graves2009novel} or \cite{bluche2013tandem}). In recent years, especially with deep neural network architectures, some authors such as Huang et al. \cite{huang2020end} or Kang et al. \cite{kang2020unsupervised}, propose recognition algorithms in which no prior normalization of the input images is applied to the model, except for a rescaling and adjustment to a fixed size of all images. This is usually due to some specific requirements of models that need that the input data have a particular structure.

In this Thesis, experiments have been carried out both with original images without normalization and with normalized images following the image preprocessing algorithm described in Subsection \ref{C6 - section - Image Preprocesing}.

As far as the comparison of different results is concerned, data normalization plays a similar role to data augmentation during training. Since specific models only work correctly if the data are normalized in a certain way, it is understood that data normalization is part of each of the problem-solving strategies proposed by each author. Therefore, comparing results from two approaches that apply different normalizations or comparing with solutions that do not include a normalization step provides interpretable conclusions about the compared results.

If the normalization step is optional in the proposed approach, it is recommended to include, as part of the results, an ablation study comparing the results with and without the use of data normalization.

\subsection{Decoding with lexicons and language models}

As detailed in Section \ref{SubSect:Decoding_predictions}, in the HTR problem is a common practice to apply a decoding process to the result, provided by the visual model that encodes the input image so that the context and language information of the text to be decoded are taken into consideration. For this purpose, language models (LM) or lexicon search are usually used, especially in cases where an architecture based on Hidden Markov Models (HMM) was proposed. 

As discussed in the previous section on data augmentation, it is correct to compare different publications that address the problem with or without decoding. It is especially true when the decoding process is integrated as a part of the model that is trained together. It happens, for example, in models based on HMM that need subsequent decoding through an LM to obtain results.

In cases where decoding with lexicons or LM is an optional subsequent step applied to the output of the visual model, comparisons are more controversial. If it is done to compare with another that also does it, the most appropriate is to use the same lexicon or the same LM, but for that, there should be a standard way to use or share the language model. If there are licensing issues to share the object, instructions or the source code to create the object should be shared. If comparing with another approach that does not employ an LM, it is most appropriate to compare the results of the visual models without using the LM.

The objective is to measure and compare the different publication contributions separately to determine the relevance of each one in terms of error. For example, a publication proposes a new visual model, uses a new language model, and only provides a single error measure that improves the state-of-the-art. In that case, it is impossible to determine whether the improvement comes from the visual model or the LM or the combination of both. 

In the case of this Thesis, we employ two lexicons by database to evaluate the models. First, a standard lexicon commonly used in the literature for this database, and second, the test lexicon, to get an error measure without out-of-vocabulary (OOV) words. In our case, whenever error results are provided using a lexicon, the same result is also provided without using any lexicon.

\section{Comparative of results}
\label{section:Results comparative}

This section compares the results obtained by the proposed model with results from other publications, given the difficulties and considerations made in the previous section. First of all, some recommendations aimed at comparing results easily interpretable are proposed in Subsection \ref{section:Common framework to compare results}. Subsequently, the results are compared on RIMES database in Subsection \ref{subsection - Rimes historical results} and on IAM database in Subsection \ref{subsection - IAM historical results}, respectively.

\subsection{Best practices recommendations to results comparative}
\label{section:Common framework to compare results}

The main objective of these best practices suggested is to be able to quantify as precisely as possible each of the contributions of a publication providing different error measures for each of the innovations introduced, and detailing all the elements necessary to guarantee the reproducibility of the results.

First of all, it is always recommended to publish the results for the standard metrics of CER and WER calculated on a well-identified test data set. In general, it is recommended not to alter the databases used, i.e., not to filter any data and to use the complete character set of the databases, differentiating between uppercase, lowercase, digits, and special characters. 

When including tables of comparative results with other authors, make sure that they can be compared without bias because they are calculated not only on the same datasets but using the same set of valid characters and with the same image selection for training and testing (the same error cases are always filtered out). In general, these tables compare the best final results of each publication, including all the contributions made together so the recognition method can be evaluated entirely. This data is usually easily identifiable in all available publications. If the publication makes relevant contributions in different aspects of the problem, it is proposed to publish comparative tables with other authors that focus on each of these aspects separately. For example, it is recommended to construct an additional comparative table using only the visual model and comparing it with results from other authors that have also been obtained using only visual models, if contributions are made on both the visual model and the language model.

The presentation of partial results to measure the relative impact on the error of the different inputs is also proposed. It is suggested first to provide the results obtained directly by the visual model exclusively. Defining a baseline model and then providing results on the absolute and relative improvement of the inclusion of each of the published inputs to the baseline model.

It is also recommended to publish sufficient details of the database used, the proposed model, the lexicons or language models used, and the experimental setup implemented to facilitate the reproducibility of the results obtained. Ideally, it is proposed to publish the source code used in the experiments.

The following subsections include detailed tables of results from various publications for IAM and RIMES databases at both word and line levels. In the case of IAM database, the results obtained with the two available partitions have been separated. Other results not included in the tables are also detailed, explaining the reasons and the differences between these cases and those appearing in the tables.

\subsection{Historical results comparative for RIMES database}
\label{subsection - Rimes historical results}

The following are the published results for the HTR continuous problem using RIMES database at word level. Values are provided for the CER and WER metrics, calculated on the test partition defined for the ICDAR2011 competition when available. This test partition is composed of a total of 100 lines and 7,464 words.


Our results included in the tables correspond to those provided in Subsection \ref{C6 - Experiments - Final model}, corresponding to the final model selected. The most appropriate parameterization has been selected according to the results in the validation partitions obtained in the previous experiments.

The results obtained using the complete set of characters available in the database for decoding are included in the tables. Publications using a reduced character set and publications using test time data augmentation are included separately.

Differentiated tables are provided for three cases depending on the use of lexicons or language models to obtain the results:
\begin{itemize}
  \item Without using lexicons or LMs.
  \item Using lexicons or LMs that do not include the test partition words (so, there are OOV words usually).
  \item Using lexicons or LMs that do include the test partition words (so, there are no OOV words).
\end{itemize}

The following abbreviations are used in the notes to the different tables.
\begin{itemize}
  \item \textbf{DA}: train data augmentation is used
  \item \textbf{Synt}: A pretrain with synthetic data is performed.
  \item \textbf{Norm}: Normalization is used or not (No norm.).
  \item \textbf{LM}: Language model is used to decoding.
\end{itemize}

Table \ref{table:Rimes_word_visual} includes the WER and CER values obtained on the test partition of RIMES database corresponding to the ICDAR2011 competition. The results have been obtained by directly applying a visual model and are not decoded by any lexicon or language model. 

\begin{table}[!ht]
\centering
\begin{tabular}{l l l l r r} 
\toprule
 Reference & Year & Model type & Notes & CER & WER \\
\midrule

 Bluche et al. \cite{bluche2015deep}     & 2015 & MLP &         & 17.8 & 59.5 \\ 
 Bluche et al. \cite{bluche2015deep}     & 2015 & RNN &         &  5.6 & 20.9  \\
 Bluche et al. \cite{bluche2016joint}    & 2015 &     &         &  2.9 & 12.6 \\
 Kang et al. \cite{kang2021candidate}    & 2021 & seq2seq & No norm, DA, Synt & \textbf{2.65} & \textbf{8.71} \\
 Ours                                    & 2021 & seq2seq & Norm, DA & 4.5 & 13.2 \\
 
\bottomrule
\end{tabular}
\caption{RIMES word level results. Visual model. No lexicon or LM.}
\label{table:Rimes_word_visual}
\end{table}

Table \ref{table:Rimes_word_lexicon} includes the WER and CER values obtained on the test partition of the RIMES database corresponding to the ICDAR2011 competition. The results have been obtained by decoding the output of the visual model by a lexicon or language model that does not use the test word set.

\begin{table}[!ht]
\centering
\begin{tabular}{l l l l r r} 
\toprule
 Reference & Year & Model type & Notes & CER & WER \\
\midrule

 Bluche et al. \cite{bluche2015deep}  & 2015 & MLP     &  & 7.2 & 26.1 \\ 
 Bluche et al. \cite{bluche2015deep}  & 2015 & RNN     &  & 4.3 & 16.4 \\
 Kang et al. \cite{kang2021candidate} & 2021 & seq2seq & No norm, DA, Synt, LM  & \textbf{2.32} & \textbf{7.47} \\
 Ours                                 & 2021 & seq2seq & Norm, DA & 5.9 & 12.2 \\

\bottomrule
\end{tabular}
\caption{RIMES word level results. Lexicon or LM not including test data.}
\label{table:Rimes_word_lexicon}
\end{table}

Table \ref{table:Rimes_word_lexicon_test} includes the WER and CER values obtained on the test partition of the RIMES database corresponding to the ICDAR2011 competition. The results have been obtained by decoding the output of the visual model by a lexicon or language model that explicitly uses the set of test words.

\begin{table}[!ht]
\centering
\begin{tabular}{l l l l r r} 
\toprule
 Reference & Year & Model type & Notes & CER & WER \\
\midrule

 Bluche et al. \cite{bluche2013tandem}  & 2013 & CNN + HMM &  &     & 9.2 \\
 Kang et al. \cite{kang2021candidate} & 2021 & seq2seq & No norm, DA, Synt, LM  & \textbf{1.45} & \textbf{3.75} \\
 Ours                                   & 2021 & seq2seq   & Norm, DA     & 2.7 & 5.4 \\
 
\bottomrule
\end{tabular}
\caption{RIMES word level results. Test data included in lexicon or LM.}
\label{table:Rimes_word_lexicon_test}
\end{table}

Error metrics from other publications are listed below, and the reason for not including them in the above tables of RIMES results is given:

\begin{itemize}

    \item Menasri et al. \cite{menasri2012a2ia}, in 2012, use a character set case insensitive. It obtains a WER of 4.75 on the ICDAR2011 test partition using a lexicon that includes the test words.

    \item Pham et al. \cite{pham2014dropout}, in 2014, use the RIMES database partition corresponding to the ICDAR2009 competition (instead of the ICDAR2011 partition). Since the test data of these two partitions do not match, comparisons between the two partitions cannot be established. The model implemented is MDLSTM type and obtains a WER of 27.01 and a CER of 8.62 without using lexicons or LM.

    \item Poznanshi et al. \cite{poznanski2016CVPR}, in 2016, indicate that the character set employed for RIMES contains the lower and upper case Latin alphabet, digits and accented letters. Punctuation marks are ignored. The model used are CNN-based and achieved a CER of 1.90 and a WER of 3.90 for RIMES using the test lexicon for decoding. 
  
    \item Wigington et al. \cite{wigington2017data}, in 2017, use an MDLSTM with CTC, employing test time data augmentation and a reduced character set without punctuation marks and upper case. It obtained a WER of 11.29 and a CER of 3.09 without any lexicon or LM, and a WER of 2.85, and a CER of 1.36 using the test lexicon.
    
    \item Dutta et al. \cite{dutta2018improving}, in 2018, use a charset that includes only lowercase letters. They implement a pre-trained seq2seq type model with synthetic images and obtained a WER of 7.04 and a CER of 2.32 for the RIMES test partition without employing any lexicon or LM. When using a lexicon that includes the test words they obtained a WER of 1.86 and a CER of 0.65.

\end{itemize}

\subsection{Historical results comparative for IAM database}
\label{subsection - IAM historical results}

The results obtained in the word recognition problems on the IAM database are detailed in the tables below. The results that are comparable in the terms discussed in preceding subsections are grouped.

These results are shown differentiated by partition (Official and Aachen). If any publication uses a test set different from those of these partitions, it is listed separately. Results obtained in papers using the entire set of characters available in the database for decoding are included in the table. Papers using a reduced character set and publications using test time data augmentation are also listed separately.

For each of the two partitions, separate tables are provided for three cases depending on the use of lexicons or language models:
\begin{itemize}
  \item The result has been obtained without using lexicons or LMs.
  \item Lexicons or LMs that \textbf{do not} include the test sample have been used.
  \item Lexicons or LM that \textbf{do} include the test sample have been used.
\end{itemize}

The publication year, the model type, and the necessary notes to identify the particularities of a specific result are detailed in the tables for each reference. The CER and WER metrics are included when available as results. In the case that data are not known, these are left blank.

Table \ref{table:IAM_aachen_word_visual} includes the WER and CER values obtained on the test data of IAM database corresponding to the \textit{Aachen} partition. The results have been obtained by directly applying a visual model and are not decoded by any lexicon or language model. 

\begin{table}[!ht]
\centering
\begin{tabular}{l l l l r r} 
\toprule
 Citation & Year & Model type & Notes & CER & WER \\
\midrule

 Pham et al. \cite{pham2014dropout}     & 2014 & MDLSTM & No LM & 13.92 & 31.44 \\
 Bluche et al. \cite{bluche2015deep}    & 2015 & MLP &  & 15.6 & 54.2 \\
 Bluche et al. \cite{bluche2015deep}    & 2015 & RNN &  &  7.3 & 24.7 \\
 Bluche et al. \cite{bluche2016joint}   & 2015 &     &  &  7.9 & 24.6 \\
 Kang et al. \cite{kang2021candidate} & 2021 & seq2seq & No norm, DA, Synt  & \textbf{5.79} & \textbf{15.15} \\
 Ours                                   & 2021 & seq2seq & Norm, DA & 9.3 & 21.0 \\

\bottomrule
\end{tabular}
\caption{IAM-Aachen test partition results. Word level results. Visual model, no lexicon or LM.}
\label{table:IAM_aachen_word_visual}
\end{table}

Table \ref{table:IAM_aachen_word_lexicon} includes the WER and CER values obtained on the test data of IAM database corresponding to the \textit{Aachen} partition. The results have been obtained by decoding the output of the visual model by a lexicon or language model that does not use the set of test words.

\begin{table}[!ht]
\centering
\begin{tabular}{l l l l r r} 
\toprule
 Citation & Year & Model type & Notes & CER & WER \\
\midrule
 Kang et al. \cite{kang2021candidate} & 2021 & seq2seq & No norm, DA, Synt, LM  & \textbf{5.74} & \textbf{15.11} \\
 Ours                                 & 2021 & seq2seq & Norm, DA               & 11.2 & 19.6 \\
\bottomrule
\end{tabular}
\caption{IAM-Aachen test partition results. Word level results. Lexicon or LM not include test data.}
\label{table:IAM_aachen_word_lexicon}
\end{table}

Table \ref{table:IAM_aachen_word_lexicon_test} includes the WER and CER values obtained on the test data of IAM database corresponding to the \textit{Aachen} partition. The results have been obtained by decoding the output of the visual model by a lexicon or language model that explicitly uses the set of test words. In this case, we have not been able to identify another publication that meets the conditions identified for inclusion in the table. Several cases could be included, but they use a reduced character set or use test time data augmentation.

\begin{table}[!ht]
\centering
\begin{tabular}{l l l l r r} 
\toprule
 Citation & Year & Model type & Notes & CER & WER \\
\midrule
 Kang et al. \cite{kang2021candidate} & 2021 & seq2seq & No norm, DA, Synt, LM  & \textbf{4.27} & \textbf{8.36} \\
 Ours                                 & 2021 & seq2seq & Norm, DA               & 7.6  & 13.0 \\
\bottomrule
\end{tabular}
\caption{IAM-Aachen test partition results. Word level results. Test data included in lexicon or LM.}
\label{table:IAM_aachen_word_lexicon_test}
\end{table}

Listed below are the results obtained in the publications using the official partition of the IAM database. As indicated in Subsection \ref{subsection: c6-choice_train_test_partitions} this partition has undergone variations over time. For example, publications prior to 2010 declare a different number of lines in the official test partition. When including results from different publications in the tables, we have followed the criterion that the number of lines or test words match the data published in the official IAM database \footnote{\url{http://www.fki.inf.unibe.ch/databases/iam-handwriting-database} (last access: April 2021)} which corresponds to a total of 1,861 lines, or 17,616 words. In some cases, as in \cite{puigcerver2017multidimensional}, the counts of the train or validation partitions do not match exactly with those available on the database official website, although the test count does. These cases have also been included in the tables because, as previously argued, the comparison is considered valid if it has been calculated on the same test data, regardless of whether there are differences in the other partitions.

Table \ref{table:IAM_Official_word_visual} includes the WER and CER values obtained on the test data of IAM database corresponding to the \textit{Official} partition. The results have been obtained by directly applying a visual model and are not decoded by any lexicon or language model. 

\begin{table}[!ht]
\centering
\begin{tabular}{l l l l r r} 
\toprule
 Citation & Year & Model type & Notes & CER & WER \\
\midrule
 Bluche et al. \cite{bluche2013tandem} & 2013 & CNN + HMM &  &  & 23.7 \\
 Ours & 2021 & seq2seq & Norm, DA & \textbf{9.4} & \textbf{19.8} \\
\bottomrule
\end{tabular}
\caption{IAM-Official test partition results. Word level results. Visual model, no lexicon or LM.}
\label{table:IAM_Official_word_visual}
\end{table}

Table \ref{table:IAM_Official_word_lexicon} includes the WER and CER values obtained on the test data from the IAM database corresponding to the \textit{Official} partition. The results have been obtained by decoding the output of the visual model by a lexicon or language model that does not use the set of test words.

\begin{table}[!ht]
\centering
\begin{tabular}{l l l l r r} 
\toprule
 Citation & Year & Model type & Notes & CER & WER \\
\midrule
 Bluche et al. \cite{bluche2013tandem} & 2013 & CNN + HMM & LM &  & 20.5 \\
 Ours & 2021 & seq2seq & Norm, DA & \textbf{11.3} & \textbf{18.7} \\
\bottomrule
\end{tabular}
\caption{IAM-Official test partition results. Word level results. Lexicon or LM not include test data.}
\label{table:IAM_Official_word_lexicon}
\end{table}

Table \ref{table:IAM_Official_word_lexicon_test} includes the WER and CER values obtained on the test data of IAM database corresponding to the \textit{Official} partition. The results have been obtained by decoding the output of the visual model by a lexicon or language model that explicitly uses the set of test words.

\begin{table}[!ht]
\centering
\begin{tabular}{l l l l r r} 
\toprule
 Citation & Year & Model type & Notes & CER & WER \\
\midrule
 Bluche et al. \cite{bluche2013tandem}  & 2014 & CNN + HMM  &   & 9.2    & 20.5 \\
 Almazan et al. \cite{almazan2014word}  & 2014 & k-NN       &   & 11.27  & 20.01 \\
 Ours                                   & 2021 & seq2seq    & Norm, DA      & \textbf{7.4}    & \textbf{12.0} \\
\bottomrule
\end{tabular}
\caption{IAM-Official test partition results. Word level results. Test data included in lexicon or LM.}
\label{table:IAM_Official_word_lexicon_test}
\end{table}

Error metrics from other publications are listed below, and the reason for not including them in the above tables of IAM results is given:

\begin{itemize}

  \item Graves et al. \cite{graves2009novel}, in 2009, use the previous official partition of IAM databases, with 2,781 lines in the test set. The authors use a LSTM with the CTC decoder and provides a WER of 25.9 and a CER of 18.2 using a standard lexicon that not include the test words. 
 
  \item Krishnan et al. \cite{krishnan2016deep}, in 2016, indicate that the evaluation of the performance is done in a case-insensitive manner. This work provides a WER of 6.61 and a CER of 3.72 decoding with the test lexicon over the official IAM partition. 

  \item Poznanski and Wolf \cite{poznanski2016CVPR}, in 2016, indicate that the character set employed contains the lower and upper case Latin alphabet. They also point out that digits were not included but are considered in the accounting of errors. Punctuation marks are ignored. The model used are CNN based and get a CER of 3.44 and a WER of 6.45 using the official partition and the test lexicon for decoding. 

  \item Wigington et al. \cite{wigington2017data}, in 2017, use a MDLSTM with CTC, employing test time data augmentation and a reduced character set without punctuation and case. For IAM dataset, they obtain a WER of 19.07 and a CER of 6.07 without Lexicon or LM, and a WER of 5.71 and a CER of 3.03 using the test lexicon. It is not possible to determine which partition has been used.
 
  \item Dutta et al. \cite{dutta2018improving}, in 2018, use a charset that includes only lowercase letters. They implement a pre-trained seq2seq type model with synthetic images and obtains a WER of 12.61 and a CER of 4.88 for the official IAM test partition without employing any lexicon or LM. Using the test lexicon, they obtain a WER of 4.07 and a CER of 2.17.
 
  \item Kang et al. \cite{kang2018convolve}, in 2018, propose a seq2seq model and they achieve a CER of 6.88 and a WER of 17.45 without using any lexicon or LM using the Aachen test partition, but exclude cases not marked as 'OK', using 20,305 words out of 25,920 in the original approval.

\end{itemize}

\section{Discussion}
\label{C6 - section - Results discussion}

In this chapter, we have presented a system for recognizing offline handwritten words. The proposed architecture is inspired in the way that the humans try to identify handwritten words and it is based in the seq2seq architecture with the addition of a convolutional network. 
A detailed set of experiments has been performed to analyze the possibilities of this new architecture. The main results and conclusions obtained from these experiments are the following ones:

\begin{itemize}

\item The analysis performed at the character recognition level concluded that VGG convolutional architecture was the most suitable one for this problem. However, in the experiments carried out at the word level, it is observed that LeNet and VGG architectures analyzed obtain different results depending on the database and whether or not the image partitioning strategy is applied in patches. From the above, it is not easy to establish a conclusion that favors one architecture over the other, especially in the standard IAM and RIMES databases. What is observed is that LeNet architecture performs better than VGG in the smaller Osborne database.

\item From the results obtained, it is concluded that dividing the original image into patches does not have a significant advantage over the strategy of directly applying the convolutional component to the original image without dividing it. In any case, it is considered that this patching strategy provides a more robust way to connect the output of the convolutional component with the RNN of the encoder. It is true that the patching strategy requires an adjustment of the patch size and step size parameters to work optimally, but it has been observed in the experiments that results very close to the optimum can be obtained for wide ranges of patch sizes and step sizes. It is therefore considered to be a robust architecture concerning these two parameters.

\item In general, the optimal parameterization identified is quite similar for the IAM and RIMES databases. In the case of Osborne database, there are significant differences, especially in the encoder parameterization. It is considered that the smaller size of this database contributes to the fact that the results of the experiments performed with it have less stability than those performed with the other two databases. Therefore, conclusions concerning Osborne database should be taken with caution.

\item Both the pre-normalization of the text images and the train data augmentation strategy significantly improve the model accuracy. In the experiments carried out, it has been observed that the combination of both techniques is the most influential factor in reducing the error consistently in the three databases analyzed.

\item The error analysis indicates that the model performs worse with long texts. It could be caused by the fact that the decoder identifies the characters present in the image sequentially (i.e., for identifying the next character, the prediction of the previous characters is used as input). However, the attention mechanism allows access to each decoder association to the complete information encoded in the input image. It is also likely that the low frequency of long text images in the databases is a factor that explains the low accuracy of the recognition of these long texts.

\item The main advantage of the proposed model is the integrated management of local aspects of a word image related to the characters that compose the word (convolutional network), the sequential horizontal aspects of the word image (encoder), and the sequence of characters in the word (decoder). When applying our visual model on two standard databases, the obtained results are competitive with those others published in the literature. 

\end{itemize}

\chapter{Conclusions}
\label{chapter_conclusions}

In this Thesis, we have focused on investigating the problem of offline continuous handwritten text recognition, consisting of transforming cursive text images into their transcription. In particular, we have worked with word images for which no character segmentation is available.

The motivation and specific objectives of this thesis are detailed in the Chapters \ref{chapter:introduction} and \ref{chapter_analysis}. In addition, an introduction to the HTR problem in general and a more detailed definition of the offline continuous HTR problem, the specific subject of this thesis, are provided. It also introduces the general concepts of the problem, such as image processing techniques, decoding alternatives, or performance evaluation metrics. 

Chapter \ref{chapter_Handwriting_Databases} provides detailed descriptions of the databases available to address the problems of recognizing isolated handwriting characters and continuous HTR. By studying the databases available for isolated characters and words, we identified that none of the character databases include examples of special characters or punctuation marks present in the continuous text databases. To fill this gap, we presented the new database COUT of handwritten isolated character images that includes a wide range of up to 93 different characters, including all printable characters of the ASCII standard. To our best knowledge, it is the database of isolated handwritten characters with more different categories. Additionally, an algorithm is provided for altering the original characters by incorporating artifacts that emulate the context in which the character is found when it is part of a word. This facilitates the study of isolated character recognition models that can be incorporated into continuous word recognition models.

Chapter \ref{chapter_background} provides an extensive review of the HTR problem research over the years. We detail the various contributions that have made significant advances in resolving the problem by putting them in context, and we review in particular detail the aspects most related to the work completed in this Thesis. This review allows us to have a clear perspective of the main research lines in the continuous HTR problem.

Next, in Chapter \ref{chapter_character_models} we explore several models with different convolutional architectures applied to the problem of recognition of isolated handwritten characters. We provide a detailed analysis of the capabilities and limitations of each model when are used on different character sets, including upper and lower case letters, special characters, and punctuation marks. We have consistently identified that the VGG architecture as produced the lowest error in recognition of isolated handwriting characters.

In Chapter \ref{chapter_word_models} we propose a novel model architecture for the continuous offline HTR at the word level. It is based on the combination of a Convolutional Neural Network (CNN) and the sequence-to-sequence (seq-to-seq) algorithm. We provide an extensive set of experiments analyzing the impact of each of the model components on the model error. This proposed model has demostrated that it obtains competitive results comparable to the state-of-the-art with the main offline continuous HTR databases. 

We found that the strategies of applying the proposed model directly on the image of each word or applying it on a sequence of patches extracted from each image provide results that favor the chunk partitioning strategy as the length of the image text increases.

We found that normalizing the handwritten text images improves the accuracy of the proposed recognition model. We proposed a novel normalization algorithm for handwriting images that significantly reduces the variability of handwritten text. As part of this algorithm, we presented a new slant angle identification algorithm and a new ascenders and descenders normalization height method. 

We found that applying an integrated data augmentation strategy in the trial process significantly improves the results of the proposed model.  We also propose a new data augmentation pipeline that includes the most relevant handwritten text image transformations from the last few years. We found that data normalization and data augmentation strategies are not mutually exclusive and that the combination of both provides better results than the use of each one separately.

We analyzed the contributions of other authors and identified the particularities of each published paper that make it difficult to compare the results between different models. We proposed a comparative framework that avoids the previous problems and that facilitates the fair comparison of results in the clearest possible way. In this line, we presented a wide set of results that facilitate comparing the proposed model with those of other authors, eliminating the identified biases. Numerous tables with partial results that allow us to evaluate the individual contribution of each of the proposed improvements in the results of the new model have also been provided.

Finally, we have publicly distributed \footnote{https://github.com/sueiras} all the source code used in the experiments performed in this Thesis under an open-source license. We have also provided the instructions and scripts necessary to replicate all the experiments performed in this Thesis.

\section{Future work}

We identified the following lines of future work on the construction of \textbf{isolated character recognition} models:

\begin{itemize}
    \item Experiment with other more recent network architectures that give good results on other Computer Vision problems, such as the EfficientNet \cite{tan2019efficientnet} models. 

    \item Experiment with other data augmentation techniques to improve the generality of the models built. For example, morphological erode and dilate, or elastic distortion techniques such as those suggested in \cite{ciresan2011convolutional}.

    \item Carry out cross-validations between the databases to better understand their generalization capacity. For this purpose, it should be taken into account that the different databases in origin have essential differences in terms of image size, border size, and whether they are binarized or not.

    \item Analyze the generalization capacity of the models built in real cases. One of the applications of recognizing isolated handwritten characters is the automatic reading of forms with writing templates that separate each character. In these cases, once the forms have been digitized and the characters have been isolated, their images can present artifacts coming from the template. Additionally, they may be cropped by the segmentation process. Analyzing the performance of models built on these templates can be of industrial relevance.

    \item It has been observed that the significant difference in distribution between the train and test partitions of the NIST database can distort the mean error. It is considered that balancing the training sample could correct these distortions. This balancing could be done by stratified sampling or by data augmentation applied on the less frequent characters.

\end{itemize}

The lines of future work identified on the construction of \textbf{continuous HTR models} at word level are the following:

\begin{itemize}

    \item Data analysis in this Thesis has shown that normalization and data augmentation processes are relevant in improving the accuracy. However, these are transformations on the text image that act in the opposite way. Thus data augmentation alters the previously normalized images at each training step. It is suspected that a data augmentation strategy that keeps as much as possible the characteristics of the normalized text may be more effective when used in conjunction with normalization. So further analysis of this interaction may be useful.

    \item In analyzing the different strategies of how to include the convolutional component in the model, it was hypothesized that dividing the image into patches might be more beneficial for large words. This hypothesis could be verified in future work.

    \item The use of pre-trained convolutional components on which training performs fine-tuning is a strategy that has worked well in other cases \cite{7426826} \cite{kang2018convolve}. Therefore, it will be explored as an alternative to the full training that has been used in this Thesis. 

    \item The use of an RNN type network in the decoder opens the door to integrating RNN-based language models as part of the decoder. It would have advantages over the usual option of applying the language model to the output of the visual model, as the visual and language models would be trained simultaneously.

    \item The model needs also to be tested at line level. It will also allow us to analyze the behavior of the patching strategy using online images. The modeling of the databases at line level also has the advantage that it allows us to compare the results with other authors who have only published results at that level, especially in the last few years \cite{wigington2018start}, \cite{ingle2019scalable}, \cite{kang2020pay}.

    \item The limitation observed in the performance of the current models due to the small size of the available annotated databases will be analyzed. Therefore, synthetic data generation strategies that have worked well in other Computer Vision problems could be explored. Based on the initial tests that have been performed, the most promising approach is the use of digital typographic fonts that emulate handwritten text as a mean of generating infinite samples of annotated handwriting data. This line of work is one of the most promising ones. 

    \item It is planned to test if this model can also work with other languages that do not use Latin characters, like for example, Chinese or Arabian ones. The proposed model can be applied directly to languages that read from right to left, or with minimal changes on vertically typed languages. 

\end{itemize}

\section{Publications}
The research carried out in this thesis has produced the following publications:

\begin{itemize}
\item Sueiras, J., Ruiz, V., Sanchez, A., and Velez, J. F. (2018). Offline continuous handwriting recognition using sequence to sequence neural networks. Neurocomputing, 289, 119-128.

\item Sueiras, J., Ruiz, V., Sánchez, Á., and Vélez, J. F. (2016, December). Using a Synthetic Character Database for Training Deep Learning Models Applied to Offline Handwritten Recognition. In International Conference on Intelligent Systems Design and Applications (pp. 299-308). Springer, Cham.

\item Ruiz, V., de Lena, M. T. G., Sueiras, J., Sanchez, A., and Velez, J. F. (2017, June). A Deep Learning Approach to Handwritten Number Recognition. In International Work-Conference on the Interplay Between Natural and Artificial Computation (pp. 193-202). Springer, Cham.

\end{itemize}

\printbibliography[heading=bibintoc]

\seeonlyindex{CNN}{Convolutional neural network}
\seeonlyindex{RNN}{Recurrent neural network}
\seeonlyindex{HMM}{Hidden Markov model}
\seeonlyindex{LSTM}{Long short-term memory network}
\seeonlyindex{GRU}{Gated recurrent unit}
\seeonlyindex{MLP}{Multilayer preceptron}
\seeonlyindex{CER}{Metrics: Character error rate}
\seeonlyindex{WER}{Metrics: Word error rate}
\seeonlyindex{CTC}{Connectionist temporal classification}
\seeonlyindex{MDRNN}{Multidimensional recurrent neural network}
\seeonlyindex{MDLSTM}{Multidimensional LSTM}
\seeonlyindex{PCA}{Principal component analysis}
\seeonlyindex{GMM}{Gaussian mixture model}
\seeonlyindex{BRNN}{Bidirectional recurrent neural networks}
\seeonlyindex{BPTT}{Back-propagation through time}
\seeonlyindex{LDA}{Linear discriminant analysis}
\seeonlyindex{SVM}{Support vector machines}
\seeonlyindex{seq2seq}{Sequence-to-sequence}
\seeonlyindex{RANSAC}{Random sample consensus}

\cleardoublepage 
\phantomsection
\setlength{\columnsep}{0.75cm} 
\addcontentsline{toc}{chapter}{\textcolor{black!90!white}{Index}} 
\printindex

\end{document}